%% file: main.tex
\documentclass[10pt,twocolumn,letterpaper]{article}

\usepackage[pagenumbers]{cvpr} 

\input{setup}

\begin{document}
\maketitle
\input{v3_sections/front}
\input{v3_sections/related}
\input{v3_sections/method}
\input{v3_sections/experiments}

\input{v3_sections/analysis}
\input{v3_sections/conclusion}

\input{llm_statement}

{
    \small
    \bibliographystyle{ieeenat_fullname}
    \bibliography{references}
}

\clearpage
\onecolumn
\setcounter{page}{1}
\renewcommand{\thepage}{S\arabic{page}}
\begin{center}
  {\Large\textbf{\thetitle}}\\[0.5em]
  {\Large Supplementary Material}
\end{center}
\vspace{1em}
\appendix
\setcounter{table}{0}\setcounter{figure}{0}\setcounter{equation}{0}\setcounter{algorithm}{0}
\renewcommand{\thetable}{S\arabic{table}}
\renewcommand{\thefigure}{S\arabic{figure}}
\renewcommand{\theequation}{S\arabic{equation}}
\renewcommand{\thealgorithm}{S\arabic{algorithm}}
\renewcommand{\theHtable}{S\arabic{table}}\renewcommand{\theHfigure}{S\arabic{figure}}
\renewcommand{\theHequation}{S\arabic{equation}}\renewcommand{\theHalgorithm}{S\arabic{algorithm}}
\input{v3_sections/appendix}

\end{document}

%% file: setup.tex
\usepackage[utf8]{inputenc}
\usepackage[T1]{fontenc}
\usepackage{microtype}
\usepackage{multirow}
\usepackage{array}
\usepackage{nicefrac}
\usepackage{amsthm}
\usepackage{algorithm}
\usepackage{algpseudocode}
\usepackage{float}
\usepackage{placeins}
\input{macros}

\usepackage{xr-hyper}
\definecolor{cvprblue}{rgb}{0.21,0.49,0.74}
\usepackage[pagebackref,breaklinks,colorlinks,allcolors=cvprblue]{hyperref}

\def\confName{CVPR}
\def\confYear{2027}

\title{HeadMem: Turning an LLM's Retrieval Heads into a Compact Memory for Later Question}

\author{
  Qingsen Ma$^{1}$ \quad
  Dianyun Wang$^{2}$ \quad
  Jiahong Liu$^{1}$ \quad
  Yaoye Wang$^{2}$ \quad
  Irwin King$^{1}$ \\[3pt]
  $^{1}$The Chinese University of Hong Kong \qquad
  $^{2}$Beijing University of Posts and Telecommunications \\
  {\tt\small \{qings.ma, jiahong.liu21\}@gmail.com, dianyunwang@bupt.edu.cn, king@cse.cuhk.edu.hk}
}

%% file: macros.tex
\newcommand{\ours}{HeadMem}                     
\newcommand{\ourstwo}{HeadMem-2}                
\newcommand{\Sthree}{S\textsuperscript{\smash{3}}}      
\newcommand{\bench}{RULER-MQ}                   
\newcommand{\pqc}{PQ-32}                        
\newcommand{\saec}{SAE-32}                      
\newcommand{\nr}{n/r}                           

\newif\ifdraft \draftfalse

\newcommand{\ci}[2]{{\scriptsize[$#1$,\,$#2$]}}   

\newcommand{\R}{\mathbb{R}}
\newcommand{\vq}{\mathbf{q}}
\newcommand{\vk}{\mathbf{k}}
\newcommand{\vkh}{\hat{\mathbf{k}}}
\newcommand{\vz}{\mathbf{z}}
\newcommand{\vc}{\mathbf{c}}
\newcommand{\Hset}{\mathcal{H}}       
\newcommand{\Gset}{\mathcal{G}}       
\newcommand{\Codec}{\mathcal{C}}
\newcommand{\Sel}{\mathcal{S}}        
\newcommand{\enc}{\operatorname{enc}}
\newcommand{\dec}{\operatorname{dec}}
\newcommand{\softmax}{\operatorname{softmax}}
\newcommand{\TopK}{\operatorname{TopK}}
\newcommand{\massq}{m^{\mathrm{q}}}       
\newcommand{\masslast}{m^{\mathrm{last}}} 
\newcommand{\massp}{m^{\mathrm{p}}}       
\newcommand{\sink}{\langle\mathrm{s}\rangle}
\newcommand{\lmax}{\ell_{\max}}

%% file: v3_sections/front.tex
\begin{abstract}
Documents, codebases and agent histories are read long before their questions arrive and are then queried many times.
Their memory may serve any later question.
Compressing the KV cache before the questions keeps this memory small but loses the evidence.
Storing the full cache keeps the evidence but costs gigabytes per document, and dense retrievers bring an embedding model and an index of their own.
We show that the LLM already contains a retriever that keeps the memory small and still finds the evidence.
Its retrieval heads, detected once per model, find the evidence even when they read the question alone. They are token-matching heads, and masking them hurts the model's own long-context reading far more than masking random heads.
\ours{} stores only the compressed keys of these heads, using $290\times$ less memory than the KV cache of Llama-3.1-8B.
For each question, the heads' own softmax selects the tokens the model re-reads.
It needs no training, no model besides the LLM and no pass over the context per question.
On multi-question RULER-style contexts, \ours{} alone beats four KV compression methods by a wide margin.
Because it only proposes tokens to re-read, it combines readily with the memories already in use.
On Llama-3.1-8B it can share the token budget of a KV cache compressed by StreamingLLM or SnapKV and keep over 90\% of its full-budget score at every length.
Fused with BM25 or a dense retriever, it beats BM25 and reranked dense RAG at 128k tokens on both this model and Qwen2.5-7B.
\ours{} also works when question and evidence share no words, on long chat histories and beyond the model's context window.
\end{abstract}

\section{Introduction}\label{sec:intro}

\begin{figure*}[t]
\centering
\includegraphics[width=\linewidth]{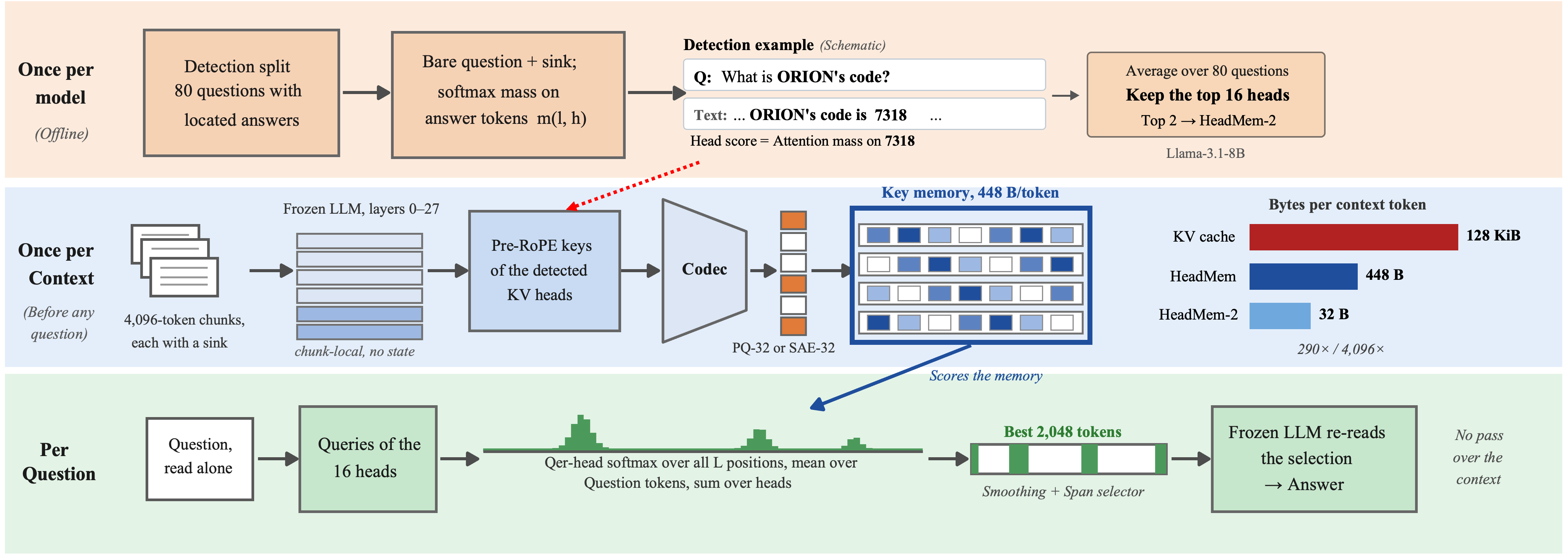}
\caption{\textbf{Overview of \ours{}.} Detection picks the heads once per model, and the grid shows the 16 detected heads of Llama-3.1-8B. Their key codes are stored once per context. Each question, read alone, scores every position with these heads' softmax, and the model re-reads the best 2{,}048 tokens. The score strip and the codes are schematic.}
\label{fig:overview}
\end{figure*}

Many long-context workloads read first and ask later.
Documents, codebases and agent histories are ingested before their questions are known, and are then queried many times~\cite{li2025scbench,wu2025longmemeval}.
Their memory may therefore serve later questions.
The natural memory is the KV cache, but it is large and slow to build.
The bf16 KV cache of Llama-3.1-8B~\cite{dubey2024llama3} takes 128\,KiB per token, or 16\,GiB at 128k tokens.
Prefilling those 128k tokens takes 17.7\,s on an H200 GPU.

Existing memories compress the context before the questions, keep it reachable for them or search it with the model's own attention.
KV compression keeps a fixed subset of the cache~\cite{xiao2024efficientstreaminglanguagemodels,zhang2023h2o,li2024snapkv,zefan2024pyramidkv,kim2025kvzip}.
Reachable memories keep the full cache, or its values, in GPU or host memory~\cite{tang2024quest,xiao2024infllm,liu2024retrievalattention,sun2024shadowkv,zhang2025pqcache}, or index the text for an external retriever~\cite{robertson2009bm25,xiao2024cpack,chen2024m3}.
Attention-based retrievers use the answering model's own attention layers~\cite{zhang2025query,chen2025icr,shi2024gemfilter,ye2025infinite,soh2025youra,song2025reform,ilre2025}, and \Sthree{}-Attention~\cite{ma2026s3attention} made this idea query-agnostic by indexing keys before any question.

None of these memories serves later questions well.
Compression decides what to keep before any question exists.
On 128k-token multi-question contexts (\bench{}) with Llama, four compression methods score 6.9--10.3 where the full cache scores 69.4 (Section~\ref{sec:exp-e2}).
Reachable memories pay for recall with the full cache or a second model, or limit it to word overlap.
Attention-based retrievers read the question with the context, so each new question touches the context or a carried state again.
Query-agnostic key indexes~\cite{ma2026s3attention} avoid this, but voting over sparse features of head-averaged keys retrieves almost at random on \bench{} (Section~\ref{sec:ana-ladder}).
No current memory is small, built before the questions and able to find evidence with the model's own attention.

We show that the model already contains the retriever such a memory needs.
Its retrieval heads~\cite{wu2024retrieval}, detected with bare questions, keep retrieving when their keys are computed before any question and the question is read alone.
At 128k tokens, the 16 detected heads of Llama-3.1-8B place the answer inside a 2{,}048-token selection for 80.9\% of development questions, against 69.3\% for BM25.

\ours{} turns these heads into a memory in three steps (Figure~\ref{fig:overview}).
Once per model, it ranks all heads by the attention that bare questions put on their answers and keeps the top 16 by default.
Once per context, a chunk-local forward stores product-quantized codes of these heads' keys at 448\,B per token.
For each question, read alone, every detected head scores all stored positions with its own softmax, and the model re-reads the best 2{,}048 tokens.
\ourstwo{} keeps only Llama's top two heads at 32\,B per token.
Like REFORM~\cite{song2025reform}, \ours{} compresses the context, gathers tokens and re-reads them, but it keeps no KV cache and reads each question alone.

\ours{} extends the context accessible to a frozen LLM using its own retrieval signal, with no training or additional neural retriever. Because it only proposes tokens to re-read, it combines readily with KV compression and text retrievers.
A pipeline that already keeps a compressed cache can hand part of its budget to \ours{}, and \ours{}'s scores can be fused with a retriever's.
On \bench{} with Llama, a compressed cache that gives half of its 2{,}048 entries to \ours{} gains 15.52 to 59.77 points.
With StreamingLLM and SnapKV, \ours{} then keeps over 90\% of its full-budget score at every length (Section~\ref{sec:exp-stack}).
Fused with BM25 or with a dense retriever, \ours{} beats BM25 and reranked dense RAG at 128k on both Llama-3.1-8B and Qwen2.5-7B.
There the dense fusion also scores above \ours{} alone.
On its own it beats all four compression methods, and with BM25 it tops every baseline in Table~\ref{tab:e2} that reads 2{,}048 tokens.
With Llama, \ourstwo{} answers eight questions on a 128k context $13\times$ faster than the full cache, at a score not significantly different from it.
We make four contributions.

\begin{itemize}
\item We find that detected retrieval heads locate the evidence even when their keys are computed before any question and the question is read alone.
\item We propose \ours{}, a training-free memory for the query-agnostic setting that needs no model besides the LLM. It stores the product-quantized keys of the detected heads, 448\,B per token on Llama-3.1-8B or 32\,B with two heads. It scores positions with each head's own softmax and combines with KV compression and text retrievers.
\item We evaluate it alone and combined with existing memories. The comparison covers KV compression, text retrievers, the full cache and ported attention-based retrievers on six frozen models, five benchmarks and up to 1M tokens, with Holm-corrected tests. \ours{} beats BM25 by up to 24.2 points when needle and question share no words.
\item We show that the detected heads are token-matching heads, induction-like on Llama and Mistral. On Llama and Qwen2.5, masking them cuts the model's own \bench{} score far more than masking random heads of the same layers. An ablation ladder shows that head selection and the per-head softmax carry the signal. Compressing the keys eightfold changes accuracy significantly in only one of nine paired tests.
\end{itemize}

%% file: v3_sections/related.tex
\section{Related work}\label{sec:related}

\paragraph{Compressing the cache before the questions.}
Eviction keeps a fixed subset of the cache~\cite{xiao2024efficientstreaminglanguagemodels,zhang2023h2o,li2024snapkv,zefan2024pyramidkv,kim2025kvzip}, and DuoAttention, RazorAttention and HeadKV keep more cache for retrieval-like heads~\cite{xiao2024duoattention,tang2025razor,fu2025headkv}.
Infini-attention trains a compressive memory~\cite{munkhdalai2024infini}.
These memories choose what survives before any question exists.
\ours{} keeps no values and can fill part of their budget with the tokens each question needs (Section~\ref{sec:exp-stack}).

\paragraph{Keeping the context reachable.}
Offloading, sparse attention and PQCache keep the full cache and select from it at each decoding step~\cite{tang2024quest,xiao2024infllm,liu2024retrievalattention,sun2024shadowkv,chen2025magicpig,zhang2025pqcache}.
RAG ranks passages by word overlap~\cite{robertson2009bm25} or with a separate model~\cite{lewis2020retrieval,xiao2024cpack,chen2024m3,li2023making}.
\ours{} keeps 448\,B per token on Llama-3.1-8B instead of the cache and fuses with a retriever's scores (Section~\ref{sec:exp-stack}).

\paragraph{The model's attention as a retriever.}
Retrieval heads are sparse, universal and causal~\cite{wu2024retrieval}, and synthetic context extension acts through them~\cite{zhao2024understanding}.
Attention re-ranks documents in ICR~\cite{chen2025icr}, QRHead~\cite{zhang2025query} and CoRe~\cite{core2025heads}, and filters the input in GemFilter~\cite{shi2024gemfilter}.
InfiniRetri~\cite{ye2025infinite} slides the question over chunks.
YOURA~\cite{soh2025youra} and ILRe~\cite{ilre2025} select tokens with the question's attention and re-read them, and MRAG embeds chunks with the heads of a separate model~\cite{besta2024mrag}.
REFORM~\cite{song2025reform} is the closest design. It compresses a recurrent cache to build few-head embeddings, gathers the tokens most similar to the end of the input and recomputes their cache.
A dense retriever adds to it, as it does to \ours{} at 128k (Section~\ref{sec:exp-stack}).
All methods above except MRAG read the question together with the context, so a new question needs the KV cache, a compressed recurrent cache or a new prefill.
\Sthree{}-Attention~\cite{ma2026s3attention} made the setting query-agnostic. It indexes head-averaged keys in one chunk-local pass before any question and votes over their sparse features.
\ours{} keeps this setting and replaces the scoring with retrieval heads detected once per model, their own per-head softmax and a product-quantized key store (Section~\ref{sec:ana-ladder}).
Section~\ref{sec:exp-attn} compares the scoring operators of REFORM and ILRe inside one query-agnostic pipeline, and Appendix~\ref{app:related} compares all these methods (Table~\ref{tab:prior-art}).

%% file: v3_sections/method.tex
\section{\ours{}: memory read by retrieval heads}
\label{sec:method}

\ours{} uses a frozen decoder LLM with grouped query attention, in which query head $h$ of layer $\ell$ reads KV head $g(h)$ and heads have dimension $d=128$. A context $C=(c_1,\dots,c_L)$ is available before any question. Each later question is answered from the same memory by reading $B=2{,}048$ selected context tokens. Heads are detected once per model, the index is built once per context, and positions are scored once per question.

\subsection{Retrieval-head detection}
\label{sec:method-detect}

\begin{figure}[t]
\centering
\includegraphics[width=\linewidth]{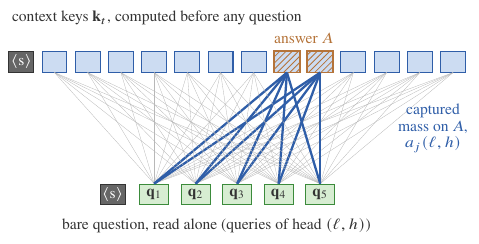}
\caption{\textbf{What detection captures} (schematic, one head $(\ell,h)$, line widths illustrative). Context keys are computed before any question, and the bare question is forwarded alone, each after a sink $\sink$. Each question token's softmax over the context positions puts mass $a_j(\ell,h)$ on the answer tokens $A$ (thick lines). Eq.~\eqref{eq:detect} averages this mass, and the 16 heads with the largest average are kept.}
\label{fig:detect}
\end{figure}

Most attention heads attend to sinks and nearby tokens, and only a few copy evidence from far back in the context~\cite{wu2024retrieval,tang2025razor}.
\ours{} scores its memory with these few heads, and it finds them once per model with a test that mirrors how the memory is read later (Figure~\ref{fig:detect}).

The test uses a separate detection split of 80 questions over 16k-token \bench{} contexts, two episodes of eight questions from each of the five task families.
Its QA items come from the development partitions of SQuAD and HotpotQA.
The answer positions $A$ of a question are the context tokens that overlap an occurrence of one of its gold answers.

Detection reads every head under the conditions of the memory.
Exact keys come from the chunk-local forward of Section~\ref{sec:method-index}, with a sink at the start of each 4{,}096-token chunk.
Keys and queries are pre-RoPE.
Queries $\vq_j^{\ell,h}$ come from the bare question, without task template, forwarded alone after a sink token $\sink$.
The question therefore never meets the context in one forward pass, as at query time.

For each head $(\ell,h)$ and question token $j$, a softmax over all $L$ context positions gives the mass $a_j(\ell,h)$ that the token puts on $A$ (Figure~\ref{fig:detect}).
A retrieval head puts more of its softmax mass on $A$ than other heads do.
The detection score averages $a_j$ over question tokens, over the questions $\mathcal D_f$ of each family $f\in\mathcal F$ and then over families,
\begin{equation}\label{eq:detect}
\begin{split}
\massq(\ell,h)&=\frac{1}{|\mathcal F|}\sum_{f\in\mathcal F}\frac{1}{|\mathcal D_f|}\sum_{(C,Q,A)\in\mathcal D_f}\frac{1}{|Q|}\sum_{j=1}^{|Q|}a_j(\ell,h),\\
a_j(\ell,h)&=\sum_{t\in A}\softmax_{t}\Big(\frac{\vq_j^{\ell,h\top}\vk_t^{\ell,g(h)}}{\sqrt d}\Big).
\end{split}
\end{equation}
$\Hset$ holds the $H=16$ query heads with the largest $\massq$, as in QRHead~\cite{zhang2025query}.
$\Gset$ holds the KV heads that host them, and $\lmax$ is their deepest layer.
For Llama-3.1-8B, $\Hset$ is drawn from 1{,}024 query heads, spans layers 2--27 ($\lmax=27$) and occupies 14 of the model's 256 KV heads.
Its top two heads, (2,22) and (2,21), share one KV head of layer 2.
Section~\ref{sec:ana-heads} compares $\massq$ with $\masslast$, which queries with the last prompt token.
There, heads re-detected from a new seed or 32k contexts are not significantly different from $\Hset$ on test at 128k.

\subsection{Building the index}
\label{sec:method-index}

\ours{} forwards $C$ in chunks of $B_c=4{,}096$ tokens, each as $[\sink;\,\text{chunk}]$ with positions restarting per chunk, and stops after layer $\lmax$. For every KV head $g\in\Gset$ it stores the code $\vz_t^{g}=\enc_g(\vk_t^{g})$ of the pre-RoPE key $\vk_t^{g}\in\R^d$ of every context position (sink row dropped) next to its token id. It keeps no values, no keys of other heads and no state across chunks. The index is thus appendable and parallel. No forward reads more than $B_c+1$ tokens, so contexts can exceed the model's window.
A sink token is prepended to each chunk and question and later dropped. It absorbs the attention sink that each chunk's first token would otherwise become~\cite{xiao2024efficientstreaminglanguagemodels,gu2025attentionsink,yang2025ape}.

\subsection{Key codecs}
\label{sec:method-codec}

A codec is a per-KV-head pair $(\enc_g,\dec_g)$ with $b_\Codec$ bytes per head and token. It must give a fast evaluation of $\vq^{\top}\vkh_t^g$ for all $t$, where $\vkh_t^g=\dec_g(\vz_t^g)$. This is all that Eq.~\eqref{eq:score} needs. Exact bf16 keys (256\,B per head) and per-token int8, int4 and int2 keys serve as references (Appendix~\ref{app:memory-full}).

\paragraph{\pqc{} (default).}
Product quantization~\cite{jegou2011pq,douze2024faiss} splits $\vk_t$ into $M=32$ sub-vectors $\vk_t^{(m)}$ of $d/M=4$ dimensions. It stores the index of the nearest of 256 centroids for each sub-vector (32\,B per head),
\begin{equation}\label{eq:pq}
\begin{split}
z_{t,m}&=\arg\min_{i\le 256}\big\|\vk_t^{(m)}-\vc_i^{(m)}\big\|_2,\\
\vq^{\top}\vkh_t&=\sum_{m=1}^{M}\vq^{(m)\top}\vc^{(m)}_{z_{t,m}}.
\end{split}
\end{equation}
Codebooks are fit per KV head on 204k WikiText-2 keys~\cite{merity2017pointer}. Scoring takes one $32\times256$ inner-product table per query vector and 32 lookups per position. $M=32$ is the smallest $M\in\{16,32,64\}$ that stays within 0.02 of the development answer presence of exact keys at 32k and 128k (Section~\ref{sec:ana-codec}).

\paragraph{\saec{} and \ourstwo{}.}
The design is codec-agnostic. \saec{}, a per-KV-head Top-$k$ sparse autoencoder~\cite{gao2024scaling}, writes each key as $k=32$ of 8{,}192 atoms in 128\,B per head (Appendix~\ref{app:sae-codec}). It is the analysis lens of the ablations (Section~\ref{sec:ana-ladder}). \ourstwo{} is the throughput configuration. It keeps the top two heads of Llama-3.1-8B, which share a KV head of layer 2, at 32\,B per token with \pqc{}. Its index and question passes stop at layer 2.

\subsection{Scoring and selection}
\label{sec:method-score}

For a question $Q$, \ours{} forwards the bare question $[\sink;\,Q]$ through layer $\lmax$. It then scores every context position $t=1,\dots,L$ with the pre-RoPE queries $\vq_j^{\ell,h}$ of the heads in $\Hset$,
\begin{equation}\label{eq:score}
s(t)=\sum_{(\ell,h)\in\Hset}\frac{1}{|Q|}\sum_{j=1}^{|Q|}\softmax_{t}\Big(\frac{\vq_j^{\ell,h\top}\vkh_t^{\ell,g(h)}}{\sqrt d}\Big).
\end{equation}
Per-head normalization keeps heads with large logit norms from dominating~\cite{wu2024tokenselect}.

\paragraph{Selector.}
The selector is shared with every retrieval baseline that scores positions, as in the common setup of Section~\ref{sec:exp}. It smooths $s$ with a width-8 box kernel into $\bar s$ and takes the first and last $B/16=128$ tokens. It then adds 128-token spans centred on the highest remaining peaks of $\bar s$ and fills the leftover budget with the highest-scoring tokens, so that $|\Sel_B(\bar s)|=\min(B,L)$. The model re-reads the selection in original order inside the task prompt, followed by the question, and answers greedily.

\paragraph{Fusion.}
\ours{}+BM25 mixes $\bar s$ with the BM25 score $b(t)$ of the 128-token chunk that contains $t$~\cite{robertson2009bm25}. With the BM25 weight fixed at 0.5 and never tuned, the fused score is
\begin{equation}\label{eq:fuse}
s_{\mathrm{fuse}}(t)=\tfrac12\,\widetilde{\bar s}(t)+\tfrac12\,\tilde b(t),
\end{equation}
where $\tilde x=(x-\min x)/(\max x-\min x)$ normalizes a score over the context.

\subsection{Memory and compute}
\label{sec:method-memory}

The index takes $|\Gset|\cdot b_{\Codec}$ bytes per token (Eq.~\eqref{eq:bytes} in Appendix~\ref{app:memory-full}) plus one token id. For Llama-3.1-8B with \pqc{} this is 448\,B per token (56\,MiB at 128k), $290\times$ below the bf16 KV cache. \ourstwo{} takes 32\,B per token ($4{,}096\times$ below the KV cache), \saec{} 1{,}792\,B and exact keys 3{,}584\,B (Table~\ref{tab:memory}).
Building the index costs one chunk-local LLM forward through layer $\lmax$ plus key encoding, paid once per context. A question costs a forward of the bare question through layer $\lmax$, $|\Hset|\cdot|Q|\cdot L$ code scores and an answer pass over $B$ tokens plus the prompt. It never runs the LLM over the context. Section~\ref{sec:exp-kv} measures both costs.

%% file: v3_sections/experiments.tex
\section{Experiments}\label{sec:exp}

The experiments answer five research questions.
\textbf{RQ1} asks whether one memory built before the questions answers them at a fixed reading budget (Section~\ref{sec:exp-e2}).
\textbf{RQ2} asks whether it carries over to other tasks and domains (Section~\ref{sec:exp-nolima}).
\textbf{RQ3} asks whether it works inside pipelines that already keep a compressed cache or a retriever (Section~\ref{sec:exp-stack}).
\textbf{RQ4} asks which design choices carry the signal (Section~\ref{sec:ana-ladder}).
\textbf{RQ5} asks what the memory costs to build, store and query (Section~\ref{sec:exp-kv}).
Section~\ref{sec:analysis} then asks why the detected heads retrieve.

\subsection*{Common setup and baselines}\label{sec:exp-setup}

Llama-3.1-8B-Instruct~\cite{dubey2024llama3} and Qwen2.5-7B-Instruct~\cite{qwen2024qwen25} with YaRN $\times4$~\cite{peng2023yarn} have 128k-token windows, and Mistral-7B-Instruct-v0.3~\cite{jiang2023mistral} and Qwen2-7B-Instruct~\cite{yang2024qwen2} have 32k windows.
All models are frozen and decode greedily (Appendix~\ref{app:config}).
FullKV within the supported window is the same-model reference for retained answering ability.
Every \bench{} episode is one context with $Q{=}8$ independent questions answered from the same memory.
\bench{} has five RULER-template families~\cite{hsieh2024ruler}, namely multi-key and multi-value NIAH, variable tracking, and QA over real SQuAD~\cite{rajpurkar2016squad} and HotpotQA~\cite{yang2018hotpotqa} paragraphs.
Llama and Qwen2.5 have 10 test episodes per family and length at 32k--128k, or 400 questions per length.
Mistral and Qwen2 have 6, and the runs at 256k--1M have 3 (Appendix~\ref{app:protocol-mq}).

All methods except FullKV read $B{=}2{,}048$ context tokens in the same prompt.
The text baselines are BM25~\cite{robertson2009bm25} over 128-token chunks through our selector and reranked dense RAG~\cite{lewis2020retrieval} with bge-small~\cite{xiao2024cpack} and bge-reranker-v2-m3~\cite{chen2024m3,li2023making}.
We also fuse BM25 with bge-small, lexical density or BGE-M3~\cite{chen2024m3}, with weights tuned on development data.
StreamingLLM~\cite{xiao2024efficientstreaminglanguagemodels}, SnapKV~\cite{li2024snapkv}, PyramidKV~\cite{zefan2024pyramidkv} and KVzip~\cite{kim2025kvzip} (kvpress~\cite{nvidia2024kvpress}) compress the cache once, to 2,048 entries, before the questions.
The \Sthree{}-Attention-style index reimplements the feature-voting index of \Sthree{}-Attention~\cite{ma2026s3attention}, and \Sthree{}-Hybrid is its union with BM25.
REFORM-style and ILRe-style rows port the scoring operators of REFORM~\cite{song2025reform} and ILRe~\cite{ilre2025} into \ours{}'s pipeline (Appendices~\ref{app:protocol-baselines} and~\ref{app:protocol-ports}).
The default 16-head configuration was fixed on Llama development data, and the BM25 weight in Eq.~\eqref{eq:fuse} is 0.5 and was never tuned.
$\Delta$ is a paired difference, first-named minus second, with a 95\% bootstrap CI.
\emph{Significant} means Holm $p<0.05$~\cite{holm1979simple}, marked $^*$, and a CI including zero means \emph{not significantly different}, not equivalent (Appendix~\ref{app:stats}).
Appendix~\ref{app:limits} collects the settings where \ours{} does not lead.

\subsection{RQ1: Can one memory answer later questions?}\label{sec:exp-e2}

\begin{table*}[t]
\centering
\caption{\textbf{One memory per context, eight questions each (\bench{}).} Macro score ($\times100$), 400 questions per length. All methods except FullKV read $B{=}2{,}048$ context tokens, and eviction keeps 2,048 entries. B/token is the state kept per context token on Llama. Port rows move the scoring operators of REFORM and ILRe into \ours{}'s pipeline, and further variants are in Appendix~\ref{app:e2-full}. Bold marks the top score among methods reading $B$ tokens. Brackets give the 95\% CI of each score, with episodes bootstrapped within family (Appendix~\ref{app:stats}). Paired differences with CIs are in Tables~\ref{tab:e2-deltas-pq-llama} and~\ref{tab:e2-deltas-pq-qwen}.}
\label{tab:e2}
\footnotesize
\setlength{\tabcolsep}{3.2pt}
\renewcommand{\arraystretch}{1}
\resizebox{\ifdim\width>\linewidth\linewidth\else\width\fi}{!}{%
\begin{tabular}{@{}lr*{6}{r@{\,}l}@{}}
\toprule
 & & \multicolumn{6}{c}{Llama-3.1-8B-Instruct} & \multicolumn{6}{c}{Qwen2.5-7B-Instruct (YaRN)} \\
\cmidrule(lr){3-8}\cmidrule(l){9-14}
Method & B/token & \multicolumn{2}{c}{32k} & \multicolumn{2}{c}{64k} & \multicolumn{2}{c}{128k} & \multicolumn{2}{c}{32k} & \multicolumn{2}{c}{64k} & \multicolumn{2}{c}{128k} \\
\midrule
FullKV & 131,072 & 83.8 & {\scriptsize[81.6, 85.9]} & 80.4 & {\scriptsize[77.5, 83.2]} & 69.4 & {\scriptsize[66.3, 72.5]} & 71.9 & {\scriptsize[68.7, 75.0]} & 62.6 & {\scriptsize[59.5, 65.9]} & 55.8 & {\scriptsize[52.8, 58.8]} \\
StreamingLLM & -- & 14.1 & {\scriptsize[11.3, 17.0]} & 12.1 & {\scriptsize[9.6, 14.8]} & 6.9 & {\scriptsize[4.6, 9.5]} & 11.4 & {\scriptsize[8.6, 14.5]} & 10.6 & {\scriptsize[7.9, 13.4]} & 7.4 & {\scriptsize[5.3, 9.5]} \\
SnapKV & -- & 12.5 & {\scriptsize[9.3, 16.1]} & 13.4 & {\scriptsize[9.9, 17.8]} & 10.3 & {\scriptsize[7.7, 12.8]} & 7.1 & {\scriptsize[4.3, 10.2]} & 9.3 & {\scriptsize[5.8, 14.3]} & 6.7 & {\scriptsize[4.5, 8.8]} \\
PyramidKV & -- & 12.4 & {\scriptsize[9.4, 15.7]} & 12.9 & {\scriptsize[9.3, 17.8]} & 9.8 & {\scriptsize[7.3, 12.1]} & 5.7 & {\scriptsize[3.5, 8.0]} & 9.7 & {\scriptsize[6.1, 14.7]} & 6.0 & {\scriptsize[4.0, 8.0]} \\
KVzip & -- & 57.1 & {\scriptsize[53.3, 60.9]} & 22.3 & {\scriptsize[19.1, 25.5]} & 8.5 & {\scriptsize[6.8, 10.3]} & 39.7 & {\scriptsize[36.2, 43.1]} & 20.6 & {\scriptsize[18.0, 23.2]} & 12.4 & {\scriptsize[10.2, 14.5]} \\
\midrule
BM25, same selector & -- & 67.7 & {\scriptsize[64.1, 71.1]} & 65.3 & {\scriptsize[62.1, 68.3]} & 66.0 & {\scriptsize[63.3, 68.9]} & 64.0 & {\scriptsize[60.1, 67.8]} & 62.9 & {\scriptsize[59.8, 65.9]} & 62.4 & {\scriptsize[58.6, 66.0]} \\
Dense RAG + reranker & -- & 72.6 & {\scriptsize[69.8, 75.4]} & 66.7 & {\scriptsize[63.7, 69.7]} & 66.2 & {\scriptsize[63.0, 69.3]} & 67.0 & {\scriptsize[63.9, 70.0]} & 66.7 & {\scriptsize[64.0, 69.3]} & 61.8 & {\scriptsize[57.9, 65.5]} \\
BM25 + dense & -- & 72.3 & {\scriptsize[68.8, 75.5]} & 69.5 & {\scriptsize[66.5, 72.5]} & 72.5 & {\scriptsize[69.9, 75.2]} & 66.3 & {\scriptsize[62.7, 69.9]} & 66.8 & {\scriptsize[63.4, 69.9]} & 66.6 & {\scriptsize[63.0, 70.0]} \\
BM25 + lexical density & -- & 73.0 & {\scriptsize[70.2, 75.6]} & 69.5 & {\scriptsize[66.6, 72.3]} & 69.5 & {\scriptsize[66.8, 71.7]} & 68.9 & {\scriptsize[65.6, 72.0]} & 67.3 & {\scriptsize[64.9, 69.7]} & 67.8 & {\scriptsize[64.6, 71.2]} \\
\midrule
\multicolumn{14}{@{}l}{\textit{The model's own states, question read alone}} \\
\Sthree{}-Attention-style index & -- & 11.0 & {\scriptsize[8.6, 13.5]} & 7.8 & {\scriptsize[5.6, 10.1]} & 5.8 & {\scriptsize[3.8, 8.2]} & 11.5 & {\scriptsize[9.1, 14.1]} & 7.5 & {\scriptsize[5.4, 9.6]} & 6.5 & {\scriptsize[4.5, 8.5]} \\
\Sthree{}-Hybrid (index $\cup$ BM25) & -- & 67.2 & {\scriptsize[64.2, 70.0]} & 65.8 & {\scriptsize[63.2, 68.2]} & 66.2 & {\scriptsize[64.0, 68.3]} & 64.5 & {\scriptsize[60.6, 68.2]} & 63.1 & {\scriptsize[60.3, 65.9]} & 62.1 & {\scriptsize[59.2, 65.1]} \\
ILRe-style (port) & 2,048 & 23.3 & {\scriptsize[20.2, 26.6]} & 15.7 & {\scriptsize[13.3, 18.2]} & 14.3 & {\scriptsize[11.3, 17.3]} & 33.9 & {\scriptsize[30.7, 37.2]} & 19.8 & {\scriptsize[16.7, 22.9]} & 15.3 & {\scriptsize[11.5, 19.1]} \\
REFORM-style (port) & 768 & 66.6 & {\scriptsize[64.0, 69.2]} & 62.6 & {\scriptsize[59.8, 65.3]} & 64.5 & {\scriptsize[61.9, 67.1]} & 66.4 & {\scriptsize[63.1, 69.6]} & 61.4 & {\scriptsize[58.9, 63.9]} & 64.6 & {\scriptsize[62.1, 67.1]} \\
\ours{} (\pqc{}) & 448 & 74.0 & {\scriptsize[71.0, 76.9]} & 71.6 & {\scriptsize[68.9, 74.3]} & 72.1 & {\scriptsize[69.0, 74.9]} & 66.6 & {\scriptsize[63.4, 69.5]} & 65.5 & {\scriptsize[62.0, 68.8]} & 60.9 & {\scriptsize[57.8, 64.1]} \\
\ours{}+BM25 & 448 & \textbf{76.0} & {\scriptsize[73.3, 78.5]} & \textbf{72.9} & {\scriptsize[70.2, 75.5]} & \textbf{73.9} & {\scriptsize[71.4, 76.3]} & \textbf{72.1} & {\scriptsize[69.2, 75.0]} & \textbf{69.9} & {\scriptsize[67.2, 72.5]} & \textbf{68.2} & {\scriptsize[65.6, 70.8]} \\
\bottomrule
\end{tabular}}
\end{table*}

\ours{}+BM25, one memory built before the questions, scores highest among the methods of Table~\ref{tab:e2} that read $B$ tokens, on both models at every length.
It significantly beats BM25 at every length.
It also significantly beats reranked dense RAG at 64k and 128k on Llama and at 32k on Qwen2.5.
On Llama its margin over dense RAG grows from 32k to 128k, where it reaches $+7.72$~[$+4.29$, $+11.10$].
On Llama \ours{} alone already significantly beats BM25 at every length, and no text retriever, tuned fusions included, scores significantly above it.

Compressing the cache before the questions loses the evidence that later questions need.
All four compressed caches score far below \ours{} alone on both models, and all 24 of these gaps are significant.
KVzip leads the four at 32k and 64k but falls as the context grows.
The full cache keeps all the evidence at $290\times$ the memory.
\ours{}+BM25 scores above it at 128k on Llama and significantly above it at 64k and 128k on Qwen2.5.

\label{sec:exp-attn}%
Other memories built from the model's own states fall behind.
The feature-voting index of \Sthree{}-Attention finds little of the evidence, and \ours{}+BM25 significantly beats its union with BM25 at every length on both models (Table~\ref{tab:holm-main}).
REFORM and ILRe read the question with the context, and \ours{} significantly beats our ports of both at every length on Llama, with less memory than either.
On Qwen2.5 it significantly beats ILRe-style, and \ours{}+BM25 significantly beats REFORM-style at every length.

The two QA families answer their questions from real SQuAD and HotpotQA paragraphs among distractors (Appendix~\ref{app:qa-subset}).
There \ours{}+BM25 scores above BM25 at 32k and 128k on Llama and above the full cache at 128k on both models, all CIs excluding zero.

\label{sec:exp-beyond}%
Chunk-local indexing lets the memory reach past the model's window (Appendices~\ref{app:e2-32k} and~\ref{app:beyond-full}).
In a new five-family stress test at 1M indexed tokens, \ours{}+BM25 scores 67.33--72.93 across Qwen2.5-7B/14B/32B, with 16/39/52 heads and a 2k reading budget (Appendix~\ref{app:scaling}).

\subsection{RQ2: Does it carry over to other tasks?}\label{sec:exp-nolima}\label{sec:exp-lb}

\begin{table*}[t]
\centering
\caption{\textbf{Tasks and domains} (all retrieval methods read $B{=}2{,}048$ tokens). (a) NoLiMa, no lexical overlap between question and needle (Llama-3.1-8B, 290 samples per length, FullKV truncated to the window at 128k). Paired $\Delta$s in the text average per-needle means. (b) LongBench QA, macro F1 ($\times100$) on six tasks, 920 test examples per model, FullKV middle-truncated. (c) LongMemEval-S, chat histories of about 105k tokens (Llama-3.1-8B), evidence-turn recall of the selection (479 questions) and answer containment (470 non-abstention questions), in \%. Parentheses give bytes per context token, the same on all three models. Bold marks the top score among methods reading $B$ tokens. Each $\pm$ is half the width of the score's 95\% bootstrap CI (Appendix~\ref{app:stats}).}
\label{tab:real}
\label{tab:nolima}
\footnotesize
\setlength{\tabcolsep}{4.2pt}
\renewcommand{\arraystretch}{1}
\resizebox{\ifdim\width>\linewidth\linewidth\else\width\fi}{!}{%
\begin{tabular}{@{}lcccccccc@{}}
\toprule
 & \multicolumn{3}{c}{(a) NoLiMa, Llama} & \multicolumn{3}{c}{(b) LongBench QA} & \multicolumn{2}{c}{(c) LongMemEval-S} \\
\cmidrule(lr){2-4}\cmidrule(lr){5-7}\cmidrule(l){8-9}
Method & 8k & 32k & 128k & Llama & Mistral & Qwen2 & evid.\ recall & answer \\
\midrule
FullKV & 31.7{\scriptsize$\,\pm$5.2} & 15.5{\scriptsize$\,\pm$4.1} & 1.4{\scriptsize$\,\pm$1.2} & 44.97{\scriptsize$\,\pm$2.5} & 38.95{\scriptsize$\,\pm$2.5} & 37.93{\scriptsize$\,\pm$2.4} & (all) & 34.0{\scriptsize$\,\pm$4.4} \\
BM25, same selector & 9.0{\scriptsize$\,\pm$3.3} & 0.3{\scriptsize$\,\pm$0.5} & 0.3{\scriptsize$\,\pm$0.5} & 39.88{\scriptsize$\,\pm$2.6} & 36.77{\scriptsize$\,\pm$2.4} & 37.53{\scriptsize$\,\pm$2.4} & 72.5{\scriptsize$\,\pm$3.5} & 38.7{\scriptsize$\,\pm$4.4} \\
Dense RAG + reranker & 20.7{\scriptsize$\,\pm$4.8} & 9.3{\scriptsize$\,\pm$3.3} & 4.8{\scriptsize$\,\pm$2.6} & \textbf{43.14}{\scriptsize$\,\pm$2.5} & \textbf{39.40}{\scriptsize$\,\pm$2.4} & \textbf{39.43}{\scriptsize$\,\pm$2.5} & 80.8{\scriptsize$\,\pm$2.9} & \textbf{45.3}{\scriptsize$\,\pm$4.5} \\
\ours{} (448) & \textbf{34.8}{\scriptsize$\,\pm$5.3} & \textbf{17.9}{\scriptsize$\,\pm$4.5} & \textbf{5.5}{\scriptsize$\,\pm$2.6} & 41.93{\scriptsize$\,\pm$2.6} & 35.56{\scriptsize$\,\pm$2.5} & 37.93{\scriptsize$\,\pm$2.5} & 83.8{\scriptsize$\,\pm$2.9} & 43.0{\scriptsize$\,\pm$4.5} \\
\ours{}+BM25 (448) & 21.4{\scriptsize$\,\pm$4.7} & 8.3{\scriptsize$\,\pm$3.1} & 0.0{\scriptsize$\,\pm$0.0} & 41.17{\scriptsize$\,\pm$2.5} & 36.20{\scriptsize$\,\pm$2.4} & 38.69{\scriptsize$\,\pm$2.5} & \textbf{87.5}{\scriptsize$\,\pm$2.5} & 43.2{\scriptsize$\,\pm$4.5} \\
\bottomrule
\end{tabular}}
\end{table*}

The memory carries over to evidence without shared words and to long chat histories (Table~\ref{tab:real}).
Among the methods that read $B$ tokens, \ours{} significantly beats all others on NoLiMa at 8k, and \ours{}+BM25 finds the most evidence in chat histories.

NoLiMa~\cite{modarressi2025nolima} links each question to its needle only through a latent association, so lexical scores carry no signal there (Appendix~\ref{app:nolima}).
\ours{} significantly beats BM25 at every length ($+24.22$ at 8k) and reranked dense RAG at 8k.
At 8k it is not significantly different from the full cache ($+3.06$~[$-2.33$, $+8.44$]), and at 128k it scores above the truncated cache ($+3.17$~[$+0.39$, $+6.00$]).

On LongBench QA~\cite{bai2024longbench}, six document QA tasks, \ours{} is not significantly different from reranked dense RAG on Llama and Qwen2 and scores above BM25 on Llama ($+2.05$~[$+0.07$, $+4.10$]).
Dense RAG, which brings its own embedding model and reranker, leads on Mistral and on the whole books of InfiniteBench~\cite{zhang2024infinitebench} (Appendices~\ref{app:lb-full} and~\ref{app:infbench}).
\ours{} beats the \Sthree{}-Attention-style index, the earlier query-agnostic key index, on all three models (Table~\ref{tab:holm-main}).

LongMemEval-S~\cite{wu2025longmemeval} gives each question its own history of about 48 user--assistant sessions and measures both the evidence found and the answer.
\ours{}+BM25 places the most evidence turns in its 2{,}048 tokens, significantly more than reranked dense RAG and BM25, and \ours{} alone already finds significantly more than BM25.
The answers of \ours{}+BM25 significantly beat those of BM25 and of the full cache (Appendix~\ref{app:lme}).

\subsection{RQ3: Does it work with existing memories?}\label{sec:exp-stack}

\input{v3_sections/tab_stack}

\ours{} works inside pipelines that already keep a compressed cache or a retriever (Table~\ref{tab:stack}).
For compression on Llama, a method that gives half of its 2{,}048 entries to \ours{} significantly beats the same method with all of them.
On the same episodes the StreamingLLM and SnapKV combinations keep over 90\% of the score that \ours{} alone reaches with all 2{,}048 tokens at every length, and the KVzip combination does so up to 64k (Appendix~\ref{app:limits}).

For retrieval, \ours{} is fused with bge-small chunk scores at a fixed weight of 0.5.
This fusion significantly beats reranked dense RAG at 128k on both models, and there it also scores above \ours{} alone ($+2.51$, $+12.65$).
With BM25 added as a third score, it reaches 75.2 and 73.8 at 128k, the highest 128k score of each model in Tables~\ref{tab:e2} and~\ref{tab:stack}.
On Qwen2.5 this third score adds a significant 5.59 points to \ours{}+BM25 (Appendix~\ref{app:stack}).

\subsection{RQ4: Which design choices carry the signal?}\label{sec:ana-ladder}

\begin{figure*}[t]
\centering
\includegraphics[width=\linewidth]{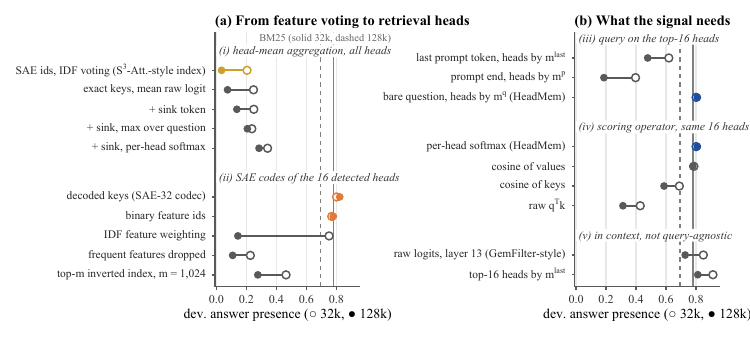}
\caption{\textbf{Ablation ladder} (Llama-3.1-8B, development answer presence, vertical lines mark BM25). $\massp$ ranks heads by the prompt-end tokens. (a) Aggregation over all heads stays near random. On the detected heads, SAE codes keep the signal (orange) until IDF, frequency pruning or an inverted index remove it. (b) What \ours{} (blue) needs. Values are in Table~\ref{tab:ladder}.}
\label{fig:ladder}
\end{figure*}

Head selection carries most of the signal, the per-head softmax comes second, and a few heads with compact keys suffice.
The ablations measure development answer presence on Llama-3.1-8B at 32k/128k (Appendix~\ref{app:dev-full}), and test results are marked as such.

Figure~\ref{fig:ladder} climbs from scoring with head-averaged keys to \ours{} one change at a time.
Feature voting over head-averaged keys, as in the \Sthree{}-Attention index~\cite{ma2026s3attention}, stays near random scores.
A sink token and a per-head softmax over all heads help but remain far below BM25.
The same softmax on the 16 detected heads with exact keys lifts presence above BM25 at both lengths.
The bare question is the best query, and the last prompt token with $\masslast$ heads falls well short.

\label{sec:ana-scoring}%
On the same 16 heads, raw $\vq^{\top}\vk$ and the cosine of values or of keys all stay below the per-head softmax (Figure~\ref{fig:ladder}b, Appendix~\ref{app:sae}).

\label{sec:ana-heads}\label{sec:ana-codec}%
With $\massq$ ranking, presence is flat for up to 64 heads, and the top head (2,\,22) alone stays close to all 16 at 256\,B per token (Figure~\ref{fig:analysis}a).
Heads ranked by $\masslast$ stay far lower for every head count.
Heads re-detected with a new seed or from 32k-token contexts are not significantly different from the default set on test at 32k and stay within 2.2 points of it at 128k (Appendix~\ref{app:robust-heads}).
\pqc{} is the smallest tested code within 0.02 of exact keys at both lengths, and PQ codes give higher presence than SAE codes of equal size (Figure~\ref{fig:analysis}b, c).
On test \pqc{} is significantly below exact keys in only one of nine paired comparisons.
On Llama it is not significantly different from \saec{} codes, which take four times the bytes (Appendix~\ref{app:codec-test}).

\subsection{RQ5: What does it cost?}\label{sec:exp-kv}

\begin{figure*}[t]
\centering
\includegraphics[width=\linewidth]{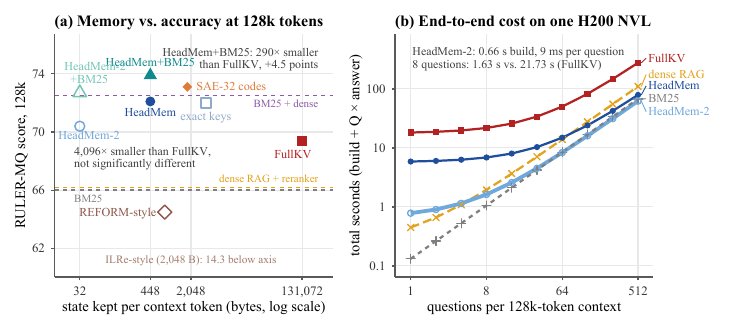}
\caption{\textbf{Memory and cost at 128k tokens} (Llama-3.1-8B). (a) State per context token vs.\ \bench{} score. Dashed lines mark text retrievers, whose index lives outside the LLM. (b) Total cost of $Q$ questions. Each method builds once (FullKV prefills) and then retrieves and answers per question (idle H200 NVL, Table~\ref{tab:e2e}).}
\label{fig:cost}
\end{figure*}

\ours{} pays for its memory once and makes every later question cheap (Figure~\ref{fig:cost}b, Table~\ref{tab:e2e}).
On a 128k context the build of \ourstwo{} stops at layer 2 and is $27\times$ faster than the prefill of the full cache, with 0.49 instead of 31.55\,GiB of peak memory.
The 16-head build of \ours{} is $3.1\times$ faster than the prefill.
Retrieval then takes 9\,ms per question for \ourstwo{} and 38\,ms for \ours{}, against 88\,ms for reranked dense RAG.
Eight questions take 1.63\,s with \ourstwo{}, against 1.95\,s for dense RAG and 21.73\,s for FullKV.
At 128k \ourstwo{} is faster than dense RAG from the fifth question on, and \ours{} is faster than FullKV for any number of questions.
At 1M tokens the KV cache no longer fits, while \ourstwo{} builds a 32\,MiB memory in 5.36\,s.

The memory is small enough to keep for every context.
\ours{} stores $290\times$ less than the KV cache, and \ourstwo{}, which keeps only the layer-2 KV head that Llama's two top heads share, stores $4{,}096\times$ less (Figure~\ref{fig:cost}a).
\ourstwo{} still significantly beats BM25 at 32k and scores above it at 64k and 128k, and at 128k it is not significantly different from the full cache ($+0.99$~[$-1.90$, $+3.83$]).
With BM25 it significantly beats reranked dense RAG at 128k (Appendix~\ref{app:two-head}).

%% file: v3_sections/tab_stack.tex
\begin{table}[t]
\centering
\caption{\textbf{\ours{} combines with existing memories} (\bench{}, macro score). (a) Each compression method keeps 1{,}024 cache entries before the questions, and \ours{} adds the 1{,}024 tokens it selects per question. The left value is the method with 2{,}048 entries, and the last row of (a) is \ours{} alone with all 2{,}048 tokens. Test episodes 0--5 (240 questions per length). At 128k the prompt runs up to 625 positions past the window. (b) \ours{}+BM25 against BM25, and \ours{} fused with bge-small chunk scores against reranked dense RAG (400 questions per length). $^*$ marks a gain with Holm $p<0.05$ (Table~\ref{tab:holm-main} for BM25, Appendix~\ref{app:stack} otherwise).}
\label{tab:stack}
\scriptsize
\setlength{\tabcolsep}{2.6pt}
\renewcommand{\arraystretch}{1}
\resizebox{\ifdim\width>\linewidth\linewidth\else\width\fi}{!}{%
\begin{tabular}{@{}lccc@{}}
\toprule
Memory & 32k & 64k & 128k \\
\midrule
\multicolumn{4}{@{}l}{\textit{(a) KV compression on Llama-3.1-8B, alone $\to$ with \ours{}}} \\
StreamingLLM & 14.1\,$\to$\,71.3$^*$ & 12.1\,$\to$\,68.1$^*$ & 6.9\,$\to$\,66.4$^*$ \\
SnapKV & 12.5\,$\to$\,71.6$^*$ & 13.4\,$\to$\,66.8$^*$ & 10.3\,$\to$\,68.7$^*$ \\
KVzip & 57.1\,$\to$\,77.1$^*$ & 22.3\,$\to$\,69.1$^*$ & 8.5\,$\to$\,23.5$^*$ \\
\textit{\ours{} alone, 2{,}048 tokens} & \textit{74.8} & \textit{72.1} & \textit{72.4} \\
\midrule
\multicolumn{4}{@{}l}{\textit{(b) Text retrievers, $B{=}2{,}048$, alone $\to$ fused with \ours{}}} \\
Llama-3.1-8B, BM25 & 67.7\,$\to$\,76.0$^*$ & 65.3\,$\to$\,72.9$^*$ & 66.0\,$\to$\,73.9$^*$ \\
Llama-3.1-8B, dense RAG & 72.6\,$\to$\,77.1 & 66.7\,$\to$\,74.8$^*$ & 66.2\,$\to$\,74.6$^*$ \\
Qwen2.5-7B, BM25 & 64.0\,$\to$\,72.1$^*$ & 62.9\,$\to$\,69.9$^*$ & 62.4\,$\to$\,68.2$^*$ \\
Qwen2.5-7B, dense RAG & 67.0\,$\to$\,73.0$^*$ & 66.7\,$\to$\,70.8 & 61.8\,$\to$\,73.6$^*$ \\
\bottomrule
\end{tabular}}
\end{table}

%% file: v3_sections/analysis.tex
\section{Mechanistic analysis of the retrieval heads}\label{sec:analysis}

The detected heads match question tokens, the model's own answers rely on them, and they miss evidence that shares no token with the question.

\paragraph{The detected heads match question tokens.}\label{sec:ana-mech}

On Llama and Mistral the detected heads, as a group, behave like induction heads~\cite{olsson2022context}.
On random token sequences shown twice, the detected Llama heads average an induction score of 0.195, against 0.023 for the other heads.
Across all 1,024 heads the induction score correlates with the detection score $\massq$ (Spearman 0.70 on Llama and 0.45 on Mistral).
In \ours{}'s own setting, the detected Llama heads put 8.99\% of a matched question token's attention right after its matches, against 0.38\% for the other heads and 0.05\% by chance, and Mistral's put 3.78\% there, against 0.19\%.
Qwen2.5's detected heads instead attend to the matching token itself (7.01\% against 0.31\%).
Either way, the heads look up the question's tokens in the context (Appendix~\ref{app:mechanism}).

\paragraph{The model relies on the detected heads.}\label{sec:ana-mask}

On 20 \bench{} test episodes at 32k, masking the 16 detected heads lowers Llama's full-cache score from 82.5 to 54.4.
Twenty-three random sets of 16 heads from the same layers have a median of 79.3 (detected minus the mean of the random sets $-21.05$~[$-24.88$, $-17.20$], significant).
Masking the top 64 heads by $\massq$ drops Llama to 7.1, below all ten random 64-head sets (median 48.2, detected minus the mean of the random sets $-44.39$, significant).
On Qwen2.5 masking its own 16 detected heads lowers the score from 71.5 to 44.1, below all 23 random sets (median 64.2, Appendix~\ref{app:mechanism}).

\paragraph{Chained evidence is read well once it is found.}\label{sec:ana-chain}

A lookup of the question's tokens reaches the first statement of a variable-tracking chain, which holds the queried value, but not the later statements.
\ours{}'s selection never holds all five statements of a chain (Appendix~\ref{app:oracle}).
With them forced into its own 2{,}048 tokens, the 128k score rises from 35.8 to 98.8 on Llama and from 23.0 to 92.2 on Qwen2.5, far above the full cache (60.8 and 39.5), although every chain spans several of the chunks that \ours{} encodes separately.
Forcing in HotpotQA's two supporting paragraphs raises the 128k score from 48.8 to 63.8 on Llama and significantly from 37.5 to 60.0 on Qwen2.5.
The model retains the ability to combine evidence across chunks when it is supplied; the gap on these families lies in finding it.

%% file: v3_sections/conclusion.tex
\section{Conclusion}\label{sec:conclusion}

A frozen LLM's retrieval heads, detected once per model, find the evidence even when they read the question alone.
\ours{} turns them into a query-agnostic memory of 448\,B per token on Llama-3.1-8B, or 32\,B with two heads, with no training and no model besides the LLM.
It beats four KV compression methods on \bench{} and combines readily with the memories already in use.
Fused with BM25 or a dense retriever, it beats BM25 and reranked dense RAG at 128k.
On Llama, \ourstwo{} answers eight questions on a 128k context $13\times$ faster than the full cache end to end, at a score not significantly different from it.
The LLM's own retrieval signal supports compact, reusable memory over 1M-token inputs, answered from a fixed reading budget.

%% file: llm_statement.tex
\section*{LLM Usage Statement}

Claude assisted with debugging the experiment, analysis and plotting scripts, and with checking reported numbers against the result logs.
Codex assisted with result verification and editing the discussion of model size and input length.
All experiments ran on the authors' own setup, and the authors take full responsibility for the content.

%% file: v3_sections/appendix.tex

\section{Extended related work}
\label{app:related}

This appendix expands Section~\ref{sec:related}.
Table~\ref{tab:prior-art} compares the methods closest to \ours{} on the properties that define its regime.

\begin{table}[!htbp]
\centering
\footnotesize
\setlength{\tabcolsep}{3pt}
\caption{\textbf{Methods closest to \ours{}.} \emph{Before $Q$} means that the method builds an index or state of the context before the question. A $\circ$ marks a stored state that is question-independent by construction. The paper of that method neither reuses nor evaluates this state across questions. \emph{$Q$ alone} means that the question is encoded without the context. \emph{Head attn.} means that relevance is the attention of detected retrieval heads. \emph{State per context} is what a later question needs kept to avoid a new pass of the LLM over the context. \emph{Frozen} means that no parameter of the LLM is trained.}
\label{tab:prior-art}
\begin{tabular}{@{}l c c c p{6.3cm} c@{}}
\toprule
Method & Before $Q$ & $Q$ alone & Head attn. & State per context & Frozen \\
\midrule
YOURA~\cite{soh2025youra} & \checkmark & $\times$ & $\times$ & full KV cache (all layers and heads) & \checkmark \\
REFORM~\cite{song2025reform} & $\circ$ & $\times$ & $\times$ & normalized states of 4 selected heads and the recurrently compressed KV cache, or a new pass per question & \checkmark \\
ILRe~\cite{ilre2025} & $\circ$ & $\times$ & $\times$ & key cache of layer 3, or a chunked prefill to layer 3 per question & \checkmark \\
GemFilter~\cite{shi2024gemfilter} & $\times$ & $\times$ & $\times$ & none: an early-layer pass over context and question per question & \checkmark \\
QRHead~\cite{zhang2025query} & $\times$ & $\times$ & \checkmark & none: passages and question are read together & \checkmark \\
MRAG~\cite{besta2024mrag} & \checkmark & \checkmark & $\times$ & per-head chunk vectors of a separate embedding model & \checkmark \\
\Sthree{}-Attention~\cite{ma2026s3attention} & \checkmark & \checkmark & $\times$ & inverted index of SAE feature ids of head-averaged keys at four layers & \checkmark \\
Infini-attention~\cite{munkhdalai2024infini} & $\circ$ & $\times$ & $\times$ & compressive memory per head, learned by continual pre-training & $\times$ \\
\midrule
\ours{} & \checkmark & \checkmark & \checkmark & \pqc{} codes of the detected heads' pre-RoPE keys (448\,B per token for Llama-3.1-8B) & \checkmark \\
\bottomrule
\end{tabular}
\end{table}

\paragraph{Attention-based retrievers over a preprocessed context.}
YOURA~\cite{soh2025youra} computes context-only attention once.
It amortizes this preprocessing over the requests that share the context.
Each question is appended to the cached context.
Sentences are scored by the ratio of the question's attention to the context-only attention, averaged over all heads and aggregated over the sentence.
The best sentences are re-read in their original order under a token budget.
Contexts longer than the window are split into window-sized pieces.
Every question therefore passes through all layers over the retained KV cache (128\,KiB per token for Llama-3.1-8B).
REFORM~\cite{song2025reform} encodes the context in a recurrent chunked pass that carries an H2O-compressed KV cache.
It stores normalized query, key or value states (mostly value states) of four heads selected on synthetic retrieval tasks.
It gathers tokens by cosine similarity to the question's states, with max-pooling over a 129-token window.
It keeps the first and last tokens and recomputes the gathered tokens with the full model.
It reaches 1M tokens and fuses its scores with a dense retriever.
REFORM forwards its question on top of the carried compressed cache.
A later question therefore needs that cache or a new pass over the context.
In REFORM's in-context study, query--key products ranked tokens worse than cosine similarity of value states.
ILRe~\cite{ilre2025} prefills the context in chunks up to an early layer (layer 3 for Llama-3.1-8B).
It keeps a sink and a sliding window below that layer and the full key cache at it.
It scores tokens by the maximum softmax attention over heads and question tokens, with multi-kernel pooling.
Section~\ref{sec:exp-attn} compares \ours{} with ports of the REFORM and ILRe scoring and storage into its query-agnostic pipeline (Appendix~\ref{app:protocol-ports}).
The ports share the index build, bare-question forward, selector, budget and prompt of \ours{}.
They are not the original systems, which compute the question with the context.
REFORM also re-reads 8k--16k tokens.

\paragraph{Retrievers that read the question alone.}
MRAG~\cite{besta2024mrag} indexes each chunk offline with a separate, contrastively trained embedding model.
It uses the last-token outputs of every attention head in the last layer of that model.
It embeds the question alone in the same way and retrieves chunks per head space with importance weights.
MRAG shares the decoupled question and the offline index, but it ranks chunk-level vectors of another model by vector similarity.
\ours{} scores every context position with the attention of the model that answers.
\Sthree{}-Attention~\cite{ma2026s3attention} is the closest endogenous index.
It forwards the context in independent chunks.
It encodes head-averaged keys of all heads at four fixed layers into Top-$k$ SAE feature ids~\cite{gao2024scaling} and stores them in an inverted index.
It then discards the KV cache.
It forwards the question alone through the same SAEs and scores positions by IDF-weighted co-activation of question and context features.
It selects spans around smoothed score peaks and takes the union of these spans with BM25 and lead and tail tokens.
\Sthree{}-Attention argued that attention weights cannot serve as the relevance signal directly.
It was evaluated on LongBench with one question per context, within the model's window.
\ours{} keeps the query-agnostic regime, the bare-question forward, the chunk-local build without a KV cache, span selection and BM25 fusion.
It changes five things.
The first is the relevance function.
\ours{} uses each detected head's softmax over all positions, computed from the stored keys.
The second is head selection, which uses the detected heads instead of an average over all heads.
The third is a sink token per chunk and question.
The fourth is the role of the codec.
Codes reconstruct keys and are scored against the dense question queries, instead of feature ids matched by equality.
The fifth is score fusion with a fixed weight instead of a union of index sets.
Section~\ref{sec:ana-ladder} reproduces the \Sthree{}-Attention index and locates its failure in the aggregation over all heads.
On the detected heads, binary feature ids keep the signal.
IDF weighting (at 128k), dropping frequent features and a pruned inverted-index lookup each remove most of it (Appendix~\ref{app:sae}).

\paragraph{In-context selectors and re-rankers.}
QRHead~\cite{zhang2025query} detects a small set of query-focused heads from a few labelled examples.
It re-ranks passages by their attention mass with null-query calibration.
ICR~\cite{chen2025icr} sums attention over all layers and heads and subtracts the attention of a content-free query.
CoRe~\cite{core2025heads} selects fewer than 1\% of the heads by a contrastive score.
GemFilter~\cite{shi2024gemfilter} runs the first layers over context and question.
It keeps the tokens with the largest raw logits of the last prompt token, summed over the heads of one early layer (layer 13 for Llama-3.1-8B).
InfiniRetri~\cite{ye2025infinite} processes chunks with the question appended and carries forward the sentences the question attends to.
AttentionRetriever~\cite{attentionretriever2026} scores the sentences of a long document by head-averaged attention of a small LLM that reads document and query.
It combines this score with dense similarity and entity expansion.
TokenSelect~\cite{wu2024tokenselect} normalizes query--key scores per head with a softmax before the heads vote on the tokens kept at each decoding step.
LongHeads~\cite{lu2024longheads} lets each head attend to its own chunks, scored against attention-pooled chunk representations.
All of them read the question in the context.
\ours{} detects heads with a criterion close to QRHead's, namely the softmax mass that question tokens put on answer tokens.
Unlike QRHead, it computes the keys before any question and reads the question alone.
Appendix~\ref{app:incontext} compares the two settings.

\paragraph{Retrieval heads and head-level KV.}
Retrieval heads are a sparse set of heads that copy evidence out of long contexts.
They are universal across model families and causal for long-context copying~\cite{wu2024retrieval}.
Synthetic context extension improves models through them~\cite{zhao2024understanding}.
DuoAttention~\cite{xiao2024duoattention} learns which heads keep the full cache.
RazorAttention~\cite{tang2025razor} keeps the full cache for echo and induction heads found on repeated random tokens.
HeadKV~\cite{fu2025headkv} allocates per-head budgets by retrieval and reasoning importance.
CompressKV~\cite{compresskv2025} lets heads that attend to the answer span during generation choose which tokens keep their cache at prefill.
These methods keep a cache and decide per head how much of it survives.
\ours{} keeps no values.
It keeps only the compressed keys of the KV heads that host the detected heads (14 of 256 in Llama-3.1-8B).

\paragraph{Trained and training-free memories.}
Infini-attention adds a compressive memory to every attention head and learns a per-head gate that mixes memory and local attention.
This design requires continual pre-training~\cite{munkhdalai2024infini}.
Memorizing Transformers attend over a $k$NN memory of past keys and values~\cite{wu2022memorizing}.
Landmark Attention trains landmark tokens that represent blocks for retrieval~\cite{mohtashami2023landmark}.
Causal retrieval attention is trained end to end for length generalization~\cite{hu2024efficient}.
Without training, Unlimiformer retrieves encoder states per cross-attention head from a $k$NN index~\cite{bertsch2023unlimiformer}.
InfLLM, also training-free, gives a frozen decoder block memories, each represented by a few of its keys~\cite{xiao2024infllm}.
\ours{} trains nothing in the model.
Its only fitted component is the offline key codec.
Detection acts as a hard gate on which heads read the memory.
The frozen model re-reads the returned positions in its prompt.

\paragraph{Relation to Infini-attention.}
Infini-attention learns, by continual pre-training, a per-head gate $A=\sigma(\beta_h)\,A_{\mathrm{mem}}+(1-\sigma(\beta_h))\,A_{\mathrm{dot}}$ between a compressive memory and local attention~\cite{munkhdalai2024infini}.
\ours{} sets this gate by detection, without gradients.
Only the heads in $\Hset$ read the memory, which holds per-token key codes of $|\Gset|\,b_\Codec$ bytes.
These heads read it with their own softmax over all $L$ positions rather than with linear attention.
The frozen model then re-reads the returned positions with all heads, so no values are stored.

\paragraph{KV-cache offloading, compression and eviction.}
Offloading and sparse-attention methods keep the full cache in GPU or host memory.
They select tokens or pages of it per query or decoding step.
This class includes Quest~\cite{tang2024quest}, RetrievalAttention~\cite{liu2024retrievalattention}, ShadowKV~\cite{sun2024shadowkv}, MagicPIG~\cite{chen2025magicpig} and RetroInfer~\cite{chen2025retroinfer}.
Squeezed Attention~\cite{hooper2024squeezed} clusters the keys of a fixed context offline and compares the queries of the appended user input with the centroids.
Loki~\cite{singhania2024loki} scores keys in a low-rank principal-component space at each decoding step.
PQCache product-quantizes the keys of every head and scores them with decode-step queries to fetch tokens from an offloaded cache~\cite{zhang2025pqcache}.
It is the closest use of the product-quantization codec~\cite{jegou2011pq,douze2024faiss} that \ours{} takes as its default.
Compression quantizes the cache~\cite{liu2024kivi,hooper2024kvquant} or codes it with a sparse dictionary~\cite{kim2025lexico}.
Eviction keeps a subset of entries chosen at compression time~\cite{xiao2024efficientstreaminglanguagemodels,zhang2023h2o,li2024snapkv,zefan2024pyramidkv,devoto2024l2}.
KVzip chooses the subset without a question, by how well the kept cache reconstructs the context~\cite{kim2025kvzip}.
SCBench evaluates several requests over one shared context~\cite{li2025scbench}, the setting that \bench{} also uses.
A study of training-free sparse attention finds that selection during decoding tolerates more sparsity than selection during prefilling~\cite{nawrot2025sparse}.
\ours{} stores no values and no keys of other heads.
It keeps every position reachable through a 448\,B per token \pqc{} index for Llama-3.1-8B (the KV cache takes 128\,KiB per token).
It selects once per question rather than at every decoding step.

\paragraph{Text retrieval and long-context RAG.}
Retrieval-augmented generation~\cite{lewis2020retrieval} ranks passages with BM25~\cite{robertson2009bm25}, dense encoders~\cite{karpukhin2020dpr,xiao2024cpack}, multi-vector models~\cite{chen2024m3} or rerankers~\cite{li2023making}, all separate from the model that answers.
OP-RAG~\cite{yu2024oprag} keeps retrieved 128-token chunks in document order.
With 16k--48k retrieved tokens, it reads InfiniteBench books better than the full context does.
On the same book QA tasks, dense retrieval scores highest among the retrieval methods of Appendix~\ref{app:infbench}.
When question and evidence share no literal words, long-context models themselves degrade~\cite{modarressi2025nolima}.
Section~\ref{sec:exp-nolima} evaluates \ours{} in this setting.

\paragraph{Attention sinks, massive activations and sparse codes.}
Independent chunk encoding makes each chunk's first token an attention sink~\cite{gu2025attentionsink,xiao2024efficientstreaminglanguagemodels,yang2025ape}.
Massive activations and fixed outlier channels in queries and pre-RoPE keys are shared across tokens.
They add a large content-independent term to raw query--key logits~\cite{sun2024massive,jin2025massive,hooper2024kvquant,barbero2025rope}.
\ours{} prepends a sink token to every chunk and to the question and drops its key.
It normalizes each head with a softmax over all positions before the heads are summed.
Section~\ref{sec:ana-ladder} measures both steps.
Top-$k$ sparse autoencoders~\cite{gao2024scaling} give sparse codes of activations whose features are often read as interpretable~\cite{cunningham2023sparse}.
In \ours{}, the SAE is one interchangeable key codec.
Its features reproduce the match (Section~\ref{sec:ana-codec} and Appendix~\ref{app:sae}), and residual-stream SAEs~\cite{he2024llamascope} link the layer-2 match to features of the matched token (Appendix~\ref{app:mechanism}).


\section{Full protocol}
\label{app:protocol}

This appendix records what was run.
Every \ours{} setting was fixed on Llama-3.1-8B-Instruct development episodes or on the detection split.
Each setting was then reused unchanged for every model and benchmark.
Test episodes, the LongBench test split, NoLiMa and InfiniteBench informed no choice.

\subsection{Runs}
\label{app:protocol-runs}

All results come from one run of the final code on H200 NVL GPUs (141\,GB), with torch 2.8.0 and transformers 4.56.2.
Result rows are put into one canonical order before any statistic is computed, so no bootstrap depends on the order in which jobs finished.
The head rankings are fixed inputs (Appendix~\ref{app:heads}).
Detection repeated in this run returns the same top-16 heads for all four models and the same two heads for \ourstwo{}.
The \pqc{} codebooks and the SAEs are trained in the same run with the recipes of Appendices~\ref{app:codecs} and~\ref{app:sae}.

\subsection{Multi-question contexts in \bench{}}
\label{app:protocol-mq}

\paragraph{Episodes.}
Five task families are generated with the templates of RULER~\cite{hsieh2024ruler}.
They are multi-key NIAH, multi-value NIAH, variable tracking, SQuAD QA~\cite{rajpurkar2016squad} and HotpotQA QA~\cite{yang2018hotpotqa}.
An episode is one long context with $Q=8$ independent questions.
The context holds the evidence of all eight.
Retrieval methods answer every question from the same index.
KV methods answer every question from the same cache, compressed before the questions.
There is no dialogue history, so the order of the questions does not matter.

\paragraph{Splits.}
Test episodes use seed 0.
At 32k, 64k and 128k tokens, the test split has 10 episodes per family and length for Llama-3.1-8B-Instruct and Qwen2.5-7B-Instruct.
This gives $10\times8=80$ questions per family cell and $80\times5=400$ per length.
Mistral-7B-Instruct-v0.3 and Qwen2-7B-Instruct have 6 episodes per family and length (240 questions per length).
At 256k, 512k and 1M tokens, the test split has 3 episodes per family and length, for Llama-3.1-8B-Instruct only.
This gives 24 questions per cell and 120 per length.
The reading-budget and chunk-size sweeps and the detection-robustness runs use test episodes 1--6 of Llama-3.1-8B-Instruct (240 questions per length).
Their numbers are never mixed with the 10-episode tables.
\ourstwo{} and \Sthree{}-Hybrid run all 10 episodes.
On Qwen2.5-7B-Instruct, every method, including FullKV and the presses, ran all 10 episodes per family and length.
Development episodes use a separate seed (1).
They have 2 episodes per family and length at 32k and 128k, which gives 16 questions per family cell and 80 per length.
They fixed every design choice of \ours{} (Appendix~\ref{app:dev-full}).
The detection split uses seed 11 at 16k tokens.
It has 2 episodes $\times$ 8 questions $\times$ 5 families $=80$ questions (Appendix~\ref{app:heads}).
QA items of the test episodes come from the test partitions.
QA items of the detection and development episodes come from the dev partitions of SQuAD and HotpotQA.
An item is in the dev partition iff the SHA-1 of its key is $0 \bmod 4$.

\paragraph{Metrics.}
The task score is RULER's string-match metric ($\times100$).
It uses all-match for the NIAH families and variable tracking and part-match for the QA families.
Answer presence is 1 if the reference answer string occurs in the tokens selected for that question and 0 otherwise.
The development tables report its mean per family cell.
The test tables report its macro mean over families (``macro presence'').
Presence is undefined for KV methods, which select no tokens.

\subsection{LongBench QA}
\label{app:protocol-lb}

We use six QA tasks of LongBench v1~\cite{bai2024longbench}.
They are NarrativeQA, Qasper, MultiFieldQA-en, HotpotQA, 2WikiMQA and MuSiQue.
We use the official prompts, the official maximum generation lengths (\texttt{dataset2maxlen}) and the official F1 metric ($\times100$).
Inputs longer than the model's window minus 512 tokens are truncated in the middle.
Each task is split into a test part of 80\% and a development part of 20\%.
The six test parts together hold 920 examples for each model (NarrativeQA 160, Qasper 160, MultiFieldQA-en 120, HotpotQA 160, 2WikiMQA 160, MuSiQue 160).
The development part is used to tune the weights of the three text fusions (Appendix~\ref{app:protocol-baselines}).
Every retrieval method reads $B=2{,}048$ context tokens inside the official prompt.
FullKV reads the whole middle-truncated context.

\subsection{NoLiMa}
\label{app:protocol-nolima}

NoLiMa~\cite{modarressi2025nolima} pairs each question with a needle that answers it only through a latent association.
Question and needle have minimal lexical overlap.
We use its needle set of 58 (entry, test, question-type) pairs at five needle depths (0.1, 0.3, 0.5, 0.7 and 0.9 of the haystack).
This gives 290 samples per length at 8k, 32k and 128k tokens, on Llama-3.1-8B-Instruct.
Each pair uses one of the shuffled haystack books (\texttt{rand\_shuffle}, with book index $=$ pair index mod 5).
The 128k haystack concatenates two books.
Prompts are the official ones with the system prompt ``You are a helpful assistant''.
The model generates up to 192 tokens.
A sample scores 100 if the prediction contains the character name of the needle.
Retrieval methods index the haystack with the needle inserted, read the question alone and re-read $B{=}2{,}048$ selected tokens.
Needle presence is 1 if the needle sentence lies inside the selection.
FullKV reads the whole haystack at 8k and 32k.
At 128k, the haystack and prompt exceed the 128k window.
FullKV therefore reads the haystack middle-truncated to the window, which keeps the needle in 0.800 of the samples.
The methods are FullKV, BM25 (same selector), dense RAG with reranker, \ours{} and \ours{}+BM25 (BM25 weight 0.5), and \ourstwo{} with and without BM25.

\subsection{LongMemEval-S}
\label{app:protocol-lme}

LongMemEval-S~\cite{wu2025longmemeval} (cleaned release, 500 questions) gives each question its own chat history.
Each history has about 48 timestamped user--assistant sessions and about 105k Llama-3.1 tokens (99k--107k).
The evidence turns are annotated (\texttt{has\_answer}) and the answer sessions are listed.
The question types are single-session user, assistant and preference questions, multi-session reasoning, knowledge updates and temporal reasoning.
The set contains 30 abstention questions (identifier ending in \texttt{\_abs}).
The history is rendered as text.
Every session gets a header ``\texttt{\#\#\# Session $i$ (date: \dots)}'' followed by its turns as ``\texttt{role: content}''.
Every retrieval method selects $B{=}2{,}048$ history tokens for the question.
BM25, \ours{}, \ourstwo{} and \ours{}+BM25 (Eq.~\eqref{eq:fuse}) select them through the selector of Section~\ref{sec:method-score}.
Dense RAG + reranker selects them by chunk filling (Appendix~\ref{app:protocol-baselines}).
The model reads the selected tokens in LongMemEval's reading prompt (``I will give you several history chats between you and a user\dots'').
The prompt ends with the current date and the question.
FullKV reads the whole history, which fits Llama-3.1-8B's window.
Answers are greedy, with at most 64 new tokens.
The metrics need no LLM judge.
\emph{Evidence-turn recall} is the fraction of a question's evidence turns with at least half of their tokens selected.
It is averaged over the 479 questions with annotated evidence turns.
\emph{Answer containment} is 1 when the normalized gold answer occurs in the prediction, as in RULER's string match.
It is defined for the 470 non-abstention questions.
\ours{} and \ourstwo{} build their index once per history, as for every other benchmark.

\subsection{InfiniteBench}
\label{app:protocol-infbench}

InfiniteBench~\cite{zhang2024infinitebench} En.QA (351 questions, token F1) and En.MC (229 questions, option match) ask questions about 86 whole books.
The books have a median length of 155{,}656 and a maximum length of 744{,}421 Llama-3.1-8B tokens.
We use the official GPT-4-style templates inside the Llama chat template, 40 new tokens and the official metrics.
Retrieval methods (BM25, dense RAG with reranker, \ours{}, \ours{}+BM25) index the whole book and re-read $B{=}2{,}048$ or $B{=}8{,}192$ selected tokens.
For En.MC, their retrieval query is the question together with its options.
FullKV middle-truncates the book to 130{,}560 tokens and reads all of them.

\subsection{Baselines}
\label{app:protocol-baselines}

All retrieval baselines receive the same budget $B=2{,}048$, the same task prompt and the same question.
Unless stated otherwise, they share the selector of Section~\ref{sec:method-score}.

\paragraph{Text retrievers.}
\emph{BM25 (same selector)} runs BM25~\cite{robertson2009bm25} over 128-token chunks.
Each chunk's score is broadcast to its tokens.
The \ours{} selector then selects from these token scores.
BM25 uses $k_1=0.9$, $b=0.4$ and word tokens with Lucene stop words.
\emph{BM25, chunk filling} first takes the same 128 lead and 128 tail tokens.
Then 128-token chunks in BM25 order fill the budget, and the last chunk is truncated at $\min(B,L)$.
\emph{Dense RAG + reranker} runs bge-small-en-v1.5~\cite{xiao2024cpack} over 256-token chunks.
The top 100 chunks are re-scored by bge-reranker-v2-m3~\cite{chen2024m3,li2023making}, and chunk filling proceeds as above.
\emph{Fusions} fuse BM25 with one of three scores.
Score (i) is the bge-small-en-v1.5 question--chunk cosine, broadcast to tokens.
Score (ii) is a token-level lexical density, the idf-weighted match of each context token against the question terms.
Score (iii) is BGE-M3 ColBERT MaxSim~\cite{chen2024m3}, the maximum cosine of each context token over the question's token vectors.
Token-level scores are smoothed with the kernel of the \ours{} selector.
The BM25 score and the other score are min--max normalized per example and mixed with a BM25 weight tuned on development data.
The \ours{} selector then selects from the mixed score.
On LongBench, the BM25 weights are 0.3 (dense), 0.3 (lexical density) and 0.9 (BGE-M3), tuned on the LongBench development split.
On \bench{}, they are 0.4, 0.8 and 0.9, tuned on the \bench{} development split.
Mistral-7B and Qwen2-7B reuse the Llama weights.
NoLiMa and InfiniteBench use no fusion baseline.
Fusion (i) scores 128-token chunks.

\paragraph{\Sthree{}-Attention-style index.}
Our implementation of an index in the style of \Sthree{}-Attention~\cite{ma2026s3attention} uses a Top-$k$ SAE ($k=128$).
It codes the head-averaged pre-RoPE keys of all heads at layers 0, 12, 16 and 29 of Llama-3.1-8B-Instruct (layers 0, 6, 19 and 26 for Qwen2.5-7B).
It stores the feature ids in an inverted index.
It scores positions by IDF-weighted feature matches with the codes of the question, which is forwarded alone.
Chunks and question are forwarded without a sink token.
\emph{\Sthree{}-Attention-style index} uses this score with the \ours{} selector.
Its fusion with BM25 uses a BM25 weight chosen on development data, 0.7 on LongBench and 1.0 on \bench{}.
On \bench{}, the development selection put all weight on BM25.
The fused rows on \bench{} therefore equal BM25 under the same selector (Tables~\ref{tab:e2-llama-full} and~\ref{tab:e2-qwen-full}).
The variants restricted to the three deeper layers or to layer 0 are fused with BM25 at the same weight.
They ran 6 test episodes per family at 32k--128k on Llama and at 32k on Qwen2.5.

\paragraph{\Sthree{}-Hybrid.}
\Sthree{}-Hybrid is the original hybrid design of \Sthree{}-Attention~\cite{ma2026s3attention} (its Eq.~8).
The \Sthree{}-Attention index above selects positions with half the budget, $B/2=1{,}024$ tokens, through the \ours{} selector.
These positions are united with 128-token BM25 chunks in BM25 order until the union holds $\min(B,L)$ tokens.
The union is read in original order.
\Sthree{}-Hybrid uses the same index, question forward, BM25 statistics, prompt and budget as the other rows.
It runs on all 10 Llama \bench{} test episodes per family and length.
Its contrasts form their own Holm family (Appendix~\ref{app:exp-holm}), so the families of Table~\ref{tab:e2} are unchanged.

\paragraph{KV-cache methods.}
We run FullKV, StreamingLLM~\cite{xiao2024efficientstreaminglanguagemodels}, SnapKV~\cite{li2024snapkv}, PyramidKV~\cite{zefan2024pyramidkv} and KVzip~\cite{kim2025kvzip}, all from kvpress 0.5.5~\cite{nvidia2024kvpress}.
In \bench{}, each press compresses the cache of the context once, to a budget of 2{,}048 entries.
This compression happens while the context is prefilled without any question.
Every question is then answered from that compressed cache.
Press settings are the kvpress defaults.
StreamingLLM keeps 4 sink tokens.
SnapKV and PyramidKV use a 64-token observation window and pooling kernel 5 (PyramidKV $\beta=20$).
KVzip uses 4 sink tokens, chunk size 2{,}048 and no layer-wise budget.
FullKV keeps the full cache.
On Llama-3.1-8B-Instruct every KV method ran 10 episodes per family at 32k, 64k and 128k.
On Qwen2.5-7B-Instruct every KV method also ran 10 episodes per family at 32k, 64k and 128k.
On Mistral-7B and Qwen2-7B, the presses ran at 32k, inside the window.
FullKV also ran at 64k and 128k on these two models, past the window with extrapolated positions.
Offloading methods (InfLLM~\cite{xiao2024infllm}, ShadowKV~\cite{sun2024shadowkv}) were not run.

\subsection{REFORM-style and ILRe-style scoring}
\label{app:protocol-ports}

REFORM~\cite{song2025reform} and ILRe~\cite{ilre2025} retrieve with the answering model's own attention states.
Both, however, compute the question together with the context.
REFORM forwards the question as the last chunk on top of a recurrently carried, compressed KV cache and re-reads 8k--16k gathered tokens.
ILRe prefills the context up to layer 3 and scores tokens with the question's attention in that layer.
We port their scoring and storage into \ours{}'s query-agnostic pipeline.
This lets us compare their relevance functions with \ours{}'s in the regime of this paper.
Both ports use \ours{}'s chunk-local index build, with chunks of $B_c=4{,}096$ tokens, a sink token and no state across chunks.
Both ports forward the question alone and pass their token scores to the same selector, budget $B=2{,}048$ and prompt.
Only what is stored and how a position is scored change.
\begin{itemize}\itemsep1pt
\item \textbf{REFORM-style} stores the bf16 value states of the KV heads that host the top-4 detected retrieval heads of Llama-3.1-8B-Instruct.
These heads are (2,22), (2,21), (14,31) and (17,23), and they sit in three KV heads.
The port stores 768\,B per token.
A context position is scored by the cosine similarity of its value state with those of the question tokens.
This similarity is maximized over question tokens and averaged over the three KV heads.
The result is then max-pooled over a 129-token window.
\item \textbf{ILRe-style} stores the bf16 key states of all 8 KV heads of layer 3, at 2{,}048\,B per token.
A context position is scored by the question's softmax attention probability on it.
The score takes the maximum of this probability over the 32 query heads of layer 3 and the question tokens.
On Qwen2.5-7B, both ports keep four KV heads (1{,}024\,B per token).
REFORM-style keeps the values of the KV heads hosting the top-4 detected heads of Qwen2.5-7B.
ILRe-style keeps the keys of the four KV heads of layer 3.
\end{itemize}
\paragraph{Deviations from the original systems.}
The ports are not the original systems.
(i) The question is forwarded alone, not on top of REFORM's compressed context cache or after ILRe's chunked prefill.
(ii) The keys and values come from \ours{}'s chunk-local build with a sink token.
They do not come from REFORM's recurrent pass with an H2O-compressed cache.
They also do not come from ILRe's streaming prefill, which keeps a sink and a sliding window below layer 3 and position-encoded keys at it.
(iii) REFORM-style uses the heads detected for \ours{} instead of the heads REFORM selects on its own synthetic retrieval split.
(iv) Both ports re-read $B=2{,}048$ tokens through \ours{}'s selector, instead of REFORM's 8k--16k gathered tokens and ILRe's multi-kernel pooling.
REFORM's 129-token max-pooling is kept.
Both ports were run on the Llama-3.1-8B-Instruct \bench{} test set with 10 episodes per family and length (Tables~\ref{tab:e2} and~\ref{tab:e2-llama-full}).
They were also run on the development split (Appendix~\ref{app:scoring-dev}).
At 32k, 64k and 128k, \ours{} (448\,B per token) exceeds REFORM-style scoring by 7.40, 9.03 and 7.56 points.
At the same lengths, it exceeds ILRe-style scoring by 50.73, 55.88 and 57.81 points.
All six margins have Holm $p\le 0.0154$.


\section{Configuration}
\label{app:config}

Table~\ref{tab:config} is the single configuration record of every run in the paper.

\begin{table}[!htbp]
\centering
\caption{\textbf{Configuration of every run.} Layers and heads are indexed from 0.}
\label{tab:config}
\footnotesize
\renewcommand{\arraystretch}{1.1}
\begin{tabular}{@{}>{\raggedright\arraybackslash}p{0.17\linewidth}>{\raggedright\arraybackslash}p{0.79\linewidth}@{}}
\toprule
Models & Llama-3.1-8B-Instruct~\cite{dubey2024llama3}: 32 layers, 32 query / 8 KV heads, $d=128$, 128k window, KV cache 128\,KiB/token; main model (\bench{} 32k--1M, LongBench). Qwen2.5-7B-Instruct~\cite{qwen2024qwen25} with YaRN $\times4$~\cite{peng2023yarn}: 28 layers, 28 query / 4 KV heads, 128k window (\bench{} 32k--128k). Mistral-7B-Instruct-v0.3~\cite{jiang2023mistral} (32 layers, 32 query / 8 KV heads) and Qwen2-7B-Instruct~\cite{yang2024qwen2} (28 layers, 28 query / 4 KV heads): 32k windows; \bench{} 32k--128k (FullKV and presses at 32k only) and LongBench. All models frozen, bf16, transformers 4.56.2, SDPA attention, greedy decoding, H200 NVL GPUs (141\,GB). Timing runs use a GPU of their own (Appendix~\ref{app:eff}). \\
\midrule
Sink & BOS (Llama, Mistral); \texttt{<|endoftext|>} (Qwen2, Qwen2.5). Prepended to every chunk and to every question; its key and query rows are dropped. \\
\midrule
Detection & Seed 11, 16k-token contexts, 2 episodes $\times$ 8 questions $\times$ 5 families $=80$ questions; QA items from the dev partitions of SQuAD and HotpotQA. Score $\massq$ of Eq.~\eqref{eq:detect}, macro-averaged over families; $H=16$ heads kept. Alternative rankings $\masslast$ (last prompt token) and $\massp$ (prompt-end tokens) evaluated on development episodes (Section~\ref{sec:ana-heads}). \\
\midrule
Index & $B_c=4{,}096$ context tokens plus the sink per forward ($B_c\in\{1{,}024, 2{,}048, 8{,}192\}$ in the chunk-size sweep); positions restart in every chunk; the forward stops after layer $\lmax$ (Llama: 27; two-head variant: 2). Pre-RoPE keys of the KV heads in $\Gset$ (Llama: 14 of 256) are coded and stored with the context's token ids; no values are kept. \\
\midrule
Two-head variant & The top two detected heads of Llama-3.1-8B-Instruct, (2,22) and (2,21), which share one KV head of layer 2; \pqc{} codes, 32\,B per token; everything else as in the default (Appendix~\ref{app:two-head}). \\
\midrule
Codecs & \emph{Exact}: bf16, 256\,B per head. \emph{int8 / int4 / int2}: per-token absmax symmetric quantization with an fp16 scale, 130 / 66 / 34\,B per head. \emph{PQ-$M$}: faiss \texttt{ProductQuantizer}($d=128$, $M$, 8 bit)~\cite{jegou2011pq,douze2024faiss}, $M$\,B per head; codebooks per KV head trained on 204k WikiText-2 keys~\cite{merity2017pointer} (400 sink-prefixed 511-token windows); $M=32$ by default, $M\in\{16,32,64\}$ on development episodes. \emph{SAE-$k$}: per-KV-head Top-$k$ SAE~\cite{gao2024scaling}, $D=8{,}192$ latents (expansion 64), $k=32$ by default, $k\in\{4,8,16,32\}$ on development episodes; about 2.4M WikiText-2 keys from sink-prefixed 511-token windows; 20k Adam steps, batch 4{,}096, learning rate $10^{-3}$, MSE loss; int16 ids and fp16 values (128\,B per head for $k=32$) or uint8 values with an fp16 per-token scale (98\,B per head). \emph{Query-aware SAE}: same architecture, loss $\mathbb{E}_{\vq}\big[(\vq^{\top}(\vkh-\vk))^2\big]$ with the second moment of the retrieval heads' queries on WikiText-2. Bytes per token follow Eq.~\eqref{eq:bytes}. \\
\midrule
Scoring & Eq.~\eqref{eq:score}: the bare question (no template) after the sink; per-head softmax over all $L$ positions with $1/\sqrt{d}$; mean over question tokens; sum over the 16 heads of $\Hset$. \pqc{} is scored by asymmetric distance computation (Eq.~\eqref{eq:pq}), \saec{} by $W_{\mathrm{dec}}^{\top}\vq$ and a gather (Eq.~\eqref{eq:sae}). Null-query calibration (content-free query ``N/A'') tested on development episodes, not used. \\
\midrule
Selection & Box smoothing with kernel 8. The first and last $B/16$ tokens (128 for $B=2{,}048$) are selected first and charged to the budget; then spans $[c-64,c+64)$ around the highest remaining peak $c$, with suppression of $[c-128,c+128)$; the remaining budget is filled with the highest-scoring tokens; exactly $\min(B,L)$ tokens. The selected tokens are re-read in their original order inside the task prompt, followed by the question. \\
\midrule
Budget & $B=2{,}048$ for every retrieval method on \bench{}, LongBench, NoLiMa and InfiniteBench; on InfiniteBench also $B=8{,}192$ for the retrieval methods; $B\in\{512, 1{,}024, 2{,}048, 4{,}096\}$ in the reading-budget sweep (Appendix~\ref{app:sweeps}). FullKV reads the whole context (middle-truncated to the window minus 512 tokens on LongBench and to 130{,}560 tokens on InfiniteBench). \\
\midrule
Fusion & \ours{}+BM25: Eq.~\eqref{eq:fuse}, BM25 over 128-token chunks broadcast to tokens, weight 0.5, fixed (not tuned). \\
\midrule
Baselines & Appendix~\ref{app:protocol-baselines}; REFORM-style and ILRe-style ports: Appendix~\ref{app:protocol-ports}. \\
Statistics & Appendix~\ref{app:stats}. \\
\bottomrule
\end{tabular}
\end{table}


\section{Algorithm}
\label{app:algorithm}

Algorithm~\ref{alg:headmem} gives the complete procedure of Section~\ref{sec:method}.
Phase~1 runs once per context, before any question exists.
Phase~2 runs once per question and never forwards the context again.

\begin{algorithm}[!htbp]
\caption{Query-agnostic index construction of \ours{}, then per-question scoring and re-reading}
\label{alg:headmem}
\small
\begin{algorithmic}[1]
\Statex \textbf{Input:} frozen LLM $\Phi$; retrieval heads $\Hset$ with KV heads $\Gset$ and deepest layer $\lmax$; codec $\Codec=(\enc_g,\dec_g)$ per KV head; chunk size $B_c$; budget $B$; context $C=(c_1,\dots,c_L)$; questions $Q_1,\dots,Q_m$
\Statex \textit{Phase 1: index (once per context, before any question)}
\State $\mathcal{I}^{\ell,g}\gets[\,]$ for all $(\ell,g)\in\Gset$; \quad $T\gets[\,]$
\For{each chunk $c$ of $B_c$ consecutive tokens of $C$, in order}
  \State $\{\mathbf{K}^{\ell,g}\}\gets\Phi.\textsc{ForwardKeys}([\sink;c]$, layers $0,\dots,\lmax)$ \Comment{pre-RoPE; positions restart}
  \For{$(\ell,g)\in\Gset$}
    \State append $\enc_g(\mathbf{K}^{\ell,g})$, sink row dropped, to $\mathcal{I}^{\ell,g}$ \Comment{codes $\vz_t^{\ell,g}$}
  \EndFor
  \State append the token ids of $c$ to $T$ \Comment{no values, no KV of earlier chunks}
\EndFor
\Statex \textit{Phase 2: per question (the index is reused)}
\For{$i=1,\dots,m$}
  \State $\{\vq_j^{\ell,h}\}\gets\Phi.\textsc{ForwardQueries}([\sink;Q_i]$, layers $0,\dots,\lmax)$ \Comment{bare question; sink row dropped}
  \State $s\gets\mathbf{0}\in\R^{L}$
  \For{$(\ell,h)\in\Hset$}
    \State $s\gets s+\frac{1}{|Q_i|}\sum_{j}\softmax_{t}\big(\vq_j^{\ell,h\top}\dec_{g(h)}(\vz_t^{\ell,g(h)})/\sqrt{d}\big)$ \Comment{Eq.~\eqref{eq:score}}
  \EndFor
  \State $\bar s\gets\textsc{Box}_8(s)$; \ for \ours{}+BM25, $\bar s\gets s_{\mathrm{fuse}}$ \Comment{Eq.~\eqref{eq:fuse}}
  \State $\Sel\gets\Sel_B(\bar s)$ \Comment{lead/tail, spans, fill; $|\Sel|=\min(B,L)$}
  \State $R_i\gets\Phi.\textsc{Generate}\big(\textsc{Prompt}(T[\textsc{Sort}(\Sel)],\,Q_i)\big)$ \Comment{original order}
\EndFor
\State \Return $R_1,\dots,R_m$
\end{algorithmic}
\end{algorithm}

No forward pass reads more than $\max(B_c+1,\ B+\text{prompt})$ tokens, so the context length $L$ is not limited by the model's window.


\section{Key codecs and memory per token}
\label{app:codecs}

\subsection{Encoding and scoring with \saec{}}
\label{app:sae-codec}

\saec{} uses a per-KV-head Top-$k$ sparse autoencoder~\cite{gao2024scaling}.
It is trained with an MSE loss on about 2.4M sink-prefixed WikiText-2 keys~\cite{merity2017pointer} (training details in Appendix~\ref{app:sae}).
It writes each key as $k=32$ of $D=8{,}192$ atoms with ids $f_{t,i}$ and values $v_{t,i}$.
A code takes 128\,B per head with int16 ids and fp16 values, or 98\,B with uint8 values and an fp16 per-token scale.
\par\vskip 0pt plus 6\baselineskip\penalty-250\vskip 0pt plus -6\baselineskip 
\noindent The code, the reconstructed key and the query--key product are
\begin{equation}\label{eq:sae}
\begin{aligned}
\vz_t&=\TopK_k\big(\mathrm{ReLU}(W_{\mathrm{enc}}(\vk_t-\mathbf b_{\mathrm{dec}})+\mathbf b_{\mathrm{enc}})\big),\qquad
\vkh_t=W_{\mathrm{dec}}\vz_t+\mathbf b_{\mathrm{dec}},\\
\vq^{\top}\vkh_t&=\sum\nolimits_{i=1}^{k}v_{t,i}\,\big(W_{\mathrm{dec}}^{\top}\vq\big)_{f_{t,i}}+\vq^{\top}\mathbf b_{\mathrm{dec}}.
\end{aligned}
\end{equation}
The last term is constant in $t$ and cancels in the softmax of Eq.~\eqref{eq:score}.
Scoring therefore needs one product $W_{\mathrm{dec}}^{\top}\vq$ per question token and head and a gather over the stored ids.
$k=32$ is the smallest $k\in\{4,8,16,32\}$ within 0.02 development answer presence of exact keys at both 32k and 128k (Section~\ref{sec:ana-codec} and Table~\ref{tab:codec-full}).
It costs four times the bytes of \pqc{} (1{,}792 against 448\,B per token).
At the size of \pqc{}, 448\,B per token, SAE codes with $k=8$ give 0.746 and 0.606 development answer presence at 32k and 128k.
\pqc{} gives 0.796 and 0.804 at these lengths.

\subsection{Memory per token for every codec}
\label{app:memory-full}

Per context token, the index of \ours{} takes
\begin{equation}\label{eq:bytes}
\text{bytes per token}=|\Gset|\cdot b_{\Codec}
\end{equation}
plus one token id, where $b_\Codec$ is the size of one key code of one KV head (Section~\ref{sec:method-codec}). Table~\ref{tab:memory} gives the resulting memory for Llama-3.1-8B ($|\Gset|=14$). All memory figures are analytic.
The bf16 KV cache of Llama-3.1-8B-Instruct takes $32\times8\times128\times2\times2\,\mathrm{B}=131{,}072$\,B (128\,KiB) per token (layers $\times$ KV heads $\times$ $d$ $\times$ K and V $\times$ bytes).
This gives 16\,GiB at 128k ($2^{17}$) tokens and 128\,GiB at 1M ($2^{20}$).
KV eviction to $B=2{,}048$ entries keeps $2{,}048\times131{,}072$\,B $=256$\,MiB.
Eviction, however, needs the full cache during prefill and cannot recover dropped tokens.
\ours{} stores $|\Gset|\cdot b_{\Codec}$ bytes per token with $|\Gset|=14$.
For exact keys, $b_{\Codec}=256$ (bf16, $128\times2$).
For the scalar quantizers, it is 130 (int8, $128+2$ for an fp16 scale), 66 (int4, $64+2$) or 34 (int2, $32+2$).
For the sparse codes, it is 128 (\saec{}, $32\times2+32\times2$) or 98 (\saec{} with uint8 values, $32\times2+32+2$).
For product quantization, it is 16, 32 or 64 (PQ-16, \pqc{} and PQ-64, one byte per sub-vector).
The index therefore takes 3{,}584, 1{,}820, 924, 476, 1{,}792, 1{,}372, 224, 448 or 896\,B per token.
The two-head variant stores \pqc{} codes of the single KV head that hosts heads (2,22) and (2,21).
This index takes 32\,B per token.
The REFORM-style and ILRe-style ports of Section~\ref{sec:exp-attn} store what their scoring reads (Appendix~\ref{app:protocol-ports}).
REFORM-style stores the bf16 value states of the three KV heads that host the top four detected heads (768\,B per token).
ILRe-style stores the bf16 keys of all eight KV heads of layer 3 (2{,}048\,B per token).
Dense RAG keeps the smallest index, one 384-dimensional fp32 bge-small vector per 256-token chunk (6\,B per token).
It ranks with that encoder and the bge-reranker-v2-m3 cross-encoder, which are held next to the LLM.
\ours{} ranks with the LLM itself.
Multiplying by $2^{17}$ and $2^{20}$ tokens gives the 128k and 1M columns of Table~\ref{tab:memory}.
Dividing $131{,}072$ by the bytes per token gives the ratios.
For \pqc{}, the ratio is $292.6\times$, written $290\times$ in the main text.
Token ids, kept by every retrieval method, are not counted.
The index is held in GPU memory and grows linearly with $L$ (448\,MiB at 1M tokens with \pqc{}).
Appendix~\ref{app:eff} reports the measured peak GPU memory.
With \pqc{}, the index takes 384, 448 and 448\,B per token for Qwen2.5-7B, Mistral-7B and Qwen2-7B.
The 16 retrieval heads of these models sit in 12, 14 and 14 KV heads (Appendix~\ref{app:heads}).

\begin{table}[!htbp]
\centering
\caption{\textbf{Memory per context token to keep every position retrievable} (Llama-3.1-8B, analytic). The last column gives the ratio to the bf16 KV cache. $b_\Codec$ is the number of bytes per KV head and token. \ours{} stores keys of the 14 KV heads that host the 16 retrieval heads. The two-head variant stores one KV head. The ports store the states their scoring reads. Token ids, kept by every retrieval method, are not counted. Eviction needs the full cache during prefill and cannot recover dropped tokens.}
\label{tab:memory}
\footnotesize
\begin{tabular}{@{}lrrrrr@{}}
\toprule
Store & $b_\Codec$ & Bytes/token & 128k tokens & 1M tokens & vs.\ KV \\
\midrule
Full KV cache (32 layers $\times$ 8 KV heads, K and V) & -- & 131{,}072 & 16\,GiB & 128\,GiB & $1\times$ \\
KV eviction, $B=2{,}048$ entries & -- & -- & 256\,MiB & 256\,MiB & -- \\
Dense RAG (bge-small, 384-d fp32 per 256 tokens) & -- & 6 & 0.75\,MiB & 6\,MiB & $21{,}845\times$ \\ 
\midrule
REFORM-style port (values, 3 KV heads, bf16) & 256 & 768 & 96\,MiB & 768\,MiB & $170.7\times$ \\
ILRe-style port (keys, 8 KV heads of layer 3, bf16) & 256 & 2{,}048 & 256\,MiB & 2\,GiB & $64.0\times$ \\
\midrule
\ours{}, exact keys (bf16) & 256 & 3{,}584 & 448\,MiB & 3.5\,GiB & $36.6\times$ \\
\ours{}, int8 & 130 & 1{,}820 & 227.5\,MiB & 1.78\,GiB & $72.0\times$ \\
\ours{}, int4 & 66 & 924 & 115.5\,MiB & 0.90\,GiB & $141.9\times$ \\
\ours{}, int2 & 34 & 476 & 59.5\,MiB & 476\,MiB & $275.4\times$ \\
\ours{}, \saec{} (int16 ids, fp16 values) & 128 & 1{,}792 & 224\,MiB & 1.75\,GiB & $73.1\times$ \\
\ours{}, \saec{} (uint8 values) & 98 & 1{,}372 & 171.5\,MiB & 1.34\,GiB & $95.5\times$ \\
\ours{}, PQ-64 & 64 & 896 & 112\,MiB & 896\,MiB & $146.3\times$ \\
\textbf{\ours{}, \pqc{} (default)} & \textbf{32} & \textbf{448} & \textbf{56\,MiB} & \textbf{448\,MiB} & $\mathbf{292.6\times}$ \\
\ours{}, PQ-16 & 16 & 224 & 28\,MiB & 224\,MiB & $585.1\times$ \\
\ourstwo{}, \pqc{} (1 KV head) & 32 & 32 & 4\,MiB & 32\,MiB & $4{,}096\times$ \\
\bottomrule
\end{tabular}
\end{table}


\section{Detected heads and detection split}
\label{app:heads}

Detection runs once per model on the detection split of Appendix~\ref{app:protocol-mq}.
It computes keys as at index time and takes queries from the bare question, forwarded alone after the sink (Eq.~\eqref{eq:detect}).
For each of the four models, all 80 detection questions have a located answer in the context.
Table~\ref{tab:heads} lists the 16 heads kept per model. 

The Llama heads lie in layers 2--27 and occupy 14 of the model's 256 KV heads.
The top head is layer 2, query head 22. 
Only one of the Llama heads, (14,31), is among the five heads that Hong et al.~\cite{hong2026heads} select for Llama-3.1-8B-Instruct.
That work selects its heads with a QRHead-style attention-mass score on agent-memory tasks.
All five lie in layers 13--14.
QRHead likewise finds that query-focused heads differ from copy-paste retrieval heads~\cite{zhang2025query}.
Qwen2 and Qwen2.5 share 10 of their 16 heads. 

\begin{table}[!htbp]
\centering
\caption{\textbf{Detected retrieval heads} (top 16 by $\massq$ on the detection split), as (layer, query head) in rank order. Distinct KV heads and \pqc{} bytes per token are counted from the lists with the GQA group size of each model. Llama and Mistral have 4 query heads per KV head, and Qwen2 and Qwen2.5 have 7. The 10 heads shared by Qwen2 and Qwen2.5 are (4,3) (4,8) (4,9) (4,11) (5,10) (9,9) (14,23) (19,3) (22,17) (23,4). The five heads selected for Llama-3.1-8B-Instruct by a QRHead-style attention-mass score on agent-memory tasks are (13,6) (14,13) (14,29) (14,31) (13,1)~\cite{hong2026heads}.}
\label{tab:heads}
\footnotesize
\begin{tabular}{@{}l>{\raggedright\arraybackslash}p{0.49\linewidth}ccc@{}}
\toprule
Model & Top-16 heads (rank order) & Layers & KV heads & \pqc{} B/token \\
\midrule
Llama-3.1-8B & (2,22) (2,21) (14,31) (17,23) (25,4) (18,21) (11,4) (27,23) (8,31) (18,20) (24,24) (26,13) (5,10) (8,1) (17,0) (16,9) & 2--27 & 14 & 448 \\ 
Qwen2.5-7B & (4,3) (5,10) (23,4) (9,9) (4,11) (22,17) (4,9) (14,7) (12,24) (4,8) (14,23) (19,3) (12,20) (14,11) (14,8) (11,7) & 4--23 & 12 & 384 \\ 
Mistral-7B & (9,10) (30,17) (3,17) (2,21) (18,14) (20,17) (9,14) (9,12) (11,11) (5,27) (9,4) (12,25) (9,6) (8,27) (9,31) (12,6) & 2--30 & 14 & 448 \\ 
Qwen2-7B & (4,3) (5,10) (9,9) (23,4) (22,17) (4,9) (4,11) (14,23) (16,20) (15,27) (21,5) (4,8) (11,24) (14,0) (7,23) (19,3) & 4--23 & 14 & 448 \\ 
\bottomrule
\end{tabular}
\end{table}

\subsection{Robustness of head detection}
\label{app:robust-heads}

Detection was repeated three times for Llama-3.1-8B-Instruct.
Each run selects 16 heads by $\massq$ (Eq.~\eqref{eq:detect}) and rebuilds the \pqc{} index.
The first run uses a new detection split drawn with seed 12 (16k-token contexts, all five families).
The second run uses a new split drawn with seed 13 from 32k-token contexts.
The third run uses the default seed-11 split restricted to the three synthetic families (multi-key NIAH, multi-value NIAH, variable tracking), i.e.\ without QA questions. 
Each variant shares 11 of its 16 heads with the default set (Table~\ref{tab:robust-heads}). The seed-12 set keeps both layer-2 heads, the 32k set keeps (2,22), and the set detected without QA families keeps neither. 

On \bench{} test episodes 1--6, the seed-12 and 32k variants score above BM25 at both lengths, by 6.96 and 5.37 points at 32k and by 7.52 and 4.15 points at 128k (CIs exclude zero).
At 32k neither is significantly different from the default ($-0.08$ and $-1.67$, CIs include zero).
At 128k the seed-12 variant stays within 1.21 points of the default (CI includes zero), and the 32k variant is 2.17 points below it (CI $[-4.12, -0.02]$). 
Detection without QA families collapses to 27.2 and 19.0 at 32k and 128k.
These scores are 47.58 and 53.44 points below the default. 
On LongBench QA (Llama, 920 examples), the same variant scores 39.72 against 41.93 for the default ($-2.21$, CI $[-4.23, -0.16]$).
It is not significantly different from BM25 ($-0.16$, CI $[-2.55, +2.15]$).
It is below reranked dense RAG ($-3.42$, CI $[-5.78, -1.00]$). 
Detection therefore needs QA-like questions with located answers.
The sample and the context length of the split matter little at 128k.

\begin{table}[!htbp]
\centering
\caption{\textbf{Robustness of head detection} (Llama-3.1-8B-Instruct, \pqc{}, \bench{} test episodes 1--6 per family, 240 questions per length, $B{=}2{,}048$). Score is the macro score ($\times$100). Presence is the macro answer presence. $\Delta$ is the paired difference with 95\% bootstrap CI over the same episodes 1--6 as the scores. The default's margin over BM25 here therefore differs from its 10-episode margin in Table~\ref{tab:e2-deltas-pq-llama}. Head sets are given as (layer, query head) in rank order. ``Shared'' counts the heads in common with the default set.}
\label{tab:robust-heads}
\scriptsize
\setlength{\tabcolsep}{3.5pt}
\begin{tabular}{@{}llcccc@{}}
\toprule
Detection split & $L$ & Score & Presence & $\Delta$ vs.\ default & $\Delta$ vs.\ BM25 \\
\midrule
Default (seed 11, 16k, five families) & 32k & 74.8 & 0.835 & -- & $+7.04$ \ci{+3.54}{+10.44} \\ 
New split, seed 12 (16k, five families) & 32k & 74.8 & 0.822 & $-0.08$ \ci{-2.52}{+2.65} & $+6.96$ \ci{+3.25}{+10.77} \\ 
New split, seed 13 (32k, five families) & 32k & 73.2 & 0.813 & $-1.67$ \ci{-5.29}{+2.19} & $+5.37$ \ci{+0.81}{+9.73} \\ 
Without QA families (seed 11, 16k) & 32k & 27.2 & 0.367 & $-47.58$ \ci{-53.08}{-42.06} & $-40.54$ \ci{-46.42}{-34.17} \\ 
BM25, same selector & 32k & 67.8 & 0.748 & -- & -- \\ 
\midrule
Default (seed 11, 16k, five families) & 128k & 72.4 & 0.825 & -- & $+6.31$ \ci{+3.58}{+9.21} \\ 
New split, seed 12 (16k, five families) & 128k & 73.6 & 0.842 & $+1.21$ \ci{-0.81}{+3.29} & $+7.52$ \ci{+4.52}{+10.67} \\ 
New split, seed 13 (32k, five families) & 128k & 70.3 & 0.806 & $-2.17$ \ci{-4.12}{-0.02} & $+4.15$ \ci{+1.60}{+6.73} \\ 
Without QA families (seed 11, 16k) & 128k & 19.0 & 0.314 & $-53.44$ \ci{-58.60}{-48.10} & $-47.12$ \ci{-52.71}{-41.21} \\ 
BM25, same selector & 128k & 66.1 & 0.749 & -- & -- \\ 
\bottomrule
\end{tabular}

\vspace{6pt}
\begin{tabular}{@{}l>{\raggedright\arraybackslash}p{0.60\linewidth}cc@{}}
\toprule
Detection split & Top-16 heads (rank order) & Shared & Layer-2 heads \\
\midrule
Default & (2,22) (2,21) (14,31) (17,23) (25,4) (18,21) (11,4) (27,23) (8,31) (18,20) (24,24) (26,13) (5,10) (8,1) (17,0) (16,9) & 16/16 & both \\ 
Seed 12 & (2,22) (14,31) (18,21) (27,23) (2,21) (17,23) (26,13) (18,20) (8,31) (9,27) (24,24) (11,4) (17,29) (10,18) (27,4) (27,22) & 11/16 & both \\ 
Seed 13, 32k & (14,31) (2,22) (18,21) (27,23) (17,23) (24,24) (9,27) (27,4) (26,13) (18,20) (25,13) (11,4) (25,4) (17,29) (27,22) (16,9) & 11/16 & (2,22) \\ 
Without QA & (17,23) (25,4) (27,23) (14,31) (18,21) (24,24) (26,13) (27,22) (17,0) (18,20) (27,4) (11,4) (16,9) (9,27) (9,19) (24,25) & 11/16 & none \\ 
\bottomrule
\end{tabular}
\end{table}


\section{Statistics}
\label{app:stats}

\paragraph{Units and differences.}
Every comparison is paired.
$\Delta$ is the first-named method minus method X on the same unit.
\ours{} uses \pqc{} codes unless another codec is named.
On \bench{} the unit is an episode, scored as the mean over its 8 questions.
Llama-3.1-8B and Qwen2.5-7B at 32k--128k have 10 episodes per family, 50 per model and length.
Mistral-7B, Qwen2-7B and the experiments on test episodes 1--6 have 6 episodes per family (30 units).
Lengths beyond 128k have 3 episodes per family (15 units).
On LongBench and InfiniteBench the unit is an example.
On NoLiMa it is a sample (needle pair and depth), and on LongMemEval-S it is a question.
A paired difference averages only the units that both methods ran.
A contrast is therefore always computed on the episodes of the table that prints it.
The detection-robustness, budget and chunk-size tables (test episodes 1--6) recompute the default \ours{} on those episodes.
Their margins can differ from the 10-episode margins of Table~\ref{tab:e2-deltas-pq-llama}.

\paragraph{Confidence intervals.}
All intervals are 95\% percentile bootstrap intervals of the mean $\Delta$.
The bootstrap resamples units within each stratum and averages the stratum means.
The \bench{} strata are the task families.
On \bench{}, the main contrasts use 10{,}000 resamples.
The detection-robustness, chunk-size and \Sthree{}-Hybrid contrasts use 4{,}000.
The LongBench strata are its six tasks (4{,}000 resamples).
The NoLiMa strata are the needle entries (4{,}000 resamples).
InfiniteBench is resampled without strata (4{,}000 resamples).
The brackets of Table~\ref{tab:e2} give the same interval for a single score, and the $\pm$ of Table~\ref{tab:real} is half its width.
It uses the unit, strata and resamples of that benchmark's contrasts, except on NoLiMa and LongMemEval-S, whose printed scores are pooled means over samples or questions and are resampled without strata.

\paragraph{Tests.}
All tests are two-sided sign-flip tests on the paired differences.
Each test uses as many random sign assignments as the bootstrap uses resamples.
The $p$-values are Holm-adjusted~\cite{holm1979simple} within each family.
A family is the set of all contrasts of one method of ours against every other method run on the same units.
The methods of ours are \ours{}, \ours{}+BM25, \ours{} with \saec{} codes, \ourstwo{} and \ourstwo{}+BM25.
On \bench{} a family covers one model and length.
On LongBench QA it covers one model, on NoLiMa one length, on InfiniteBench one task and budget, and on LongMemEval-S one metric.
The contrasts with \Sthree{}-Hybrid form one family per model (four variants at three lengths).
Table~\ref{tab:holm-families} lists every family size.
It is generated from the same records as every table of differences.
The rows X of those tables are the members of the family.
A \ourstwo{} family holds the members of the corresponding \ours{} family and that \ours{} variant itself.
With 10{,}000 sign assignments the smallest raw $p$ is $1/10{,}001$.
The smallest attainable Holm-adjusted $p$ in a family of 19 is therefore about 0.0019.
The paper calls a difference significant only when its Holm-adjusted $p$ is below 0.05.

\begin{table}[!htbp]
\centering
\caption{\textbf{Holm families.} Each cell gives the number of contrasts per family.
Each column names the method of ours that the family compares with every other method.
-- marks a setting without a family.
The members are the rows X of Tables~\ref{tab:e2-deltas-pq-llama}, \ref{tab:e2-deltas-pq-qwen} and~\ref{tab:e2-deltas} (\bench{}, 32k--128k), \ref{tab:beyond-full}, \ref{tab:e2-32kmodels}, \ref{tab:lb-deltas-pq} and~\ref{tab:lb-deltas} (LongBench), \ref{tab:nolima-full}, \ref{tab:infbench} and~\ref{tab:lme-delta}.
The \Sthree{}-Hybrid members are the rows of Table~\ref{tab:holm-main}.}
\label{tab:holm-families}
\scriptsize
\setlength{\tabcolsep}{3pt}
\resizebox{\ifdim\width>\linewidth\linewidth\else\width\fi}{!}{%
\begin{tabular}{@{}lllccccc@{}}
\toprule
Benchmark & Model & Setting & \ours{} & \ours{}+BM25 & \ours{} (\saec{}) & \ourstwo{} & \ourstwo{}+BM25 \\
\midrule
\bench{} & Llama-3.1-8B & 32k, 64k, 128k & 19 & 19 & 19 & 20 & 20 \\
\bench{} & Llama-3.1-8B & 256k, 512k, 1M & 8 & 8 & 8 & -- & -- \\
\bench{} & Qwen2.5-7B & 32k, 64k, 128k & 19 & 19 & 19 & 20 & 20 \\
\bench{} & Mistral-7B, Qwen2-7B & 32k & 12 & 12 & -- & -- & -- \\
\bench{} & Mistral-7B, Qwen2-7B & 64k, 128k & 8 & 8 & -- & -- & -- \\
LongBench QA & Llama-3.1-8B & six tasks & 13 & 13 & 13 & 14 & 14 \\
LongBench QA & Mistral-7B, Qwen2-7B & six tasks & 12 & 12 & 12 & -- & -- \\
NoLiMa & Llama-3.1-8B & 8k, 32k, 128k & 4 & -- & -- & -- & -- \\
InfiniteBench & Llama-3.1-8B & each task and $B$ & 4 & 4 & -- & -- & -- \\
LongMemEval-S & Llama-3.1-8B & evidence-turn recall & 3 & 3 & -- & 4 & -- \\
LongMemEval-S & Llama-3.1-8B & answer contained & 4 & 4 & -- & 5 & -- \\
\bench{}, vs.\ \Sthree{}-Hybrid & Llama-3.1-8B & 32k--128k jointly & \multicolumn{5}{c}{12 (all four variants $\times$ three lengths, one family)} \\
\bench{}, vs.\ \Sthree{}-Hybrid & Qwen2.5-7B & 32k--128k jointly & \multicolumn{5}{c}{12 (all four variants $\times$ three lengths, one family)} \\
\bottomrule
\end{tabular}}
\end{table}

\paragraph{Why the intervals are the primary report.}
With few units per family cell, the sign-flip test has little resolution.
Holm also multiplies the smallest raw $p$-value of a family by the family size.
The adjusted $p$-values are therefore conservative.
A CI that excludes zero often comes with a large adjusted $p$.
At 256k, for example, \ours{}+BM25 scores above dense RAG by $+10.83$ (95\% CI [$+6.67$, $+14.79$]) with Holm $p=0.0894$, from 3 episodes per family. 
The text therefore uses three terms.
A difference is \emph{significant} when its Holm $p$ is below 0.05.
A Holm $p$ of exactly 0.0500 is not called significant.
A method \emph{scores above} (or \emph{below}) another when the 95\% CI of their difference excludes zero and the Holm $p$ is at least 0.05.
Two methods are \emph{not significantly different} when the CI includes zero.
This last term is not a claim of equivalence.
No equivalence margin is set.
The tables therefore print every difference with its CI.
The CI bounds the difference that remains possible.


\section{Full \bench{} results}
\label{app:e2-full}

Tables~\ref{tab:e2-llama-full} and~\ref{tab:e2-qwen-full} give every family cell behind Table~\ref{tab:e2}, with the number of test episodes per family (Ep.).
Tables~\ref{tab:e2-deltas-pq-llama} and~\ref{tab:e2-deltas-pq-qwen} give every paired difference of \ours{} and \ours{}+BM25 (\pqc{} codes).
Table~\ref{tab:e2-deltas} gives those of \ours{} with \saec{} codes.

On Llama, variable tracking separates FullKV from every method that reads $B$ tokens.
FullKV scores 96.0, 87.8 and 60.8 at 32k, 64k and 128k, and \ours{} scores 38.5, 36.5 and 35.8.
In multi-value NIAH, \ours{} is above FullKV at every length (94.1, 94.1 and 95.9 against 76.6, 75.6 and 68.8).
The REFORM-style port is within 2.6 points of \ours{} on multi-key NIAH (98.8, 97.5 and 100.0 against 96.2, 96.2 and 98.8).
It loses in multi-value NIAH (66.6, 68.4 and 74.4 against 94.1, 94.1 and 95.9).
It also loses in variable tracking (22.8, 22.0 and 22.0 against 38.5, 36.5 and 35.8).
The ILRe-style port fails on both NIAH families (at most 16.2).
The \Sthree{}-Attention-style fusion with BM25 selected BM25 weight 1.0 on the \bench{} development split.
Its rows therefore equal BM25 (same selector).
The deeper-layer and layer-0 fusions ran 6 episodes per family.
On Qwen2.5, \ours{}+BM25 gains over \ours{} mostly in multi-value NIAH and the QA families.
At 128k it scores 85.6, 51.2 and 81.2 in these families against 80.3, 37.5 and 68.8.

\begin{table}[!htbp]
\centering
\caption{\textbf{\bench{}, Llama-3.1-8B-Instruct, every family cell.} Columns give the score per family ($\times100$), the macro score and the macro answer presence of the $B{=}2{,}048$ selection.
The last column (Ep.) gives the number of test episodes per family, each with 8 questions.
In the Presence column, -- marks KV methods, which select no tokens.}
\label{tab:e2-llama-full}
\scriptsize
\def\arraystretch{.86}
\resizebox{\ifdim\width>\linewidth\linewidth\else\width\fi}{!}{%
\begin{tabular}{@{}lcccccccc@{}}
\toprule
Method & NIAH-MK & NIAH-MV & QA-Hotpot & QA-SQuAD & VT & Macro & Presence & Ep. \\
\midrule
\multicolumn{9}{@{}l}{\textit{32k tokens}} \\
FullKV & 98.8 & 76.6 & 60.0 & 87.5 & 96.0 & 83.8 & -- & 10 \\ 
StreamingLLM & 7.5 & 5.6 & 31.2 & 16.2 & 9.8 & 14.1 & -- & 10 \\ 
SnapKV & 3.8 & 5.6 & 26.2 & 11.2 & 15.8 & 12.5 & -- & 10 \\ 
PyramidKV & 3.8 & 3.8 & 23.8 & 15.0 & 16.0 & 12.4 & -- & 10 \\ 
KVzip & 80.0 & 36.9 & 33.8 & 48.8 & 86.2 & 57.1 & -- & 10 \\ 
BM25, same selector & 88.8 & 89.7 & 56.2 & 78.8 & 25.0 & 67.7 & 0.748 & 10 \\ 
BM25, chunk filling & 91.2 & 90.6 & 53.8 & 82.5 & 25.0 & 68.6 & 0.761 & 10 \\ 
Dense RAG + reranker & 91.2 & 87.2 & 72.5 & 85.0 & 27.0 & 72.6 & 0.823 & 10 \\ 
BM25 + dense & 91.2 & 88.1 & 61.2 & 77.5 & 43.2 & 72.3 & 0.848 & 10 \\ 
BM25 + lexical density & 96.2 & 94.4 & 61.2 & 87.5 & 25.8 & 73.0 & 0.822 & 10 \\ 
BM25 + BGE-M3 & 95.0 & 94.1 & 56.2 & 73.8 & 39.8 & 71.8 & 0.822 & 10 \\ 
\Sthree{}-Attention-style index & 6.2 & 5.3 & 15.0 & 10.0 & 18.2 & 11.0 & 0.151 & 10 \\ 
\Sthree{}-Attention-style index + BM25 & 88.8 & 89.7 & 56.2 & 78.8 & 25.0 & 67.7 & 0.748 & 10 \\ 
\Sthree{}-Attention-style, deeper layers, + BM25 & 87.5 & 88.5 & 56.2 & 81.2 & 25.4 & 67.8 & 0.748 & 6 \\ 
\Sthree{}-Attention-style, layer 0, + BM25 & 87.5 & 88.5 & 56.2 & 81.2 & 25.4 & 67.8 & 0.748 & 6 \\ 
\Sthree{}-Hybrid & 93.8 & 89.1 & 51.2 & 77.5 & 24.5 & 67.2 & 0.736 & 10 \\ 
ILRe-style & 16.2 & 9.7 & 27.5 & 26.2 & 36.8 & 23.3 & 0.265 & 10 \\ 
REFORM-style & 98.8 & 66.6 & 57.5 & 87.5 & 22.8 & 66.6 & 0.751 & 10 \\ 
\ours{}, exact keys & 96.2 & 94.4 & 57.5 & 86.2 & 41.5 & 75.2 & 0.842 & 10 \\ 
\ours{}, \saec{} & 96.2 & 94.4 & 53.8 & 87.5 & 42.0 & 74.8 & 0.840 & 10 \\ 
\ours{}, \pqc{} & 96.2 & 94.1 & 55.0 & 86.2 & 38.5 & 74.0 & 0.838 & 10 \\ 
\ourstwo{}, \pqc{} & 97.5 & 94.4 & 60.0 & 86.2 & 36.0 & 74.8 & 0.843 & 10 \\ 
\ours{}+BM25, \saec{} & 97.5 & 95.9 & 57.5 & 88.8 & 42.0 & 76.3 & 0.855 & 10 \\ 
\ours{}+BM25, \pqc{} & 96.2 & 96.6 & 60.0 & 88.8 & 38.5 & 76.0 & 0.854 & 10 \\ 
\ourstwo{}+BM25, \pqc{} & 96.2 & 95.9 & 65.0 & 90.0 & 36.2 & 76.7 & 0.854 & 10 \\ 
\midrule
\multicolumn{9}{@{}l}{\textit{64k tokens}} \\
FullKV & 98.8 & 75.6 & 58.8 & 81.2 & 87.8 & 80.4 & -- & 10 \\ 
StreamingLLM & 3.8 & 4.7 & 21.2 & 23.8 & 7.2 & 12.1 & -- & 10 \\ 
SnapKV & 13.8 & 0.6 & 22.5 & 25.0 & 5.0 & 13.4 & -- & 10 \\ 
PyramidKV & 11.2 & 0.3 & 21.2 & 23.8 & 8.0 & 12.9 & -- & 10 \\ 
KVzip & 21.2 & 5.9 & 22.5 & 21.2 & 40.5 & 22.3 & -- & 10 \\ 
BM25, same selector & 88.8 & 86.6 & 50.0 & 80.0 & 21.0 & 65.3 & 0.756 & 10 \\ 
BM25, chunk filling & 90.0 & 86.9 & 53.8 & 83.8 & 21.0 & 67.1 & 0.772 & 10 \\ 
Dense RAG + reranker & 87.5 & 80.6 & 57.5 & 83.8 & 24.2 & 66.7 & 0.790 & 10 \\ 
BM25 + dense & 87.5 & 86.2 & 52.5 & 78.8 & 42.5 & 69.5 & 0.850 & 10 \\ 
BM25 + lexical density & 95.0 & 94.1 & 52.5 & 83.8 & 22.0 & 69.5 & 0.811 & 10 \\ 
BM25 + BGE-M3 & 91.2 & 92.5 & 50.0 & 73.8 & 36.0 & 68.7 & 0.819 & 10 \\ 
\Sthree{}-Attention-style index & 5.0 & 0.9 & 15.0 & 13.8 & 4.2 & 7.8 & 0.069 & 10 \\ 
\Sthree{}-Attention-style index + BM25 & 88.8 & 86.6 & 50.0 & 80.0 & 21.0 & 65.3 & 0.756 & 10 \\ 
\Sthree{}-Attention-style, deeper layers, + BM25 & 87.5 & 85.9 & 43.8 & 83.3 & 20.4 & 64.2 & 0.759 & 6 \\ 
\Sthree{}-Attention-style, layer 0, + BM25 & 87.5 & 85.9 & 43.8 & 83.3 & 20.4 & 64.2 & 0.759 & 6 \\ 
\Sthree{}-Hybrid & 88.8 & 85.0 & 50.0 & 82.5 & 22.5 & 65.8 & 0.736 & 10 \\ 
ILRe-style & 2.5 & 5.9 & 22.5 & 18.8 & 29.0 & 15.7 & 0.215 & 10 \\ 
REFORM-style & 97.5 & 68.4 & 51.2 & 73.8 & 22.0 & 62.6 & 0.718 & 10 \\ 
\ours{}, exact keys & 100.0 & 92.5 & 52.5 & 83.8 & 41.5 & 74.0 & 0.864 & 10 \\ 
\ours{}, \saec{} & 98.8 & 92.8 & 53.8 & 80.0 & 41.5 & 73.4 & 0.862 & 10 \\ 
\ours{}, \pqc{} & 96.2 & 94.1 & 55.0 & 76.2 & 36.5 & 71.6 & 0.844 & 10 \\ 
\ourstwo{}, \pqc{} & 97.5 & 95.0 & 51.2 & 76.2 & 24.2 & 68.8 & 0.775 & 10 \\ 
\ours{}+BM25, \saec{} & 98.8 & 95.0 & 56.2 & 85.0 & 41.5 & 75.3 & 0.881 & 10 \\ 
\ours{}+BM25, \pqc{} & 97.5 & 94.1 & 55.0 & 81.2 & 36.5 & 72.9 & 0.863 & 10 \\ 
\ourstwo{}+BM25, \pqc{} & 96.2 & 94.1 & 58.8 & 81.2 & 24.2 & 70.9 & 0.810 & 10 \\ 
\midrule
\multicolumn{9}{@{}l}{\textit{128k tokens}} \\
FullKV & 97.5 & 68.8 & 43.8 & 76.2 & 60.8 & 69.4 & -- & 10 \\ 
StreamingLLM & 1.2 & 0.6 & 22.5 & 10.0 & 0.0 & 6.9 & -- & 10 \\ 
SnapKV & 1.2 & 0.0 & 25.0 & 18.8 & 6.5 & 10.3 & -- & 10 \\ 
PyramidKV & 0.0 & 0.0 & 26.2 & 18.8 & 3.8 & 9.8 & -- & 10 \\ 
KVzip & 0.0 & 0.0 & 6.2 & 2.5 & 33.8 & 8.5 & -- & 10 \\ 
BM25, same selector & 93.8 & 90.6 & 50.0 & 76.2 & 19.5 & 66.0 & 0.752 & 10 \\ 
BM25, chunk filling & 93.8 & 91.2 & 51.2 & 78.8 & 19.5 & 66.9 & 0.766 & 10 \\ 
Dense RAG + reranker & 82.5 & 78.1 & 57.5 & 88.8 & 24.0 & 66.2 & 0.758 & 10 \\ 
BM25 + dense & 96.2 & 90.0 & 50.0 & 80.0 & 46.2 & 72.5 & 0.857 & 10 \\ 
BM25 + lexical density & 98.8 & 95.3 & 50.0 & 82.5 & 20.8 & 69.5 & 0.796 & 10 \\ 
BM25 + BGE-M3 & 91.2 & 90.6 & 53.8 & 73.8 & 32.5 & 68.4 & 0.794 & 10 \\ 
\Sthree{}-Attention-style index & 1.2 & 1.2 & 16.2 & 7.5 & 3.0 & 5.8 & 0.052 & 10 \\ 
\Sthree{}-Attention-style index + BM25 & 93.8 & 90.6 & 50.0 & 76.2 & 19.5 & 66.0 & 0.752 & 10 \\ 
\Sthree{}-Attention-style, deeper layers, + BM25 & 93.8 & 90.6 & 52.1 & 75.0 & 19.2 & 66.1 & 0.749 & 6 \\ 
\Sthree{}-Attention-style, layer 0, + BM25 & 93.8 & 90.6 & 52.1 & 75.0 & 19.2 & 66.1 & 0.749 & 6 \\ 
\Sthree{}-Hybrid & 92.5 & 89.1 & 47.5 & 81.2 & 20.8 & 66.2 & 0.743 & 10 \\ 
ILRe-style & 5.0 & 3.1 & 17.5 & 22.5 & 23.2 & 14.3 & 0.158 & 10 \\ 
REFORM-style & 100.0 & 74.4 & 50.0 & 76.2 & 22.0 & 64.5 & 0.719 & 10 \\ 
\ours{}, exact keys & 100.0 & 94.4 & 45.0 & 82.5 & 38.0 & 72.0 & 0.827 & 10 \\ 
\ours{}, \saec{} & 98.8 & 94.4 & 53.8 & 82.5 & 36.0 & 73.1 & 0.825 & 10 \\ 
\ours{}, \pqc{} & 98.8 & 95.9 & 48.8 & 81.2 & 35.8 & 72.1 & 0.824 & 10 \\ 
\ourstwo{}, \pqc{} & 100.0 & 95.9 & 48.8 & 83.8 & 23.5 & 70.4 & 0.777 & 10 \\ 
\ours{}+BM25, \saec{} & 100.0 & 95.0 & 56.2 & 83.8 & 35.8 & 74.2 & 0.848 & 10 \\ 
\ours{}+BM25, \pqc{} & 100.0 & 95.0 & 53.8 & 85.0 & 35.8 & 73.9 & 0.857 & 10 \\ 
\ourstwo{}+BM25, \pqc{} & 100.0 & 96.2 & 57.5 & 86.2 & 23.5 & 72.7 & 0.805 & 10 \\ 
\bottomrule
\end{tabular}}
\end{table}

\begin{table}[!htbp]
\centering
\caption{\textbf{\bench{}, Qwen2.5-7B-Instruct (YaRN), every family cell.} Columns are as in Table~\ref{tab:e2-llama-full}.
Every method has 10 test episodes per family and length.}
\label{tab:e2-qwen-full}
\scriptsize
\def\arraystretch{.86}
\setlength{\tabcolsep}{4pt}
\resizebox{\ifdim\width>\linewidth\linewidth\else\width\fi}{!}{%
\begin{tabular}{@{}lcccccccc@{}}
\toprule
Method & NIAH-MK & NIAH-MV & QA-Hotpot & QA-SQuAD & VT & Macro & Presence & Ep. \\
\midrule
\multicolumn{9}{@{}l}{\textit{32k tokens}} \\
FullKV & 97.5 & 69.4 & 53.8 & 72.5 & 66.2 & 71.9 & -- & 10 \\ 
StreamingLLM & 7.5 & 5.6 & 21.2 & 15.0 & 7.8 & 11.4 & -- & 10 \\ 
SnapKV & 0.0 & 4.4 & 20.0 & 5.0 & 6.2 & 7.1 & -- & 10 \\ 
PyramidKV & 0.0 & 2.8 & 18.8 & 3.8 & 3.0 & 5.7 & -- & 10 \\ 
KVzip & 62.5 & 18.4 & 32.5 & 25.0 & 60.0 & 39.7 & -- & 10 \\ 
BM25, same selector & 90.0 & 83.4 & 50.0 & 72.5 & 24.2 & 64.0 & 0.756 & 10 \\ 
BM25, chunk filling & 91.2 & 85.9 & 50.0 & 75.0 & 24.2 & 65.3 & 0.776 & 10 \\ 
Dense RAG + reranker & 91.2 & 87.2 & 55.0 & 76.2 & 25.2 & 67.0 & 0.819 & 10 \\ 
BM25 + dense & 90.0 & 82.5 & 51.2 & 66.2 & 41.5 & 66.3 & 0.848 & 10 \\ 
BM25 + lexical density & 96.2 & 88.4 & 57.5 & 76.2 & 26.0 & 68.9 & 0.824 & 10 \\ 
BM25 + BGE-M3 & 92.5 & 87.5 & 56.2 & 62.5 & 31.0 & 66.0 & 0.796 & 10 \\ 
\Sthree{}-Attention-style index & 2.5 & 6.9 & 22.5 & 7.5 & 18.0 & 11.5 & 0.157 & 10 \\ 
\Sthree{}-Attention-style index + BM25 & 90.0 & 83.4 & 50.0 & 72.5 & 24.2 & 64.0 & 0.756 & 10 \\ 
\Sthree{}-Attention-style, deeper layers, + BM25 & 87.5 & 82.8 & 45.8 & 70.8 & 25.0 & 62.4 & 0.751 & 6 \\ 
\Sthree{}-Attention-style, layer 0, + BM25 & 87.5 & 82.8 & 45.8 & 70.8 & 25.0 & 62.4 & 0.751 & 6 \\ 
\Sthree{}-Hybrid & 91.2 & 87.2 & 46.2 & 75.0 & 22.8 & 64.5 & 0.744 & 10 \\ 
ILRe-style & 78.8 & 23.4 & 30.0 & 16.2 & 21.2 & 33.9 & 0.359 & 10 \\ 
REFORM-style & 100.0 & 71.9 & 57.5 & 77.5 & 25.2 & 66.4 & 0.788 & 10 \\ 
\ours{}, exact keys & 98.8 & 79.7 & 56.2 & 70.0 & 30.0 & 66.9 & 0.797 & 10 \\ 
\ours{}, \saec{} & 100.0 & 83.4 & 56.2 & 66.2 & 30.8 & 67.3 & 0.806 & 10 \\ 
\ours{}, \pqc{} & 100.0 & 78.1 & 52.5 & 72.5 & 29.8 & 66.6 & 0.785 & 10 \\ 
\ourstwo{}, \pqc{} & 91.2 & 39.1 & 21.2 & 6.2 & 35.0 & 38.6 & 0.422 & 10 \\ 
\ours{}+BM25, \saec{} & 100.0 & 89.1 & 55.0 & 81.2 & 30.8 & 71.2 & 0.841 & 10 \\ 
\ours{}+BM25, \pqc{} & 100.0 & 90.9 & 61.2 & 78.8 & 29.8 & 72.1 & 0.836 & 10 \\ 
\ourstwo{}+BM25, \pqc{} & 98.8 & 87.2 & 47.5 & 47.5 & 35.0 & 63.2 & 0.761 & 10 \\ 
\midrule
\multicolumn{9}{@{}l}{\textit{64k tokens}} \\
FullKV & 93.8 & 63.8 & 45.0 & 66.2 & 44.5 & 62.6 & -- & 10 \\ 
StreamingLLM & 3.8 & 4.4 & 20.0 & 18.8 & 6.0 & 10.6 & -- & 10 \\ 
SnapKV & 10.0 & 0.0 & 20.0 & 13.8 & 3.0 & 9.3 & -- & 10 \\ 
PyramidKV & 10.0 & 0.0 & 21.2 & 13.8 & 3.5 & 9.7 & -- & 10 \\ 
KVzip & 15.0 & 4.1 & 18.8 & 20.0 & 45.0 & 20.6 & -- & 10 \\ 
BM25, same selector & 90.0 & 79.4 & 48.8 & 76.2 & 20.2 & 62.9 & 0.732 & 10 \\ 
BM25, chunk filling & 90.0 & 80.9 & 51.2 & 80.0 & 20.2 & 64.5 & 0.748 & 10 \\ 
Dense RAG + reranker & 88.8 & 79.1 & 56.2 & 85.0 & 24.5 & 66.7 & 0.769 & 10 \\ 
BM25 + dense & 88.8 & 79.7 & 43.8 & 80.0 & 41.8 & 66.8 & 0.837 & 10 \\ 
BM25 + lexical density & 100.0 & 82.2 & 51.2 & 81.2 & 22.0 & 67.3 & 0.803 & 10 \\ 
BM25 + BGE-M3 & 90.0 & 86.6 & 45.0 & 68.8 & 30.2 & 64.1 & 0.793 & 10 \\ 
\Sthree{}-Attention-style index & 5.0 & 2.5 & 13.8 & 11.2 & 4.8 & 7.5 & 0.067 & 10 \\ 
\Sthree{}-Attention-style index + BM25 & 90.0 & 79.4 & 48.8 & 76.2 & 20.2 & 62.9 & 0.732 & 10 \\ 
\Sthree{}-Hybrid & 90.0 & 83.4 & 43.8 & 77.5 & 21.0 & 63.1 & 0.717 & 10 \\ 
ILRe-style & 53.8 & 9.1 & 20.0 & 11.2 & 5.0 & 19.8 & 0.201 & 10 \\ 
REFORM-style & 100.0 & 62.2 & 48.8 & 75.0 & 21.0 & 61.4 & 0.727 & 10 \\ 
\ours{}, exact keys & 100.0 & 83.8 & 40.0 & 71.2 & 28.0 & 64.6 & 0.779 & 10 \\ 
\ours{}, \saec{} & 98.8 & 86.2 & 42.5 & 72.5 & 24.5 & 64.9 & 0.761 & 10 \\ 
\ours{}, \pqc{} & 100.0 & 79.7 & 48.8 & 70.0 & 29.2 & 65.5 & 0.751 & 10 \\ 
\ourstwo{}, \pqc{} & 92.5 & 36.9 & 13.8 & 12.5 & 27.8 & 36.7 & 0.408 & 10 \\ 
\ours{}+BM25, \saec{} & 100.0 & 89.4 & 52.5 & 85.0 & 24.5 & 70.3 & 0.812 & 10 \\ 
\ours{}+BM25, \pqc{} & 100.0 & 86.6 & 52.5 & 81.2 & 29.2 & 69.9 & 0.830 & 10 \\ 
\ourstwo{}+BM25, \pqc{} & 100.0 & 84.4 & 40.0 & 65.0 & 27.8 & 63.4 & 0.766 & 10 \\ 
\midrule
\multicolumn{9}{@{}l}{\textit{128k tokens}} \\
FullKV & 92.5 & 51.9 & 42.5 & 52.5 & 39.5 & 55.8 & -- & 10 \\ 
StreamingLLM & 1.2 & 0.6 & 21.2 & 12.5 & 1.2 & 7.4 & -- & 10 \\ 
SnapKV & 0.0 & 0.0 & 17.5 & 8.8 & 7.0 & 6.7 & -- & 10 \\ 
PyramidKV & 0.0 & 0.0 & 18.8 & 8.8 & 2.8 & 6.0 & -- & 10 \\ 
KVzip & 0.0 & 0.3 & 18.8 & 12.5 & 30.5 & 12.4 & -- & 10 \\ 
BM25, same selector & 88.8 & 77.5 & 51.2 & 75.0 & 19.2 & 62.4 & 0.747 & 10 \\ 
BM25, chunk filling & 90.0 & 78.4 & 50.0 & 81.2 & 19.2 & 63.8 & 0.761 & 10 \\ 
Dense RAG + reranker & 76.2 & 76.6 & 52.5 & 81.2 & 22.2 & 61.8 & 0.737 & 10 \\ 
BM25 + dense & 88.8 & 77.8 & 48.8 & 72.5 & 45.2 & 66.6 & 0.854 & 10 \\ 
BM25 + lexical density & 100.0 & 85.0 & 52.5 & 81.2 & 20.5 & 67.8 & 0.796 & 10 \\ 
BM25 + BGE-M3 & 91.2 & 83.8 & 47.5 & 71.2 & 28.8 & 64.5 & 0.788 & 10 \\ 
\Sthree{}-Attention-style index & 1.2 & 0.3 & 20.0 & 7.5 & 3.5 & 6.5 & 0.051 & 10 \\ 
\Sthree{}-Attention-style index + BM25 & 88.8 & 77.5 & 51.2 & 75.0 & 19.2 & 62.4 & 0.747 & 10 \\ 
\Sthree{}-Hybrid & 88.8 & 82.2 & 42.5 & 76.2 & 21.0 & 62.1 & 0.734 & 10 \\ 
ILRe-style & 31.2 & 7.2 & 25.0 & 11.2 & 1.8 & 15.3 & 0.143 & 10 \\ 
REFORM-style & 100.0 & 71.6 & 48.8 & 82.5 & 20.0 & 64.6 & 0.740 & 10 \\ 
\ours{}, exact keys & 100.0 & 83.4 & 43.8 & 73.8 & 27.5 & 65.7 & 0.763 & 10 \\ 
\ours{}, \saec{} & 100.0 & 88.1 & 42.5 & 70.0 & 22.0 & 64.5 & 0.739 & 10 \\ 
\ours{}, \pqc{} & 95.0 & 80.3 & 37.5 & 68.8 & 23.0 & 60.9 & 0.709 & 10 \\ 
\ourstwo{}, \pqc{} & 66.2 & 33.8 & 13.8 & 7.5 & 26.5 & 29.6 & 0.325 & 10 \\ 
\ours{}+BM25, \saec{} & 100.0 & 87.2 & 57.5 & 78.8 & 22.0 & 69.1 & 0.805 & 10 \\ 
\ours{}+BM25, \pqc{} & 100.0 & 85.6 & 51.2 & 81.2 & 23.0 & 68.2 & 0.814 & 10 \\ 
\ourstwo{}+BM25, \pqc{} & 95.0 & 85.6 & 38.8 & 65.0 & 26.5 & 62.2 & 0.758 & 10 \\ 
\bottomrule
\end{tabular}}
\end{table}

\begin{table}[!htbp]
\centering
\caption{\textbf{Paired differences on \bench{}, Llama-3.1-8B-Instruct, \pqc{} codes.} The table gives \ours{} minus X and \ours{}+BM25 minus X as the mean over episodes.
It adds the 95\% bootstrap CI and the Holm-adjusted $p$ within each length (Appendix~\ref{app:stats}).
$^*$ marks $p<0.05$.
Rows \saec{} and exact keys are \ours{} with that key store.
-- marks the method itself.}
\label{tab:e2-deltas-pq-llama}
\scriptsize
\setlength{\tabcolsep}{3pt}
\resizebox{\ifdim\width>\linewidth\linewidth\else\width\fi}{!}{%
\begin{tabular}{@{}lrccrcc@{}}
\toprule
& \multicolumn{3}{c}{\ours{} (\pqc{}) $-$ X} & \multicolumn{3}{c}{\ours{}+BM25 (\pqc{}) $-$ X} \\
\cmidrule(lr){2-4}\cmidrule(l){5-7}
X & $\Delta$ & 95\% CI & Holm $p$ & $\Delta$ & 95\% CI & Holm $p$ \\
\midrule
\multicolumn{7}{@{}l}{\textit{32k tokens}} \\
FullKV & $-9.75$ & \ci{-12.39}{-7.11} & 0.1188 & $-7.75$ & \ci{-10.28}{-5.22} & 0.3390 \\ 
StreamingLLM & $+59.94^*$ & \ci{+56.00}{+63.84} & 0.0019 & $+61.94^*$ & \ci{+58.39}{+65.55} & 0.0019 \\ 
SnapKV & $+61.49^*$ & \ci{+57.35}{+65.64} & 0.0019 & $+63.49^*$ & \ci{+59.24}{+67.58} & 0.0019 \\ 
PyramidKV & $+61.56^*$ & \ci{+57.56}{+65.58} & 0.0019 & $+63.56^*$ & \ci{+59.67}{+67.45} & 0.0019 \\ 
KVzip & $+16.89^*$ & \ci{+12.51}{+21.27} & 0.0410 & $+18.89^*$ & \ci{+14.69}{+23.15} & 0.0200 \\ 
BM25, same selector & $+6.33^*$ & \ci{+3.33}{+9.35} & 0.0056 & $+8.32^*$ & \ci{+5.11}{+11.46} & 0.0019 \\ 
BM25, chunk filling & $+5.39^*$ & \ci{+2.55}{+8.28} & 0.0132 & $+7.39^*$ & \ci{+4.35}{+10.34} & 0.0022 \\ 
Dense RAG + reranker & $+1.43$ & \ci{-1.57}{+4.54} & 1.0000 & $+3.42$ & \ci{+0.19}{+6.70} & 0.5349 \\ 
BM25 + dense & $+1.74$ & \ci{-1.16}{+4.64} & 1.0000 & $+3.74$ & \ci{+0.93}{+6.55} & 0.2051 \\ 
BM25 + lexical density & $+0.99$ & \ci{-1.44}{+3.56} & 1.0000 & $+2.99^*$ & \ci{+1.68}{+4.36} & 0.0486 \\ 
BM25 + BGE-M3 & $+2.25$ & \ci{-0.75}{+5.34} & 1.0000 & $+4.25$ & \ci{+1.02}{+7.58} & 0.2032 \\ 
\Sthree{}-Attention-style index & $+63.05^*$ & \ci{+59.62}{+66.26} & 0.0019 & $+65.05^*$ & \ci{+61.86}{+68.20} & 0.0019 \\ 
\Sthree{}-Attention-style index + BM25 & $+6.33^*$ & \ci{+3.33}{+9.35} & 0.0056 & $+8.32^*$ & \ci{+5.11}{+11.46} & 0.0019 \\ 
ILRe-style & $+50.73^*$ & \ci{+46.91}{+54.39} & 0.0019 & $+52.73^*$ & \ci{+48.70}{+56.85} & 0.0019 \\ 
REFORM-style & $+7.40^*$ & \ci{+4.62}{+9.99} & 0.0154 & $+9.40^*$ & \ci{+6.88}{+11.75} & 0.0019 \\ 
\ours{}, exact keys & $-1.16$ & \ci{-2.79}{+0.47} & 1.0000 & $+0.84$ & \ci{-1.29}{+2.94} & 0.9896 \\ 
\ours{}, \saec{} & $-0.76$ & \ci{-2.38}{+0.85} & 1.0000 & $+1.24$ & \ci{-1.00}{+3.48} & 0.9896 \\ 
\ours{}, \pqc{} & -- &  &  & $+2.00$ & \ci{-0.50}{+4.31} & 0.5491 \\ 
\ours{}+BM25, \saec{} & $-2.33$ & \ci{-4.29}{-0.21} & 0.2512 & $-0.33$ & \ci{-1.85}{+1.20} & 0.9896 \\ 
\ours{}+BM25, \pqc{} & $-2.00$ & \ci{-4.31}{+0.50} & 0.9610 & -- &  &  \\ 
\midrule
\multicolumn{7}{@{}l}{\textit{64k tokens}} \\
FullKV & $-8.81$ & \ci{-11.60}{-6.21} & 0.1071 & $-7.56$ & \ci{-10.35}{-4.85} & 0.1945 \\ 
StreamingLLM & $+59.48^*$ & \ci{+55.95}{+62.94} & 0.0019 & $+60.73^*$ & \ci{+57.33}{+64.12} & 0.0019 \\ 
SnapKV & $+58.24^*$ & \ci{+52.88}{+62.79} & 0.0019 & $+59.49^*$ & \ci{+54.30}{+63.86} & 0.0019 \\ 
PyramidKV & $+58.70^*$ & \ci{+52.84}{+63.36} & 0.0019 & $+59.95^*$ & \ci{+54.19}{+64.51} & 0.0019 \\ 
KVzip & $+49.33^*$ & \ci{+46.04}{+52.62} & 0.0019 & $+50.58^*$ & \ci{+47.48}{+53.71} & 0.0019 \\ 
BM25, same selector & $+6.35^*$ & \ci{+4.10}{+8.60} & 0.0019 & $+7.60^*$ & \ci{+5.24}{+9.85} & 0.0019 \\ 
BM25, chunk filling & $+4.54^*$ & \ci{+2.42}{+6.64} & 0.0405 & $+5.79^*$ & \ci{+3.35}{+8.15} & 0.0020 \\ 
Dense RAG + reranker & $+4.89$ & \ci{+2.08}{+7.71} & 0.1024 & $+6.14^*$ & \ci{+3.67}{+8.61} & 0.0045 \\ 
BM25 + dense & $+2.11$ & \ci{-0.81}{+5.14} & 0.5079 & $+3.36$ & \ci{+0.46}{+6.33} & 0.2208 \\ 
BM25 + lexical density & $+2.15$ & \ci{+0.00}{+4.25} & 0.5079 & $+3.40$ & \ci{+1.45}{+5.39} & 0.0960 \\ 
BM25 + BGE-M3 & $+2.91$ & \ci{-0.16}{+6.00} & 0.4305 & $+4.16$ & \ci{+1.40}{+6.90} & 0.0518 \\ 
\Sthree{}-Attention-style index & $+63.83^*$ & \ci{+60.69}{+67.03} & 0.0019 & $+65.08^*$ & \ci{+62.10}{+68.08} & 0.0019 \\ 
\Sthree{}-Attention-style index + BM25 & $+6.35^*$ & \ci{+4.10}{+8.60} & 0.0019 & $+7.60^*$ & \ci{+5.24}{+9.85} & 0.0019 \\ 
ILRe-style & $+55.88^*$ & \ci{+52.74}{+59.11} & 0.0019 & $+57.12^*$ & \ci{+54.23}{+59.98} & 0.0019 \\ 
REFORM-style & $+9.03^*$ & \ci{+6.22}{+11.75} & 0.0022 & $+10.28^*$ & \ci{+7.90}{+12.61} & 0.0019 \\ 
\ours{}, exact keys & $-2.44$ & \ci{-4.53}{-0.34} & 0.3042 & $-1.19$ & \ci{-2.98}{+0.55} & 0.4982 \\ 
\ours{}, \saec{} & $-1.75$ & \ci{-3.88}{+0.26} & 0.5079 & $-0.50$ & \ci{-2.30}{+1.38} & 0.6464 \\ 
\ours{}, \pqc{} & -- &  &  & $+1.25$ & \ci{-0.06}{+2.62} & 0.3810 \\ 
\ours{}+BM25, \saec{} & $-3.69^*$ & \ci{-5.67}{-1.76} & 0.0150 & $-2.44^*$ & \ci{-3.94}{-0.96} & 0.0280 \\ 
\ours{}+BM25, \pqc{} & $-1.25$ & \ci{-2.62}{+0.06} & 0.5079 & -- &  &  \\ 
\midrule
\multicolumn{7}{@{}l}{\textit{128k tokens}} \\
FullKV & $+2.69$ & \ci{-0.26}{+5.65} & 1.0000 & $+4.50$ & \ci{+1.35}{+7.60} & 0.5861 \\ 
StreamingLLM & $+65.21^*$ & \ci{+62.36}{+67.89} & 0.0019 & $+67.03^*$ & \ci{+64.17}{+69.84} & 0.0019 \\ 
SnapKV & $+61.79^*$ & \ci{+58.02}{+65.49} & 0.0019 & $+63.60^*$ & \ci{+60.06}{+67.17} & 0.0019 \\ 
PyramidKV & $+62.34^*$ & \ci{+59.17}{+65.59} & 0.0019 & $+64.15^*$ & \ci{+60.91}{+67.42} & 0.0019 \\ 
KVzip & $+63.59^*$ & \ci{+59.92}{+66.95} & 0.0019 & $+65.40^*$ & \ci{+62.00}{+68.47} & 0.0019 \\ 
BM25, same selector & $+6.06^*$ & \ci{+3.48}{+8.74} & 0.0039 & $+7.88^*$ & \ci{+5.50}{+10.34} & 0.0019 \\ 
BM25, chunk filling & $+5.19^*$ & \ci{+2.39}{+8.01} & 0.0310 & $+7.00^*$ & \ci{+4.71}{+9.36} & 0.0019 \\ 
Dense RAG + reranker & $+5.91$ & \ci{+2.41}{+9.29} & 0.1611 & $+7.72^*$ & \ci{+4.29}{+11.10} & 0.0105 \\ 
BM25 + dense & $-0.41$ & \ci{-3.19}{+2.30} & 1.0000 & $+1.40$ & \ci{-1.40}{+4.24} & 1.0000 \\ 
BM25 + lexical density & $+2.62$ & \ci{+0.45}{+4.76} & 0.5018 & $+4.44^*$ & \ci{+2.77}{+6.15} & 0.0036 \\ 
BM25 + BGE-M3 & $+3.71$ & \ci{+0.79}{+6.83} & 0.2320 & $+5.53^*$ & \ci{+2.54}{+8.84} & 0.0088 \\ 
\Sthree{}-Attention-style index & $+66.24^*$ & \ci{+62.91}{+69.49} & 0.0019 & $+68.05^*$ & \ci{+65.11}{+70.94} & 0.0019 \\ 
\Sthree{}-Attention-style index + BM25 & $+6.06^*$ & \ci{+3.48}{+8.74} & 0.0039 & $+7.88^*$ & \ci{+5.50}{+10.34} & 0.0019 \\ 
ILRe-style & $+57.81^*$ & \ci{+53.50}{+62.01} & 0.0019 & $+59.62^*$ & \ci{+55.75}{+63.48} & 0.0019 \\ 
REFORM-style & $+7.56^*$ & \ci{+5.00}{+10.09} & 0.0039 & $+9.38^*$ & \ci{+7.17}{+11.64} & 0.0019 \\ 
\ours{}, exact keys & $+0.11$ & \ci{-1.57}{+1.73} & 1.0000 & $+1.93$ & \ci{-0.04}{+3.95} & 0.5861 \\ 
\ours{}, \saec{} & $-0.99$ & \ci{-2.86}{+0.86} & 1.0000 & $+0.82$ & \ci{-1.64}{+3.31} & 1.0000 \\ 
\ours{}, \pqc{} & -- &  &  & $+1.81$ & \ci{+0.06}{+3.75} & 0.5861 \\ 
\ours{}+BM25, \saec{} & $-2.06$ & \ci{-4.83}{+0.44} & 0.8579 & $-0.25$ & \ci{-1.85}{+1.24} & 1.0000 \\ 
\ours{}+BM25, \pqc{} & $-1.81$ & \ci{-3.75}{-0.06} & 0.5861 & -- &  &  \\ 
\bottomrule
\end{tabular}}
\end{table}

\begin{table}[!htbp]
\centering
\caption{\textbf{Paired differences on \bench{}, Qwen2.5-7B-Instruct (YaRN), \pqc{} codes.} Columns as in Table~\ref{tab:e2-deltas-pq-llama}.}
\label{tab:e2-deltas-pq-qwen}
\scriptsize
\setlength{\tabcolsep}{3pt}
\resizebox{\ifdim\width>\linewidth\linewidth\else\width\fi}{!}{%
\begin{tabular}{@{}lrccrcc@{}}
\toprule
& \multicolumn{3}{c}{\ours{} (\pqc{}) $-$ X} & \multicolumn{3}{c}{\ours{}+BM25 (\pqc{}) $-$ X} \\
\cmidrule(lr){2-4}\cmidrule(l){5-7}
X & $\Delta$ & 95\% CI & Holm $p$ & $\Delta$ & 95\% CI & Holm $p$ \\
\midrule
\multicolumn{7}{@{}l}{\textit{32k tokens}} \\
FullKV & $-5.30$ & \ci{-9.19}{-1.50} & 0.9668 & $+0.26$ & \ci{-3.14}{+3.75} & 1.0000 \\ 
StreamingLLM & $+55.15^*$ & \ci{+51.41}{+58.50} & 0.0019 & $+60.71^*$ & \ci{+56.11}{+65.09} & 0.0019 \\ 
SnapKV & $+59.45^*$ & \ci{+56.14}{+62.62} & 0.0019 & $+65.01^*$ & \ci{+60.92}{+69.10} & 0.0019 \\ 
PyramidKV & $+60.91^*$ & \ci{+58.09}{+63.70} & 0.0019 & $+66.47^*$ & \ci{+62.91}{+70.16} & 0.0019 \\ 
KVzip & $+26.89^*$ & \ci{+23.09}{+30.73} & 0.0019 & $+32.45^*$ & \ci{+28.85}{+36.04} & 0.0019 \\ 
BM25, same selector & $+2.54$ & \ci{-1.75}{+7.21} & 1.0000 & $+8.10^*$ & \ci{+4.41}{+11.96} & 0.0019 \\ 
BM25, chunk filling & $+1.29$ & \ci{-1.80}{+4.45} & 1.0000 & $+6.85^*$ & \ci{+3.95}{+9.75} & 0.0019 \\ 
Dense RAG + reranker & $-0.41$ & \ci{-3.23}{+2.41} & 1.0000 & $+5.15^*$ & \ci{+1.91}{+8.18} & 0.0192 \\ 
BM25 + dense & $+0.28$ & \ci{-4.10}{+4.78} & 1.0000 & $+5.84^*$ & \ci{+2.19}{+9.80} & 0.0453 \\ 
BM25 + lexical density & $-2.31$ & \ci{-5.33}{+0.88} & 1.0000 & $+3.25^*$ & \ci{+0.86}{+5.47} & 0.0352 \\ 
BM25 + BGE-M3 & $+0.62$ & \ci{-2.39}{+3.61} & 1.0000 & $+6.19^*$ & \ci{+3.15}{+9.24} & 0.0060 \\ 
\Sthree{}-Attention-style index & $+55.10^*$ & \ci{+52.00}{+58.21} & 0.0019 & $+60.66^*$ & \ci{+57.24}{+64.24} & 0.0019 \\ 
\Sthree{}-Attention-style index + BM25 & $+2.54$ & \ci{-1.75}{+7.21} & 1.0000 & $+8.10^*$ & \ci{+4.41}{+11.96} & 0.0019 \\ 
ILRe-style & $+32.64^*$ & \ci{+29.43}{+35.84} & 0.0019 & $+38.20^*$ & \ci{+34.10}{+42.17} & 0.0019 \\ 
REFORM-style & $+0.15$ & \ci{-2.36}{+2.66} & 1.0000 & $+5.71^*$ & \ci{+2.75}{+8.61} & 0.0252 \\ 
\ours{}, exact keys & $-0.36$ & \ci{-3.44}{+2.84} & 1.0000 & $+5.20^*$ & \ci{+2.00}{+8.32} & 0.0252 \\ 
\ours{}, \saec{} & $-0.76$ & \ci{-3.74}{+2.05} & 1.0000 & $+4.80^*$ & \ci{+2.14}{+7.34} & 0.0203 \\ 
\ours{}, \pqc{} & -- &  &  & $+5.56^*$ & \ci{+2.56}{+8.38} & 0.0135 \\ 
\ours{}+BM25, \saec{} & $-4.64^*$ & \ci{-6.94}{-2.30} & 0.0169 & $+0.93$ & \ci{-1.73}{+3.00} & 1.0000 \\ 
\ours{}+BM25, \pqc{} & $-5.56^*$ & \ci{-8.38}{-2.56} & 0.0180 & -- &  &  \\ 
\midrule
\multicolumn{7}{@{}l}{\textit{64k tokens}} \\
FullKV & $+2.89$ & \ci{-0.69}{+6.50} & 1.0000 & $+7.26^*$ & \ci{+3.96}{+10.50} & 0.0290 \\ 
StreamingLLM & $+54.96^*$ & \ci{+51.81}{+58.17} & 0.0019 & $+59.34^*$ & \ci{+56.27}{+62.35} & 0.0019 \\ 
SnapKV & $+56.19^*$ & \ci{+50.85}{+60.56} & 0.0019 & $+60.56^*$ & \ci{+55.31}{+64.86} & 0.0019 \\ 
PyramidKV & $+55.84^*$ & \ci{+50.54}{+60.29} & 0.0019 & $+60.21^*$ & \ci{+54.77}{+64.85} & 0.0019 \\ 
KVzip & $+44.98^*$ & \ci{+41.60}{+48.30} & 0.0019 & $+49.35^*$ & \ci{+45.96}{+52.58} & 0.0019 \\ 
BM25, same selector & $+2.61$ & \ci{+0.05}{+5.09} & 1.0000 & $+6.99^*$ & \ci{+4.30}{+9.81} & 0.0019 \\ 
BM25, chunk filling & $+1.05$ & \ci{-1.57}{+3.65} & 1.0000 & $+5.42^*$ & \ci{+2.56}{+8.24} & 0.0056 \\ 
Dense RAG + reranker & $-1.18$ & \ci{-3.83}{+1.43} & 1.0000 & $+3.20$ & \ci{+0.84}{+5.54} & 0.1584 \\ 
BM25 + dense & $-1.25$ & \ci{-4.00}{+1.48} & 1.0000 & $+3.12$ & \ci{+0.14}{+6.11} & 0.2438 \\ 
BM25 + lexical density & $-1.80$ & \ci{-4.38}{+1.02} & 1.0000 & $+2.58$ & \ci{+0.35}{+5.00} & 0.1584 \\ 
BM25 + BGE-M3 & $+1.43$ & \ci{-1.86}{+4.60} & 1.0000 & $+5.80^*$ & \ci{+3.40}{+8.20} & 0.0045 \\ 
\Sthree{}-Attention-style index & $+58.09^*$ & \ci{+54.25}{+61.60} & 0.0019 & $+62.46^*$ & \ci{+58.95}{+65.89} & 0.0019 \\ 
\Sthree{}-Attention-style index + BM25 & $+2.61$ & \ci{+0.05}{+5.09} & 1.0000 & $+6.99^*$ & \ci{+4.30}{+9.81} & 0.0019 \\ 
ILRe-style & $+45.73^*$ & \ci{+41.86}{+49.56} & 0.0019 & $+50.10^*$ & \ci{+46.16}{+53.95} & 0.0019 \\ 
REFORM-style & $+4.15$ & \ci{+0.79}{+7.35} & 0.5774 & $+8.53^*$ & \ci{+5.70}{+11.14} & 0.0020 \\ 
\ours{}, exact keys & $+0.94$ & \ci{-2.38}{+4.10} & 1.0000 & $+5.31^*$ & \ci{+3.06}{+7.74} & 0.0019 \\ 
\ours{}, \saec{} & $+0.64$ & \ci{-2.77}{+4.03} & 1.0000 & $+5.01^*$ & \ci{+2.60}{+7.65} & 0.0045 \\ 
\ours{}, \pqc{} & -- &  &  & $+4.38^*$ & \ci{+2.00}{+6.81} & 0.0096 \\ 
\ours{}+BM25, \saec{} & $-4.74$ & \ci{-8.04}{-1.62} & 0.2052 & $-0.36$ & \ci{-2.41}{+1.76} & 0.7717 \\ 
\ours{}+BM25, \pqc{} & $-4.38^*$ & \ci{-6.81}{-2.00} & 0.0208 & -- &  &  \\ 
\midrule
\multicolumn{7}{@{}l}{\textit{128k tokens}} \\
FullKV & $+5.14$ & \ci{+2.01}{+8.19} & 0.4879 & $+12.45^*$ & \ci{+8.70}{+16.01} & 0.0048 \\ 
StreamingLLM & $+53.54^*$ & \ci{+49.91}{+57.52} & 0.0019 & $+60.85^*$ & \ci{+57.99}{+63.85} & 0.0019 \\ 
SnapKV & $+54.26^*$ & \ci{+50.69}{+58.16} & 0.0019 & $+61.58^*$ & \ci{+58.76}{+64.50} & 0.0019 \\ 
PyramidKV & $+54.86^*$ & \ci{+51.60}{+58.46} & 0.0019 & $+62.17^*$ & \ci{+59.39}{+65.14} & 0.0019 \\ 
KVzip & $+48.50^*$ & \ci{+44.79}{+52.39} & 0.0019 & $+55.81^*$ & \ci{+52.75}{+58.96} & 0.0019 \\ 
BM25, same selector & $-1.44$ & \ci{-5.20}{+2.33} & 1.0000 & $+5.88^*$ & \ci{+2.64}{+9.12} & 0.0209 \\ 
BM25, chunk filling & $-2.88$ & \ci{-6.54}{+0.74} & 0.8159 & $+4.44$ & \ci{+1.21}{+7.72} & 0.0864 \\ 
Dense RAG + reranker & $-0.85$ & \ci{-5.65}{+3.92} & 1.0000 & $+6.46$ & \ci{+2.58}{+10.49} & 0.0840 \\ 
BM25 + dense & $-5.70$ & \ci{-9.20}{-2.25} & 0.1548 & $+1.61$ & \ci{-1.29}{+4.64} & 1.0000 \\ 
BM25 + lexical density & $-6.94^*$ & \ci{-10.79}{-3.16} & 0.0143 & $+0.38$ & \ci{-2.24}{+3.11} & 1.0000 \\ 
BM25 + BGE-M3 & $-3.59$ & \ci{-7.49}{+0.39} & 0.5259 & $+3.73$ & \ci{+0.56}{+6.99} & 0.2565 \\ 
\Sthree{}-Attention-style index & $+54.40^*$ & \ci{+51.19}{+57.85} & 0.0019 & $+61.71^*$ & \ci{+58.94}{+64.56} & 0.0019 \\ 
\Sthree{}-Attention-style index + BM25 & $-1.44$ & \ci{-5.20}{+2.33} & 1.0000 & $+5.88^*$ & \ci{+2.64}{+9.12} & 0.0209 \\ 
ILRe-style & $+45.62^*$ & \ci{+40.89}{+50.44} & 0.0019 & $+52.94^*$ & \ci{+48.23}{+57.70} & 0.0019 \\ 
REFORM-style & $-3.65$ & \ci{-6.89}{-0.62} & 0.4879 & $+3.66^*$ & \ci{+1.74}{+5.70} & 0.0432 \\ 
\ours{}, exact keys & $-4.78^*$ & \ci{-7.94}{-1.76} & 0.0410 & $+2.54$ & \ci{-0.07}{+5.17} & 0.4088 \\ 
\ours{}, \saec{} & $-3.61$ & \ci{-6.44}{-0.79} & 0.1720 & $+3.70$ & \ci{+1.62}{+5.74} & 0.0504 \\ 
\ours{}, \pqc{} & -- &  &  & $+7.31^*$ & \ci{+4.38}{+10.25} & 0.0039 \\ 
\ours{}+BM25, \saec{} & $-8.18^*$ & \ci{-11.22}{-5.12} & 0.0019 & $-0.86$ & \ci{-3.70}{+1.91} & 1.0000 \\ 
\ours{}+BM25, \pqc{} & $-7.31^*$ & \ci{-10.25}{-4.38} & 0.0036 & -- &  &  \\ 
\bottomrule
\end{tabular}}
\end{table}

\begin{table}[!htbp]
\centering
\caption{\textbf{Paired differences of \ours{} with \saec{} codes} (\ours{} (\saec{}) minus X) on \bench{}, 32k--128k.
Table~\ref{tab:beyond-full} gives the results beyond the window.
Columns are as in Table~\ref{tab:e2-deltas-pq-llama}.}
\label{tab:e2-deltas}
\scriptsize
\setlength{\tabcolsep}{3pt}
\resizebox{\ifdim\width>\linewidth\linewidth\else\width\fi}{!}{%
\begin{tabular}{@{}lrccrcc@{}}
\toprule
& \multicolumn{3}{c}{Llama-3.1-8B-Instruct} & \multicolumn{3}{c}{Qwen2.5-7B-Instruct (YaRN)} \\
\cmidrule(lr){2-4}\cmidrule(l){5-7}
X & $\Delta$ & 95\% CI & Holm $p$ & $\Delta$ & 95\% CI & Holm $p$ \\
\midrule
\multicolumn{7}{@{}l}{\textit{32k tokens}} \\
FullKV & $-8.99$ & \ci{-11.53}{-6.41} & 0.1638 & $-4.54$ & \ci{-8.38}{-0.78} & 1.0000 \\ 
BM25 & $+7.09^*$ & \ci{+3.70}{+10.61} & 0.0098 & $+3.30$ & \ci{-1.02}{+7.80} & 1.0000 \\ 
BM25, chunk filling & $+6.15^*$ & \ci{+2.88}{+9.55} & 0.0154 & $+2.05$ & \ci{-0.95}{+5.06} & 1.0000 \\ 
Dense RAG & $+2.19$ & \ci{-0.64}{+5.19} & 1.0000 & $+0.35$ & \ci{-2.94}{+3.67} & 1.0000 \\ 
BM25 + dense & $+2.50$ & \ci{-0.28}{+5.35} & 0.9131 & $+1.04$ & \ci{-2.92}{+5.24} & 1.0000 \\ 
BM25 + lex.\ density & $+1.75$ & \ci{-0.64}{+4.28} & 1.0000 & $-1.55$ & \ci{-3.95}{+1.09} & 1.0000 \\ 
BM25 + BGE-M3 & $+3.01$ & \ci{+0.14}{+6.05} & 0.5732 & $+1.39$ & \ci{-1.31}{+4.01} & 1.0000 \\ 
\Sthree{}-Att.-style + BM25 & $+7.09^*$ & \ci{+3.70}{+10.61} & 0.0098 & $+3.30$ & \ci{-1.02}{+7.80} & 1.0000 \\ 
\Sthree{}-Att.-style index & $+63.81^*$ & \ci{+60.46}{+66.94} & 0.0019 & $+55.86^*$ & \ci{+52.31}{+59.73} & 0.0019 \\ 
SnapKV & $+62.25^*$ & \ci{+57.91}{+66.50} & 0.0019 & $+60.21^*$ & \ci{+56.25}{+64.25} & 0.0019 \\ 
PyramidKV & $+62.33^*$ & \ci{+58.33}{+66.33} & 0.0019 & $+61.67^*$ & \ci{+58.41}{+65.33} & 0.0019 \\ 
StreamingLLM & $+60.70^*$ & \ci{+56.70}{+64.64} & 0.0019 & $+55.91^*$ & \ci{+52.21}{+59.76} & 0.0019 \\ 
KVzip & $+17.65^*$ & \ci{+13.35}{+21.85} & 0.0220 & $+27.65^*$ & \ci{+23.85}{+31.41} & 0.0019 \\ 
REFORM-style & $+8.16^*$ & \ci{+5.26}{+11.05} & 0.0120 & $+0.91$ & \ci{-2.50}{+4.45} & 1.0000 \\ 
ILRe-style & $+51.49^*$ & \ci{+47.54}{+55.35} & 0.0019 & $+33.40^*$ & \ci{+29.73}{+37.24} & 0.0019 \\ 
\ours{}, exact keys & $-0.40$ & \ci{-1.60}{+0.78} & 1.0000 & $+0.40$ & \ci{-2.58}{+3.19} & 1.0000 \\ 
\ours{}+BM25, \saec{} & $-1.56$ & \ci{-3.00}{-0.12} & 0.4112 & $-3.88^*$ & \ci{-5.88}{-1.88} & 0.0480 \\ 
\ours{}, \pqc{} & $+0.76$ & \ci{-0.85}{+2.38} & 1.0000 & $+0.76$ & \ci{-2.05}{+3.74} & 1.0000 \\ 
\ours{}+BM25, \pqc{} & $-1.24$ & \ci{-3.48}{+1.00} & 1.0000 & $-4.80^*$ & \ci{-7.34}{-2.14} & 0.0377 \\ 
\midrule
\multicolumn{7}{@{}l}{\textit{64k tokens}} \\
FullKV & $-7.06$ & \ci{-9.84}{-4.42} & 0.2268 & $+2.25$ & \ci{-1.57}{+6.01} & 1.0000 \\ 
BM25 & $+8.10^*$ & \ci{+5.54}{+10.62} & 0.0019 & $+1.98$ & \ci{-1.34}{+5.24} & 1.0000 \\ 
BM25, chunk filling & $+6.29^*$ & \ci{+3.40}{+9.05} & 0.0180 & $+0.41$ & \ci{-3.08}{+3.70} & 1.0000 \\ 
Dense RAG & $+6.64^*$ & \ci{+4.08}{+9.24} & 0.0070 & $-1.81$ & \ci{-5.16}{+1.48} & 1.0000 \\ 
BM25 + dense & $+3.86$ & \ci{+0.61}{+7.16} & 0.2268 & $-1.89$ & \ci{-5.19}{+1.30} & 1.0000 \\ 
BM25 + lex.\ density & $+3.90$ & \ci{+1.52}{+6.28} & 0.1848 & $-2.44$ & \ci{-4.94}{-0.12} & 1.0000 \\ 
BM25 + BGE-M3 & $+4.66$ & \ci{+1.21}{+8.12} & 0.1184 & $+0.79$ & \ci{-1.96}{+3.48} & 1.0000 \\ 
\Sthree{}-Att.-style + BM25 & $+8.10^*$ & \ci{+5.54}{+10.62} & 0.0019 & $+1.98$ & \ci{-1.34}{+5.24} & 1.0000 \\ 
\Sthree{}-Att.-style index & $+65.58^*$ & \ci{+62.25}{+68.75} & 0.0019 & $+57.45^*$ & \ci{+53.60}{+61.26} & 0.0019 \\ 
SnapKV & $+59.99^*$ & \ci{+55.31}{+64.21} & 0.0019 & $+55.55^*$ & \ci{+50.21}{+60.06} & 0.0019 \\ 
PyramidKV & $+60.45^*$ & \ci{+55.12}{+64.89} & 0.0019 & $+55.20^*$ & \ci{+49.67}{+60.12} & 0.0019 \\ 
StreamingLLM & $+61.23^*$ & \ci{+57.73}{+64.65} & 0.0019 & $+54.33^*$ & \ci{+51.08}{+57.60} & 0.0019 \\ 
KVzip & $+51.08^*$ & \ci{+47.75}{+54.39} & 0.0019 & $+44.34^*$ & \ci{+40.64}{+47.88} & 0.0019 \\ 
REFORM-style & $+10.78^*$ & \ci{+8.72}{+12.89} & 0.0019 & $+3.51$ & \ci{+0.65}{+6.21} & 1.0000 \\ 
ILRe-style & $+57.62^*$ & \ci{+53.88}{+61.11} & 0.0019 & $+45.09^*$ & \ci{+40.86}{+49.25} & 0.0019 \\ 
\ours{}, exact keys & $-0.69$ & \ci{-2.89}{+1.10} & 1.0000 & $+0.30$ & \ci{-1.93}{+2.55} & 1.0000 \\ 
\ours{}+BM25, \saec{} & $-1.94$ & \ci{-4.25}{+0.12} & 0.4676 & $-5.38^*$ & \ci{-8.44}{-2.50} & 0.0144 \\ 
\ours{}, \pqc{} & $+1.75$ & \ci{-0.26}{+3.88} & 0.4676 & $-0.64$ & \ci{-4.03}{+2.77} & 1.0000 \\ 
\ours{}+BM25, \pqc{} & $+0.50$ & \ci{-1.38}{+2.30} & 1.0000 & $-5.01^*$ & \ci{-7.65}{-2.60} & 0.0065 \\ 
\midrule
\multicolumn{7}{@{}l}{\textit{128k tokens}} \\
FullKV & $+3.67$ & \ci{+0.46}{+6.99} & 1.0000 & $+8.75$ & \ci{+5.54}{+11.88} & 0.0728 \\ 
BM25 & $+7.05^*$ & \ci{+4.35}{+9.91} & 0.0019 & $+2.17$ & \ci{-1.11}{+5.67} & 1.0000 \\ 
BM25, chunk filling & $+6.17^*$ & \ci{+3.25}{+9.21} & 0.0120 & $+0.74$ & \ci{-2.41}{+3.96} & 1.0000 \\ 
Dense RAG & $+6.90$ & \ci{+3.14}{+10.46} & 0.0504 & $+2.76$ & \ci{-0.65}{+6.14} & 1.0000 \\ 
BM25 + dense & $+0.57$ & \ci{-2.24}{+3.34} & 1.0000 & $-2.09$ & \ci{-5.05}{+1.02} & 1.0000 \\ 
BM25 + lex.\ density & $+3.61$ & \ci{+1.07}{+6.08} & 0.1617 & $-3.33$ & \ci{-6.71}{-0.10} & 0.7991 \\ 
BM25 + BGE-M3 & $+4.70$ & \ci{+0.99}{+8.31} & 0.1304 & $+0.03$ & \ci{-2.89}{+3.14} & 1.0000 \\ 
\Sthree{}-Att.-style + BM25 & $+7.05^*$ & \ci{+4.35}{+9.91} & 0.0019 & $+2.17$ & \ci{-1.11}{+5.67} & 1.0000 \\ 
\Sthree{}-Att.-style index & $+67.22^*$ & \ci{+64.09}{+70.45} & 0.0019 & $+58.01^*$ & \ci{+55.45}{+60.60} & 0.0019 \\ 
SnapKV & $+62.77^*$ & \ci{+59.10}{+66.45} & 0.0019 & $+57.88^*$ & \ci{+55.52}{+60.15} & 0.0019 \\ 
PyramidKV & $+63.33^*$ & \ci{+60.06}{+66.76} & 0.0019 & $+58.48^*$ & \ci{+56.40}{+60.58} & 0.0019 \\ 
StreamingLLM & $+66.20^*$ & \ci{+63.14}{+69.34} & 0.0019 & $+57.15^*$ & \ci{+54.75}{+59.52} & 0.0019 \\ 
KVzip & $+64.58^*$ & \ci{+61.04}{+67.97} & 0.0019 & $+52.11^*$ & \ci{+49.36}{+54.75} & 0.0019 \\ 
REFORM-style & $+8.55^*$ & \ci{+6.01}{+10.97} & 0.0019 & $-0.04$ & \ci{-2.51}{+2.39} & 1.0000 \\ 
ILRe-style & $+58.80^*$ & \ci{+54.45}{+62.94} & 0.0019 & $+49.24^*$ & \ci{+44.85}{+53.74} & 0.0019 \\ 
\ours{}, exact keys & $+1.10$ & \ci{-0.51}{+2.71} & 1.0000 & $-1.16$ & \ci{-3.31}{+0.71} & 1.0000 \\ 
\ours{}+BM25, \saec{} & $-1.07$ & \ci{-3.65}{+1.50} & 1.0000 & $-4.56$ & \ci{-7.44}{-1.81} & 0.0924 \\ 
\ours{}, \pqc{} & $+0.99$ & \ci{-0.86}{+2.86} & 1.0000 & $+3.61$ & \ci{+0.79}{+6.44} & 0.2150 \\ 
\ours{}+BM25, \pqc{} & $-0.82$ & \ci{-3.31}{+1.64} & 1.0000 & $-3.70$ & \ci{-5.74}{-1.62} & 0.0756 \\ 
\bottomrule
\end{tabular}}
\end{table}

\subsection{Mistral-7B and Qwen2-7B beyond their 32k windows}
\label{app:e2-32k}

Mistral-7B-Instruct-v0.3 and Qwen2-7B-Instruct have 32k-token windows.
At indexing, \ours{} reads at most one 4{,}096-token chunk plus the sink.
At answering, it reads $B$ tokens plus the prompt.
Both models therefore answer \bench{} contexts of 64k and 128k tokens without any change.
The presses run only inside the window.
FullKV, run past the window with extrapolated positions, collapses (Mistral 34.8 and 16.1, Qwen2 44.9 and 13.4 at 64k and 128k).
It falls 16.60--53.17 points below \ours{} (Holm-significant).
Table~\ref{tab:e2-32kmodels} gives the macro scores (6 test episodes per family and length, 240 questions per length) and the paired differences.
On Qwen2, \ours{}+BM25 scores above BM25 at 32k and 64k with CIs that exclude zero, significantly at 64k ($+6.62$, Holm $p=0.0021$).
At 128k it is not significantly different from BM25 and dense RAG ($+2.96$ and $+3.73$, CIs include zero), and neither is \ours{} alone from BM25 ($+4.10$, 95\% CI [$-0.04$, $+8.10$]).
Mistral is the weakest model for \ours{}.
Alone, \ours{} significantly beats BM25 at 32k ($+5.75$, Holm $p=0.0360$).
It scores below that baseline at 64k and 128k ($-11.19$ [$-16.06$, $-6.10$] and $-9.04$ [$-13.29$, $-4.90$]), mostly in the QA families (Table~\ref{tab:e2-32kmodels}).
Fused with BM25, it significantly beats BM25 at 32k ($+5.96$ [$+3.25$, $+8.62$]).
At 64k and 128k the fusion is not significantly different from that baseline ($+1.92$ and $+2.79$, with CIs that include zero).

\begin{table}[!htbp]
\centering
\caption{\textbf{\bench{} on Mistral-7B-Instruct-v0.3 and Qwen2-7B-Instruct (32k windows)}, with 6 test episodes per family and length.
The top part gives the macro score ($\times100$), with QA-Hotpot / QA-SQuAD in parentheses for \ours{}.
The bottom part gives paired differences with 95\% CI.
$^*$ marks Holm $p<0.05$.
$^\ddagger$ marks FullKV run past the 32k window (positions extrapolated).
\nr{} marks entries that were not run, because the presses are run inside the window only.}
\label{tab:e2-32kmodels}
\scriptsize
\setlength{\tabcolsep}{3pt}
\resizebox{\ifdim\width>\linewidth\linewidth\else\width\fi}{!}{%
\begin{tabular}{@{}lcccccc@{}}
\toprule
& \multicolumn{3}{c}{Mistral-7B-Instruct-v0.3} & \multicolumn{3}{c}{Qwen2-7B-Instruct} \\
\cmidrule(lr){2-4}\cmidrule(l){5-7}
Method & 32k & 64k & 128k & 32k & 64k & 128k \\
\midrule
FullKV & 68.3 & 34.8$^\ddagger$ & 16.1$^\ddagger$ & 76.0 & 44.9$^\ddagger$ & 13.4$^\ddagger$ \\ 
StreamingLLM & 10.8 & \nr{} & \nr{} & 12.1 & \nr{} & \nr{} \\ 
SnapKV & 7.2 & \nr{} & \nr{} & 8.6 & \nr{} & \nr{} \\ 
PyramidKV & 7.1 & \nr{} & \nr{} & 7.9 & \nr{} & \nr{} \\ 
KVzip & 35.3 & \nr{} & \nr{} & 41.4 & \nr{} & \nr{} \\ 
BM25, same selector & 62.4 & 62.6 & 62.6 & 62.4 & 64.0 & 62.5 \\ 
BM25, chunk filling & 63.5 & 63.6 & 63.8 & 63.6 & 66.3 & 63.7 \\ 
Dense RAG + reranker & 66.9 & 61.2 & 61.2 & 68.6 & 66.5 & 61.7 \\ 
BM25 + dense & 67.8 & 65.5 & 69.1 & 66.4 & 69.6 & 66.7 \\ 
BM25 + lexical density & 66.9 & 64.8 & 66.6 & 70.9 & 68.4 & 65.8 \\ 
BM25 + BGE-M3 & 64.5 & 65.7 & 63.7 & 66.6 & 66.1 & 67.0 \\ 
\ours{}, \pqc{} & 68.2 (45.8 / 70.8) & 51.4 (22.9 / 33.3) & 53.5 (25.0 / 35.4) & 67.0 (52.1 / 68.8) & 67.6 (47.9 / 75.0) & 66.6 (43.8 / 77.1) \\ 
\ours{}+BM25, \pqc{} & 68.4 & 64.5 & 65.4 & 69.5 & 70.6 & 65.4 \\ 
\midrule
$\Delta$ \ours{} $-$ FullKV & $-0.15$ & $+16.60^*$ & $+37.42^*$ & $-9.00$ & $+22.69^*$ & $+53.17^*$ \\ 
 & \ci{-4.25}{+3.90} & \ci{+10.23}{+22.98} & \ci{+34.33}{+40.79} & \ci{-15.42}{-3.04} & \ci{+17.94}{+27.21} & \ci{+49.92}{+56.23} \\
$\Delta$ \ours{} $-$ BM25 & $+5.75^*$ & $-11.19$ & $-9.04$ & $+4.54$ & $+3.60$ & $+4.10$ \\ 
 & \ci{+3.25}{+8.12} & \ci{-16.06}{-6.10} & \ci{-13.29}{-4.90} & \ci{-0.98}{+11.04} & \ci{-0.58}{+7.83} & \ci{-0.04}{+8.10} \\
$\Delta$ \ours{} $-$ Dense RAG & $+1.23$ & $-9.83$ & $-7.73$ & $-1.60$ & $+1.10$ & $+4.88$ \\ 
 & \ci{-2.46}{+5.42} & \ci{-14.75}{-4.62} & \ci{-11.69}{-3.73} & \ci{-6.81}{+4.04} & \ci{-3.38}{+5.27} & \ci{+0.88}{+9.04} \\
$\Delta$ \ours{} $-$ BM25 + dense & $+0.33$ & $-14.12^*$ & $-15.62^*$ & $+0.56$ & $-2.04$ & $-0.19$ \\ 
 & \ci{-2.31}{+2.87} & \ci{-19.17}{-9.02} & \ci{-19.75}{-11.40} & \ci{-4.69}{+6.35} & \ci{-5.92}{+1.96} & \ci{-3.73}{+3.44} \\
$\Delta$ \ours{} $-$ BM25 + lex.\ density & $+1.27$ & $-13.33^*$ & $-13.08^*$ & $-3.92$ & $-0.79$ & $+0.79$ \\ 
 & \ci{-1.40}{+3.79} & \ci{-17.92}{-8.62} & \ci{-16.71}{-9.65} & \ci{-7.33}{-0.71} & \ci{-4.67}{+3.02} & \ci{-2.42}{+4.08} \\
$\Delta$ \ours{}+BM25 $-$ BM25 & $+5.96^*$ & $+1.92$ & $+2.79$ & $+7.04$ & $+6.62^*$ & $+2.96$ \\ 
 & \ci{+3.25}{+8.62} & \ci{-1.52}{+5.33} & \ci{-2.15}{+7.67} & \ci{+2.10}{+12.96} & \ci{+3.69}{+9.60} & \ci{-0.65}{+6.65} \\
$\Delta$ \ours{}+BM25 $-$ Dense RAG & $+1.44$ & $+3.27$ & $+4.10$ & $+0.90$ & $+4.12$ & $+3.73$ \\ 
 & \ci{-1.79}{+5.12} & \ci{-0.96}{+7.52} & \ci{-0.31}{+8.52} & \ci{-4.25}{+6.17} & \ci{+0.90}{+7.06} & \ci{-0.44}{+7.98} \\
$\Delta$ \ours{}+BM25 $-$ BM25 + dense & $+0.54$ & $-1.02$ & $-3.79$ & $+3.06$ & $+0.98$ & $-1.33$ \\ 
 & \ci{-2.50}{+3.46} & \ci{-4.42}{+2.10} & \ci{-7.92}{+0.58} & \ci{-2.25}{+8.52} & \ci{-1.33}{+3.31} & \ci{-5.08}{+2.31} \\
$\Delta$ \ours{}+BM25 $-$ BM25 + lex.\ density & $+1.48$ & $-0.23$ & $-1.25$ & $-1.42$ & $+2.23$ & $-0.35$ \\ 
 & \ci{-1.62}{+4.35} & \ci{-3.56}{+3.40} & \ci{-4.94}{+2.25} & \ci{-3.56}{+0.75} & \ci{-0.10}{+4.44} & \ci{-3.63}{+2.94} \\
\bottomrule
\end{tabular}}
\end{table}

\subsection{Two-head variant (\ourstwo{})}
\label{app:two-head}

The two-head variant keeps only the top-2 detected heads of Llama-3.1-8B-Instruct, (2,22) and (2,21).
Both heads share one KV head of layer 2.
The variant's \pqc{} index takes 32\,B per token, and its index forward stops at layer 2.
Everything else (sink, chunking, scoring, selector, budget, prompt, BM25 weight 0.5) is unchanged.
It was run on all 10 \bench{} test episodes per family (400 questions per length) and on LongBench QA (920 examples).
It was also run on NoLiMa, on LongMemEval-S and in the end-to-end benchmark of Appendix~\ref{app:eff}.
Table~\ref{tab:two-head} gives its scores, paired differences and costs.
At 128k its index build takes 0.66\,s against 5.73\,s for 16 heads.
Its retrieval takes 9\,ms per question against 38\,ms.
The shortcut is specific to Llama, whose two strongest heads share one KV head of layer 2.
On Qwen2.5-7B the top two heads, (4,3) and (5,10), occupy two KV heads of layers 4 and 5 (64\,B per token).
They score 38.6, 36.7 and 29.6 at 32k, 64k and 128k, 28.01--31.36 points below the 16 heads (Holm-significant).
With BM25 they score 63.2, 63.4 and 62.2 (Table~\ref{tab:e2-qwen-full}), not significantly different from BM25 ($-0.85$, $+0.50$, $-0.17$). 

\begin{table}[!htbp]
\centering
\caption{\textbf{Two-head variant} (Llama-3.1-8B-Instruct, heads (2,22) and (2,21), one KV head, 32\,B per token).
\bench{} rows use 10 test episodes per family (400 questions per length) and give macro score / macro presence.
Paired differences come with 95\% CI and Holm $p$ (families of Appendix~\ref{app:stats}).
LongBench QA rows give macro F1, computed on 920 examples.
Costs are measured on an idle H200 NVL GPU, in the end-to-end benchmark of Appendix~\ref{app:eff}.}
\label{tab:two-head}
\scriptsize
\setlength{\tabcolsep}{3.5pt}
\resizebox{\ifdim\width>\linewidth\linewidth\else\width\fi}{!}{%
\begin{tabular}{@{}lccc@{}}
\toprule
\bench{} & 32k & 64k & 128k \\
\midrule
\ourstwo{} & 74.8 / 0.843 & 68.8 / 0.775 & 70.4 / 0.777 \\ 
\ourstwo{}+BM25 & 76.7 / 0.854 & 70.9 / 0.810 & 72.7 / 0.805 \\ 
\midrule
$\Delta$ \ourstwo{} $-$ \ours{} (16 heads) & $+0.81$ \ci{-0.97}{+2.64}, 1.0000 & $-2.76$ \ci{-4.74}{-0.85}, 0.1890 & $-1.70$ \ci{-3.73}{+0.21}, 1.0000 \\ 
$\Delta$ \ourstwo{} $-$ FullKV & $-8.94$ \ci{-11.25}{-6.59}, 0.3000 & $-11.57$ \ci{-14.59}{-8.70}, 0.0675 & $+0.99$ \ci{-1.90}{+3.83}, 1.0000 \\ 
$\Delta$ \ourstwo{} $-$ BM25 & $+7.14^*$ \ci{+4.30}{+10.10}, 0.0020 & $+3.59$ \ci{+1.32}{+5.79}, 0.0675 & $+4.36$ \ci{+1.70}{+7.17}, 0.0559 \\ 
$\Delta$ \ourstwo{} $-$ dense RAG & $+2.24$ \ci{-0.44}{+4.95}, 1.0000 & $+2.12$ \ci{-0.79}{+4.97}, 1.0000 & $+4.21$ \ci{+0.82}{+7.72}, 0.6463 \\ 
$\Delta$ \ourstwo{} $-$ BM25 + dense & $+2.55$ \ci{-0.34}{+5.49}, 1.0000 & $-0.65$ \ci{-3.35}{+2.00}, 1.0000 & $-2.11$ \ci{-4.95}{+0.72}, 1.0000 \\ 
$\Delta$ \ourstwo{}+BM25 $-$ BM25 & $+9.00^*$ \ci{+5.95}{+12.09}, 0.0020 & $+5.65^*$ \ci{+3.54}{+7.75}, 0.0020 & $+6.67^*$ \ci{+4.22}{+9.20}, 0.0020 \\ 
$\Delta$ \ourstwo{}+BM25 $-$ dense RAG & $+4.10$ \ci{+0.79}{+7.41}, 0.2758 & $+4.19$ \ci{+1.40}{+7.04}, 0.1539 & $+6.53^*$ \ci{+3.04}{+9.99}, 0.0468 \\ 
$\Delta$ \ourstwo{}+BM25 $-$ FullKV & $-7.08$ \ci{-9.79}{-4.47}, 0.4810 & $-9.51$ \ci{-12.44}{-6.74}, 0.2312 & $+3.30$ \ci{+0.24}{+6.50}, 1.0000 \\ 
$\Delta$ \ourstwo{}+BM25 $-$ \ours{}+BM25 & $+0.68$ \ci{-0.56}{+2.04}, 0.8479 & $-1.95$ \ci{-3.66}{-0.20}, 0.6365 & $-1.20$ \ci{-2.62}{+0.21}, 1.0000 \\ 
\midrule
LongBench QA (Llama) & \multicolumn{3}{l}{Two heads 40.33; + BM25 41.03; 16 heads 41.93} \\ 
$\Delta$ two heads $-$ 16 heads (\pqc{}) & \multicolumn{3}{l}{$-1.59$ \ci{-3.43}{+0.14}, Holm $p=0.7221$} \\ 
$\Delta$ two heads $-$ BM25 & \multicolumn{3}{l}{$+0.45$ \ci{-1.49}{+2.41}, Holm $p=1.0000$} \\ 
$\Delta$ two heads $-$ dense RAG & \multicolumn{3}{l}{$-2.80$ \ci{-4.77}{-0.84}, Holm $p=0.0750$} \\ 
\midrule
Cost & \multicolumn{3}{l}{build (s) / retrieval per question (s) / peak GPU while building (GiB)} \\
32k & \multicolumn{3}{l}{two heads 0.164 / 0.006 / 0.489; dense RAG 0.655 / 0.113 / 0.143; BM25 -- / 0.006 / --} \\ 
128k & \multicolumn{3}{l}{two heads 0.664 / 0.009 / 0.492; dense RAG 0.234 / 0.088 / 0.153; BM25 -- / 0.023 / --} \\ 
1M & \multicolumn{3}{l}{two heads 5.358 / 0.058 / 0.519; dense RAG 1.720 / 0.085 / 0.158; BM25 -- / 0.200 / --} \\ 
\bottomrule
\end{tabular}}
\end{table}


\section{Holm-adjusted \texorpdfstring{$p$}{p} and key stores on test}\label{app:exp-codecs}

Section~\ref{sec:exp} reports \ours{} with its default codec, \pqc{} (448\,B per token on Llama-3.1-8B).
This appendix gives the Holm-adjusted $p$ of every contrast of the main text (Appendix~\ref{app:exp-holm}).
It also gives the test results of the two other key stores.
These are \saec{} codes (1{,}792\,B per token) and exact bf16 keys (3{,}584\,B per token).
Both were run with the same 16 retrieval heads, scoring, selector and budget $B{=}2{,}048$.
\ours{}+BM25 always uses BM25 weight 0.5.
The paired differences between the three key stores are in Table~\ref{tab:codec-test} (Appendix~\ref{app:codec-test}).

\subsection{Holm-adjusted \texorpdfstring{$p$}{p} of the quoted contrasts}\label{app:exp-holm}

Tables~\ref{tab:e2-deltas-pq-llama} and~\ref{tab:e2-deltas-pq-qwen} give the paired differences behind Table~\ref{tab:e2}, with CIs and Holm $p$.
Table~\ref{tab:holm-main} collects the Holm-adjusted two-sided sign-flip $p$ of the differences that Section~\ref{sec:exp-e2} quotes from them, of the LongBench contrasts of Table~\ref{tab:real} and of the runs beyond the window (Table~\ref{tab:beyond-full}).
It also gives the $p$ of the NoLiMa contrasts of Section~\ref{sec:exp-nolima}.
For every further contrast that Sections~\ref{sec:exp} and~\ref{sec:analysis} or Appendix~\ref{app:limits} quote, it gives the difference, 95\% CI and $p$.
The natural-text QA contrasts are in Appendix~\ref{app:qa-subset}, and the LongMemEval contrasts are in Appendix~\ref{app:lme}.
Every $p$ is adjusted within its family as defined in Appendix~\ref{app:stats} and listed in Table~\ref{tab:holm-families}.
The \Sthree{}-Hybrid contrasts form one family of 12 per model.
Each such family holds \ours{}, \ours{}+BM25, \ourstwo{} and \ourstwo{}+BM25 minus \Sthree{}-Hybrid at 32k, 64k and 128k.
With 4{,}000 resamples, 0.0030 is the smallest attainable Holm $p$ in this family.

\begin{table}[!htbp]
\centering
\caption{\textbf{Holm-adjusted $p$ of the quoted contrasts.} The table gives the $p$ of the Table~\ref{tab:e2} contrasts quoted in Section~\ref{sec:exp-e2}, of the LongBench contrasts of Table~\ref{tab:real} and of the runs beyond the window (Table~\ref{tab:beyond-full}).
It also gives the $p$ of the \ourstwo{} $-$ FullKV and REFORM-style contrasts and of the NoLiMa contrasts.
The continuation lists the contrasts quoted only in the text.
Llama (including \ourstwo{}) and Qwen2.5 at 32k--128k have 10 episodes per family and length.
Mistral and Qwen2 have 6 episodes per family and length, and lengths beyond 128k have 3.
\nr{} marks entries that were not run.}
\label{tab:holm-main}
\scriptsize
\setlength{\tabcolsep}{3.5pt}
\resizebox{\ifdim\width>\linewidth\linewidth\else\width\fi}{!}{%
\begin{tabular}{@{}lcccccc@{}}
\toprule
Table~\ref{tab:e2} & \multicolumn{3}{c}{Llama-3.1-8B-Instruct} & \multicolumn{3}{c}{Qwen2.5-7B-Instruct (YaRN)} \\
\cmidrule(lr){2-4}\cmidrule(l){5-7}
 & 32k & 64k & 128k & 32k & 64k & 128k \\
\midrule
\ours{} $-$ BM25 & 0.0056 & 0.0019 & 0.0039 & 1.0000 & 1.0000 & 1.0000 \\ 
\ours{}+BM25 $-$ BM25 & 0.0019 & 0.0019 & 0.0019 & 0.0019 & 0.0019 & 0.0209 \\ 
\ours{}+BM25 $-$ dense RAG & 0.5349 & 0.0045 & 0.0105 & 0.0192 & 0.1584 & 0.0840 \\ 
\ours{}+BM25 $-$ FullKV & 0.3390 & 0.1945 & 0.5861 & 1.0000 & 0.0290 & 0.0048 \\ 
\ourstwo{} $-$ FullKV & 0.3000 & 0.0675 & 1.0000 & 0.0020 & 0.0020 & 0.0020 \\ 
\ours{} $-$ REFORM-style & 0.0154 & 0.0022 & 0.0039 & 1.0000 & 0.5774 & 0.4879 \\ 
\midrule
NoLiMa (Llama, Section~\ref{sec:exp-nolima}) & 8k & 32k & 128k & & & \\
\midrule
\ours{} $-$ BM25 & 0.0010 & 0.0010 & 0.0015 &  &  &  \\ 
\ours{} $-$ dense RAG & 0.0045 & 0.4784 & 0.3527 &  &  &  \\ 
\ours{} $-$ FullKV & 0.2962 & 0.9178 & 0.0600 &  &  &  \\ 
\midrule
Tables~\ref{tab:real}, \ref{tab:beyond-full} & \multicolumn{3}{c}{(a) LongBench QA} & \multicolumn{3}{c}{(b) beyond the window, Llama} \\
\cmidrule(lr){2-4}\cmidrule(l){5-7}
 & Llama & Mistral & Qwen2 & 256k & 512k & 1M \\
\midrule
\ours{} $-$ BM25 & 0.3949 & 1.0000 & 1.0000 & 1.0000 & 0.7991 & 1.0000 \\ 
\ours{} $-$ dense RAG & 1.0000 & 0.0082 & 1.0000 & 1.0000 & 0.7991 & 0.9014 \\ 
\bottomrule
\end{tabular}}
\end{table}

\begin{table}[!htbp]
\centering
\caption*{\textbf{Table~\ref{tab:holm-main}, continued.} This part gives the \Sthree{}-Hybrid contrasts of Section~\ref{sec:exp-e2}, which form their own family per model.
It also gives the contrasts quoted only in the text of Sections~\ref{sec:exp} and~\ref{sec:analysis} or of Appendix~\ref{app:limits}.
$\Delta$ is the first-named minus the second method, with its 95\% bootstrap CI and Holm $p$.
$^*$ marks $p<0.05$.}
\scriptsize
\def\arraystretch{.84}
\setlength{\tabcolsep}{3.5pt}
\resizebox{\ifdim\width>\linewidth\linewidth\else\width\fi}{!}{%
\begin{tabular}{@{}llccc@{}}
\toprule
Contrast quoted in the text & Setting & $\Delta$ & 95\% CI & Holm $p$ \\
\midrule
\ours{}+BM25 $-$ \Sthree{}-Hybrid & Llama, 32k & $+8.80^*$ & [$+6.16$, $+11.51$] & 0.0030 \\ 
 & Llama, 64k & $+7.11^*$ & [$+4.61$, $+9.59$] & 0.0030 \\ 
 & Llama, 128k & $+7.69^*$ & [$+5.34$, $+10.19$] & 0.0030 \\ 
 & Qwen2.5, 32k & $+7.65^*$ & [$+3.61$, $+11.89$] & 0.0070 \\ 
 & Qwen2.5, 64k & $+6.78^*$ & [$+3.99$, $+9.78$] & 0.0030 \\ 
 & Qwen2.5, 128k & $+6.09^*$ & [$+3.19$, $+9.00$] & 0.0030 \\ 
\ours{} $-$ \Sthree{}-Hybrid & Llama, 32k & $+6.80^*$ & [$+3.88$, $+9.64$] & 0.0030 \\ 
 & Llama, 64k & $+5.86^*$ & [$+3.39$, $+8.34$] & 0.0040 \\ 
 & Llama, 128k & $+5.88^*$ & [$+2.81$, $+8.86$] & 0.0040 \\ 
\ours{} $-$ dense RAG & Llama, 64k & $+4.89$ & [$+2.08$, $+7.71$] & 0.1024 \\ 
 & Llama, 128k & $+5.91$ & [$+2.41$, $+9.29$] & 0.1611 \\ 
\ours{} $-$ FullKV & Llama, 32k & $-9.75$ & [$-12.39$, $-7.11$] & 0.1188 \\ 
 & Llama, 64k & $-8.81$ & [$-11.60$, $-6.21$] & 0.1071 \\ 
 & Llama, 128k & $+2.69$ & [$-0.26$, $+5.65$] & 1.0000 \\ 
\ours{}+BM25 $-$ BM25, chunk filling & Llama, 32k & $+7.39^*$ & [$+4.35$, $+10.34$] & 0.0022 \\ 
 & Llama, 64k & $+5.79^*$ & [$+3.35$, $+8.15$] & 0.0020 \\ 
 & Llama, 128k & $+7.00^*$ & [$+4.71$, $+9.36$] & 0.0019 \\ 
\ours{}+BM25 $-$ BM25 + dense & Llama, 32k & $+3.74$ & [$+0.93$, $+6.55$] & 0.2051 \\ 
 & Llama, 64k & $+3.36$ & [$+0.46$, $+6.33$] & 0.2208 \\ 
 & Llama, 128k & $+1.40$ & [$-1.40$, $+4.24$] & 1.0000 \\ 
\ours{}+BM25 $-$ BM25 + lexical density & Llama, 32k & $+2.99^*$ & [$+1.68$, $+4.36$] & 0.0486 \\ 
 & Llama, 64k & $+3.40$ & [$+1.45$, $+5.39$] & 0.0960 \\ 
 & Llama, 128k & $+4.44^*$ & [$+2.77$, $+6.15$] & 0.0036 \\ 
 & Qwen2.5, 32k & $+3.25^*$ & [$+0.86$, $+5.47$] & 0.0352 \\ 
 & Qwen2.5, 64k & $+2.58$ & [$+0.35$, $+5.00$] & 0.1584 \\ 
 & Qwen2.5, 128k & $+0.38$ & [$-2.24$, $+3.11$] & 1.0000 \\ 
\ours{}+BM25 $-$ BM25 + BGE-M3 & Llama, 32k & $+4.25$ & [$+1.02$, $+7.58$] & 0.2032 \\ 
 & Llama, 64k & $+4.16$ & [$+1.40$, $+6.90$] & 0.0518 \\ 
 & Llama, 128k & $+5.53^*$ & [$+2.54$, $+8.84$] & 0.0088 \\ 
\ours{} $-$ BM25 + lexical density & Qwen2.5, 32k & $-2.31$ & [$-5.33$, $+0.88$] & 1.0000 \\ 
 & Qwen2.5, 64k & $-1.80$ & [$-4.38$, $+1.02$] & 1.0000 \\ 
 & Qwen2.5, 128k & $-6.94^*$ & [$-10.79$, $-3.16$] & 0.0143 \\ 
\ours{} $-$ KVzip & Llama, 32k & $+16.89^*$ & [$+12.51$, $+21.27$] & 0.0410 \\ 
\ours{} $-$ StreamingLLM & Llama, 128k & $+65.21^*$ & [$+62.36$, $+67.89$] & 0.0019 \\ 
\ours{} $-$ \Sthree{}-Attention-style index & Llama, 32k & $+63.05^*$ & [$+59.62$, $+66.26$] & 0.0019 \\ 
 & Llama, 64k & $+63.83^*$ & [$+60.69$, $+67.03$] & 0.0019 \\ 
 & Llama, 128k & $+66.24^*$ & [$+62.91$, $+69.49$] & 0.0019 \\ 
 & LongBench, Llama & $+7.52^*$ & [$+5.07$, $+10.04$] & 0.0032 \\ 
 & LongBench, Mistral & $+7.44^*$ & [$+5.07$, $+9.80$] & 0.0030 \\ 
 & LongBench, Qwen2 & $+6.23^*$ & [$+3.73$, $+8.59$] & 0.0030 \\ 
\ours{} $-$ ILRe-style & Llama, 32k & $+50.73^*$ & [$+46.91$, $+54.39$] & 0.0019 \\ 
 & Llama, 64k & $+55.88^*$ & [$+52.74$, $+59.11$] & 0.0019 \\ 
 & Llama, 128k & $+57.81^*$ & [$+53.50$, $+62.01$] & 0.0019 \\ 
 & Qwen2.5, 32k & $+32.64^*$ & [$+29.43$, $+35.84$] & 0.0019 \\ 
 & Qwen2.5, 64k & $+45.73^*$ & [$+41.86$, $+49.56$] & 0.0019 \\ 
 & Qwen2.5, 128k & $+45.62^*$ & [$+40.89$, $+50.44$] & 0.0019 \\ 
\ours{}+BM25 $-$ REFORM-style & Qwen2.5, 64k & $+8.53^*$ & [$+5.70$, $+11.14$] & 0.0020 \\ 
 & Qwen2.5, 128k & $+3.66^*$ & [$+1.74$, $+5.70$] & 0.0432 \\ 
\ourstwo{} $-$ BM25 & Llama, 32k & $+7.14^*$ & [$+4.30$, $+10.10$] & 0.0020 \\ 
 & Llama, 128k & $+4.36$ & [$+1.70$, $+7.17$] & 0.0559 \\ 
\ourstwo{}+BM25 $-$ dense RAG & Llama, 128k & $+6.53^*$ & [$+3.04$, $+9.99$] & 0.0468 \\ 
\ourstwo{} $-$ \ours{} (16 heads) & Llama, 32k & $+0.81$ & [$-0.97$, $+2.64$] & 1.0000 \\ 
 & Llama, 64k & $-2.76$ & [$-4.74$, $-0.85$] & 0.1890 \\ 
 & Llama, 128k & $-1.70$ & [$-3.73$, $+0.21$] & 1.0000 \\ 
\ours{} $-$ FullKV (beyond the window) & Mistral, 128k & $+37.42^*$ & [$+34.33$, $+40.79$] & 0.0008 \\ 
 & Qwen2, 128k & $+53.17^*$ & [$+49.92$, $+56.23$] & 0.0008 \\ 
\ours{} $-$ BM25 & Mistral, 64k & $-11.19$ & [$-16.06$, $-6.10$] & 0.0524 \\ 
 & Mistral, 128k & $-9.04$ & [$-13.29$, $-4.90$] & 0.0778 \\ 
\ours{}+BM25 $-$ BM25 & Mistral, 32k & $+5.96^*$ & [$+3.25$, $+8.62$] & 0.0328 \\ 
 & Mistral, 64k & $+1.92$ & [$-1.52$, $+5.33$] & 1.0000 \\ 
 & Mistral, 128k & $+2.79$ & [$-2.15$, $+7.67$] & 1.0000 \\ 
 & Qwen2, 128k & $+2.96$ & [$-0.65$, $+6.65$] & 1.0000 \\ 
 & Llama, 256k & $+8.83^*$ & [$+5.33$, $+12.37$] & 0.0140 \\ 
\ours{}+BM25 $-$ dense RAG & Qwen2, 128k & $+3.73$ & [$-0.44$, $+7.98$] & 1.0000 \\ 
 & Llama, 256k & $+10.83$ & [$+6.67$, $+14.79$] & 0.0894 \\ 
 & Llama, 512k & $+14.12$ & [$+8.29$, $+20.12$] & 0.2984 \\ 
 & Llama, 1M & $+14.00$ & [$+7.58$, $+20.67$] & 0.1134 \\ 
\ours{}+BM25 $-$ BM25 + dense & Llama, 256k & $+0.25$ & [$-3.50$, $+4.00$] & 1.0000 \\ 
 & Llama, 512k & $-4.42$ & [$-8.42$, $-0.42$] & 1.0000 \\ 
 & Llama, 1M & $-3.54$ & [$-11.00$, $+3.29$] & 1.0000 \\ 
\ours{} $-$ FullKV & LongBench, Llama & $-3.04^*$ & [$-5.06$, $-0.98$] & 0.0330 \\ 
 & LongBench, Mistral & $-3.38^*$ & [$-5.50$, $-1.25$] & 0.0082 \\ 
 & LongBench, Qwen2 & $-0.00$ & [$-2.09$, $+1.97$] & 1.0000 \\ 
\ours{} $-$ \ours{}+BM25 & NoLiMa, 8k & $+12.11^*$ & [$+7.89$, $+16.33$] & 0.0010 \\ 
 & NoLiMa, 32k & $+8.00^*$ & [$+4.67$, $+11.67$] & 0.0010 \\ 
 & NoLiMa, 128k & $+4.61^*$ & [$+2.44$, $+7.06$] & 0.0010 \\ 
\ours{}+BM25 $-$ FullKV & InfiniteBench En.MC{,} $B{=}2{,}048$ & $+5.24$ & [$-2.18$, $+12.23$] & 0.3779 \\ 
\bottomrule
\end{tabular}}
\end{table}

\subsection{\bench{}, 32k--128k}\label{app:exp-codecs-e2}

Table~\ref{tab:e2-codecs} compares the key stores on \bench{}.
On Llama, \ours{} with \saec{} codes significantly exceeds BM25 under the same selector at every length ($+7.09$, $+8.10$ and $+7.05$, all Holm-significant).
It is not significantly different from the strongest text retriever at 32k and 128k ($+1.75$ and $+0.57$, with CIs that include zero) and scores above it at 64k ($+3.86$~\ci{+0.61}{+7.16}).
It is not significantly different from exact keys at any length.
It scores below FullKV at 32k and 64k ($-8.99$ and $-7.06$, with CIs that exclude zero).
At 128k it scores above FullKV ($+3.67$~\ci{+0.46}{+6.99}).
On Qwen2.5 it is not significantly different from BM25 at any length.
It is below BM25 + lexical density at 64k and 128k with CIs that exclude zero ($-2.44$, $-3.33$).
Its own BM25 fusion scores higher at every length (CIs exclude zero).
With \pqc{} codes on Qwen2.5, \ours{} scores above BM25 at 64k ($+2.61$~\ci{+0.05}{+5.09}) and below BM25 + lexical density at 128k ($-6.94$, Holm $p=0.0143$), with CIs that include zero at the other lengths (Table~\ref{tab:e2-deltas-pq-qwen}).
With \pqc{} codes, \ours{}+BM25 significantly exceeds \ours{} alone at every Qwen2.5 length ($+5.56$, $+4.38$ and $+7.31$, all Holm-significant) and scores above it at Llama 128k ($+1.81$~\ci{+0.06}{+3.75}).

\begin{table}[!htbp]
\centering
\caption{\textbf{\bench{} with every key store} (macro score $\times100$, 400 questions per entry, $B{=}2{,}048$).
The best text retriever is the text retriever with the highest macro score in the column.
It is BM25 + lexical density for Llama 32k and for Qwen2.5, and BM25 + dense for Llama 64k and 128k.
$\Delta$ is the first-named minus the second method, with the 95\% CI below.
$^*$ marks Holm $p<0.05$ (Appendix~\ref{app:stats}).}
\label{tab:e2-codecs}
\footnotesize
\setlength{\tabcolsep}{3pt}
\renewcommand{\arraystretch}{0.9}
\resizebox{\ifdim\width>\linewidth\linewidth\else\width\fi}{!}{%
\begin{tabular}{@{}lcccccc@{}}
\toprule
 & \multicolumn{3}{c}{Llama-3.1-8B-Instruct} & \multicolumn{3}{c}{Qwen2.5-7B-Instruct (YaRN)} \\
\cmidrule(lr){2-4}\cmidrule(l){5-7}
Method & 32k & 64k & 128k & 32k & 64k & 128k \\
\midrule
BM25, same selector & 67.7 & 65.3 & 66.0 & 64.0 & 62.9 & 62.4 \\ 
\midrule
\ours{}, exact keys (3{,}584\,B/token) & 75.2 & 74.0 & 72.0 & 66.9 & 64.6 & 65.7 \\ 
\ours{}, \saec{} (1{,}792\,B/token) & 74.8 & 73.4 & 73.1 & 67.3 & 64.9 & 64.5 \\ 
\ours{}, \pqc{} (448\,B/token) & 74.0 & 71.6 & 72.1 & 66.6 & 65.5 & 60.9 \\ 
\ours{}+BM25, \saec{} & 76.3 & 75.3 & 74.2 & 71.2 & 70.3 & 69.1 \\ 
\ours{}+BM25, \pqc{} & 76.0 & 72.9 & 73.9 & 72.1 & 69.9 & 68.2 \\ 
\midrule
\multicolumn{7}{@{}l}{\textit{\saec{} codes: \ours{} (\saec{}) minus X}} \\
BM25, same selector & $+7.09^*$ & $+8.10^*$ & $+7.05^*$ & $+3.30$ & $+1.98$ & $+2.17$ \\ 
 & \ci{+3.70}{+10.61} & \ci{+5.54}{+10.62} & \ci{+4.35}{+9.91} & \ci{-1.02}{+7.80} & \ci{-1.34}{+5.24} & \ci{-1.11}{+5.67} \\
best text retriever & $+1.75$ & $+3.86$ & $+0.57$ & $-1.55$ & $-2.44$ & $-3.33$ \\ 
 & \ci{-0.64}{+4.28} & \ci{+0.61}{+7.16} & \ci{-2.24}{+3.34} & \ci{-3.95}{+1.09} & \ci{-4.94}{-0.12} & \ci{-6.71}{-0.10} \\
FullKV & $-8.99$ & $-7.06$ & $+3.67$ & $-4.54$ & $+2.25$ & $+8.75$ \\ 
 & \ci{-11.53}{-6.41} & \ci{-9.84}{-4.42} & \ci{+0.46}{+6.99} & \ci{-8.38}{-0.78} & \ci{-1.57}{+6.01} & \ci{+5.54}{+11.88} \\
exact keys & $-0.40$ & $-0.69$ & $+1.10$ & $+0.40$ & $+0.30$ & $-1.16$ \\ 
 & \ci{-1.60}{+0.78} & \ci{-2.89}{+1.10} & \ci{-0.51}{+2.71} & \ci{-2.58}{+3.19} & \ci{-1.93}{+2.55} & \ci{-3.31}{+0.71} \\
\ours{}+BM25 (\saec{}) & $-1.56$ & $-1.94$ & $-1.07$ & $-3.88^*$ & $-5.38^*$ & $-4.56$ \\ 
 & \ci{-3.00}{-0.12} & \ci{-4.25}{+0.12} & \ci{-3.65}{+1.50} & \ci{-5.88}{-1.88} & \ci{-8.44}{-2.50} & \ci{-7.44}{-1.81} \\
\midrule
\multicolumn{7}{@{}l}{\textit{\pqc{} codes}} \\
\ours{} minus best text retriever & $+0.99$ & $+2.11$ & $-0.41$ & $-2.31$ & $-1.80$ & $-6.94^*$ \\ 
 & \ci{-1.44}{+3.56} & \ci{-0.81}{+5.14} & \ci{-3.19}{+2.30} & \ci{-5.33}{+0.88} & \ci{-4.38}{+1.02} & \ci{-10.79}{-3.16} \\
\ours{}+BM25 minus \ours{} & $+2.00$ & $+1.25$ & $+1.81$ & $+5.56^*$ & $+4.38^*$ & $+7.31^*$ \\ 
 & \ci{-0.50}{+4.31} & \ci{-0.06}{+2.62} & \ci{+0.06}{+3.75} & \ci{+2.56}{+8.38} & \ci{+2.00}{+6.81} & \ci{+4.38}{+10.25} \\
\bottomrule
\end{tabular}}
\end{table}

\subsection{LongBench QA}\label{app:exp-codecs-lb}

Table~\ref{tab:lb-codecs} compares the key stores on LongBench QA.
On Llama, \ours{} with \saec{} codes is not significantly different from BM25 ($+0.78$~\ci{-1.23}{+2.83}) or from the three fusions.
It scores above the \Sthree{}-Attention-style index fused with BM25 ($+2.33$~\ci{+0.26}{+4.51}).
It significantly exceeds that index alone ($+6.25$~\ci{+3.85}{+8.71}, Holm $p=0.0032$).
On Mistral it is not significantly different from BM25, BM25 chunk filling or the three fusions.
Dense RAG scores significantly higher ($-3.31$, Holm $p=0.0302$) and FullKV scores higher ($-2.87$~\ci{-4.97}{-0.82}, Holm $p=0.0675$).
Exact keys do no better (35.71).
On Qwen2 it is not significantly different from FullKV or any text retriever (CIs include zero).
The same holds for \ours{} with \pqc{} codes (Table~\ref{tab:real} and Appendix~\ref{app:limits}).

\begin{table}[!htbp]
\centering
\caption{\textbf{LongBench QA with every key store} (macro F1 $\times100$, six QA tasks, 920 test examples per model, $B{=}2{,}048$ for retrieval methods).
FullKV middle-truncates to the window.
$\Delta$ is the first-named minus the second method, with 95\% CI.
$^*$ marks Holm $p<0.05$.
\nr{} marks entries that were not run.}
\label{tab:lb-codecs}
\footnotesize
\setlength{\tabcolsep}{4pt}
\renewcommand{\arraystretch}{0.9}
\resizebox{\ifdim\width>\linewidth\linewidth\else\width\fi}{!}{%
\begin{tabular}{@{}lccc@{}}
\toprule
Method & Llama-3.1-8B & Mistral-7B & Qwen2-7B \\
\midrule
FullKV & 44.97 & 38.95 & 37.93 \\ 
BM25, same selector & 39.88 & 36.77 & 37.53 \\ 
Dense RAG + reranker & 43.14 & 39.40 & 39.43 \\ 
\midrule
\ours{}, exact keys & 42.89 & 35.71 & 38.91 \\ 
\ours{}, \saec{} & 40.66 & 36.08 & 38.14 \\ 
\ours{}, \pqc{} & 41.93 & 35.56 & 37.93 \\ 
\ours{}+BM25, \saec{} & 40.34 & 36.58 & 38.78 \\ 
\ours{}+BM25, \pqc{} & 41.17 & 36.20 & 38.69 \\ 
\midrule
\multicolumn{4}{@{}l}{\textit{\saec{} codes: \ours{} (\saec{}) minus X}} \\
BM25, same selector & $+0.78$ \ci{-1.23}{+2.83} & $-0.69$ \ci{-2.96}{+1.57} & $+0.61$ \ci{-1.12}{+2.31} \\ 
dense RAG & $-2.48$ \ci{-4.54}{-0.40} & $-3.31^*$ \ci{-5.54}{-1.30} & $-1.29$ \ci{-3.33}{+0.66} \\ 
FullKV & $-4.31^*$ \ci{-6.43}{-2.25} & $-2.87$ \ci{-4.97}{-0.82} & $+0.22$ \ci{-1.79}{+2.23} \\ 
exact keys & $-2.23^*$ \ci{-3.65}{-0.87} & $+0.37$ \ci{-0.83}{+1.61} & $-0.77$ \ci{-2.00}{+0.43} \\ 
\ours{}+BM25 (\saec{}) & $+0.31$ \ci{-1.57}{+2.12} & $-0.49$ \ci{-2.43}{+1.40} & $-0.64$ \ci{-2.05}{+0.85} \\ 
\midrule
\multicolumn{4}{@{}l}{\textit{\pqc{} codes}} \\
\ours{} minus \ours{}+BM25 & $+0.76$ \ci{-0.91}{+2.52} & $-0.64$ \ci{-2.51}{+1.21} & $-0.76$ \ci{-2.50}{+0.91} \\ 
\ours{}+BM25 minus dense RAG & $-1.97$ \ci{-3.76}{-0.06} & $-3.20^*$ \ci{-5.46}{-1.16} & $-0.75$ \ci{-2.63}{+1.20} \\ 
\bottomrule
\end{tabular}}
\end{table}


\section{Codec agreement on test episodes}
\label{app:codec-test}

Section~\ref{sec:ana-codec} compares key codecs on development episodes.
Table~\ref{tab:codec-test} gives the paired test differences of \ours{} with \pqc{} codes minus exact bf16 keys and minus \saec{} codes.
The heads, selector, budget and paired bootstrap are the same as in Section~\ref{sec:exp}.
\pqc{} is significantly below exact keys only on Qwen2.5-7B at 128k ($-4.78$~\ci{-7.94}{-1.76}, Holm $p=0.0410$).
It uses one eighth of the bytes of exact keys.
In the other eight comparisons no difference is significant, and seven of their CIs include zero.
The intervals bound a possible loss rather than establish equivalence.
On Llama-3.1-8B \bench{}, \pqc{} minus \saec{} is $-0.76$~\ci{-2.38}{+0.85}, $-1.75$~\ci{-3.88}{+0.26} and $-0.99$~\ci{-2.86}{+0.86} points at 32k, 64k and 128k.
On LongBench QA it is $+1.27$~\ci{-0.31}{+2.90}, $-0.52$~\ci{-1.83}{+0.81} and $-0.22$~\ci{-1.60}{+1.16} for Llama-3.1-8B, Mistral-7B and Qwen2-7B.
Every one of these CIs includes zero.
On Qwen2.5-7B \bench{} it is $-0.76$~\ci{-3.74}{+2.05} and $+0.64$~\ci{-2.77}{+4.03} at 32k and 64k, with CIs that include zero.
At 128k it is $-3.61$~\ci{-6.44}{-0.79} (Holm $p=0.1720$).

\begin{table}[!htbp]
\centering
\caption{\textbf{Key stores on test.} The table gives paired differences of \ours{} with \pqc{} codes (448\,B per token) minus exact bf16 keys (3{,}584\,B) and minus \saec{} codes (1{,}792\,B).
Each difference comes with its 95\% CI and Holm $p$.
\bench{} uses 10 episodes per family and length.
LongBench uses 920 examples per model.}
\label{tab:codec-test}
\scriptsize
\setlength{\tabcolsep}{3.5pt}
\resizebox{\ifdim\width>\linewidth\linewidth\else\width\fi}{!}{%
\begin{tabular}{@{}llrccrcc@{}}
\toprule
& & \multicolumn{3}{c}{\pqc{} $-$ exact keys} & \multicolumn{3}{c}{\pqc{} $-$ \saec{}} \\
\cmidrule(lr){3-5}\cmidrule(l){6-8}
Benchmark & Setting & $\Delta$ & 95\% CI & Holm $p$ & $\Delta$ & 95\% CI & Holm $p$ \\
\midrule
\bench{} & Llama, 32k & $-1.16$ & \ci{-2.79}{+0.47} & 1.0000 & $-0.76$ & \ci{-2.38}{+0.85} & 1.0000 \\ 
\bench{} & Llama, 64k & $-2.44$ & \ci{-4.53}{-0.34} & 0.3042 & $-1.75$ & \ci{-3.88}{+0.26} & 0.5079 \\ 
\bench{} & Llama, 128k & $+0.11$ & \ci{-1.57}{+1.73} & 1.0000 & $-0.99$ & \ci{-2.86}{+0.86} & 1.0000 \\ 
\bench{} & Qwen2.5, 32k & $-0.36$ & \ci{-3.44}{+2.84} & 1.0000 & $-0.76$ & \ci{-3.74}{+2.05} & 1.0000 \\ 
\bench{} & Qwen2.5, 64k & $+0.94$ & \ci{-2.38}{+4.10} & 1.0000 & $+0.64$ & \ci{-2.77}{+4.03} & 1.0000 \\ 
\bench{} & Qwen2.5, 128k & $-4.78^*$ & \ci{-7.94}{-1.76} & 0.0410 & $-3.61$ & \ci{-6.44}{-0.79} & 0.1720 \\ 
LongBench QA & Llama & $-0.96$ & \ci{-2.49}{+0.59} & 1.0000 & $+1.27$ & \ci{-0.31}{+2.90} & 0.9335 \\ 
LongBench QA & Mistral & $-0.15$ & \ci{-1.41}{+1.16} & 1.0000 & $-0.52$ & \ci{-1.83}{+0.81} & 1.0000 \\ 
LongBench QA & Qwen2 & $-0.99$ & \ci{-2.43}{+0.36} & 1.0000 & $-0.22$ & \ci{-1.60}{+1.16} & 1.0000 \\ 
\bottomrule
\end{tabular}}
\end{table}


\section{Full LongBench results}
\label{app:lb-full}

Table~\ref{tab:lb-full} gives every task behind Table~\ref{tab:real}b, for every key store that was run.
Table~\ref{tab:lb-deltas-pq} gives every paired difference of \ours{} and \ours{}+BM25 with \pqc{} codes.
Table~\ref{tab:lb-deltas} gives every paired difference of \ours{} with \saec{} codes.

On Llama, \ours{} (\pqc{}) scores above BM25 ($\Delta=+2.05$, 95\% CI [$+0.07$, $+4.10$]).
It is not significantly different from dense RAG and the three fusions.
It significantly exceeds the \Sthree{}-Attention-style index alone ($+7.52$, Holm $p=0.0032$) and fused with BM25 ($+3.60$, Holm $p=0.0090$).
\pqc{} codes (41.93) are not significantly different from exact keys (42.89) or from \saec{} codes (40.66) (Table~\ref{tab:codec-test}).
On Qwen2-7B, \ours{} (\pqc{}) at 2{,}048 tokens scores 37.93, as does FullKV on the middle-truncated context ($\Delta=-0.00$, CI [$-2.09$, $+1.97$]).
\ours{} is not significantly different from any text retriever.
FullKV on Mistral-7B scores 38.95.
This is below dense RAG + reranker (39.40) and above every other method in the Mistral block.
Per task, FullKV is highest on NarrativeQA (25.4, against 13.3--20.6 for the retrieval methods) and on Qasper (38.7, against at most 37.9).
\ours{} (\pqc{}) scores 35.56 on Mistral-7B, and \saec{} codes and exact keys give 36.08 and 35.71.
Its CIs against BM25, chunk filling and the three fusions include zero.
FullKV scores significantly higher ($\Delta=-3.38$, 95\% CI [$-5.50$, $-1.25$], Holm $p=0.0082$), and so does dense RAG + reranker ($\Delta=-3.83$, Holm $p=0.0082$).

\begin{table}[!htbp]
\centering
\caption{\textbf{LongBench QA, every task} (F1 $\times100$, 920 test examples per model). FullKV middle-truncates to the window. Every other method reads $B{=}2{,}048$ context tokens. Appendices~\ref{app:two-head} and~\ref{app:robust-heads} discuss the two-head variant and the heads detected without QA.}
\label{tab:lb-full}
\scriptsize
\setlength{\tabcolsep}{3.5pt}
\resizebox{\ifdim\width>\linewidth\linewidth\else\width\fi}{!}{%
\begin{tabular}{@{}lccccccc@{}}
\toprule
Method & 2WikiMQA & HotpotQA & MultiFieldQA-en & MuSiQue & NarrativeQA & Qasper & Macro \\
\midrule
\multicolumn{8}{@{}l}{\textit{Llama-3.1-8B-Instruct}} \\
FullKV & 49.8 & 56.1 & 57.1 & 33.1 & 27.5 & 46.1 & 44.97 \\ 
BM25, same selector & 40.7 & 52.6 & 54.9 & 29.3 & 18.1 & 43.7 & 39.88 \\ 
BM25, chunk filling & 43.6 & 55.0 & 56.3 & 28.2 & 18.7 & 43.4 & 40.87 \\ 
Dense RAG + reranker & 48.2 & 56.1 & 55.8 & 32.8 & 21.0 & 44.8 & 43.14 \\ 
BM25 + dense & 40.7 & 56.9 & 55.7 & 32.7 & 17.6 & 42.8 & 41.06 \\ 
BM25 + lexical density & 42.2 & 55.1 & 56.0 & 27.8 & 18.4 & 42.9 & 40.40 \\ 
BM25 + BGE-M3 & 43.2 & 53.0 & 56.3 & 31.0 & 21.3 & 43.4 & 41.37 \\ 
\Sthree{}-Attention-style index & 36.8 & 46.5 & 41.6 & 25.7 & 18.7 & 37.1 & 34.41 \\ 
\Sthree{}-Attention-style index + BM25 & 39.7 & 49.6 & 50.2 & 27.8 & 19.9 & 42.7 & 38.33 \\ 
\Sthree{}-Attention-style, deeper layers, + BM25 & 37.9 & 49.6 & 51.5 & 30.0 & 19.9 & 43.1 & 38.68 \\ 
\Sthree{}-Attention-style, layer 0, + BM25 & 40.7 & 53.9 & 53.7 & 29.8 & 21.9 & 42.7 & 40.45 \\ 
\ours{}, exact keys & 47.4 & 56.7 & 55.7 & 31.7 & 20.6 & 45.3 & 42.89 \\ 
\ours{}, \saec{} & 45.8 & 52.7 & 54.0 & 29.9 & 18.4 & 43.1 & 40.66 \\ 
\ours{}+BM25, \saec{} & 40.9 & 55.4 & 54.4 & 29.8 & 17.4 & 44.2 & 40.34 \\ 
\ours{}, \pqc{} & 44.7 & 55.2 & 54.8 & 33.5 & 20.1 & 43.2 & 41.93 \\ 
\ours{}+BM25, \pqc{} & 42.7 & 57.1 & 54.5 & 30.2 & 18.8 & 43.7 & 41.17 \\ 
\ourstwo{} & 45.4 & 53.0 & 54.4 & 27.8 & 18.5 & 42.9 & 40.33 \\ 
\ourstwo{}+BM25 & 45.5 & 54.8 & 56.9 & 29.3 & 16.9 & 42.7 & 41.03 \\ 
\ours{}, heads detected without QA & 41.8 & 53.4 & 48.7 & 33.4 & 19.5 & 41.4 & 39.72 \\ 
\ours{}+BM25, heads detected without QA & 43.2 & 55.7 & 54.7 & 33.1 & 19.4 & 41.5 & 41.26 \\ 
\midrule
\multicolumn{8}{@{}l}{\textit{Mistral-7B-Instruct-v0.3}} \\
FullKV & 38.3 & 51.6 & 51.4 & 28.3 & 25.4 & 38.7 & 38.95 \\ 
BM25, same selector & 41.3 & 52.0 & 49.7 & 25.0 & 17.4 & 35.3 & 36.77 \\ 
BM25, chunk filling & 39.8 & 53.5 & 52.2 & 25.5 & 16.9 & 35.7 & 37.26 \\ 
Dense RAG + reranker & 45.6 & 52.2 & 50.9 & 29.3 & 20.6 & 37.9 & 39.40 \\ 
BM25 + dense & 42.3 & 52.0 & 50.7 & 25.9 & 17.7 & 36.9 & 37.59 \\ 
BM25 + lexical density & 39.7 & 50.7 & 51.4 & 28.6 & 17.0 & 35.9 & 37.25 \\ 
BM25 + BGE-M3 & 40.4 & 50.8 & 49.4 & 25.6 & 16.0 & 37.1 & 36.54 \\ 
\Sthree{}-Attention-style index & 36.2 & 37.4 & 34.8 & 19.8 & 13.3 & 27.2 & 28.13 \\ 
\ours{}, exact keys & 38.3 & 49.6 & 51.9 & 23.1 & 15.4 & 36.0 & 35.71 \\ 
\ours{}, \saec{} & 38.1 & 51.2 & 49.9 & 26.1 & 15.2 & 35.9 & 36.08 \\ 
\ours{}+BM25, \saec{} & 39.0 & 50.2 & 52.3 & 26.7 & 14.8 & 36.4 & 36.58 \\ 
\ours{}, \pqc{} & 39.8 & 46.5 & 50.3 & 25.7 & 15.2 & 36.0 & 35.56 \\ 
\ours{}+BM25, \pqc{} & 37.9 & 50.4 & 51.3 & 25.9 & 17.2 & 34.5 & 36.20 \\ 
\midrule
\multicolumn{8}{@{}l}{\textit{Qwen2-7B-Instruct}} \\
FullKV & 45.1 & 43.0 & 46.8 & 26.1 & 22.7 & 43.9 & 37.93 \\ 
BM25, same selector & 39.0 & 50.7 & 49.4 & 26.6 & 18.8 & 40.7 & 37.53 \\ 
BM25, chunk filling & 43.6 & 49.5 & 50.3 & 26.2 & 18.0 & 42.2 & 38.30 \\ 
Dense RAG + reranker & 43.7 & 50.4 & 49.2 & 28.2 & 21.4 & 43.6 & 39.43 \\ 
BM25 + dense & 46.4 & 50.3 & 51.5 & 24.6 & 19.4 & 39.7 & 38.65 \\ 
BM25 + lexical density & 46.3 & 49.1 & 49.4 & 28.7 & 17.7 & 40.7 & 38.66 \\ 
BM25 + BGE-M3 & 42.6 & 46.3 & 49.5 & 31.7 & 20.7 & 40.1 & 38.49 \\ 
\Sthree{}-Attention-style index & 41.5 & 41.4 & 39.5 & 16.8 & 15.2 & 35.7 & 31.69 \\ 
\ours{}, exact keys & 44.0 & 49.8 & 51.5 & 29.6 & 18.2 & 40.5 & 38.91 \\ 
\ours{}, \saec{} & 42.2 & 48.9 & 49.7 & 28.8 & 18.7 & 40.6 & 38.14 \\ 
\ours{}+BM25, \saec{} & 44.2 & 49.4 & 50.2 & 29.0 & 18.0 & 41.9 & 38.78 \\ 
\ours{}, \pqc{} & 41.5 & 46.9 & 50.8 & 30.0 & 19.2 & 39.2 & 37.93 \\ 
\ours{}+BM25, \pqc{} & 45.9 & 47.5 & 50.0 & 29.0 & 18.5 & 41.2 & 38.69 \\ 
\bottomrule
\end{tabular}}
\end{table}

\begin{table}[!htbp]
\centering
\caption{\textbf{Paired differences on LongBench QA, \pqc{} codes.} The columns give \ours{} minus X and \ours{}+BM25 minus X per example. Each difference has a 95\% bootstrap CI (4{,}000 resamples, stratified by task) and a Holm $p$ within each model. $^*$ marks $p<0.05$. The table continues with Mistral and Qwen2 below.}
\label{tab:lb-deltas-pq}
\scriptsize
\setlength{\tabcolsep}{3pt}
\resizebox{\ifdim\width>\linewidth\linewidth\else\width\fi}{!}{%
\begin{tabular}{@{}lrccrcc@{}}
\toprule
& \multicolumn{3}{c}{\ours{} (\pqc{}) $-$ X} & \multicolumn{3}{c}{\ours{}+BM25 (\pqc{}) $-$ X} \\
\cmidrule(lr){2-4}\cmidrule(l){5-7}
X & $\Delta$ & 95\% CI & Holm $p$ & $\Delta$ & 95\% CI & Holm $p$ \\
\midrule
\multicolumn{7}{@{}l}{\textit{Llama-3.1-8B-Instruct}} \\
FullKV & $-3.04^*$ & \ci{-5.06}{-0.98} & 0.0330 & $-3.80^*$ & \ci{-5.95}{-1.59} & 0.0032 \\ 
BM25 & $+2.05$ & \ci{+0.07}{+4.10} & 0.3949 & $+1.29$ & \ci{-0.50}{+3.18} & 1.0000 \\ 
BM25, chunk filling & $+1.06$ & \ci{-0.69}{+2.92} & 1.0000 & $+0.30$ & \ci{-1.30}{+1.90} & 1.0000 \\ 
Dense RAG & $-1.21$ & \ci{-3.07}{+0.74} & 1.0000 & $-1.97$ & \ci{-3.76}{-0.06} & 0.3774 \\ 
BM25 + dense & $+0.86$ & \ci{-1.08}{+2.91} & 1.0000 & $+0.10$ & \ci{-1.75}{+2.03} & 1.0000 \\ 
BM25 + lex.\ density & $+1.53$ & \ci{-0.31}{+3.36} & 0.9335 & $+0.77$ & \ci{-0.85}{+2.36} & 1.0000 \\ 
BM25 + BGE-M3 & $+0.56$ & \ci{-1.39}{+2.59} & 1.0000 & $-0.20$ & \ci{-2.05}{+1.73} & 1.0000 \\ 
\Sthree{}-Att.-style index & $+7.52^*$ & \ci{+5.07}{+10.04} & 0.0032 & $+6.76^*$ & \ci{+4.29}{+9.22} & 0.0032 \\ 
\Sthree{}-Att.-style + BM25 & $+3.60^*$ & \ci{+1.59}{+5.73} & 0.0090 & $+2.84^*$ & \ci{+0.87}{+4.80} & 0.0302 \\ 
\ours{}, exact keys & $-0.96$ & \ci{-2.49}{+0.59} & 1.0000 & $-1.72$ & \ci{-3.55}{+0.07} & 0.5466 \\ 
\ours{}, \saec{} & $+1.27$ & \ci{-0.31}{+2.90} & 0.9335 & $+0.51$ & \ci{-1.31}{+2.40} & 1.0000 \\ 
\ours{}+BM25, \saec{} & $+1.58$ & \ci{-0.23}{+3.47} & 0.9335 & $+0.82$ & \ci{-0.49}{+2.16} & 1.0000 \\ 
\ours{}, \pqc{} & -- &  &  & $-0.76$ & \ci{-2.52}{+0.91} & 1.0000 \\ 
\ours{}+BM25, \pqc{} & $+0.76$ & \ci{-0.91}{+2.52} & 1.0000 & -- &  &  \\ 
\midrule
\multicolumn{7}{@{}l}{\textit{Mistral-7B-Instruct-v0.3}} \\
FullKV & $-3.38^*$ & \ci{-5.50}{-1.25} & 0.0082 & $-2.75$ & \ci{-4.97}{-0.57} & 0.1200 \\ 
BM25 & $-1.21$ & \ci{-3.51}{+1.08} & 1.0000 & $-0.57$ & \ci{-2.55}{+1.40} & 1.0000 \\ 
BM25, chunk filling & $-1.70$ & \ci{-3.97}{+0.60} & 0.9218 & $-1.06$ & \ci{-2.91}{+0.80} & 1.0000 \\ 
Dense RAG & $-3.83^*$ & \ci{-6.01}{-1.71} & 0.0082 & $-3.20^*$ & \ci{-5.46}{-1.16} & 0.0467 \\ 
BM25 + dense & $-2.02$ & \ci{-4.22}{+0.18} & 0.5736 & $-1.39$ & \ci{-3.42}{+0.59} & 1.0000 \\ 
BM25 + lex.\ density & $-1.68$ & \ci{-3.81}{+0.41} & 0.9218 & $-1.04$ & \ci{-2.77}{+0.62} & 1.0000 \\ 
BM25 + BGE-M3 & $-0.98$ & \ci{-3.04}{+1.21} & 1.0000 & $-0.34$ & \ci{-2.14}{+1.44} & 1.0000 \\ 
\Sthree{}-Att.-style index & $+7.44^*$ & \ci{+5.07}{+9.80} & 0.0030 & $+8.07^*$ & \ci{+5.62}{+10.43} & 0.0030 \\ 
\ours{}, exact keys & $-0.15$ & \ci{-1.41}{+1.16} & 1.0000 & $+0.49$ & \ci{-1.31}{+2.31} & 1.0000 \\ 
\ours{}, \saec{} & $-0.52$ & \ci{-1.83}{+0.81} & 1.0000 & $+0.12$ & \ci{-1.72}{+1.99} & 1.0000 \\ 
\ours{}+BM25, \saec{} & $-1.01$ & \ci{-2.98}{+0.92} & 1.0000 & $-0.38$ & \ci{-1.67}{+0.88} & 1.0000 \\ 
\ours{}, \pqc{} & -- &  &  & $+0.64$ & \ci{-1.21}{+2.51} & 1.0000 \\ 
\ours{}+BM25, \pqc{} & $-0.64$ & \ci{-2.51}{+1.21} & 1.0000 & -- &  &  \\ 
\midrule
\multicolumn{7}{@{}l}{\textit{Qwen2-7B-Instruct}} \\
FullKV & $-0.00$ & \ci{-2.09}{+1.97} & 1.0000 & $+0.76$ & \ci{-1.30}{+2.80} & 1.0000 \\ 
BM25 & $+0.40$ & \ci{-1.42}{+2.13} & 1.0000 & $+1.16$ & \ci{-0.53}{+2.81} & 1.0000 \\ 
BM25, chunk filling & $-0.37$ & \ci{-2.20}{+1.41} & 1.0000 & $+0.38$ & \ci{-1.29}{+2.08} & 1.0000 \\ 
Dense RAG & $-1.50$ & \ci{-3.45}{+0.44} & 1.0000 & $-0.75$ & \ci{-2.63}{+1.20} & 1.0000 \\ 
BM25 + dense & $-0.73$ & \ci{-2.55}{+1.03} & 1.0000 & $+0.03$ & \ci{-1.62}{+1.77} & 1.0000 \\ 
BM25 + lex.\ density & $-0.74$ & \ci{-2.70}{+1.22} & 1.0000 & $+0.02$ & \ci{-1.64}{+1.68} & 1.0000 \\ 
BM25 + BGE-M3 & $-0.56$ & \ci{-2.51}{+1.35} & 1.0000 & $+0.20$ & \ci{-1.46}{+1.82} & 1.0000 \\ 
\Sthree{}-Att.-style index & $+6.23^*$ & \ci{+3.73}{+8.59} & 0.0030 & $+6.99^*$ & \ci{+4.60}{+9.27} & 0.0030 \\ 
\ours{}, exact keys & $-0.99$ & \ci{-2.43}{+0.36} & 1.0000 & $-0.23$ & \ci{-1.87}{+1.46} & 1.0000 \\ 
\ours{}, \saec{} & $-0.22$ & \ci{-1.60}{+1.16} & 1.0000 & $+0.54$ & \ci{-1.09}{+2.19} & 1.0000 \\ 
\ours{}+BM25, \saec{} & $-0.86$ & \ci{-2.51}{+0.82} & 1.0000 & $-0.10$ & \ci{-1.43}{+1.27} & 1.0000 \\ 
\ours{}, \pqc{} & -- &  &  & $+0.76$ & \ci{-0.91}{+2.50} & 1.0000 \\ 
\ours{}+BM25, \pqc{} & $-0.76$ & \ci{-2.50}{+0.91} & 1.0000 & -- &  &  \\ 
\bottomrule
\end{tabular}}
\end{table}

\begin{table}[!htbp]
\centering
\caption{\textbf{Paired differences on LongBench QA, \ours{} with \saec{} codes} (\ours{} (\saec{}) minus X). The columns are as in Table~\ref{tab:lb-deltas-pq}.}
\label{tab:lb-deltas}
\scriptsize
\setlength{\tabcolsep}{3pt}
\resizebox{\ifdim\width>\linewidth\linewidth\else\width\fi}{!}{%
\begin{tabular}{@{}lrcc@{}}
\toprule
X & $\Delta$ & 95\% CI & Holm $p$ \\
\midrule
\multicolumn{4}{@{}l}{\textit{Llama-3.1-8B-Instruct}} \\
FullKV & $-4.31^*$ & \ci{-6.43}{-2.25} & 0.0032 \\ 
BM25 & $+0.78$ & \ci{-1.23}{+2.83} & 1.0000 \\ 
BM25, chunk filling & $-0.21$ & \ci{-2.13}{+1.79} & 1.0000 \\ 
Dense RAG & $-2.48$ & \ci{-4.54}{-0.40} & 0.2224 \\ 
BM25 + dense & $-0.41$ & \ci{-2.48}{+1.75} & 1.0000 \\ 
BM25 + lex.\ density & $+0.26$ & \ci{-1.69}{+2.24} & 1.0000 \\ 
BM25 + BGE-M3 & $-0.71$ & \ci{-2.81}{+1.41} & 1.0000 \\ 
\Sthree{}-Att.-style index & $+6.25^*$ & \ci{+3.85}{+8.71} & 0.0032 \\ 
\Sthree{}-Att.-style + BM25 & $+2.33$ & \ci{+0.26}{+4.51} & 0.3374 \\ 
\ours{}, exact keys & $-2.23^*$ & \ci{-3.65}{-0.87} & 0.0192 \\ 
\ours{}+BM25, \saec{} & $+0.31$ & \ci{-1.57}{+2.12} & 1.0000 \\ 
\ours{}, \pqc{} & $-1.27$ & \ci{-2.90}{+0.31} & 0.9798 \\ 
\ours{}+BM25, \pqc{} & $-0.51$ & \ci{-2.40}{+1.31} & 1.0000 \\ 
\midrule
\multicolumn{4}{@{}l}{\textit{Mistral-7B-Instruct-v0.3}} \\
FullKV & $-2.87$ & \ci{-4.97}{-0.82} & 0.0675 \\ 
BM25 & $-0.69$ & \ci{-2.96}{+1.57} & 1.0000 \\ 
BM25, chunk filling & $-1.18$ & \ci{-3.35}{+1.05} & 1.0000 \\ 
Dense RAG & $-3.31^*$ & \ci{-5.54}{-1.30} & 0.0302 \\ 
BM25 + dense & $-1.50$ & \ci{-3.75}{+0.60} & 1.0000 \\ 
BM25 + lex.\ density & $-1.16$ & \ci{-3.21}{+0.90} & 1.0000 \\ 
BM25 + BGE-M3 & $-0.46$ & \ci{-2.58}{+1.58} & 1.0000 \\ 
\Sthree{}-Att.-style index & $+7.96^*$ & \ci{+5.55}{+10.31} & 0.0030 \\ 
\ours{}, exact keys & $+0.37$ & \ci{-0.83}{+1.61} & 1.0000 \\ 
\ours{}+BM25, \saec{} & $-0.49$ & \ci{-2.43}{+1.40} & 1.0000 \\ 
\ours{}, \pqc{} & $+0.52$ & \ci{-0.81}{+1.83} & 1.0000 \\ 
\ours{}+BM25, \pqc{} & $-0.12$ & \ci{-1.99}{+1.72} & 1.0000 \\ 
\midrule
\multicolumn{4}{@{}l}{\textit{Qwen2-7B-Instruct}} \\
FullKV & $+0.22$ & \ci{-1.79}{+2.23} & 1.0000 \\ 
BM25 & $+0.61$ & \ci{-1.12}{+2.31} & 1.0000 \\ 
BM25, chunk filling & $-0.16$ & \ci{-1.96}{+1.60} & 1.0000 \\ 
Dense RAG & $-1.29$ & \ci{-3.33}{+0.66} & 1.0000 \\ 
BM25 + dense & $-0.51$ & \ci{-2.22}{+1.16} & 1.0000 \\ 
BM25 + lex.\ density & $-0.52$ & \ci{-2.28}{+1.29} & 1.0000 \\ 
BM25 + BGE-M3 & $-0.35$ & \ci{-2.32}{+1.59} & 1.0000 \\ 
\Sthree{}-Att.-style index & $+6.45^*$ & \ci{+4.10}{+8.73} & 0.0030 \\ 
\ours{}, exact keys & $-0.77$ & \ci{-2.00}{+0.43} & 1.0000 \\ 
\ours{}+BM25, \saec{} & $-0.64$ & \ci{-2.05}{+0.85} & 1.0000 \\ 
\ours{}, \pqc{} & $+0.22$ & \ci{-1.16}{+1.60} & 1.0000 \\ 
\ours{}+BM25, \pqc{} & $-0.54$ & \ci{-2.19}{+1.09} & 1.0000 \\ 
\bottomrule
\end{tabular}}
\end{table}


\section{\bench{} beyond the context window}
\label{app:beyond-full}

\begin{figure}[!htbp]
\centering
\includegraphics[width=\linewidth]{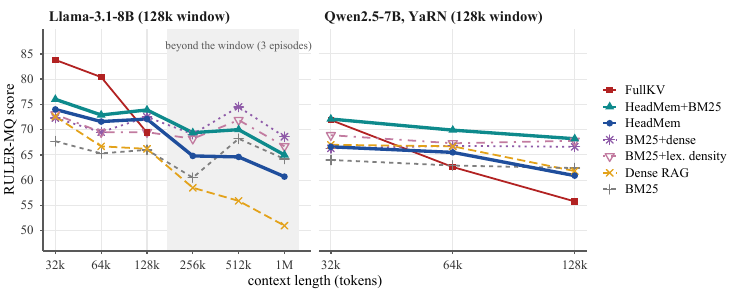}
\caption{\textbf{\bench{} score against context length} (macro score). Each family has 10 test episodes up to 128k and 3 beyond. Qwen2.5 FullKV uses the episodes of Table~\ref{tab:e2}. FullKV stops at the window. \ours{} and the text retrievers read $B{=}2{,}048$ tokens at every length.}
\label{fig:scaling}
\end{figure}

Table~\ref{tab:beyond-full} gives every family cell and paired difference at 256k, 512k and 1M tokens.
The setting is Llama-3.1-8B-Instruct with 3 test episodes per family and 120 questions per length.
With \pqc{} codes, \ours{} scores above dense RAG at every length (CIs exclude zero).
It is not significantly different from BM25 at any of the three lengths ($+4.25$, $-3.58$ and $-3.50$, with CIs that include zero).
\ours{}+BM25 significantly exceeds BM25 at 256k ($+8.83$, Holm $p=0.0140$) and scores above dense RAG at every length (CIs exclude zero).
It is not significantly different from BM25 + dense, the strongest text retriever, at 256k and 1M ($+0.25$ and $-3.54$).
\saec{} codes hold at 256k (66.8) but degrade at 1M (39.4, with variable tracking at 0.0 and multi-value NIAH at 26.0).
At this length, \pqc{} holds (60.7), and \pqc{} $-$ \saec{} is $+21.33$ (95\% CI [$+11.33$, $+33.83$]).
This is one reason \pqc{} is the default codec.
Figure~\ref{fig:scaling} plots the macro scores of Tables~\ref{tab:e2} and~\ref{tab:beyond-full} against the context length.

\begin{table}[!htbp]
\centering
\caption{\textbf{\bench{} beyond the window} (Llama-3.1-8B-Instruct, 3 test episodes per family and length, $B{=}2{,}048$). The top block gives the score per family, the macro score and the macro presence. The bottom block gives paired differences (first-named minus X) with 95\% CI and Holm $p$. $^*$ marks $p<0.05$.}
\label{tab:beyond-full}
\scriptsize
\setlength{\tabcolsep}{3.5pt}
\resizebox{\ifdim\width>\linewidth\linewidth\else\width\fi}{!}{%
\begin{tabular}{@{}lccccccc@{}}
\toprule
Method & NIAH-MK & NIAH-MV & QA-Hotpot & QA-SQuAD & VT & Macro & Presence \\
\midrule
\multicolumn{8}{@{}l}{\textit{256k tokens}} \\
BM25, same selector & 87.5 & 88.5 & 54.2 & 54.2 & 18.3 & 60.5 & 0.778 \\ 
BM25, chunk filling & 87.5 & 90.6 & 50.0 & 54.2 & 18.3 & 60.1 & 0.782 \\ 
Dense RAG + reranker & 70.8 & 71.9 & 58.3 & 66.7 & 25.0 & 58.5 & 0.725 \\ 
BM25 + dense & 87.5 & 90.6 & 58.3 & 66.7 & 42.5 & 69.1 & 0.866 \\ 
BM25 + lexical density & 100.0 & 95.8 & 58.3 & 66.7 & 20.0 & 68.2 & 0.830 \\ 
\ours{}, \saec{} & 100.0 & 95.8 & 50.0 & 66.7 & 21.7 & 66.8 & 0.811 \\ 
\ours{}+BM25, \saec{} & 95.8 & 92.7 & 62.5 & 62.5 & 25.0 & 67.7 & 0.832 \\ 
\ours{}, \pqc{} & 100.0 & 94.8 & 50.0 & 54.2 & 25.0 & 64.8 & 0.798 \\ 
\ours{}+BM25, \pqc{} & 100.0 & 92.7 & 58.3 & 70.8 & 25.0 & 69.4 & 0.840 \\ 
\midrule
\multicolumn{8}{@{}l}{\textit{512k tokens}} \\
BM25, same selector & 95.8 & 88.5 & 54.2 & 83.3 & 19.2 & 68.2 & 0.770 \\ 
BM25, chunk filling & 95.8 & 89.6 & 54.2 & 87.5 & 19.2 & 69.2 & 0.770 \\ 
Dense RAG + reranker & 70.8 & 52.1 & 50.0 & 83.3 & 23.3 & 55.9 & 0.646 \\ 
BM25 + dense & 95.8 & 90.6 & 58.3 & 83.3 & 44.2 & 74.5 & 0.895 \\ 
BM25 + lexical density & 100.0 & 97.9 & 50.0 & 91.7 & 20.0 & 71.9 & 0.798 \\ 
\ours{}, \saec{} & 100.0 & 95.8 & 50.0 & 75.0 & 0.8 & 64.3 & 0.718 \\ 
\ours{}+BM25, \saec{} & 100.0 & 95.8 & 50.0 & 87.5 & 20.0 & 70.7 & 0.748 \\ 
\ours{}, \pqc{} & 95.8 & 94.8 & 41.7 & 75.0 & 15.8 & 64.6 & 0.723 \\ 
\ours{}+BM25, \pqc{} & 100.0 & 96.9 & 50.0 & 83.3 & 20.0 & 70.0 & 0.782 \\ 
\midrule
\multicolumn{8}{@{}l}{\textit{1M tokens}} \\
BM25, same selector & 91.7 & 92.7 & 37.5 & 79.2 & 20.0 & 64.2 & 0.748 \\ 
BM25, chunk filling & 91.7 & 92.7 & 41.7 & 79.2 & 20.0 & 65.0 & 0.757 \\ 
Dense RAG + reranker & 66.7 & 63.5 & 37.5 & 70.8 & 16.7 & 51.0 & 0.651 \\ 
BM25 + dense & 91.7 & 93.8 & 45.8 & 70.8 & 40.8 & 68.6 & 0.858 \\ 
BM25 + lexical density & 100.0 & 96.9 & 41.7 & 75.0 & 20.0 & 66.7 & 0.773 \\ 
\ours{}, \saec{} & 66.7 & 26.0 & 37.5 & 66.7 & 0.0 & 39.4 & 0.469 \\ 
\ours{}+BM25, \saec{} & 91.7 & 80.2 & 37.5 & 75.0 & 20.0 & 60.9 & 0.725 \\ 
\ours{}, \pqc{} & 100.0 & 96.9 & 37.5 & 54.2 & 15.0 & 60.7 & 0.705 \\ 
\ours{}+BM25, \pqc{} & 95.8 & 96.9 & 41.7 & 70.8 & 20.0 & 65.0 & 0.765 \\ 
\bottomrule
\end{tabular}}

\vspace{6pt}
\resizebox{\ifdim\width>\linewidth\linewidth\else\width\fi}{!}{%
\begin{tabular}{@{}lccc@{}}
\toprule
$\Delta$ (95\% CI, Holm $p$) & 256k & 512k & 1M \\
\midrule
\multicolumn{4}{@{}l}{\textit{\ours{} (\pqc{}) $-$ X}} \\
BM25, same selector & $+4.25$ \ci{-0.75}{+9.25}, 1.0000 & $-3.58$ \ci{-7.25}{+0.25}, 0.7991 & $-3.50$ \ci{-6.83}{+0.25}, 1.0000 \\ 
BM25, chunk filling & $+4.67$ \ci{-0.58}{+9.67}, 1.0000 & $-4.62$ \ci{-10.38}{+1.12}, 0.7991 & $-4.33$ \ci{-8.08}{-0.58}, 1.0000 \\ 
Dense RAG + reranker & $+6.25$ \ci{+1.25}{+11.54}, 1.0000 & $+8.71$ \ci{+3.37}{+14.25}, 0.7991 & $+9.67$ \ci{+2.17}{+16.88}, 0.9014 \\ 
BM25 + dense & $-4.33$ \ci{-9.33}{+0.46}, 1.0000 & $-9.83$ \ci{-13.33}{-6.17}, 0.1416 & $-7.88$ \ci{-13.88}{-1.46}, 0.6923 \\ 
BM25 + lexical density & $-3.38$ \ci{-8.38}{+1.00}, 1.0000 & $-7.29^*$ \ci{-9.83}{-4.83}, 0.0128 & $-6.00$ \ci{-9.33}{-3.50}, 0.1016 \\ 
\ours{}+BM25, \pqc{} & $-4.58$ \ci{-7.71}{-1.88}, 0.9599 & $-5.42$ \ci{-9.37}{-1.21}, 0.1895 & $-4.33$ \ci{-10.58}{+1.29}, 1.0000 \\ 
\ours{}, \saec{} & $-2.04$ \ci{-6.83}{+1.00}, 1.0000 & $+0.29$ \ci{-2.00}{+2.62}, 0.9097 & $+21.33$ \ci{+11.33}{+33.83}, 0.3745 \\ 
\ours{}+BM25, \saec{} & $-2.92$ \ci{-7.08}{+1.87}, 1.0000 & $-6.04$ \ci{-9.12}{-3.04}, 0.0532 & $-0.17$ \ci{-5.17}{+4.83}, 1.0000 \\ 
\midrule
\multicolumn{4}{@{}l}{\textit{\ours{}+BM25 (\pqc{}) $-$ X}} \\
BM25, same selector & $+8.83^*$ \ci{+5.33}{+12.37}, 0.0140 & $+1.83$ \ci{-2.54}{+6.17}, 1.0000 & $+0.83$ \ci{-5.21}{+6.67}, 1.0000 \\ 
BM25, chunk filling & $+9.25^*$ \ci{+5.92}{+12.58}, 0.0128 & $+0.79$ \ci{-5.42}{+6.63}, 1.0000 & $+0.00$ \ci{-6.46}{+6.04}, 1.0000 \\ 
Dense RAG + reranker & $+10.83$ \ci{+6.67}{+14.79}, 0.0894 & $+14.12$ \ci{+8.29}{+20.12}, 0.2984 & $+14.00$ \ci{+7.58}{+20.67}, 0.1134 \\ 
BM25 + dense & $+0.25$ \ci{-3.50}{+4.00}, 1.0000 & $-4.42$ \ci{-8.42}{-0.42}, 1.0000 & $-3.54$ \ci{-11.00}{+3.29}, 1.0000 \\ 
BM25 + lexical density & $+1.21$ \ci{-1.92}{+3.29}, 1.0000 & $-1.88$ \ci{-5.42}{+1.46}, 1.0000 & $-1.67$ \ci{-6.04}{+2.50}, 1.0000 \\ 
\ours{}, \pqc{} & $+4.58$ \ci{+1.88}{+7.71}, 0.5999 & $+5.42$ \ci{+1.21}{+9.37}, 0.2984 & $+4.33$ \ci{-1.29}{+10.58}, 1.0000 \\ 
\ours{}, \saec{} & $+2.54$ \ci{-1.29}{+6.29}, 1.0000 & $+5.71$ \ci{+1.33}{+9.83}, 0.5027 & $+25.67$ \ci{+14.42}{+39.21}, 0.0872 \\ 
\ours{}+BM25, \saec{} & $+1.67$ \ci{-1.67}{+5.00}, 1.0000 & $-0.62$ \ci{-2.29}{+0.42}, 1.0000 & $+4.17$ \ci{-1.67}{+10.00}, 1.0000 \\ 
\midrule
\multicolumn{4}{@{}l}{\textit{\ours{} (\saec{}) $-$ X}} \\
BM25, same selector & $+6.29$ \ci{+1.29}{+11.29}, 0.5627 & $-3.88$ \ci{-6.50}{-1.17}, 0.6815 & $-24.83^*$ \ci{-35.04}{-16.29}, 0.0246 \\ 
BM25, chunk filling & $+6.71$ \ci{+1.33}{+11.71}, 0.4552 & $-4.92$ \ci{-9.58}{-0.46}, 0.6815 & $-25.67^*$ \ci{-35.88}{-17.12}, 0.0168 \\ 
Dense RAG + reranker & $+8.29$ \ci{+3.29}{+13.46}, 0.7481 & $+8.42$ \ci{+3.83}{+13.83}, 0.6815 & $-11.67$ \ci{-23.54}{-0.62}, 0.1336 \\ 
BM25 + dense & $-2.29$ \ci{-7.46}{+2.67}, 1.0000 & $-10.12$ \ci{-12.67}{-7.38}, 0.4044 & $-29.21^*$ \ci{-40.67}{-18.58}, 0.0246 \\ 
BM25 + lexical density & $-1.33$ \ci{-4.38}{+1.00}, 1.0000 & $-7.58$ \ci{-9.25}{-6.33}, 0.1216 & $-27.33^*$ \ci{-39.83}{-17.54}, 0.0168 \\ 
\ours{}, \pqc{} & $+2.04$ \ci{-1.00}{+6.83}, 1.0000 & $-0.29$ \ci{-2.62}{+2.00}, 0.9097 & $-21.33$ \ci{-33.83}{-11.33}, 0.1070 \\ 
\ours{}+BM25, \pqc{} & $-2.54$ \ci{-6.29}{+1.29}, 1.0000 & $-5.71$ \ci{-9.83}{-1.33}, 0.4190 & $-25.67^*$ \ci{-39.21}{-14.42}, 0.0327 \\ 
\ours{}+BM25, \saec{} & $-0.88$ \ci{-4.83}{+2.92}, 1.0000 & $-6.33$ \ci{-9.71}{-3.00}, 0.3339 & $-21.50^*$ \ci{-32.12}{-11.71}, 0.0246 \\ 
\bottomrule
\end{tabular}}
\end{table}


\section{NoLiMa}
\label{app:nolima}

Table~\ref{tab:real}a summarizes these results in the main text.
Appendix~\ref{app:protocol-nolima} gives the protocol (Llama-3.1-8B-Instruct, 290 samples per length at 8k, 32k and 128k).
At 128k, FullKV middle-truncates each context to the window, which keeps the needle in 0.800 of the samples.
Table~\ref{tab:nolima-full} gives scores, needle presence and every paired difference.
\ours{} significantly exceeds BM25 at every length and dense RAG at 8k (Holm $p\le 0.0045$).
It is not significantly different from dense RAG at 32k and 128k.
It is not significantly different from the full cache at 8k and 32k ($+3.06$, CI [$-2.33$, $+8.44$], and $-0.33$, CI [$-5.22$, $+4.61$]).
It scores above the full cache at 128k ($+3.17$, CI [$+0.39$, $+6.00$], Holm $p=0.0600$), where the full cache scores 1.4.
BM25 fusion lowers \ours{} at every length (Holm-significant).
BM25 alone rarely places the needle into its selection (presence 0.162, 0.000 and 0.007, against 0.731, 0.355 and 0.079 for \ours{}).
\ourstwo{} keeps the top two heads (32\,B per token).
It scores 16.6, 5.9 and 1.7 at 8k, 32k and 128k (needle presence 0.366, 0.114 and 0.017).
With BM25, it scores 12.1, 6.6 and 0.0. \ours{} with all 16 heads and no fusion scores 34.8, 17.9 and 5.5.
Associative retrieval thus needs more than the two layer-2 heads. 

\begin{table}[!htbp]
\centering
\caption{\textbf{NoLiMa} (Llama-3.1-8B-Instruct, 290 samples per length, $B{=}2{,}048$). Each cell gives the score ($\times100$) / needle presence of the selection. FullKV reads the whole context. At 128k it is middle-truncated to the window, with presence 0.800. Paired differences are \ours{} minus X. They are averaged within each needle entry and then over entries, so they are not the difference of the per-sample scores above. Each difference has a 95\% bootstrap CI and a Holm $p$ within each length. $^*$ marks $p<0.05$. \nr{} means not run.}
\label{tab:nolima-full}
\scriptsize
\setlength{\tabcolsep}{4pt}
\resizebox{\ifdim\width>\linewidth\linewidth\else\width\fi}{!}{%
\begin{tabular}{@{}lccc@{}}
\toprule
Method & 8k & 32k & 128k \\
\midrule
FullKV & 31.7 / 1.000 & 15.5 / 1.000 & 1.4 / 0.800 \\ 
BM25, same selector & 9.0 / 0.162 & 0.3 / 0.000 & 0.3 / 0.007 \\ 
Dense RAG + reranker & 20.7 / 0.362 & 9.3 / 0.155 & 4.8 / 0.062 \\ 
\ours{} (\pqc{}) & 34.8 / 0.731 & 17.9 / 0.355 & 5.5 / 0.079 \\ 
\ours{}+BM25 & 21.4 / 0.455 & 8.3 / 0.134 & 0.0 / 0.010 \\ 
\midrule
$\Delta$ \ours{} $-$ BM25 & $+24.22^*$ \ci{+18.89}{+29.67}, 0.0010 & $+14.61^*$ \ci{+11.17}{+18.61}, 0.0010 & $+4.11^*$ \ci{+1.72}{+6.72}, 0.0015 \\ 
$\Delta$ \ours{} $-$ dense RAG & $+10.61^*$ \ci{+4.67}{+16.61}, 0.0045 & $+3.44$ \ci{-1.50}{+8.61}, 0.4784 & $-1.94$ \ci{-5.83}{+1.67}, 0.3527 \\ 
$\Delta$ \ours{} $-$ FullKV & $+3.06$ \ci{-2.33}{+8.44}, 0.2962 & $-0.33$ \ci{-5.22}{+4.61}, 0.9178 & $+3.17$ \ci{+0.39}{+6.00}, 0.0600 \\ 
$\Delta$ \ours{} $-$ \ours{}+BM25 & $+12.11^*$ \ci{+7.89}{+16.33}, 0.0010 & $+8.00^*$ \ci{+4.67}{+11.67}, 0.0010 & $+4.61^*$ \ci{+2.44}{+7.06}, 0.0010 \\ 
\bottomrule
\end{tabular}}
\end{table}


\FloatBarrier
\section{Natural-text QA families of \bench{}}
\label{app:qa-subset}

The SQuAD and HotpotQA families of \bench{} embed real paragraphs of the two datasets among distractor paragraphs.
They ask question-specific questions (Appendix~\ref{app:protocol-mq}).
The three other families use synthetic needles.
Table~\ref{tab:qa-subset} reports the macro score of the two QA families, which is their mean, for every model.
It uses the test episodes of the main tables.
Llama-3.1-8B and Qwen2.5-7B have 10 episodes per family and length, with 160 questions per length.
Mistral-7B and Qwen2-7B have 6, with 96 questions per length.
Table~\ref{tab:qa-subset-delta} gives the paired differences, with episodes bootstrapped within family.
Its $p$ is the two-sided sign-flip $p$ without multiplicity adjustment.

\begin{table}[!htbp]
\centering
\caption{\textbf{\bench{}, natural-text QA families only} (macro score over SQuAD and HotpotQA, $\times100$). Every method but FullKV reads $B{=}2{,}048$ tokens. -- marks a cell that was not run, or one beyond the window for FullKV.}
\label{tab:qa-subset}
\scriptsize
\setlength{\tabcolsep}{3.2pt}
\resizebox{\ifdim\width>\linewidth\linewidth\else\width\fi}{!}{%
\begin{tabular}{@{}lcccccccccccc@{}}
\toprule
 & \multicolumn{3}{c}{Llama-3.1-8B} & \multicolumn{3}{c}{Qwen2.5-7B} & \multicolumn{3}{c}{Mistral-7B} & \multicolumn{3}{c}{Qwen2-7B} \\
\cmidrule(lr){2-4}\cmidrule(lr){5-7}\cmidrule(lr){8-10}\cmidrule(l){11-13}
Method & 32k & 64k & 128k & 32k & 64k & 128k & 32k & 64k & 128k & 32k & 64k & 128k \\
\midrule
FullKV & 73.8 & 70.0 & 60.0 & 63.1 & 55.6 & 47.5 & 64.6 & 21.9 & 20.8 & 64.6 & 35.4 & 12.5 \\
BM25, same selector & 67.5 & 65.0 & 63.1 & 61.2 & 62.5 & 63.1 & 60.4 & 63.5 & 61.5 & 57.3 & 64.6 & 60.4 \\
Dense RAG + reranker & 78.8 & 70.6 & 73.1 & 65.6 & 70.6 & 66.9 & 63.5 & 57.3 & 64.6 & 71.9 & 67.7 & 64.6 \\
BM25 + dense & 69.4 & 65.6 & 65.0 & 58.8 & 61.9 & 60.6 & 61.5 & 57.3 & 63.5 & 60.4 & 65.6 & 60.4 \\
BM25 + lexical density & 74.4 & 68.1 & 66.2 & 66.9 & 66.2 & 66.9 & 61.5 & 61.5 & 63.5 & 66.7 & 66.7 & 60.4 \\
REFORM-style & 72.5 & 62.5 & 63.1 & 67.5 & 61.9 & 65.6 & -- & -- & -- & -- & -- & -- \\
\Sthree{}-Hybrid & 64.4 & 66.2 & 64.4 & 60.6 & 60.6 & 59.4 & -- & -- & -- & -- & -- & -- \\
\ourstwo{} & 73.1 & 63.8 & 66.2 & 13.8 & 13.1 & 10.6 & -- & -- & -- & -- & -- & -- \\
\ours{} & 70.6 & 65.6 & 65.0 & 62.5 & 59.4 & 53.1 & 58.3 & 28.1 & 30.2 & 60.4 & 61.5 & 60.4 \\
\ours{}+BM25 & 74.4 & 68.1 & 69.4 & 70.0 & 66.9 & 66.2 & 60.4 & 59.4 & 60.4 & 62.5 & 67.7 & 57.3 \\
\bottomrule
\end{tabular}}
\end{table}

\begin{table}[!htbp]
\centering
\caption{\textbf{Paired differences on the natural-text QA families.} Each cell gives the first-named minus the second method, with a 95\% bootstrap CI and the unadjusted two-sided sign-flip $p$.}
\label{tab:qa-subset-delta}
\scriptsize
\setlength{\tabcolsep}{3pt}
\resizebox{\ifdim\width>\linewidth\linewidth\else\width\fi}{!}{%
\begin{tabular}{@{}llccc@{}}
\toprule
Model & Contrast & 32k & 64k & 128k \\
\midrule
Llama-3.1-8B & \ours{}+BM25 $-$ Dense RAG + reranker & $-4.38$ \ci{-10.62}{+1.88}, 0.3327 & $-2.50$ \ci{-5.62}{+0.62}, 0.2872 & $-3.75$ \ci{-9.38}{+2.50}, 0.3439 \\
 & \ours{}+BM25 $-$ BM25, same selector & $+6.88$ \ci{+0.62}{+13.12}, 0.0815 & $+3.12$ \ci{-1.25}{+6.88}, 0.2672 & $+6.25$ \ci{+1.88}{+10.62}, 0.0355 \\
 & \ours{}+BM25 $-$ FullKV & $+0.62$ \ci{-3.75}{+5.00}, 1.0000 & $-1.88$ \ci{-7.50}{+3.12}, 0.6563 & $+9.38$ \ci{+2.50}{+16.25}, 0.0235 \\
 & \ours{} $-$ Dense RAG + reranker & $-8.12$ \ci{-14.38}{-1.88}, 0.0660 & $-5.00$ \ci{-9.38}{-0.62}, 0.0957 & $-8.12$ \ci{-15.00}{-1.88}, 0.0365 \\
 & \ours{} $-$ BM25, same selector & $+3.12$ \ci{-3.75}{+9.38}, 0.4846 & $+0.62$ \ci{-3.75}{+5.00}, 1.0000 & $+1.88$ \ci{-3.75}{+7.50}, 0.6653 \\
\midrule
Qwen2.5-7B & \ours{}+BM25 $-$ Dense RAG + reranker & $+4.38$ \ci{-3.12}{+11.25}, 0.3489 & $-3.75$ \ci{-8.12}{+0.00}, 0.1602 & $-0.62$ \ci{-8.75}{+6.88}, 1.0000 \\
 & \ours{}+BM25 $-$ BM25, same selector & $+8.75$ \ci{+0.00}{+17.50}, 0.0990 & $+4.38$ \ci{-0.62}{+10.00}, 0.2192 & $+3.12$ \ci{-3.75}{+10.00}, 0.5019 \\
 & \ours{}+BM25 $-$ FullKV & $+6.88$ \ci{-0.62}{+14.38}, 0.1475 & $+11.25$ \ci{+5.00}{+16.88}, 0.0067 & $+18.75$ \ci{+10.62}{+26.88}, 0.0017 \\
 & \ours{} $-$ Dense RAG + reranker & $-3.12$ \ci{-9.38}{+3.12}, 0.4531 & $-11.25$ \ci{-16.88}{-6.88}, 0.0010 & $-13.75$ \ci{-23.12}{-5.62}, 0.0117 \\
 & \ours{} $-$ BM25, same selector & $+1.25$ \ci{-8.75}{+11.88}, 0.9030 & $-3.12$ \ci{-8.75}{+1.88}, 0.4051 & $-10.00$ \ci{-17.50}{-2.50}, 0.0322 \\
\midrule
Mistral-7B & \ours{}+BM25 $-$ Dense RAG + reranker & $-3.12$ \ci{-10.42}{+5.21}, 0.6498 & $+2.08$ \ci{-6.25}{+9.38}, 0.7996 & $-4.17$ \ci{-11.46}{+3.12}, 0.4651 \\
 & \ours{}+BM25 $-$ BM25, same selector & $+0.00$ \ci{-5.21}{+5.21}, 1.0000 & $-4.17$ \ci{-11.46}{+3.12}, 0.4709 & $-1.04$ \ci{-11.46}{+10.42}, 1.0000 \\
 & \ours{}+BM25 $-$ FullKV & $-4.17$ \ci{-11.46}{+3.12}, 0.4754 & $+37.50$ \ci{+26.04}{+51.04}, 0.0020 & $+39.58$ \ci{+28.12}{+51.04}, 0.0020 \\
 & \ours{} $-$ Dense RAG + reranker & $-5.21$ \ci{-13.54}{+4.17}, 0.4231 & $-29.17$ \ci{-39.58}{-18.75}, 0.0012 & $-34.38$ \ci{-40.62}{-28.12}, 0.0010 \\
 & \ours{} $-$ BM25, same selector & $-2.08$ \ci{-7.29}{+3.12}, 0.7646 & $-35.42$ \ci{-46.88}{-23.96}, 0.0020 & $-31.25$ \ci{-39.58}{-21.88}, 0.0012 \\
\midrule
Qwen2-7B & \ours{}+BM25 $-$ Dense RAG + reranker & $-9.38$ \ci{-21.88}{+3.12}, 0.2312 & $+0.00$ \ci{-6.25}{+5.21}, 1.0000 & $-7.29$ \ci{-13.54}{-1.04}, 0.1287 \\
 & \ours{}+BM25 $-$ BM25, same selector & $+5.21$ \ci{-6.25}{+18.75}, 0.7186 & $+3.12$ \ci{-2.08}{+8.33}, 0.5014 & $-3.12$ \ci{-9.38}{+3.12}, 0.5854 \\
 & \ours{}+BM25 $-$ FullKV & $-2.08$ \ci{-13.54}{+10.42}, 0.8755 & $+32.29$ \ci{+20.83}{+42.71}, 0.0037 & $+44.79$ \ci{+36.46}{+53.12}, 0.0010 \\
 & \ours{} $-$ Dense RAG + reranker & $-11.46$ \ci{-23.96}{+1.04}, 0.1607 & $-6.25$ \ci{-15.62}{+2.08}, 0.3624 & $-4.17$ \ci{-9.38}{+2.08}, 0.3534 \\
 & \ours{} $-$ BM25, same selector & $+3.12$ \ci{-9.38}{+18.75}, 0.8445 & $-3.12$ \ci{-11.46}{+6.25}, 0.6763 & $+0.00$ \ci{-8.33}{+8.33}, 1.0000 \\
\bottomrule
\end{tabular}}
\end{table}


\section{LongMemEval-S}
\label{app:lme}

\begin{figure}[!htbp]
\centering
\includegraphics[width=\linewidth]{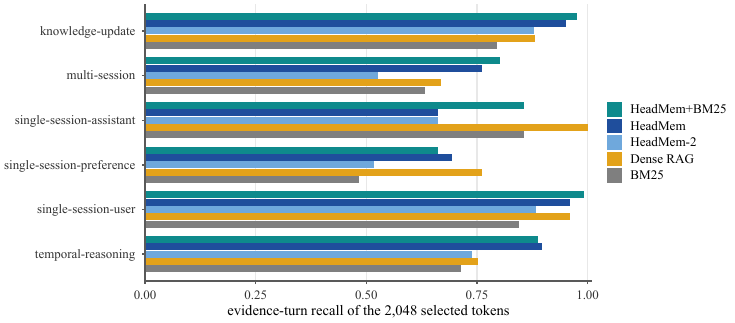}
\caption{\textbf{LongMemEval-S evidence-turn recall by question type} (Llama-3.1-8B-Instruct, $B{=}2{,}048$). Within each group the bars are, from top to bottom, \ours{}+BM25, \ours{}, \ourstwo{}, dense RAG and BM25. Table~\ref{tab:lme-types} gives the values.}
\label{fig:lme-types}
\end{figure}

Table~\ref{tab:lme-full} gives the LongMemEval-S results of Section~\ref{sec:exp-nolima}.
Appendix~\ref{app:protocol-lme} gives the protocol.
FullKV reads the whole history, so its recall is 100 by construction.
Figure~\ref{fig:lme-types} breaks the evidence-turn recall down by question type.

\begin{table}[!htbp]
\centering
\caption{\textbf{LongMemEval-S} (Llama-3.1-8B-Instruct, 500 questions). Every method but FullKV reads $B{=}2{,}048$ history tokens. Recall is the share of evidence turns (sessions) with at least half of an evidence turn's tokens selected. Answer contained means that the normalized gold answer is contained in the prediction (470 non-abstention questions). All scores are $\times100$.}
\label{tab:lme-full}
\scriptsize
\setlength{\tabcolsep}{4pt}
\begin{tabular}{@{}lcccc@{}}
\toprule
Method & evidence-turn recall & evidence-session recall & answer contained & questions with evidence \\
\midrule
FullKV & 100.0 & 100.0 & 34.0 & 479 \\
BM25, same selector & 72.5 & 71.5 & 38.7 & 479 \\
Dense RAG + reranker & 80.8 & 79.1 & 45.3 & 479 \\
\ourstwo{} & 70.1 & 69.5 & 39.1 & 479 \\
\ours{} & 83.8 & 81.9 & 43.0 & 479 \\
\ours{}+BM25 & 87.5 & 85.4 & 43.2 & 479 \\
\bottomrule
\end{tabular}
\end{table}

\begin{table}[!htbp]
\centering
\caption{\textbf{LongMemEval-S evidence-turn recall by question type} ($\times100$).}
\label{tab:lme-types}
\scriptsize
\setlength{\tabcolsep}{4pt}
\begin{tabular}{@{}lccccc@{}}
\toprule
Question type & BM25, same selector & Dense RAG + reranker & \ourstwo{} & \ours{} & \ours{}+BM25 \\
\midrule
knowledge-update & 79.6 & 88.2 & 88.0 & 95.1 & 97.5 \\
multi-session & 63.3 & 66.9 & 52.7 & 76.2 & 80.2 \\
single-session-assistant & 85.7 & 100.0 & 66.1 & 66.1 & 85.7 \\
single-session-preference & 48.3 & 76.1 & 51.7 & 69.4 & 66.1 \\
single-session-user & 84.4 & 96.1 & 88.3 & 96.1 & 99.2 \\
temporal-reasoning & 71.5 & 75.3 & 73.9 & 89.8 & 88.9 \\
\bottomrule
\end{tabular}
\end{table}

\begin{table}[!htbp]
\centering
\caption{\textbf{LongMemEval-S paired differences} (first-named minus second method, $\times100$). The bootstrap resamples questions. The Holm adjustment runs within each method and metric.}
\label{tab:lme-delta}
\scriptsize
\setlength{\tabcolsep}{4pt}
\begin{tabular}{@{}llccc@{}}
\toprule
Metric & Contrast & $\Delta$ & 95\% CI & Holm $p$ \\
\midrule
recall & \ours{} $-$ BM25, same selector & $+11.33^*$ & [$+7.52$, $+15.13$] & 0.0007 \\
recall & \ours{} $-$ Dense RAG + reranker & $+3.08$ & [$-0.42$, $+6.60$] & 0.0907 \\
recall & \ours{} $-$ \ours{}+BM25 & $-3.63^*$ & [$-6.05$, $-1.27$] & 0.0040 \\
recall & \ours{}+BM25 $-$ BM25, same selector & $+14.96^*$ & [$+12.09$, $+17.86$] & 0.0007 \\
recall & \ours{}+BM25 $-$ Dense RAG + reranker & $+6.71^*$ & [$+3.87$, $+9.53$] & 0.0007 \\
recall & \ours{}+BM25 $-$ \ours{} & $+3.63^*$ & [$+1.27$, $+6.05$] & 0.0020 \\
recall & \ourstwo{} $-$ BM25, same selector & $-2.43$ & [$-6.02$, $+1.18$] & 0.2012 \\
recall & \ourstwo{} $-$ Dense RAG + reranker & $-10.68^*$ & [$-14.23$, $-7.16$] & 0.0010 \\
recall & \ourstwo{} $-$ \ours{} & $-13.76^*$ & [$-16.67$, $-10.92$] & 0.0010 \\
recall & \ourstwo{} $-$ \ours{}+BM25 & $-17.39^*$ & [$-20.35$, $-14.59$] & 0.0010 \\
answer & \ours{} $-$ BM25, same selector & $+4.26$ & [$+1.06$, $+7.66$] & 0.0562 \\
answer & \ours{} $-$ Dense RAG + reranker & $-2.34$ & [$-5.53$, $+0.85$] & 0.4194 \\
answer & \ours{} $-$ \ours{}+BM25 & $-0.21$ & [$-3.19$, $+2.77$] & 1.0000 \\
answer & \ours{} $-$ FullKV & $+8.94^*$ & [$+4.68$, $+13.19$] & 0.0010 \\
answer & \ours{}+BM25 $-$ BM25, same selector & $+4.47^*$ & [$+1.28$, $+7.45$] & 0.0202 \\
answer & \ours{}+BM25 $-$ Dense RAG + reranker & $-2.13$ & [$-5.11$, $+0.64$] & 0.3764 \\
answer & \ours{}+BM25 $-$ \ours{} & $+0.21$ & [$-2.77$, $+3.19$] & 1.0000 \\
answer & \ours{}+BM25 $-$ FullKV & $+9.15^*$ & [$+4.89$, $+13.19$] & 0.0010 \\
answer & \ourstwo{} $-$ BM25, same selector & $+0.43$ & [$-2.98$, $+3.83$] & 0.8965 \\
answer & \ourstwo{} $-$ Dense RAG + reranker & $-6.17^*$ & [$-9.36$, $-2.98$] & 0.0050 \\
answer & \ourstwo{} $-$ \ours{} & $-3.83$ & [$-6.81$, $-0.85$] & 0.0680 \\
answer & \ourstwo{} $-$ \ours{}+BM25 & $-4.04$ & [$-7.23$, $-0.85$] & 0.0680 \\
answer & \ourstwo{} $-$ FullKV & $+5.11$ & [$+1.06$, $+9.15$] & 0.0680 \\
\bottomrule
\end{tabular}
\end{table}


\section{InfiniteBench book QA}
\label{app:infbench}

Appendix~\ref{app:protocol-infbench} gives the protocol.
The model is Llama-3.1-8B-Instruct, and the tasks are En.QA (351 questions) and En.MC (229 questions) on 86 whole books.
Retrieval runs at $B{=}2{,}048$ and $B{=}8{,}192$.
FullKV reads the book middle-truncated to 130{,}560 tokens.
Table~\ref{tab:infbench} gives the results.
These questions need global reading of a whole book.
The middle-truncated full cache leads on En.QA (F1 31.70).
Dense semantic retrieval leads among the retrieval methods on both tasks.
\ours{} is not significantly different from BM25 on either task at either budget (CIs include zero).
At $B{=}2{,}048$ on En.MC, \ours{}+BM25 scores above BM25 ($+5.24$, 95\% CI [$+0.44$, $+10.04$]).
It is not significantly different there from the truncated full cache (69.87 vs.\ 64.63) or from dense RAG.

\begin{table}[!htbp]
\centering
\caption{\textbf{InfiniteBench} (Llama-3.1-8B-Instruct, scores $\times100$). En.QA has 351 questions and uses token F1. En.MC has 229 questions and uses option match. Each cell gives the score / answer presence of the selection. FullKV middle-truncates the book to 130{,}560 tokens. Paired differences are per question, with 95\% bootstrap CI and Holm $p$ within each (task, budget). $^*$ marks $p<0.05$.}
\label{tab:infbench}
\scriptsize
\setlength{\tabcolsep}{3.5pt}
\resizebox{\ifdim\width>\linewidth\linewidth\else\width\fi}{!}{%
\begin{tabular}{@{}lcccc@{}}
\toprule
& \multicolumn{2}{c}{En.QA} & \multicolumn{2}{c}{En.MC} \\
\cmidrule(lr){2-3}\cmidrule(l){4-5}
Method & $B{=}2{,}048$ & $B{=}8{,}192$ & $B{=}2{,}048$ & $B{=}8{,}192$ \\
\midrule
FullKV (truncated) & \multicolumn{2}{c}{31.70 / 0.584} & \multicolumn{2}{c}{64.63 / 0.428} \\ 
BM25, same selector & 13.73 / 0.345 & 21.51 / 0.493 & 64.63 / 0.397 & 72.05 / 0.432 \\ 
Dense RAG + reranker & 17.15 / 0.427 & 25.07 / 0.513 & 73.80 / 0.345 & 76.42 / 0.397 \\ 
\ours{} (\pqc{}) & 13.88 / 0.348 & 20.40 / 0.481 & 62.88 / 0.358 & 67.69 / 0.410 \\ 
\ours{}+BM25 & 14.33 / 0.379 & 21.62 / 0.504 & 69.87 / 0.397 & 69.43 / 0.419 \\ 
\midrule
$\Delta$ \ours{} $-$ BM25 & $+0.15$ \ci{-1.88}{+2.25} & $-1.11$ \ci{-4.00}{+1.73} & $-1.75$ \ci{-8.30}{+4.80} & $-4.37$ \ci{-9.61}{+0.44} \\ 
$\Delta$ \ours{} $-$ dense RAG & $-3.27^*$ \ci{-5.67}{-0.84} & $-4.67^*$ \ci{-7.22}{-2.25} & $-10.92^*$ \ci{-17.90}{-3.93} & $-8.73^*$ \ci{-14.85}{-3.06} \\ 
$\Delta$ \ours{} $-$ FullKV & $-17.82^*$ \ci{-21.62}{-13.97} & $-11.30^*$ \ci{-15.02}{-7.57} & $-1.75$ \ci{-9.17}{+5.24} & $+3.06$ \ci{-3.49}{+9.61} \\ 
$\Delta$ \ours{} $-$ \ours{}+BM25 & $-0.44$ \ci{-2.38}{+1.51} & $-1.22$ \ci{-3.59}{+1.16} & $-6.99$ \ci{-13.10}{-0.87} & $-1.75$ \ci{-6.99}{+3.49} \\ 
$\Delta$ \ours{}+BM25 $-$ BM25 & $+0.60$ \ci{-1.42}{+2.73} & $+0.11$ \ci{-2.40}{+2.64} & $+5.24$ \ci{+0.44}{+10.04} & $-2.62$ \ci{-7.42}{+2.18} \\ 
$\Delta$ \ours{}+BM25 $-$ dense RAG & $-2.82$ \ci{-5.18}{-0.49} & $-3.45^*$ \ci{-5.78}{-1.29} & $-3.93$ \ci{-10.04}{+2.18} & $-6.99^*$ \ci{-12.23}{-2.18} \\ 
$\Delta$ \ours{}+BM25 $-$ FullKV & $-17.37^*$ \ci{-21.11}{-13.60} & $-10.08^*$ \ci{-13.77}{-6.50} & $+5.24$ \ci{-2.18}{+12.23} & $+4.80$ \ci{-1.31}{+11.35} \\ 
\bottomrule
\end{tabular}}
\end{table}


\FloatBarrier
\input{v3_sections/app_stack}

\FloatBarrier
\input{v3_sections/app_limits}


\FloatBarrier
\section{Reading budget and index chunk size}
\label{app:sweeps}

Both sweeps use Llama-3.1-8B-Instruct \bench{} test episodes 1--6 per family (240 questions per length).
With 10 episodes the default \ours{} scores differ (Table~\ref{tab:e2}), so the two sets are not mixed.
Table~\ref{tab:budget-full} varies the number $B$ of re-read tokens for six methods.
No paired CIs were computed for this sweep.
\ours{} is above BM25 at every budget and length.
At 32k, \ours{}+BM25 is highest at every budget, from 71.5 at $B{=}512$ to 81.9 at $B{=}4{,}096$.
At 128k, \ours{} rises from 63.3 at $B{=}512$ to 75.6 at $B{=}4{,}096$.
Table~\ref{tab:chunk} varies the index chunk size $B_c$ with \pqc{} codes.
At 32k every change stays within 2.54 points of the default $B_c{=}4{,}096$, and $B_c{=}2{,}048$ scores above it ($+2.54$~\ci{+0.19}{+5.10}).
At 128k only $B_c{=}1{,}024$ scores below the default ($-3.83$~\ci{-6.25}{-1.58}), and the other sizes stay within 0.44 points.

\begin{figure}[!htbp]
\centering
\includegraphics[width=\linewidth]{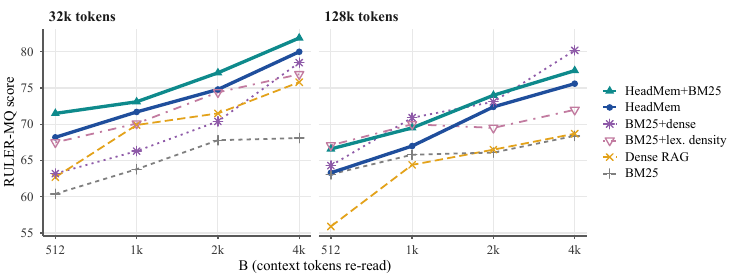}
\caption{\textbf{Reading budget} (\bench{}, Llama-3.1-8B-Instruct, test episodes 1--6). Table~\ref{tab:budget-full} gives the values.}
\label{fig:budget}
\end{figure}

\begin{table}[!htbp]
\centering
\caption{\textbf{Reading budget} (Llama-3.1-8B-Instruct, \bench{} test episodes 1--6, 240 questions per length). Each cell gives the macro score / macro answer presence for $B$ re-read context tokens.}
\label{tab:budget-full}
\scriptsize
\setlength{\tabcolsep}{4pt}
\resizebox{\ifdim\width>\linewidth\linewidth\else\width\fi}{!}{%
\begin{tabular}{@{}llcccc@{}}
\toprule
Length & Method & $B{=}512$ & $B{=}1{,}024$ & $B{=}2{,}048$ & $B{=}4{,}096$ \\
\midrule
32k & \ours{} (\pqc{}) & 68.2 / 0.763 & 71.7 / 0.807 & 74.8 / 0.835 & 80.0 / 0.906 \\ 
 & \ours{}+BM25 & 71.5 / 0.783 & 73.1 / 0.813 & 77.1 / 0.853 & 81.9 / 0.917 \\ 
 & BM25, same selector & 60.4 / 0.661 & 63.8 / 0.698 & 67.8 / 0.748 & 68.1 / 0.773 \\ 
 & Dense RAG + reranker & 62.7 / 0.687 & 69.9 / 0.772 & 71.5 / 0.814 & 75.8 / 0.879 \\ 
 & BM25 + dense & 63.2 / 0.682 & 66.3 / 0.759 & 70.4 / 0.836 & 78.5 / 0.905 \\ 
 & BM25 + lexical density & 67.5 / 0.741 & 70.1 / 0.767 & 74.4 / 0.826 & 76.9 / 0.847 \\ 
\midrule
128k & \ours{} (\pqc{}) & 63.3 / 0.684 & 67.0 / 0.745 & 72.4 / 0.825 & 75.6 / 0.895 \\ 
 & \ours{}+BM25 & 66.6 / 0.738 & 69.5 / 0.778 & 74.0 / 0.861 & 77.4 / 0.906 \\ 
 & BM25, same selector & 63.1 / 0.673 & 65.8 / 0.724 & 66.1 / 0.749 & 68.4 / 0.762 \\ 
 & Dense RAG + reranker & 55.9 / 0.618 & 64.4 / 0.712 & 66.5 / 0.756 & 68.7 / 0.819 \\ 
 & BM25 + dense & 64.3 / 0.708 & 70.9 / 0.766 & 73.1 / 0.856 & 80.2 / 0.933 \\ 
 & BM25 + lexical density & 67.1 / 0.734 & 70.0 / 0.769 & 69.5 / 0.794 & 72.0 / 0.807 \\ 
\bottomrule
\end{tabular}}
\end{table}

\begin{table}[!htbp]
\centering
\caption{\textbf{Index chunk size $B_c$} (Llama-3.1-8B-Instruct, \ours{} with \pqc{} codes, \bench{} test episodes 1--6, $B{=}2{,}048$). $\Delta$ is the paired difference to the default $B_c{=}4{,}096$, with 95\% CI.}
\label{tab:chunk}
\scriptsize
\setlength{\tabcolsep}{4pt}
\resizebox{\ifdim\width>\linewidth\linewidth\else\width\fi}{!}{%
\begin{tabular}{@{}lcccc@{}}
\toprule
& \multicolumn{2}{c}{32k} & \multicolumn{2}{c}{128k} \\
\cmidrule(lr){2-3}\cmidrule(l){4-5}
$B_c$ & macro / presence & $\Delta$ & macro / presence & $\Delta$ \\
\midrule
1{,}024 & 74.9 / 0.877 & $+0.04$ \ci{-3.10}{+3.19} & 68.6 / 0.760 & $-3.83$ \ci{-6.25}{-1.58} \\ 
2{,}048 & 77.4 / 0.872 & $+2.54$ \ci{+0.19}{+5.10} & 72.0 / 0.807 & $-0.44$ \ci{-2.69}{+1.87} \\ 
4{,}096 (default) & 74.8 / 0.835 & -- & 72.4 / 0.825 & -- \\ 
8{,}192 & 73.6 / 0.846 & $-1.21$ \ci{-3.83}{+1.33} & 72.8 / 0.809 & $+0.33$ \ci{-2.02}{+2.62} \\ 
\bottomrule
\end{tabular}}
\end{table}


\section{Development diagnostics}
\label{app:dev-full}

\begin{table}[!htbp]
\centering
\caption{\textbf{Ablation ladder} (Llama-3.1-8B-Instruct, development answer presence, $B{=}2{,}048$, exact keys unless stated). Blocks (i)--(iv) are query-agnostic. Block (i) uses all heads of layers 0, 12, 16, 29. Block (v) is in context.}
\label{tab:ladder}
\footnotesize
\setlength{\tabcolsep}{4pt}\renewcommand{\arraystretch}{0.94}
\begin{tabular}[t]{@{}lcc@{}}
\toprule
Signal & 32k & 128k \\
\midrule
Random scores & 0.171 & 0.013 \\ 
BM25, same selector & 0.779 & 0.693 \\ 
\midrule
\multicolumn{3}{@{}l}{\textit{(i) \Sthree{}-Attention-style aggregation}} \\
SAE ids, IDF voting (\Sthree{}-Att.-style index) & 0.204 & 0.034 \\ 
exact keys, mean raw logit & 0.247 & 0.074 \\ 
\quad + sink & 0.249 & 0.136 \\ 
\quad + sink, max over question & 0.236 & 0.206 \\ 
\quad + sink, per-head softmax & 0.341 & 0.284 \\ 
\midrule
\multicolumn{3}{@{}l}{\textit{(ii) SAE codes of the 16 detected heads}} \\
decoded keys (\saec{}) & 0.801 & 0.822 \\ 
binary feature ids & 0.769 & 0.779 \\ 
IDF feature weighting & 0.751 & 0.143 \\ 
features with df${>}0.05L$ dropped & 0.226 & 0.108 \\ 
top-$m$ inverted index, $m{=}1{,}024$ & 0.464 & 0.276 \\ 
\bottomrule
\end{tabular}\hfill
\begin{tabular}[t]{@{}lcc@{}}
\toprule
Signal & 32k & 128k \\
\midrule
\multicolumn{3}{@{}l}{\textit{(iii) Query, top-16 heads}} \\
last prompt token, heads by $\masslast$ & 0.619 & 0.478 \\ 
prompt end, heads by $\massp$ & 0.397 & 0.186 \\ 
\textbf{bare question, heads by $\massq$ (\ours{})} & \textbf{0.801} & \textbf{0.809} \\ 
\midrule
\multicolumn{3}{@{}l}{\textit{(iv) Scoring operator, same 16 heads, bare question}} \\
per-head softmax (\ours{}) & 0.801 & 0.809 \\ 
cosine of values & 0.786 & 0.779 \\ 
cosine of keys & 0.689 & 0.587 \\ 
raw $\vq^{\top}\vk$ & 0.428 & 0.313 \\ 
\midrule
\multicolumn{3}{@{}l}{\textit{(v) In context (not query-agnostic)}} \\
raw logits, layer 13 (GemFilter-style) & 0.849 & 0.728 \\ 
top-16 heads by $\masslast$ & 0.913 & 0.812 \\ 
\bottomrule
\end{tabular}
\end{table}

\begin{figure}[!htbp]
\centering
\includegraphics[width=\linewidth]{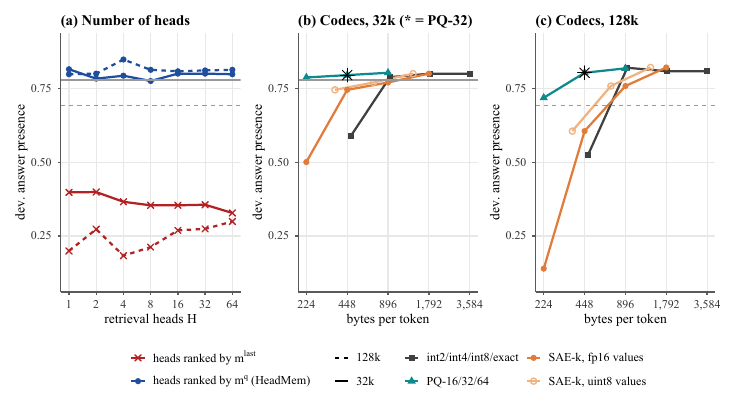}
\caption{\textbf{Heads and codecs} (Llama-3.1-8B-Instruct, development answer presence). Grey lines mark BM25 (solid 32k, dashed 128k). (a) Heads ranked by $\massq$ or $\masslast$. Solid lines show 32k and dashed lines show 128k. (b, c) int8 and \pqc{} stay within 0.005 of exact keys. Below 1{,}372\,B per token, SAE codes trail PQ codes of no larger size. At 1{,}792\,B, \saec{} reaches them.}
\label{fig:analysis}
\end{figure}

Unless stated otherwise, numbers in this section are development answer presence of the $B{=}2{,}048$ selection.
They are measured on Llama-3.1-8B-Instruct at 32k and 128k tokens.
Every signal goes through the same selector.
Every family cell has 16 questions, so differences of a few hundredths are not interpreted.

\paragraph{Ablation ladder (Table~\ref{tab:ladder-full}).}
Every variant of the \Sthree{}-Attention-style aggregation stays at or below 0.341 at 32k and 0.284 at 128k.
The best of them is the per-head softmax summed over all heads of the four layers (attention mass).
With the last prompt token as query, the top-16 heads ranked by $\masslast$ reach 0.619 and 0.478.
The bare question on the top-16 heads ranked by $\massq$ reaches 0.801 and 0.809, above BM25 (0.779 and 0.693).

\begin{table}[!htbp]
\centering
\caption{\textbf{Development ladder per task family} (answer presence, Llama-3.1-8B-Instruct, exact keys unless stated). Blocks (a)--(c) are query-agnostic. Table~\ref{tab:ic-full} gives the in-context control. Last-token rows rank heads by $\masslast$. The prompt-end row ranks heads by $\massp$. Bare-question rows rank heads by $\massq$. $n$ is the number of questions per family cell.}
\label{tab:ladder-full}
\scriptsize
\setlength{\tabcolsep}{4.5pt}
\begin{tabular}{@{}lccccccc@{}}
\toprule
Signal & NIAH-MK & NIAH-MV & VT & QA-SQuAD & QA-Hotpot & Mean & $n$ \\
\midrule
\multicolumn{8}{@{}l}{\textbf{32k tokens}} \\
Random scores & 0.000 & 0.078 & 0.150 & 0.375 & 0.250 & 0.171 & 16 \\ 
BM25, same selector & 0.938 & 0.969 & 0.238 & 0.875 & 0.875 & 0.779 & 16 \\ 
\multicolumn{8}{@{}l}{\textit{(a) \Sthree{}-Attention-style aggregation: mean raw logit over 32 heads $\times$ layers \{0,12,16,29\} and question tokens}} \\
\Sthree{}-Attention-style index (head-mean SAE, IDF) & 0.000 & 0.047 & 0.350 & 0.375 & 0.250 & 0.204 & 16 \\ 
same aggregation, exact keys & 0.062 & 0.047 & 0.312 & 0.375 & 0.438 & 0.247 & 16 \\ 
\quad + sink token & 0.125 & 0.047 & 0.325 & 0.312 & 0.438 & 0.249 & 16 \\ 
\quad + sink, \Sthree{}-Attention-style SAE index & 0.125 & 0.062 & 0.475 & 0.312 & 0.375 & 0.270 & 16 \\ 
\quad + sink, max over question tokens & 0.125 & 0.031 & 0.338 & 0.312 & 0.375 & 0.236 & 16 \\ 
\quad + sink, per-head softmax (attention mass) & 0.312 & 0.156 & 0.363 & 0.375 & 0.500 & 0.341 & 16 \\ 
\multicolumn{8}{@{}l}{\textit{(b) last prompt token as query, out of context}} \\
raw logit sum, layer 13 (GemFilter-style) & 0.125 & 0.109 & 0.413 & 0.250 & 0.250 & 0.229 & 16 \\ 
per-head softmax vote, layer 13 & 0.000 & 0.094 & 0.463 & 0.438 & 0.250 & 0.249 & 16 \\ 
per-head softmax vote, all heads & 0.375 & 0.156 & 0.400 & 0.375 & 0.438 & 0.349 & 16 \\ 
top-1 head ($\masslast$) & 1.000 & 0.438 & 0.525 & 0.562 & 0.625 & 0.630 & 16 \\ 
top-4 heads ($\masslast$) & 1.000 & 0.422 & 0.575 & 0.750 & 0.500 & 0.649 & 16 \\ 
top-16 heads ($\masslast$) & 0.938 & 0.422 & 0.487 & 0.750 & 0.500 & 0.619 & 16 \\ 
top-64 heads ($\masslast$) & 0.938 & 0.438 & 0.488 & 0.562 & 0.500 & 0.585 & 16 \\ 
\multicolumn{8}{@{}l}{\textit{(c) question as query, top-16 retrieval heads}} \\
prompt-end tokens ($\massp$) & 0.688 & 0.234 & 0.438 & 0.312 & 0.312 & 0.397 & 16 \\ 
bare question ($\massq$) $=$ \ours{}, exact keys & 1.000 & 0.906 & 0.413 & 1.000 & 0.688 & 0.801 & 16 \\ 
\midrule
\multicolumn{8}{@{}l}{\textbf{128k tokens}} \\
Random scores & 0.000 & 0.000 & 0.000 & 0.062 & 0.000 & 0.013 & 16 \\ 
BM25, same selector & 0.938 & 0.953 & 0.200 & 0.812 & 0.562 & 0.693 & 16 \\ 
\multicolumn{8}{@{}l}{\textit{(a) \Sthree{}-Attention-style aggregation}} \\
\Sthree{}-Attention-style index (head-mean SAE, IDF) & 0.000 & 0.047 & 0.000 & 0.125 & 0.000 & 0.034 & 16 \\ 
same aggregation, exact keys & 0.000 & 0.031 & 0.025 & 0.250 & 0.062 & 0.074 & 16 \\ 
\quad + sink token & 0.062 & 0.031 & 0.087 & 0.375 & 0.125 & 0.136 & 16 \\ 
\quad + sink, \Sthree{}-Attention-style SAE index & 0.000 & 0.031 & 0.375 & 0.062 & 0.000 & 0.094 & 16 \\ 
\quad + sink, max over question tokens & 0.125 & 0.016 & 0.450 & 0.312 & 0.125 & 0.206 & 16 \\ 
\quad + sink, per-head softmax (attention mass) & 0.125 & 0.094 & 0.388 & 0.375 & 0.438 & 0.284 & 16 \\ 
\multicolumn{8}{@{}l}{\textit{(b) last prompt token as query, out of context}} \\
raw logit sum, layer 13 (GemFilter-style) & 0.062 & 0.047 & 0.588 & 0.188 & 0.062 & 0.189 & 16 \\ 
per-head softmax vote, layer 13 & 0.062 & 0.047 & 0.400 & 0.125 & 0.000 & 0.127 & 16 \\ 
per-head softmax vote, all heads & 0.125 & 0.156 & 0.412 & 0.188 & 0.062 & 0.189 & 16 \\ 
top-1 head ($\masslast$) & 0.938 & 0.359 & 0.438 & 0.125 & 0.312 & 0.434 & 16 \\ 
top-4 heads ($\masslast$) & 0.875 & 0.359 & 0.425 & 0.188 & 0.250 & 0.419 & 16 \\ 
top-16 heads ($\masslast$) & 0.875 & 0.375 & 0.387 & 0.375 & 0.375 & 0.478 & 16 \\ 
top-64 heads ($\masslast$) & 0.750 & 0.406 & 0.438 & 0.188 & 0.312 & 0.419 & 16 \\ 
\multicolumn{8}{@{}l}{\textit{(c) question as query, top-16 retrieval heads}} \\
prompt-end tokens ($\massp$) & 0.188 & 0.156 & 0.400 & 0.188 & 0.000 & 0.186 & 16 \\ 
bare question ($\massq$) $=$ \ours{}, exact keys & 1.000 & 0.969 & 0.512 & 0.875 & 0.688 & 0.809 & 16 \\ 
\bottomrule
\end{tabular}
\end{table}

\paragraph{Why the head-mean aggregation fails.}
Raw logit sums are dominated by the heads and channels with the largest norms~\cite{wu2024tokenselect,hooper2024kvquant,jin2025massive}.
A sink token per chunk and question~\cite{gu2025attentionsink,yang2025ape}, the maximum over question tokens and a per-head softmax before summing~\cite{wu2024tokenselect,chen2025icr} address this.
Over all heads of the four layers, however, they reach at most 0.341 and 0.284 (Table~\ref{tab:ladder-full}).
This is far below BM25 (0.779 and 0.693).
The signal sits in a few heads that this aggregation dilutes. 
On this aggregation with exact keys, prepending the sink token raises answer presence at 128k from 0.074 to 0.136.
It leaves 32k unchanged (0.247 and 0.249).
The sink is part of every \ours{} configuration.
It was not ablated on the 16 detected heads.

\paragraph{In-context control (Table~\ref{tab:ic-full}).}
This control prefills the context under full causal attention and appends the prompt end, so it is not query-agnostic.
Here the top-16 heads reach 0.913 at 32k and 0.812 at 128k.
Query-agnostic \ours{} with exact keys reaches 0.801 and 0.809.
In context, the raw logit sum at layer 13 in the style of GemFilter~\cite{shi2024gemfilter} reaches 0.849 and 0.728. 
Voting over all heads falls from 0.853 to 0.362 between 32k and 128k even in context.
The out-of-context last-token query against the in-context keys gives 0.634 and 0.285.
In context, the top-16 heads reach 0.800 and 0.825 on variable tracking, against 0.413 and 0.512 for \ours{} with exact keys. 

\begin{table}[!htbp]
\centering
\caption{\textbf{In-context control per task family} (answer presence). The context is prefilled with full causal attention, and the prompt end is appended. The model computes the post-RoPE logits. This control is not query-agnostic. Top-$H$ rows rank heads by $\masslast$ and use the last prompt token as query. The out-of-context row computes the last-token query out of context, rotates it to the same position and scores it against the in-context keys. The \ours{} row repeats Table~\ref{tab:ladder-full} for reference.}
\label{tab:ic-full}
\scriptsize
\setlength{\tabcolsep}{4.5pt}
\begin{tabular}{@{}lccccccc@{}}
\toprule
Signal & NIAH-MK & NIAH-MV & VT & QA-SQuAD & QA-Hotpot & Mean & $n$ \\
\midrule
\multicolumn{8}{@{}l}{\textbf{32k tokens}} \\
raw logit sum, layer 13 (GemFilter) & 1.000 & 0.859 & 0.638 & 1.000 & 0.750 & 0.849 & 16 \\ 
per-head softmax vote, layer 13 & 1.000 & 0.828 & 0.625 & 1.000 & 0.812 & 0.853 & 16 \\ 
per-head softmax vote, all heads & 1.000 & 0.766 & 0.562 & 1.000 & 0.938 & 0.853 & 16 \\ 
top-1 head & 1.000 & 0.812 & 0.825 & 1.000 & 0.938 & 0.915 & 16 \\ 
top-4 heads & 1.000 & 0.734 & 0.800 & 1.000 & 0.875 & 0.882 & 16 \\ 
top-16 heads & 1.000 & 0.891 & 0.800 & 1.000 & 0.875 & 0.913 & 16 \\ 
top-64 heads & 1.000 & 0.891 & 0.788 & 1.000 & 0.938 & 0.923 & 16 \\ 
out-of-context last-token query, top-16 & 1.000 & 0.281 & 0.575 & 0.750 & 0.562 & 0.634 & 16 \\ 
\textit{\ours{}, exact keys (query-agnostic)} & 1.000 & 0.906 & 0.413 & 1.000 & 0.688 & 0.801 & 16 \\ 
\midrule
\multicolumn{8}{@{}l}{\textbf{128k tokens}} \\
raw logit sum, layer 13 (GemFilter) & 1.000 & 0.766 & 0.750 & 0.750 & 0.375 & 0.728 & 16 \\ 
per-head softmax vote, layer 13 & 0.938 & 0.781 & 0.500 & 0.500 & 0.250 & 0.594 & 16 \\ 
per-head softmax vote, all heads & 0.688 & 0.109 & 0.325 & 0.562 & 0.125 & 0.362 & 16 \\ 
top-1 head & 1.000 & 0.594 & 0.750 & 0.875 & 0.438 & 0.731 & 16 \\ 
top-4 heads & 1.000 & 0.641 & 0.800 & 0.875 & 0.562 & 0.776 & 16 \\ 
top-16 heads & 1.000 & 0.672 & 0.825 & 0.938 & 0.625 & 0.812 & 16 \\ 
top-64 heads & 1.000 & 0.531 & 0.688 & 0.875 & 0.500 & 0.719 & 16 \\ 
out-of-context last-token query, top-16 & 0.375 & 0.125 & 0.363 & 0.438 & 0.125 & 0.285 & 16 \\ 
\textit{\ours{}, exact keys (query-agnostic)} & 1.000 & 0.969 & 0.512 & 0.875 & 0.688 & 0.809 & 16 \\ 
\bottomrule
\end{tabular}
\end{table}

\paragraph{Number of heads and ranking criterion (Table~\ref{tab:nheads-full}).}
The scoring is that of \ours{} (bare-question queries, per-head softmax, exact keys).
Only the set of heads changes.
With $\massq$ ranking, the mean is 0.776--0.816 at 32k and 0.799--0.849 at 128k for $H$ from 1 to 64.
With $\masslast$ ranking, it is 0.328--0.399 and 0.183--0.299.
The single top head, (2,22), stores 256\,B per token of exact keys and reaches 0.816 and 0.799. 

\begin{table}[!htbp]
\centering
\caption{\textbf{Number of retrieval heads and ranking criterion} (exact keys, development answer presence). Bytes per token count the distinct KV heads hosting the selected query heads (256\,B per KV head).}
\label{tab:nheads-full}
\scriptsize
\setlength{\tabcolsep}{4.5pt}
\begin{tabular}{@{}lrrcccccc@{}}
\toprule
Ranking & $H$ & B/token & NIAH-MK & NIAH-MV & VT & QA-SQuAD & QA-Hotpot & Mean \\
\midrule
\multicolumn{9}{@{}l}{\textbf{32k tokens}} \\
$\massq$ & 1 & 256 & 1.000 & 0.906 & 0.487 & 1.000 & 0.688 & 0.816 \\ 
$\massq$ & 2 & 256 & 1.000 & 0.906 & 0.450 & 1.000 & 0.562 & 0.784 \\ 
$\massq$ & 4 & 768 & 1.000 & 0.906 & 0.375 & 1.000 & 0.688 & 0.794 \\ 
$\massq$ & 8 & 1{,}792 & 1.000 & 0.906 & 0.413 & 1.000 & 0.562 & 0.776 \\ 
$\massq$ & 16 & 3{,}584 & 1.000 & 0.906 & 0.413 & 1.000 & 0.688 & 0.801 \\ 
$\massq$ & 32 & 5{,}888 & 1.000 & 0.906 & 0.413 & 1.000 & 0.688 & 0.801 \\ 
$\massq$ & 64 & 12{,}544 & 1.000 & 0.906 & 0.400 & 0.938 & 0.750 & 0.799 \\ 
$\masslast$ & 1 & 256 & 0.375 & 0.188 & 0.425 & 0.375 & 0.625 & 0.398 \\ 
$\masslast$ & 2 & 512 & 0.438 & 0.156 & 0.400 & 0.438 & 0.562 & 0.399 \\ 
$\masslast$ & 4 & 768 & 0.250 & 0.094 & 0.425 & 0.562 & 0.500 & 0.366 \\ 
$\masslast$ & 8 & 1{,}792 & 0.188 & 0.109 & 0.413 & 0.562 & 0.500 & 0.354 \\ 
$\masslast$ & 16 & 3{,}584 & 0.188 & 0.094 & 0.425 & 0.562 & 0.500 & 0.354 \\ 
$\masslast$ & 32 & 6{,}144 & 0.250 & 0.094 & 0.438 & 0.500 & 0.500 & 0.356 \\ 
$\masslast$ & 64 & 11{,}264 & 0.125 & 0.078 & 0.438 & 0.438 & 0.562 & 0.328 \\ 
\midrule
\multicolumn{9}{@{}l}{\textbf{128k tokens}} \\
$\massq$ & 1 & 256 & 1.000 & 0.969 & 0.400 & 0.938 & 0.688 & 0.799 \\ 
$\massq$ & 2 & 256 & 1.000 & 0.969 & 0.350 & 1.000 & 0.688 & 0.801 \\ 
$\massq$ & 4 & 768 & 1.000 & 0.969 & 0.525 & 1.000 & 0.750 & 0.849 \\ 
$\massq$ & 8 & 1{,}792 & 1.000 & 0.969 & 0.475 & 0.938 & 0.688 & 0.814 \\ 
$\massq$ & 16 & 3{,}584 & 1.000 & 0.969 & 0.512 & 0.875 & 0.688 & 0.809 \\ 
$\massq$ & 32 & 5{,}888 & 1.000 & 0.984 & 0.513 & 0.875 & 0.688 & 0.812 \\ 
$\massq$ & 64 & 12{,}544 & 1.000 & 0.984 & 0.525 & 0.875 & 0.688 & 0.814 \\ 
$\masslast$ & 1 & 256 & 0.062 & 0.109 & 0.200 & 0.250 & 0.375 & 0.199 \\ 
$\masslast$ & 2 & 512 & 0.125 & 0.203 & 0.413 & 0.250 & 0.375 & 0.273 \\ 
$\masslast$ & 4 & 768 & 0.000 & 0.078 & 0.338 & 0.250 & 0.250 & 0.183 \\ 
$\masslast$ & 8 & 1{,}792 & 0.000 & 0.062 & 0.375 & 0.312 & 0.312 & 0.212 \\ 
$\masslast$ & 16 & 3{,}584 & 0.000 & 0.047 & 0.425 & 0.500 & 0.375 & 0.269 \\ 
$\masslast$ & 32 & 6{,}144 & 0.125 & 0.031 & 0.400 & 0.438 & 0.375 & 0.274 \\ 
$\masslast$ & 64 & 11{,}264 & 0.125 & 0.031 & 0.400 & 0.562 & 0.375 & 0.299 \\ 
\bottomrule
\end{tabular}
\end{table}

\paragraph{Key codecs (Table~\ref{tab:codec-full}).}
All codecs store the same 14 KV heads and are scored identically.
int8 gives the presence of exact keys at both lengths (0.801, 0.809).
\pqc{} reaches 0.796 and 0.804 at 448\,B per token, where SAE-8 reaches 0.746 and 0.606.
At 224\,B, PQ-16 reaches 0.788 and 0.719, and SAE-4 reaches 0.501 and 0.139.
SAE codes reach the presence of exact keys only from \saec{} on.
\saec{} stores 1{,}792\,B per token, or 1{,}372\,B with uint8 values.
The quantized values leave the mean unchanged for $k=8,16,32$.

\begin{table}[!htbp]
\centering
\caption{\textbf{Rate, reconstruction error and development answer presence of key codecs} (16 retrieval heads $=$ 14 KV heads, Llama-3.1-8B-Instruct, scoring of Eq.~\eqref{eq:score}). Bytes per token follow Eq.~\eqref{eq:bytes}. FVU is measured on WikiText-2 validation keys and averaged over KV heads. -- marks values that were not measured. ``v8'' denotes uint8 values with an fp16 per-token scale. Each family cell has 16 questions.}
\label{tab:codec-full}
\scriptsize
\setlength{\tabcolsep}{3.5pt}
\begin{tabular}{@{}lrccccccc@{}}
\toprule
Codec & B/token & FVU & NIAH-MK & NIAH-MV & VT & QA-SQuAD & QA-Hotpot & Mean \\
\midrule
\multicolumn{9}{@{}l}{\textbf{32k tokens}} \\
exact (bf16) & 3{,}584 & -- & 1.000 & 0.906 & 0.413 & 1.000 & 0.688 & 0.801 \\ 
int8 & 1{,}820 & -- & 1.000 & 0.906 & 0.413 & 1.000 & 0.688 & 0.801 \\ 
int4 & 924 & -- & 1.000 & 0.906 & 0.413 & 1.000 & 0.625 & 0.789 \\ 
int2 & 476 & -- & 1.000 & 0.609 & 0.400 & 0.688 & 0.250 & 0.589 \\ 
PQ-64 & 896 & -- & 1.000 & 0.906 & 0.425 & 1.000 & 0.688 & 0.804 \\ 
\pqc{} (default) & 448 & -- & 1.000 & 0.906 & 0.388 & 1.000 & 0.688 & 0.796 \\ 
PQ-16 & 224 & -- & 1.000 & 0.891 & 0.425 & 0.938 & 0.688 & 0.788 \\ 
\saec{} & 1{,}792 & 0.069 & 1.000 & 0.906 & 0.413 & 1.000 & 0.688 & 0.801 \\ 
\saec{}, v8 & 1{,}372 & -- & 1.000 & 0.906 & 0.413 & 1.000 & 0.688 & 0.801 \\ 
SAE-16 & 896 & 0.168 & 1.000 & 0.906 & 0.388 & 0.938 & 0.625 & 0.771 \\ 
SAE-16, v8 & 700 & -- & 1.000 & 0.906 & 0.388 & 0.938 & 0.625 & 0.771 \\ 
SAE-8 & 448 & 0.286 & 1.000 & 0.906 & 0.388 & 0.875 & 0.562 & 0.746 \\ 
SAE-8, v8 & 364 & -- & 1.000 & 0.906 & 0.388 & 0.875 & 0.562 & 0.746 \\ 
SAE-4 & 224 & 0.419 & 0.562 & 0.406 & 0.350 & 0.750 & 0.438 & 0.501 \\ 
\midrule
\multicolumn{9}{@{}l}{\textbf{128k tokens}} \\
exact (bf16) & 3{,}584 & -- & 1.000 & 0.969 & 0.512 & 0.875 & 0.688 & 0.809 \\ 
int8 & 1{,}820 & -- & 1.000 & 0.969 & 0.512 & 0.875 & 0.688 & 0.809 \\ 
int4 & 924 & -- & 1.000 & 0.969 & 0.512 & 0.875 & 0.750 & 0.821 \\ 
int2 & 476 & -- & 0.812 & 0.766 & 0.362 & 0.500 & 0.188 & 0.526 \\ 
PQ-64 & 896 & -- & 1.000 & 0.969 & 0.500 & 0.875 & 0.750 & 0.819 \\ 
\pqc{} (default) & 448 & -- & 1.000 & 0.969 & 0.487 & 0.875 & 0.688 & 0.804 \\ 
PQ-16 & 224 & -- & 1.000 & 0.984 & 0.300 & 0.750 & 0.562 & 0.719 \\ 
\saec{} & 1{,}792 & 0.069 & 1.000 & 0.984 & 0.500 & 0.875 & 0.750 & 0.822 \\ 
\saec{}, v8 & 1{,}372 & -- & 1.000 & 0.984 & 0.500 & 0.875 & 0.750 & 0.822 \\ 
SAE-16 & 896 & 0.168 & 1.000 & 0.969 & 0.450 & 0.750 & 0.625 & 0.759 \\ 
SAE-16, v8 & 700 & -- & 1.000 & 0.969 & 0.450 & 0.750 & 0.625 & 0.759 \\ 
SAE-8 & 448 & 0.286 & 0.875 & 0.891 & 0.388 & 0.500 & 0.375 & 0.606 \\ 
SAE-8, v8 & 364 & -- & 0.875 & 0.891 & 0.388 & 0.500 & 0.375 & 0.606 \\ 
SAE-4 & 224 & 0.419 & 0.062 & 0.109 & 0.025 & 0.188 & 0.312 & 0.139 \\ 
\bottomrule
\end{tabular}
\end{table}

\paragraph{SAE size and fusion (Table~\ref{tab:v3-dev}).}
These runs fixed $k=32$.
The fusion was run only at weight 0.5.
\saec{} gives the presence of exact keys at 32k (0.801) and is above them at 128k (0.822 vs 0.809).
SAE-16 gives 0.771 and 0.759.
Fusion with BM25 gives 0.820 and 0.812. 

\begin{table}[!htbp]
\centering
\caption{\textbf{SAE size and fusion} on development episodes (Llama-3.1-8B-Instruct, answer presence, 16 questions per family cell). Fusion uses Eq.~\eqref{eq:fuse} with weight 0.5.}
\label{tab:v3-dev}
\scriptsize
\setlength{\tabcolsep}{4.5pt}
\begin{tabular}{@{}lcccccc@{}}
\toprule
Variant & NIAH-MK & NIAH-MV & VT & QA-SQuAD & QA-Hotpot & Mean \\
\midrule
\multicolumn{7}{@{}l}{\textbf{32k tokens}} \\
BM25, same selector & 0.938 & 0.969 & 0.238 & 0.875 & 0.875 & 0.779 \\ 
\ours{}, exact keys & 1.000 & 0.906 & 0.413 & 1.000 & 0.688 & 0.801 \\ 
\ours{}, \saec{} & 1.000 & 0.906 & 0.413 & 1.000 & 0.688 & 0.801 \\ 
\ours{}, SAE-16 & 1.000 & 0.906 & 0.388 & 0.938 & 0.625 & 0.771 \\ 
\ours{}+BM25, \saec{} & 1.000 & 0.938 & 0.413 & 1.000 & 0.750 & 0.820 \\ 
\ours{}+BM25, SAE-16 & 1.000 & 0.938 & 0.388 & 0.938 & 0.688 & 0.790 \\ 
\midrule
\multicolumn{7}{@{}l}{\textbf{128k tokens}} \\
BM25, same selector & 0.938 & 0.953 & 0.200 & 0.812 & 0.562 & 0.693 \\ 
\ours{}, exact keys & 1.000 & 0.969 & 0.512 & 0.875 & 0.688 & 0.809 \\ 
\ours{}, \saec{} & 1.000 & 0.984 & 0.500 & 0.875 & 0.750 & 0.822 \\ 
\ours{}, SAE-16 & 1.000 & 0.969 & 0.450 & 0.750 & 0.625 & 0.759 \\ 
\ours{}+BM25, \saec{} & 1.000 & 1.000 & 0.500 & 0.875 & 0.688 & 0.812 \\ 
\ours{}+BM25, SAE-16 & 1.000 & 1.000 & 0.450 & 0.812 & 0.688 & 0.790 \\ 
\bottomrule
\end{tabular}
\end{table}

\paragraph{Qwen2.5 (Table~\ref{tab:qwen-dev}).}
Qwen2.5 follows the Llama pattern.
Voting over all heads with the last token gives 0.341 and 0.223.
The top-16 heads with the last token give 0.629 and 0.600.
The top-16 heads with the bare question give 0.784 and 0.743, against 0.723 and 0.677 for BM25. 
On Qwen2.5, null-query calibration raises presence with both key stores.
With exact keys it gives 0.813 and 0.786, against 0.784 and 0.743 without it.
With \saec{} it gives 0.811 and 0.789, against 0.787 and 0.736 without it.
The test runs keep one configuration for every model and do not use calibration. 

\begin{table}[!htbp]
\centering
\caption{\textbf{Qwen2.5-7B-Instruct development diagnostics} (answer presence, 16 questions per family cell). Heads are detected on the Qwen2.5 detection split (Table~\ref{tab:heads}). The top-16 bare-question row and the exact-keys row are the same configuration, recorded by the diagnostic and by the pipeline run.}
\label{tab:qwen-dev}
\scriptsize
\setlength{\tabcolsep}{4.5pt}
\begin{tabular}{@{}lccccccc@{}}
\toprule
Signal & NIAH-MK & NIAH-MV & VT & QA-SQuAD & QA-Hotpot & Mean & $n$ \\
\midrule
\multicolumn{8}{@{}l}{\textbf{32k tokens}} \\
Random scores & 0.000 & 0.031 & 0.137 & 0.312 & 0.188 & 0.134 & 16 \\ 
BM25, same selector & 0.938 & 0.891 & 0.225 & 0.875 & 0.688 & 0.723 & 16 \\ 
last token, vote over all heads & 0.562 & 0.141 & 0.438 & 0.250 & 0.312 & 0.341 & 16 \\ 
last token, top-16 heads ($\masslast$) & 1.000 & 0.406 & 0.487 & 0.688 & 0.562 & 0.629 & 16 \\ 
bare question, top-16 heads ($\massq$) & 1.000 & 0.844 & 0.325 & 1.000 & 0.750 & 0.784 & 16 \\ 
\quad + null-query calibration & 1.000 & 0.891 & 0.425 & 1.000 & 0.750 & 0.813 & 16 \\ 
\ours{}, exact keys & 1.000 & 0.844 & 0.325 & 1.000 & 0.750 & 0.784 & 16 \\ 
\ours{}, \saec{} & 1.000 & 0.859 & 0.325 & 1.000 & 0.750 & 0.787 & 16 \\ 
\ours{}, \saec{} + calibration & 1.000 & 0.891 & 0.413 & 1.000 & 0.750 & 0.811 & 16 \\ 
\ours{}+BM25, \saec{} & 1.000 & 0.953 & 0.325 & 1.000 & 0.750 & 0.806 & 16 \\ 
\midrule
\multicolumn{8}{@{}l}{\textbf{128k tokens}} \\
Random scores & 0.000 & 0.062 & 0.025 & 0.250 & 0.062 & 0.080 & 16 \\ 
BM25, same selector & 0.875 & 0.938 & 0.200 & 0.812 & 0.562 & 0.677 & 16 \\ 
last token, vote over all heads & 0.250 & 0.078 & 0.537 & 0.188 & 0.062 & 0.223 & 16 \\ 
last token, top-16 heads ($\masslast$) & 1.000 & 0.500 & 0.625 & 0.562 & 0.312 & 0.600 & 16 \\ 
bare question, top-16 heads ($\massq$) & 1.000 & 0.953 & 0.325 & 0.812 & 0.625 & 0.743 & 16 \\ 
\quad + null-query calibration & 1.000 & 0.953 & 0.475 & 0.812 & 0.688 & 0.786 & 16 \\ 
\ours{}, exact keys & 1.000 & 0.953 & 0.325 & 0.812 & 0.625 & 0.743 & 16 \\ 
\ours{}, \saec{} & 1.000 & 0.969 & 0.213 & 0.875 & 0.625 & 0.736 & 16 \\ 
\ours{}, \saec{} + calibration & 1.000 & 0.969 & 0.475 & 0.812 & 0.688 & 0.789 & 16 \\ 
\ours{}+BM25, \saec{} & 1.000 & 0.984 & 0.213 & 0.812 & 0.625 & 0.727 & 16 \\ 
\bottomrule
\end{tabular}
\end{table}

\paragraph{Question specificity (Table~\ref{tab:qspec-full}).}
Every \bench{} context holds the evidence of all eight questions.
A selection can therefore contain the answer without being specific to its question.
The control measures three quantities for each question.
Own is the presence of its answer in its own selection.
Cross is the presence of its answer in the selections of the other seven questions.
Jaccard is the overlap of two questions' selections, with lead and tail excluded.
On multi-key NIAH, the answer is in another question's selection as often as in its own for \ours{} (1.00 / 1.00 at both lengths).
The same nearly holds for BM25 (0.94 / 0.94 at 32k, 0.94 / 0.88 at 128k).
All eight needles share one template, so this family does not test question-specific retrieval.
On variable tracking, \ours{}'s own and cross presence are close (0.41 / 0.39 at 32k, 0.51 / 0.35 at 128k).
BM25's cross presence there is 0.05 and 0.01.
\ours{}'s advantage in this family is mostly question-agnostic.
On the QA families, cross presence is far below own presence for both methods.
For \ours{}, own / cross is 1.00 / 0.24 and 0.69 / 0.21 at 32k, and 0.88 / 0.16 and 0.69 / 0.03 at 128k.
These families therefore test question-specific retrieval.
There, \ours{}'s own presence is above BM25's at 128k (0.88 vs 0.81 on SQuAD, 0.69 vs 0.56 on HotpotQA).
At 32k it is above on SQuAD (1.00 vs 0.88) and below on HotpotQA (0.69 vs 0.88).
The top head alone shows the same pattern. 

\begin{table}[!htbp]
\centering
\caption{\textbf{Question-specificity control} (Llama-3.1-8B-Instruct, development episodes, exact keys). Each cell gives own / cross / Jaccard. Own is the presence of question $i$'s answer in question $i$'s selection. Cross is its presence in another question's selection (mean over $i\neq j$). Jaccard is the overlap of two questions' selections, with lead and tail excluded. \ours{}-16 uses the 16 retrieval heads. \ours{}-1 uses the top head only.}
\label{tab:qspec-full}
\scriptsize
\setlength{\tabcolsep}{4pt}
\begin{tabular}{@{}llccccc@{}}
\toprule
$L$ & Method & NIAH-MK & NIAH-MV & VT & QA-SQuAD & QA-Hotpot \\
\midrule
32k & \ours{}-16 & 1.00/1.00/0.89 & 0.91/0.28/0.58 & 0.41/0.39/0.89 & 1.00/0.24/0.07 & 0.69/0.21/0.07 \\ 
 & \ours{}-1 & 1.00/1.00/0.85 & 0.91/0.37/0.62 & 0.49/0.39/0.77 & 1.00/0.22/0.03 & 0.69/0.17/0.02 \\ 
 & BM25 & 0.94/0.94/0.95 & 0.97/0.41/0.64 & 0.24/0.05/0.87 & 0.88/0.21/0.05 & 0.88/0.16/0.01 \\ 
 & Random & 0.00/0.06/0.03 & 0.08/0.06/0.03 & 0.15/0.11/0.03 & 0.38/0.28/0.03 & 0.25/0.26/0.03 \\ 
\midrule
128k & \ours{}-16 & 1.00/1.00/0.88 & 0.97/0.34/0.62 & 0.51/0.35/0.87 & 0.88/0.16/0.08 & 0.69/0.03/0.06 \\ 
 & \ours{}-1 & 1.00/1.00/0.92 & 0.97/0.33/0.62 & 0.40/0.26/0.70 & 0.94/0.20/0.07 & 0.69/0.01/0.02 \\ 
 & BM25 & 0.94/0.88/0.79 & 0.95/0.34/0.64 & 0.20/0.01/0.87 & 0.81/0.14/0.01 & 0.56/0.04/0.00 \\ 
 & Random & 0.00/0.04/0.01 & 0.00/0.00/0.01 & 0.00/0.02/0.00 & 0.06/0.15/0.01 & 0.00/0.04/0.00 \\ 
\bottomrule
\end{tabular}
\end{table}

\subsection{Scoring operators on the same heads}
\label{app:scoring-dev}

Table~\ref{tab:scoring-dev} keeps the chunk-local build, the bare-question forward, the selector and the budget of \ours{} fixed.
It changes only the stored states and the operator that turns them into a score per position.
On the 16 detected heads with exact states, \ours{}'s per-head softmax (Eq.~\eqref{eq:score}) gives 0.801 and 0.809.
Cosine similarity of value states, maximized over question tokens, comes closest (0.786 and 0.779).
Cosine similarity of keys falls behind (0.689 and 0.587).
The mean raw $\vq^{\top}\vk$ falls far behind (0.428 and 0.313), below BM25 (0.779 and 0.693).
REFORM found raw query--key products inferior to cosine similarity of value states in its in-context setting~\cite{song2025reform}.
The same holds with the question read alone.
The per-head softmax is at or above cosine of values at both lengths.
The REFORM-style port uses cosine of values on the top-4 heads with 129-token max-pooling (768\,B per token).
It gives 0.712 and 0.683.
The ILRe-style port uses all heads of layer 3, scored by the maximum softmax probability (2{,}048\,B per token).
It gives 0.274 and 0.208.
\ours{}'s own score on all heads of layer 3 gives 0.199 and 0.150.
The ILRe-style port thus fails through its heads, not only through its operator.
The test runs of both ports agree (Table~\ref{tab:e2} and Appendix~\ref{app:protocol-ports}).

\begin{table}[!htbp]
\centering
\caption{\textbf{Scoring operators with the question read alone} (Llama-3.1-8B-Instruct, development answer presence of the $B{=}2{,}048$ selection, 16 questions per family cell). All rows use the chunk-local build with a sink token, the bare question and the \ours{} selector. States are exact (bf16). B/token gives the bytes stored per context token. The ports are defined in Appendix~\ref{app:protocol-ports}.}
\label{tab:scoring-dev}
\scriptsize
\setlength{\tabcolsep}{3pt}
\resizebox{\ifdim\width>\linewidth\linewidth\else\width\fi}{!}{%
\begin{tabular}{@{}llccccccr@{}}
\toprule
Scoring & Heads & NIAH-MK & NIAH-MV & VT & QA-SQuAD & QA-Hotpot & Mean & B/token \\
\midrule
\multicolumn{9}{@{}l}{\textbf{32k tokens}} \\
per-head softmax, mean over question, sum over heads (\ours{}) & 16 detected & 1.000 & 0.906 & 0.413 & 1.000 & 0.688 & 0.801 & 3{,}584 \\ 
cosine of value states, max over question & 16 detected & 1.000 & 0.891 & 0.350 & 0.938 & 0.750 & 0.786 & 3{,}584 \\ 
cosine of keys, max over question & 16 detected & 1.000 & 0.672 & 0.213 & 0.938 & 0.625 & 0.689 & 3{,}584 \\ 
raw $\vq^{\top}\vk$, mean over question, sum over heads & 16 detected & 0.438 & 0.188 & 0.450 & 0.500 & 0.562 & 0.428 & 3{,}584 \\ 
REFORM-style: cosine of values, max over question, 129-token max-pool & top-4 detected & 1.000 & 0.609 & 0.263 & 1.000 & 0.688 & 0.712 & 768 \\ 
ILRe-style: max softmax probability over heads and question & all of layer 3 & 0.062 & 0.109 & 0.388 & 0.500 & 0.312 & 0.274 & 2{,}048 \\ 
per-head softmax as in \ours{} & all of layer 3 & 0.000 & 0.047 & 0.450 & 0.250 & 0.250 & 0.199 & 2{,}048 \\ 
\midrule
\multicolumn{9}{@{}l}{\textbf{128k tokens}} \\
per-head softmax, mean over question, sum over heads (\ours{}) & 16 detected & 1.000 & 0.969 & 0.512 & 0.875 & 0.688 & 0.809 & 3{,}584 \\ 
cosine of value states, max over question & 16 detected & 1.000 & 0.984 & 0.288 & 0.938 & 0.688 & 0.779 & 3{,}584 \\ 
cosine of keys, max over question & 16 detected & 0.688 & 0.672 & 0.200 & 0.750 & 0.625 & 0.587 & 3{,}584 \\ 
raw $\vq^{\top}\vk$, mean over question, sum over heads & 16 detected & 0.125 & 0.141 & 0.550 & 0.438 & 0.312 & 0.313 & 3{,}584 \\ 
REFORM-style: cosine of values, max over question, 129-token max-pool & top-4 detected & 1.000 & 0.703 & 0.213 & 0.875 & 0.625 & 0.683 & 768 \\ 
ILRe-style: max softmax probability over heads and question & all of layer 3 & 0.000 & 0.000 & 0.287 & 0.438 & 0.312 & 0.208 & 2{,}048 \\ 
per-head softmax as in \ours{} & all of layer 3 & 0.000 & 0.000 & 0.312 & 0.312 & 0.125 & 0.150 & 2{,}048 \\ 
\bottomrule
\end{tabular}}
\end{table}


\FloatBarrier
\section{Mechanism of the retrieval heads}
\label{app:mechanism}

\paragraph{Token matching.}
Two measurements test whether the detected heads match question tokens.
The first is the classical induction score~\cite{olsson2022context}.
A sequence of 200 random token ids follows a sink and is then repeated, and eight such sequences are averaged.
The induction score of a head is its mean attention from a token of the second copy to the token that followed the same token in the first copy.
The duplicate-token score is its attention to that earlier occurrence itself.
The second measurement uses \ours{}'s own setting, with chunk-local keys, a sink, and the bare question read alone.
For every content token X of a question that also occurs in the context, it records two attention masses of X.
\emph{Same} is the mass on the positions of X in the context, and \emph{next} is the mass on the positions right after them.
It uses 32k development episodes with 16 questions per family, over all five families on Llama and all but variable tracking on Qwen2.5 and Mistral. Mistral uses test episodes, which were never used for tuning.
Table~\ref{tab:mech-ind} compares the 16 detected heads with all other heads of each model.
The Spearman column correlates the induction score with the detection score $\massq$ over all heads, and every permutation $p$ is at most $10^{-4}$.
On Llama and Mistral the detected heads, as a group, read the token after a match (induction pattern).
On Qwen2.5 they read the matching token itself (duplicate-token pattern).
On these episodes, which detection never saw, the heads' answer mass correlates with their detection score across all heads (Spearman 0.98 on Llama, 0.80 on Mistral and 0.75 on Qwen2.5).

\begin{table}[!htbp]
\centering
\caption{\textbf{Token matching by the detected heads.} Induction and duplicate-token scores on repeated random tokens (in context), and attention mass in \ours{}'s setting (percent of a matched question token's attention, mean over questions). Each cell pair gives the 16 detected heads and all other heads, and the difference has a 95\% bootstrap CI over questions stratified by family. Chance mass is 0.05--0.06\%.}
\label{tab:mech-ind}
\scriptsize
\setlength{\tabcolsep}{2.6pt}
\resizebox{\linewidth}{!}{%
\begin{tabular}{@{}lcccccccc@{}}
\toprule
 & & \multicolumn{3}{c}{repeated random tokens} & \multicolumn{4}{c}{\ours{} setting, attention mass (\%)} \\
\cmidrule(lr){3-5}\cmidrule(l){6-9}
Model & heads & Spearman (induction) & induction & duplicate & next & next, difference & same & same, difference \\
\midrule
Llama-3.1-8B & 1024 & 0.70 & 0.195 / 0.023 & 0.013 / 0.011 & 8.99 / 0.38 & $+8.60$ \ci{+8.02}{+9.21} & 0.27 / 0.23 & $+0.04$ \ci{+0.02}{+0.07} \\ 
Mistral-7B & 1024 & 0.45 & 0.123 / 0.026 & 0.012 / 0.012 & 3.78 / 0.19 & $+3.59$ \ci{+3.42}{+3.77} & 0.15 / 0.16 & $-0.01$ \ci{-0.03}{+0.00} \\ 
Qwen2.5-7B & 784 & 0.28 & 0.009 / 0.039 & 0.153 / 0.018 & 0.29 / 0.24 & $+0.05$ \ci{+0.01}{+0.09} & 7.01 / 0.31 & $+6.70$ \ci{+6.32}{+7.09} \\ 
\bottomrule
\end{tabular}}
\end{table}

\paragraph{Causal ablation.}
The ablation zeroes the attention output of a set of query heads at every position, during the prefill and the decoding.
The full cache then reads the first four \bench{} test episodes of each family at 32k, with the same prompts and decoding as in Section~\ref{sec:exp}.
The unmasked runs reproduce the full-cache predictions of Table~\ref{tab:e2}.
A random head set draws one head from the layer of each detected head and never a detected head.
Llama-3.1-8B and Qwen2.5-7B each get 23 random sets of 16 heads.
Llama also masks its top 64 heads by $\massq$, against ten random sets of 64 heads.
Table~\ref{tab:mech-abl} places the detected heads in this distribution.
Its $\Delta$ is the paired difference between the detected-head score and the mean over the random sets, tested as in Appendix~\ref{app:stats}.
The detected heads cost more than every random set on Qwen2.5 and with 64 heads on Llama.
With 16 heads on Llama they cost more than 20 of the 23 random sets.
The five lowest of these random sets are exactly the five that contain one of two critical heads, (8,9) or (8,11) (Table~\ref{tab:mech-head}).
Masking (8,9) or (8,11) alone costs 16.94 or 10.31 points, while six other heads from the random sets cost at most 0.78.
Without their critical head the five sets score 69.0--86.1, all above the detected heads (54.4).

\begin{table}[!htbp]
\centering
\caption{\textbf{Masking heads in the full cache against random head sets} (\bench{} test at 32k, first four episodes per family). Each random set draws one head from the layer of each detected head. The random column gives the median score and the range over the sets, and the next column counts the sets that score at or below the detected heads. $\Delta$ is the paired difference between the detected heads and the mean over the random sets, with a 95\% CI and Holm $p$ within each model, and $^*$ marks $p<0.05$.}
\label{tab:mech-abl}
\scriptsize
\setlength{\tabcolsep}{3pt}
\begin{tabular}{@{}lcccccccc@{}}
\toprule
Model & heads & none & detected & random, median [range] & random $\le$ detected & $\Delta$ & 95\% CI & Holm $p$ \\
\midrule
Llama-3.1-8B & 16 & 82.5 & 54.4 & 79.3 [52.6, 85.5] & 3/23 & $-21.05^*$ & \ci{-24.88}{-17.20} & 0.0002 \\ 
Llama-3.1-8B & 64 & 82.5 & 7.1 & 48.2 [37.8, 70.7] & 0/10 & $-44.39^*$ & \ci{-48.22}{-40.54} & 0.0002 \\ 
Qwen2.5-7B & 16 & 71.5 & 44.1 & 64.2 [52.8, 71.2] & 0/23 & $-19.40^*$ & \ci{-24.75}{-14.24} & 0.0001 \\ 
\bottomrule
\end{tabular}
\end{table}

\begin{table}[!htbp]
\centering
\caption{\textbf{Single heads behind the most harmful random sets} (Llama-3.1-8B, episodes of Table~\ref{tab:mech-abl}, unmasked score 82.5). The upper rows mask one head, and the other single heads come from the random sets. The lower rows mask a random 16-head set without the critical head, (8,9) or (8,11), that it contains. $\Delta$ is the paired difference to no mask, with a 95\% CI and Holm $p$ over all rows, and $^*$ marks $p<0.05$.}
\label{tab:mech-head}
\scriptsize
\setlength{\tabcolsep}{4pt}
\begin{tabular}{@{}lcccc@{}}
\toprule
Masked heads & score & $\Delta$ & 95\% CI & Holm $p$ \\
\midrule
head (8,9) alone & 65.5 & $-16.94^*$ & \ci{-20.34}{-13.59} & 0.0065 \\ 
head (8,11) alone & 72.2 & $-10.31$ & \ci{-16.25}{-4.38} & 0.1140 \\ 
head (2,16) alone & 82.6 & $+0.16$ & \ci{-1.94}{+2.28} & 1.0000 \\ 
head (8,16) alone & 82.9 & $+0.41$ & \ci{-1.31}{+2.12} & 1.0000 \\ 
head (8,19) alone & 83.4 & $+0.94$ & \ci{-0.16}{+2.34} & 1.0000 \\ 
head (8,24) alone & 81.7 & $-0.78$ & \ci{-3.16}{+1.50} & 1.0000 \\ 
head (17,19) alone & 83.4 & $+0.94$ & \ci{+0.16}{+2.19} & 1.0000 \\ 
head (27,2) alone & 82.3 & $-0.16$ & \ci{-0.47}{+0.00} & 1.0000 \\ 
\midrule
random set 0 without (8,9) & 72.0 & $-10.50^*$ & \ci{-14.81}{-6.12} & 0.0077 \\ 
random set 4 without (8,11) & 80.2 & $-2.28$ & \ci{-4.81}{+0.12} & 1.0000 \\ 
random set 11 without (8,9) & 80.5 & $-1.97$ & \ci{-5.59}{+1.50} & 1.0000 \\ 
random set 15 without (8,11) & 69.0 & $-13.50^*$ & \ci{-17.88}{-9.00} & 0.0072 \\ 
random set 18 without (8,9) & 86.1 & $+3.62$ & \ci{+0.06}{+7.75} & 1.0000 \\ 
\bottomrule
\end{tabular}
\end{table}

\paragraph{SAE view of the match.}
A residual-stream SAE from Llama Scope~\cite{he2024llamascope}, trained on the base model, decomposes the residual that layer 2 of the Instruct model reads. It has 32{,}768 features and an FVU of 0.15 on the first 4{,}096 context positions of a development episode.
With the layer's RMSNorm folded in, the pre-RoPE logit of a layer-2 head splits exactly into feature pairs plus bias and error terms.
For each question and head, we take the single position right after a context match that receives the most attention from any matched question token X.
We then decompose its logit advantage over 512 random positions of the same context, keeping the 2{,}048 context-side features whose activation differs most from those positions.
Table~\ref{tab:mech-sae} shows that SAE feature pairs carry most of this advantage.
The leading pair usually joins a current-token feature of X on the question side with a context feature that over-represents positions after X.
The advantage is spread over many pairs rather than a few.

\begin{table}[!htbp]
\centering
\caption{\textbf{SAE view of the two top Llama heads} (32k development episodes). The share is the part of the logit advantage over random positions carried by feature pairs, as a median over questions. The top-5 column gives the mean share of the positive pair mass in the five largest pairs. The current-token column gives the fraction of questions whose leading question-side feature puts at least half its mass on X. The lift column gives, for the leading context-side feature, the median over-representation of positions after X relative to uniform activation. The last column gives the fraction of questions with a lift of at least 10.}
\label{tab:mech-sae}
\scriptsize
\setlength{\tabcolsep}{4pt}
\begin{tabular}{@{}lcccccc@{}}
\toprule
Head & questions & share by feature pairs & top-5 share & current token X & lift (median) & lift $\ge10$ \\
\midrule
(2,22) & 80 & 0.86 & 0.087 & 0.64 & 14.3 & 0.70 \\ 
(2,21) & 80 & 0.78 & 0.076 & 0.54 & 14.8 & 0.60 \\ 
\bottomrule
\end{tabular}
\end{table}


\FloatBarrier
\section{Chained and multi-paragraph evidence}
\label{app:oracle}

This appendix asks whether \ours{}'s errors on variable tracking and HotpotQA come from missing the evidence or from reading it.
Table~\ref{tab:presence} gives, for four retrieval methods, the share of questions whose full answer lies in the $B{=}2{,}048$ selected tokens and the score among them.
On SQuAD a question whose answer \ours{} selects is answered at least as well as from dense RAG's chunks, on Llama 92.0 against 88.3 at 32k and 91.5 against 91.0 at 128k.
On variable tracking the answer consists of all five variables of a chain, and no method selects all five for more than 1.3\% of the questions at 32k and 128k.
The later statements of a chain share no token with the question, so a single lookup of the question does not reach them.

Table~\ref{tab:oracle} then puts the gold evidence into \ours{}'s own selection and keeps the budget, prompt and decoding.
\ours{}'s selection never holds all five statements of a chain.
With them forced in, variable tracking scores 98.8 to 100.0 on Llama and 89.5 to 92.2 on Qwen2.5, significantly above \ours{} alone and above the full cache at every length.
At 128k every chain spans more than one of the 4{,}096-token chunks that \ours{} encodes separately, 3.1 on average.
The model therefore combines evidence across chunks when it re-reads the selected tokens.
HotpotQA needs two supporting paragraphs, and both lie fully inside \ours{}'s selection for at most 18.8\% of the questions.
Forcing them in raises the HotpotQA score by 8.75 to 22.50 points.
The gap to the full cache on these families thus lies in finding the evidence, not in reading it.

\begin{table}[!htbp]
\centering
\caption{\textbf{Oracle evidence} (\bench{} test episodes, 10 per family, $B{=}2{,}048$, RULER score). \ours{} is its own selection. Forced in keeps \ours{}'s selector and budget but puts the question's gold units into the selection first (VT the five chain statements, HotpotQA the supporting paragraphs, SQuAD the paragraph). Gold only reads the gold units and the selector's lead and tail tokens. All units is the share of questions whose gold units lie fully inside \ours{}'s selection. $\Delta$ is forced in minus \ours{}, with 95\% CI over episodes. $^*$ marks Holm $p<0.05$.}
\label{tab:oracle}
\scriptsize
\setlength{\tabcolsep}{3.5pt}
\resizebox{\ifdim\width>\linewidth\linewidth\else\width\fi}{!}{%
\begin{tabular}{@{}lllcccccc@{}}
\toprule
Model & Family & $L$ & \ours{} & Forced in & Gold only & FullKV & All units & $\Delta$ forced in $-$ \ours{} \\
\midrule
Llama-3.1-8B & VT & 32k & 38.5 & 99.5 & 100.0 & 96.0 & 0.000 & $+61.00^*$ \ci{+58.00}{+64.25} \\ 
 & VT & 64k & 36.5 & 100.0 & 100.0 & 87.8 & 0.000 & $+63.50^*$ \ci{+60.75}{+66.75} \\ 
 & VT & 128k & 35.8 & 98.8 & 100.0 & 60.8 & 0.000 & $+63.00^*$ \ci{+60.50}{+65.25} \\ 
 & HotpotQA & 32k & 55.0 & 67.5 & 73.8 & 60.0 & 0.188 & $+12.50$ \ci{+5.00}{+18.75} \\ 
 & HotpotQA & 64k & 55.0 & 63.8 & 67.5 & 58.8 & 0.087 & $+8.75$ \ci{+1.25}{+16.25} \\ 
 & HotpotQA & 128k & 48.8 & 63.8 & 65.0 & 43.8 & 0.125 & $+15.00$ \ci{+5.00}{+26.25} \\ 
 & SQuAD & 32k & 86.2 & 90.0 & 85.0 & 87.5 & 0.013 & $+3.75$ \ci{-2.50}{+10.00} \\ 
 & SQuAD & 64k & 76.2 & 87.5 & 91.2 & 81.2 & 0.075 & $+11.25^*$ \ci{+6.25}{+16.25} \\ 
 & SQuAD & 128k & 81.2 & 91.2 & 97.5 & 76.2 & 0.087 & $+10.00^*$ \ci{+5.00}{+15.00} \\ 
\midrule
Qwen2.5-7B & VT & 32k & 29.8 & 91.2 & 99.8 & 66.2 & 0.000 & $+61.50^*$ \ci{+57.25}{+65.25} \\ 
 & VT & 64k & 29.2 & 89.5 & 100.0 & 44.5 & 0.000 & $+60.25^*$ \ci{+55.00}{+65.50} \\ 
 & VT & 128k & 23.0 & 92.2 & 100.0 & 39.5 & 0.000 & $+69.25^*$ \ci{+65.25}{+72.75} \\ 
 & HotpotQA & 32k & 52.5 & 67.5 & 72.5 & 53.8 & 0.125 & $+15.00^*$ \ci{+6.25}{+25.00} \\ 
 & HotpotQA & 64k & 48.8 & 67.5 & 71.2 & 45.0 & 0.075 & $+18.75^*$ \ci{+12.50}{+25.00} \\ 
 & HotpotQA & 128k & 37.5 & 60.0 & 68.8 & 42.5 & 0.037 & $+22.50^*$ \ci{+11.25}{+33.75} \\ 
 & SQuAD & 32k & 72.5 & 81.2 & 85.0 & 72.5 & 0.013 & $+8.75$ \ci{+2.50}{+15.00} \\ 
 & SQuAD & 64k & 70.0 & 83.8 & 93.8 & 66.2 & 0.087 & $+13.75^*$ \ci{+6.25}{+22.50} \\ 
 & SQuAD & 128k & 68.8 & 86.2 & 90.0 & 52.5 & 0.050 & $+17.50^*$ \ci{+8.75}{+27.50} \\ 
\bottomrule
\end{tabular}}
\end{table}

\begin{table}[!htbp]
\centering
\caption{\textbf{Answer presence and score given it} (\bench{} test episodes, 80 questions per cell, $B{=}2{,}048$). Full answer is the share of questions whose reference answers all occur in the selected tokens. If full is the RULER score among those questions, and -- marks an empty cell.}
\label{tab:presence}
\scriptsize
\setlength{\tabcolsep}{3.5pt}
\resizebox{\ifdim\width>\linewidth\linewidth\else\width\fi}{!}{%
\begin{tabular}{@{}lllcccccc@{}}
\toprule
& & & \multicolumn{3}{c}{32k} & \multicolumn{3}{c}{128k} \\
\cmidrule(lr){4-6}\cmidrule(l){7-9}
Model & Family & Method & Score & Full answer & If full & Score & Full answer & If full \\
\midrule
Llama-3.1-8B & VT & \ours{} & 38.5 & 0.000 & -- & 35.8 & 0.013 & 60.0 \\ 
 & VT & \ours{}+BM25 & 38.5 & 0.000 & -- & 35.8 & 0.013 & 60.0 \\ 
 & VT & Dense RAG + reranker & 27.0 & 0.013 & 100.0 & 24.0 & 0.000 & -- \\ 
 & VT & BM25, same selector & 25.0 & 0.000 & -- & 19.5 & 0.000 & -- \\ 
\cmidrule(l){2-9}
 & HotpotQA & \ours{} & 55.0 & 0.887 & 62.0 & 48.8 & 0.762 & 54.1 \\ 
 & HotpotQA & \ours{}+BM25 & 60.0 & 0.925 & 63.5 & 53.8 & 0.850 & 57.4 \\ 
 & HotpotQA & Dense RAG + reranker & 72.5 & 0.950 & 75.0 & 57.5 & 0.875 & 61.4 \\ 
 & HotpotQA & BM25, same selector & 56.2 & 0.787 & 68.3 & 50.0 & 0.787 & 58.7 \\ 
\cmidrule(l){2-9}
 & SQuAD & \ours{} & 86.2 & 0.938 & 92.0 & 81.2 & 0.887 & 91.5 \\ 
 & SQuAD & \ours{}+BM25 & 88.8 & 0.963 & 92.2 & 85.0 & 0.963 & 88.3 \\ 
 & SQuAD & Dense RAG + reranker & 85.0 & 0.963 & 88.3 & 88.8 & 0.975 & 91.0 \\ 
 & SQuAD & BM25, same selector & 78.8 & 0.875 & 90.0 & 76.2 & 0.875 & 87.1 \\ 
\midrule
Qwen2.5-7B & VT & \ours{} & 29.8 & 0.000 & -- & 23.0 & 0.013 & 60.0 \\ 
 & VT & \ours{}+BM25 & 29.8 & 0.000 & -- & 23.0 & 0.013 & 60.0 \\ 
 & VT & Dense RAG + reranker & 25.2 & 0.013 & 60.0 & 22.2 & 0.000 & -- \\ 
 & VT & BM25, same selector & 24.2 & 0.000 & -- & 19.2 & 0.000 & -- \\ 
\cmidrule(l){2-9}
 & HotpotQA & \ours{} & 52.5 & 0.838 & 61.2 & 37.5 & 0.650 & 53.8 \\ 
 & HotpotQA & \ours{}+BM25 & 61.2 & 0.900 & 63.9 & 51.2 & 0.863 & 55.1 \\ 
 & HotpotQA & Dense RAG + reranker & 55.0 & 0.938 & 57.3 & 52.5 & 0.887 & 57.7 \\ 
 & HotpotQA & BM25, same selector & 50.0 & 0.863 & 55.1 & 51.2 & 0.863 & 55.1 \\ 
\cmidrule(l){2-9}
 & SQuAD & \ours{} & 72.5 & 0.875 & 82.9 & 68.8 & 0.775 & 87.1 \\ 
 & SQuAD & \ours{}+BM25 & 78.8 & 0.950 & 82.9 & 81.2 & 0.925 & 87.8 \\ 
 & SQuAD & Dense RAG + reranker & 76.2 & 0.975 & 78.2 & 81.2 & 0.963 & 84.4 \\ 
 & SQuAD & BM25, same selector & 72.5 & 0.863 & 84.1 & 75.0 & 0.912 & 82.2 \\ 
\bottomrule
\end{tabular}}
\end{table}


\input{v3_sections/app_scaling}

\section{SAE codec and features}
\label{app:sae}

\paragraph{Training.}
One Top-$k$ SAE~\cite{gao2024scaling} is trained per KV head in $\Gset$ (14 for Llama-3.1-8B-Instruct) on that head's pre-RoPE keys.
The training set holds about 2.4M keys from sink-prefixed 511-token WikiText-2 windows~\cite{merity2017pointer}.
Each SAE has $D=8{,}192$ latents (expansion 64 over $d=128$) and $k=32$.
It is trained with the MSE loss for 20k Adam steps with batch 4{,}096 and learning rate $10^{-3}$.
Encoding and decoding follow Eq.~\eqref{eq:sae}.
At query time, scoring needs only $W_{\mathrm{dec}}^{\top}\vq$ per question token and a gather over the stored ids.
The term $\vq^{\top}\mathbf{b}_{\mathrm{dec}}$ cancels in the softmax.
Storage is 32 int16 ids and 32 fp16 values per head (128\,B), or uint8 values with an fp16 per-token scale (98\,B).
Mean FVU on WikiText-2 validation keys is 0.419, 0.286, 0.168 and 0.069 for $k=4,8,16,32$. 
Table~\ref{tab:codec-full} gives the resulting development presence. 

\paragraph{Use as a codec.}
\saec{} is an alternative sparse codec from the same code family as the \Sthree{}-Attention index.
On development episodes, it reaches the presence of exact keys (0.801 and 0.822 against 0.801 and 0.809) at four times the bytes of \pqc{}.
It is not significantly different from \pqc{} in eight of the nine test comparisons (Appendix~\ref{app:codec-test}). 

\paragraph{Feature voting on the detected heads (Tables~\ref{tab:sae-scoring} and~\ref{tab:invidx}).}
The \Sthree{}-Attention index~\cite{ma2026s3attention} stores SAE feature ids in an inverted index.
It weights the votes of the question's features by inverse document frequency and drops very frequent features.
Table~\ref{tab:sae-scoring} applies these choices one at a time to the SAE codes of the 16 detected heads.
The reweighted codes are scored with \ours{}'s exact question queries and per-head softmax.
Decoded \saec{} codes reproduce exact keys (0.801 and 0.822 against 0.801 and 0.809).
Binary feature ids value each active feature at the head's mean activation.
They keep most of the signal (0.769 and 0.779).
IDF weighting lowers presence at 32k (0.751) and removes the signal at 128k (0.143).
The IDF form of the \Sthree{}-Attention index behaves the same way (0.784 and 0.238).
IDF on binary ids lowers presence to 0.613 at 32k and removes the signal at 128k (0.108).
Dropping the features active at more than 5\% of the positions removes it at both lengths (0.226 and 0.108).
The same holds for a threshold of 1\% (0.210 and 0.079).
SAE-16 and SAE-8 codes follow the same pattern.
Table~\ref{tab:invidx} serves the binary ids from an inverted index.
For each question token, the index reads the postings of only the top-$m$ features ranked by $\vq^{\top}\mathbf{d}_f$ (the decoder atom of feature $f$).
With $m=1{,}024$, 256 and 64 the index reads 8.6\%, 3.3\% and 1.5\% of the postings at 32k.
It keeps 0.464, 0.340 and 0.314 at 32k and 0.276, 0.220 and 0.160 at 128k.
The full scan of the same ids keeps 0.769 and 0.779.
On an idle GPU at 128k, the lookup with $m=64$ takes 0.029\,s per question against 0.053\,s for the full scan.
A one-time postings build of 0.163\,s precedes it.
With 1M tokens, these times are 0.049\,s, 0.190\,s and 0.393\,s.
Feature voting therefore fails through its aggregation over all heads (Section~\ref{sec:ana-ladder}), its IDF and stop-feature weighting and its pruned lookup.
It does not fail through the SAE codes themselves.

\begin{table}[!htbp]
\centering
\caption{\textbf{SAE feature scoring on the 16 detected heads} (Llama-3.1-8B-Instruct, development answer presence of the $B{=}2{,}048$ selection, 16 questions per family cell). All rows use chunk-local keys, the bare question and a per-head softmax. Rows with stored activations use the decoded SAE key. Binary feature ids value every active feature at the head's mean activation. This is the storage of the \Sthree{}-Attention index. IDF multiplies the activations by $\log\frac{L+1}{\mathrm{df}_f+1}$ and normalizes them. \Sthree{}-Attention IDF multiplies them by $\frac{1}{\log(1+\mathrm{df}_f)+1}$. Rows with df $>xL$ dropped remove the features active at more than a share $x$ of the $L$ positions.}
\label{tab:sae-scoring}
\scriptsize
\setlength{\tabcolsep}{4pt}
\begin{tabular}{@{}lcccccc@{}}
\toprule
Codes and scoring & NIAH-MK & NIAH-MV & VT & QA-SQuAD & QA-Hotpot & Mean \\
\midrule
\multicolumn{7}{@{}l}{\textbf{32k tokens}} \\
exact keys (reference) & 1.000 & 0.906 & 0.413 & 1.000 & 0.688 & 0.801 \\ 
\saec{}, stored activations (decoded keys) & 1.000 & 0.906 & 0.413 & 1.000 & 0.688 & 0.801 \\ 
\saec{}, binary feature ids & 1.000 & 0.906 & 0.438 & 0.938 & 0.562 & 0.769 \\ 
\saec{}, IDF-weighted & 1.000 & 0.906 & 0.413 & 1.000 & 0.438 & 0.751 \\ 
\saec{}, \Sthree{}-Attention IDF & 1.000 & 0.906 & 0.388 & 1.000 & 0.625 & 0.784 \\ 
\saec{}, binary ids, IDF-weighted & 1.000 & 0.641 & 0.425 & 0.625 & 0.375 & 0.613 \\ 
\saec{}, binary ids, \Sthree{}-Attention IDF & 1.000 & 0.906 & 0.413 & 0.875 & 0.562 & 0.751 \\ 
\saec{}, features with df $>0.05L$ dropped & 0.062 & 0.078 & 0.363 & 0.438 & 0.188 & 0.226 \\ 
\saec{}, features with df $>0.01L$ dropped & 0.000 & 0.062 & 0.425 & 0.375 & 0.188 & 0.210 \\ 
SAE-16, stored activations & 1.000 & 0.906 & 0.388 & 0.938 & 0.625 & 0.771 \\ 
SAE-16, IDF-weighted & 1.000 & 0.891 & 0.400 & 1.000 & 0.562 & 0.771 \\ 
SAE-16, \Sthree{}-Attention IDF & 1.000 & 0.906 & 0.388 & 1.000 & 0.562 & 0.771 \\ 
SAE-16, df $>0.05L$ dropped & 0.125 & 0.141 & 0.212 & 0.750 & 0.562 & 0.358 \\ 
SAE-16, df $>0.01L$ dropped & 0.062 & 0.047 & 0.400 & 0.500 & 0.562 & 0.314 \\ 
SAE-8, stored activations & 1.000 & 0.906 & 0.388 & 0.875 & 0.562 & 0.746 \\ 
SAE-8, IDF-weighted & 0.875 & 0.500 & 0.400 & 1.000 & 0.562 & 0.667 \\ 
SAE-8, \Sthree{}-Attention IDF & 1.000 & 0.891 & 0.338 & 0.938 & 0.625 & 0.758 \\ 
SAE-8, df $>0.05L$ dropped & 0.188 & 0.156 & 0.087 & 0.562 & 0.750 & 0.349 \\ 
SAE-8, df $>0.01L$ dropped & 0.062 & 0.047 & 0.387 & 0.375 & 0.375 & 0.249 \\ 
\midrule
\multicolumn{7}{@{}l}{\textbf{128k tokens}} \\
exact keys (reference) & 1.000 & 0.969 & 0.512 & 0.875 & 0.688 & 0.809 \\ 
\saec{}, stored activations (decoded keys) & 1.000 & 0.984 & 0.500 & 0.875 & 0.750 & 0.822 \\ 
\saec{}, binary feature ids & 1.000 & 0.984 & 0.537 & 0.875 & 0.500 & 0.779 \\ 
\saec{}, IDF-weighted & 0.000 & 0.016 & 0.325 & 0.312 & 0.062 & 0.143 \\ 
\saec{}, \Sthree{}-Attention IDF & 0.000 & 0.078 & 0.300 & 0.500 & 0.312 & 0.238 \\ 
\saec{}, binary ids, IDF-weighted & 0.000 & 0.016 & 0.400 & 0.062 & 0.062 & 0.108 \\ 
\saec{}, binary ids, \Sthree{}-Attention IDF & 0.062 & 0.047 & 0.375 & 0.500 & 0.250 & 0.247 \\ 
\saec{}, features with df $>0.05L$ dropped & 0.000 & 0.016 & 0.213 & 0.250 & 0.062 & 0.108 \\ 
\saec{}, features with df $>0.01L$ dropped & 0.000 & 0.047 & 0.162 & 0.188 & 0.000 & 0.079 \\ 
SAE-16, stored activations & 1.000 & 0.969 & 0.450 & 0.750 & 0.625 & 0.759 \\ 
SAE-16, IDF-weighted & 0.000 & 0.047 & 0.338 & 0.375 & 0.188 & 0.189 \\ 
SAE-16, \Sthree{}-Attention IDF & 0.000 & 0.047 & 0.287 & 0.375 & 0.250 & 0.192 \\ 
SAE-16, df $>0.05L$ dropped & 0.000 & 0.000 & 0.125 & 0.250 & 0.188 & 0.113 \\ 
SAE-16, df $>0.01L$ dropped & 0.000 & 0.000 & 0.250 & 0.125 & 0.062 & 0.087 \\ 
SAE-8, stored activations & 0.875 & 0.891 & 0.388 & 0.500 & 0.375 & 0.606 \\ 
SAE-8, IDF-weighted & 0.000 & 0.031 & 0.338 & 0.188 & 0.125 & 0.136 \\ 
SAE-8, \Sthree{}-Attention IDF & 0.000 & 0.031 & 0.288 & 0.375 & 0.188 & 0.176 \\ 
SAE-8, df $>0.05L$ dropped & 0.000 & 0.094 & 0.075 & 0.188 & 0.188 & 0.109 \\ 
SAE-8, df $>0.01L$ dropped & 0.000 & 0.000 & 0.350 & 0.375 & 0.000 & 0.145 \\ 
\bottomrule
\end{tabular}
\end{table}

\begin{table}[!htbp]
\centering
\caption{\textbf{Serving binary SAE ids from an inverted index} (\saec{} feature ids of the 16 detected heads, Llama-3.1-8B-Instruct, development answer presence, $B{=}2{,}048$, 16 questions per family cell). For each question token, the index reads only the postings of the top-$m$ features by $\vq^{\top}\mathbf{d}_f$. The dense scan reads every posting. The Read column gives the postings read divided by the postings of the dense scan.}
\label{tab:invidx}
\scriptsize
\setlength{\tabcolsep}{4pt}
\begin{tabular}{@{}lccccccc@{}}
\toprule
Lookup & NIAH-MK & NIAH-MV & VT & QA-SQuAD & QA-Hotpot & Mean & Read \\
\midrule
\multicolumn{8}{@{}l}{\textbf{32k tokens}} \\
binary ids, every posting (dense scan) & 1.000 & 0.906 & 0.438 & 0.938 & 0.562 & 0.769 & 1.000 \\ 
inverted index, top-1{,}024 features per question token & 0.688 & 0.297 & 0.463 & 0.438 & 0.438 & 0.464 & 0.086 \\ 
inverted index, top-256 & 0.375 & 0.188 & 0.513 & 0.438 & 0.188 & 0.340 & 0.033 \\ 
inverted index, top-64 & 0.375 & 0.156 & 0.537 & 0.312 & 0.188 & 0.314 & 0.015 \\ 
\midrule
\multicolumn{8}{@{}l}{\textbf{128k tokens}} \\
binary ids, every posting (dense scan) & 1.000 & 0.984 & 0.537 & 0.875 & 0.500 & 0.779 & 1.000 \\ 
inverted index, top-1{,}024 features per question token & 0.438 & 0.219 & 0.413 & 0.250 & 0.062 & 0.276 & 0.085 \\ 
inverted index, top-256 & 0.250 & 0.125 & 0.413 & 0.250 & 0.062 & 0.220 & 0.032 \\ 
inverted index, top-64 & 0.125 & 0.062 & 0.425 & 0.125 & 0.062 & 0.160 & 0.015 \\ 
\bottomrule
\end{tabular}
\end{table}


\section{Question specificity and the in-context reference}\label{sec:ana-incontext}\label{app:incontext}

This appendix complements Section~\ref{sec:ana-ladder} (development answer presence, Llama-3.1-8B-Instruct, $B{=}2{,}048$).

Each \bench{} context holds all eight answers.
We therefore compare a question's own presence with its presence in the other questions' selections (exact keys).
Multi-key NIAH uses one needle template and is question-agnostic for both methods (\ours{} 1.00/1.00, BM25 0.94/0.94 and 0.94/0.88).
\ours{}'s variable-tracking edge is mostly question-agnostic (0.41/0.39, 0.51/0.35).
QA selections are question-specific, and \ours{} has the higher own presence at 128k (SQuAD 0.88 vs.\ 0.81, HotpotQA 0.69 vs.\ 0.56).
In context (block (v), GemFilter-style~\cite{shi2024gemfilter}, not query-agnostic), the top-16 heads reach 0.913/0.812 and raw layer-13 logits reach 0.849/0.728.
At 128k, \ours{} (0.809) reaches the in-context heads and exceeds the in-context layer-13 logits.
At 32k, the gap comes from the query, not the keys.
Indeed, an out-of-context last-token query on in-context keys gives 0.634/0.285, close to or below its presence on chunk-local keys (0.619/0.478).


\section{Efficiency}
\label{app:eff}

Two benchmarks run on one otherwise idle H200 NVL GPU (141\,GB).
Both use Llama-3.1-8B-Instruct with one HotpotQA \bench{} test episode of 8 questions per length.
Timings are CUDA-synchronised.
Peak memory is the maximum allocated GPU memory during a step minus the memory allocated before it.
The LLM weights are therefore excluded.

\paragraph{End to end (Figure~\ref{fig:cost}).}
Table~\ref{tab:e2e} times every stage in the configuration of the accuracy runs.
The build runs once per context.
Questions are processed one at a time.
For each question, the table times the retrieval and the greedy answer over the $B{=}2{,}048$ selected tokens inside the task prompt.
FullKV prefills the prompt prefix and the context once and decodes each question on its cache.
FullKV is not run at 1M, where its cache takes 128\,GiB.
The total for $Q$ questions is build $+\,Q\cdot$(retrieval $+$ answer).
\ourstwo{}'s build plus retrieval (without the answer) undercuts dense RAG from the first question at 32k and from six questions at 128k.
With the answer, its total undercuts dense RAG from the first question at 32k and from five questions at 128k.
It undercuts FullKV at every $Q$ at both lengths.
At 32k, decoding on the cache is cheap (0.157\,s per question), yet the 16-head \ours{} also has a lower total than FullKV at every $Q$.
Building several chunks per forward (8 for 16 heads, 32 for two heads) barely shortens the build.
With 16 heads at 128k, the batched build takes 5.685\,s vs.\ 5.729\,s for one chunk per forward.
With two heads, it takes 0.619 vs.\ 0.664\,s, at 14.78 instead of 0.49\,GiB peak memory.
The time gain is small because each 4{,}097-token chunk already occupies the GPU.
The batched builds select the same tokens as one chunk per forward for all eight questions.
Every reported run builds one chunk per forward.

\begin{table}[!htbp]
\centering
\caption{\textbf{End-to-end cost} (Llama-3.1-8B-Instruct, idle H200 NVL 141\,GB, one HotpotQA \bench{} test episode of 8 questions per length, $B{=}2{,}048$). Build runs once per context. For FullKV, it is the prefill of prompt prefix and context. Retrieval and answer are means per question. For FullKV, these columns give decoding on the cache. Peak is the GPU memory of the build without weights. Total$(Q)$ = build $+\,Q\cdot$(retrieval $+$ answer), in seconds.}
\label{tab:e2e}
\scriptsize
\setlength{\tabcolsep}{3pt}
\resizebox{\ifdim\width>\linewidth\linewidth\else\width\fi}{!}{%
\begin{tabular}{@{}llrrrrrrrrrr@{}}
\toprule
 & & & & & & & \multicolumn{5}{c}{Total for $Q$ questions (s)} \\
\cmidrule(l){8-12}
$L$ (tokens) & Method & B/token & Build (s) & Retrieval (s) & Answer (s) & Peak (GiB) & $Q{=}1$ & 8 & 32 & 128 & 512 \\
\midrule
32k (31{,}988) & FullKV & 131{,}072 & 1.764 & -- & 0.157 & 8.63 & 1.92 & 3.02 & 6.80 & 21.90 & 82.30 \\ 
 & Dense RAG + reranker & -- & 0.655 & 0.113 & 0.125 & 0.14 & 0.89 & 2.56 & 8.26 & 31.07 & 122.32 \\ 
 & BM25 & -- & -- & 0.006 & 0.108 & -- & 0.11 & 0.91 & 3.63 & 14.54 & 58.16 \\ 
 & \ours{} (\pqc{}) & 448 & 1.395 & 0.024 & 0.115 & 0.67 & 1.53 & 2.51 & 5.84 & 19.17 & 72.50 \\ 
 & \ourstwo{} & 32 & 0.164 & 0.006 & 0.149 & 0.49 & 0.32 & 1.40 & 5.11 & 19.97 & 79.37 \\ 
\midrule
128k (130{,}296) & FullKV & 131{,}072 & 17.695 & -- & 0.505 & 31.55 & 18.20 & 21.73 & 33.84 & 82.29 & 276.06 \\ 
 & Dense RAG + reranker & -- & 0.234 & 0.088 & 0.126 & 0.15 & 0.45 & 1.95 & 7.08 & 27.63 & 109.82 \\ 
 & BM25 & -- & -- & 0.023 & 0.109 & -- & 0.13 & 1.05 & 4.21 & 16.86 & 67.43 \\ 
 & \ours{} (\pqc{}) & 448 & 5.729 & 0.038 & 0.105 & 0.71 & 5.87 & 6.87 & 10.30 & 24.01 & 78.85 \\ 
 & \ourstwo{} & 32 & 0.664 & 0.009 & 0.111 & 0.49 & 0.78 & 1.63 & 4.51 & 16.04 & 62.19 \\ 
\midrule
1M (1{,}047{,}931) & Dense RAG + reranker & -- & 1.720 & 0.085 & 0.140 & 0.16 & 1.94 & 3.52 & 8.91 & 30.47 & 116.71 \\ 
 & BM25 & -- & -- & 0.200 & 0.175 & -- & 0.38 & 3.00 & 12.01 & 48.03 & 192.11 \\ 
 & \ours{} (\pqc{}) & 448 & 46.264 & 0.341 & 0.170 & 1.09 & 46.77 & 50.35 & 62.60 & 111.60 & 307.61 \\ 
 & \ourstwo{} & 32 & 5.358 & 0.058 & 0.176 & 0.52 & 5.59 & 7.23 & 12.85 & 35.32 & 125.22 \\ 
\bottomrule
\end{tabular}}
\end{table}

\paragraph{Components.}
Table~\ref{tab:eff} comes from a separate run under the same protocol, with its own baselines.
It adds the \saec{} codec and the peak memory of retrieval.
With 16 heads (deepest layer 27), the build takes 1.401\,s, 5.720\,s and 46.359\,s at 32k, 128k and 1M with \pqc{} codes, and 1.467\,s, 6.000\,s and 48.528\,s with \saec{} codes.
The \pqc{} builds agree with Table~\ref{tab:e2e} within 1\%, and their per-question retrieval times within 25\%.
Section~\ref{sec:exp-kv} quotes Table~\ref{tab:e2e}.

\begin{table}[!htbp]
\centering
\caption{\textbf{Cost per context and per question on an idle GPU} (Llama-3.1-8B-Instruct, one H200 NVL 141\,GB running no other job, one HotpotQA test episode of 8 questions per length). This is a separate run from Table~\ref{tab:e2e}, whose numbers the main text quotes. Index is the one-time build per context. For \ours{}, it is the chunk-local forward up to the deepest retrieval layer plus key encoding. For dense RAG, it covers the bge-small-en-v1.5 embeddings of 256-token chunks. Retrieval / q is the time per question. For dense RAG, it covers question embedding and reranking of the top 100 chunks. BM25 recomputes its chunk statistics per question. Peak GPU is the maximum allocated memory during the step minus the memory before it, with LLM weights excluded. -- marks values that are not recorded or not applicable.}
\label{tab:eff}
\scriptsize
\setlength{\tabcolsep}{4pt}
\resizebox{\ifdim\width>\linewidth\linewidth\else\width\fi}{!}{%
\begin{tabular}{@{}llrrrrr@{}}
\toprule
& & & & \multicolumn{2}{c}{Peak GPU (GiB)} & \\
\cmidrule(lr){5-6}
$L$ (tokens) & Method & Index (s) & Retrieval / q (s) & index & retrieval & B/token \\
\midrule
32k (31{,}988) & \ours{}, \pqc{} & 1.401 & 0.029 & 0.670 & 0.021 & 448 \\ 
 & \ours{}, \saec{} & 1.467 & 0.019 & 0.701 & 0.132 & 1{,}792 \\ 
 & Dense RAG + reranker & 0.125 & 0.089 & 0.143 & -- & -- \\ 
 & BM25 & -- & 0.005 & -- & -- & -- \\ 
\midrule
128k (130{,}296) & \ours{}, \pqc{} & 5.720 & 0.038 & 0.711 & 0.075 & 448 \\ 
 & \ours{}, \saec{} & 6.000 & 0.040 & 0.865 & 0.525 & 1{,}792 \\ 
 & Dense RAG + reranker & 0.228 & 0.087 & 0.153 & -- & -- \\ 
 & BM25 & -- & 0.021 & -- & -- & -- \\ 
\midrule
1M (1{,}047{,}931) & \ours{}, \pqc{} & 46.359 & 0.326 & 1.094 & 0.573 & 448 \\ 
 & \ours{}, \saec{} & 48.528 & 0.184 & 3.579 & 4.133 & 1{,}792 \\ 
 & Dense RAG + reranker & 1.697 & 0.085 & 0.158 & -- & -- \\ 
 & BM25 & -- & 0.207 & -- & -- & -- \\ 
\bottomrule
\end{tabular}}
\end{table}

%% file: v3_sections/app_stack.tex
\section{Combining \ours{} with KV compression and dense retrieval}
\label{app:stack}

These runs support Section~\ref{sec:exp-stack} and Table~\ref{tab:stack}.
All of them reuse the \ours{} index, heads, selector and decoding of the main results.

\paragraph{KV compression.}
A kvpress compression method (a \emph{press}) with default settings compresses the prefilled context to 1{,}024 entries before any question.
For each question, \ours{} selects 1{,}024 tokens with the lead and tail of the selector scaled to this budget.
These tokens are fed after the compressed cache in 64-token pieces, with positions continuing after the context, and are followed by the question.
The cache is cut back after each answer.
The press alone at 1{,}024 entries uses the same cache without the selected tokens, and the press at 2{,}048 entries is its Llama row of Table~\ref{tab:e2} restricted to the same episodes.
At 128k every combined prompt decodes up to 625 positions past the 131{,}072-position window.
Table~\ref{tab:stack-kv} gives the results.

\begin{table}[!htbp]
\centering
\caption{\textbf{KV compression + \ours{}} (Llama-3.1-8B, \bench{} test episodes 0--5 of every family, 240 questions per length, macro score). A press @\,2{,}048 keeps 2{,}048 cache entries, @\,1{,}024 keeps 1{,}024, and + \ours{} @\,1{,}024+1{,}024 adds the 1{,}024 tokens \ours{} selects. Each $\Delta$ cell gives the paired difference, its 95\% CI and Holm $p$ within each length. $^*$ marks Holm $p<0.05$. Every row uses the same episodes.}
\label{tab:stack-kv}
\scriptsize
\setlength{\tabcolsep}{3.5pt}
\resizebox{\ifdim\width>\linewidth\linewidth\else\width\fi}{!}{%
\begin{tabular}{@{}lccc@{}}
\toprule
Method & 32k & 64k & 128k \\
\midrule
\ours{} @\,2{,}048 & 74.8 & 72.1 & 72.4 \\
\ours{} @\,1{,}024 & 71.7 & 68.6 & 67.0 \\
StreamingLLM @\,2{,}048 & 14.1 & 12.1 & 6.9 \\
StreamingLLM @\,1{,}024 & 9.6 & 8.5 & 5.1 \\
StreamingLLM + \ours{} @\,1{,}024+1{,}024 & 71.3 & 68.1 & 66.4 \\
SnapKV @\,2{,}048 & 12.5 & 13.4 & 10.3 \\
SnapKV @\,1{,}024 & 8.1 & 7.9 & 9.0 \\
SnapKV + \ours{} @\,1{,}024+1{,}024 & 71.6 & 66.8 & 68.7 \\
KVzip @\,2{,}048 & 57.1 & 22.3 & 8.5 \\
KVzip @\,1{,}024 & 30.7 & 12.8 & 7.6 \\
KVzip + \ours{} @\,1{,}024+1{,}024 & 77.1 & 69.1 & 23.5 \\
\midrule
$\Delta$ KVzip + \ours{} @\,1{,}024+1{,}024 $-$ KVzip @\,2{,}048 & $+22.33^*$ \ci{+17.60}{+26.81} 0.0012 & $+48.04^*$ \ci{+43.92}{+52.12} 0.0012 & $+15.52^*$ \ci{+10.85}{+20.48} 0.0020 \\
$\Delta$ KVzip + \ours{} @\,1{,}024+1{,}024 $-$ KVzip @\,1{,}024 & $+46.46^*$ \ci{+41.83}{+50.71} 0.0012 & $+56.25^*$ \ci{+52.12}{+60.23} 0.0012 & $+15.94^*$ \ci{+11.69}{+20.21} 0.0012 \\
$\Delta$ KVzip + \ours{} @\,1{,}024+1{,}024 $-$ \ours{} @\,2{,}048 & $+2.29$ \ci{-1.81}{+6.25} 1.0000 & $-3.04$ \ci{-6.69}{+0.58} 0.8075 & $-48.92^*$ \ci{-54.29}{-43.81} 0.0012 \\
$\Delta$ KVzip + \ours{} @\,1{,}024+1{,}024 $-$ \ours{} @\,1{,}024 & $+5.46$ \ci{+0.92}{+9.85} 0.8031 & $+0.46$ \ci{-2.56}{+3.44} 1.0000 & $-43.44^*$ \ci{-49.25}{-37.52} 0.0012 \\
$\Delta$ SnapKV + \ours{} @\,1{,}024+1{,}024 $-$ SnapKV @\,2{,}048 & $+59.77^*$ \ci{+55.08}{+64.17} 0.0012 & $+56.56^*$ \ci{+51.96}{+60.83} 0.0012 & $+58.79^*$ \ci{+53.48}{+63.96} 0.0012 \\
$\Delta$ SnapKV + \ours{} @\,1{,}024+1{,}024 $-$ SnapKV @\,1{,}024 & $+63.46^*$ \ci{+59.71}{+67.10} 0.0012 & $+58.83^*$ \ci{+54.35}{+63.00} 0.0012 & $+59.71^*$ \ci{+54.38}{+64.65} 0.0012 \\
$\Delta$ SnapKV + \ours{} @\,1{,}024+1{,}024 $-$ \ours{} @\,2{,}048 & $-3.23$ \ci{-6.31}{-0.42} 0.3750 & $-5.38^*$ \ci{-7.92}{-2.92} 0.0078 & $-3.73$ \ci{-6.69}{-0.98} 0.1314 \\
$\Delta$ SnapKV + \ours{} @\,1{,}024+1{,}024 $-$ \ours{} @\,1{,}024 & $-0.06$ \ci{-2.06}{+1.98} 1.0000 & $-1.88$ \ci{-3.98}{+0.19} 0.6299 & $+1.75$ \ci{-0.42}{+3.92} 0.5883 \\
$\Delta$ StreamingLLM + \ours{} @\,1{,}024+1{,}024 $-$ StreamingLLM @\,2{,}048 & $+59.56^*$ \ci{+56.12}{+62.92} 0.0012 & $+56.73^*$ \ci{+52.00}{+61.25} 0.0012 & $+58.77^*$ \ci{+56.12}{+61.21} 0.0012 \\
$\Delta$ StreamingLLM + \ours{} @\,1{,}024+1{,}024 $-$ StreamingLLM @\,1{,}024 & $+61.62^*$ \ci{+58.25}{+65.02} 0.0012 & $+59.58^*$ \ci{+54.42}{+64.35} 0.0012 & $+61.27^*$ \ci{+57.98}{+64.62} 0.0012 \\
$\Delta$ StreamingLLM + \ours{} @\,1{,}024+1{,}024 $-$ \ours{} @\,2{,}048 & $-3.56$ \ci{-7.35}{+0.12} 0.5194 & $-4.02^*$ \ci{-6.44}{-1.79} 0.0370 & $-6.06^*$ \ci{-9.08}{-3.02} 0.0340 \\
$\Delta$ StreamingLLM + \ours{} @\,1{,}024+1{,}024 $-$ \ours{} @\,1{,}024 & $-0.40$ \ci{-3.25}{+2.08} 1.0000 & $-0.52$ \ci{-2.35}{+1.29} 1.0000 & $-0.58$ \ci{-3.98}{+3.02} 0.7631 \\
\bottomrule
\end{tabular}}
\end{table}

\paragraph{Dense retrieval.}
\ours{}+dense mixes the min--max normalized \ours{} score and the bge-small cosine of the 128-token chunk that contains each position with weight 0.5 each.
\ours{}+BM25+dense gives the \ours{}, BM25 and bge-small scores one third each.
\ours{} $\cup$ dense RAG fills half of the budget with \ours{} spans and the rest with reranked dense RAG chunks in rank order.
None of these weights was tuned.
The \ours{}, \ours{}+BM25, BM25 and dense RAG rows were recomputed in the same run and reproduce Table~\ref{tab:e2}.
Tables~\ref{tab:stack-dense-llama} and~\ref{tab:stack-dense-qwen} give \bench{}, and Table~\ref{tab:stack-ib} gives InfiniteBench with the protocol of Appendix~\ref{app:protocol-infbench}.

\begin{table}[!htbp]
\centering
\caption{\textbf{\ours{} fused with dense retrieval} (Llama-3.1-8B, \bench{}, $B{=}2{,}048$, 400 questions per length). Notation as in Table~\ref{tab:stack-kv}, with Holm $p$ within each length.}
\label{tab:stack-dense-llama}
\scriptsize
\setlength{\tabcolsep}{3.5pt}
\resizebox{\ifdim\width>\linewidth\linewidth\else\width\fi}{!}{%
\begin{tabular}{@{}lccc@{}}
\toprule
Method & 32k & 64k & 128k \\
\midrule
\ours{} & 74.0 & 71.6 & 72.1 \\
\ours{}+BM25 & 76.0 & 72.9 & 73.9 \\
\ours{}+dense & 77.1 & 74.8 & 74.6 \\
\ours{}+BM25+dense & 77.9 & 74.3 & 75.2 \\
\ours{} $\cup$ dense RAG & 76.2 & 72.9 & 72.8 \\
dense RAG + reranker & 72.6 & 66.7 & 66.2 \\
BM25 & 67.7 & 65.3 & 66.0 \\
\midrule
$\Delta$ \ours{}+dense $-$ dense RAG + reranker & $+4.49$ \ci{+1.73}{+7.31} 0.0896 & $+8.05^*$ \ci{+4.92}{+11.15} 0.0010 & $+8.43^*$ \ci{+5.26}{+11.62} 0.0015 \\
$\Delta$ \ours{}+dense $-$ \ours{} & $+3.06$ \ci{+0.79}{+5.46} 0.0896 & $+3.16^*$ \ci{+1.02}{+5.19} 0.0216 & $+2.51$ \ci{+0.33}{+4.89} 0.1461 \\
$\Delta$ \ours{}+dense $-$ BM25 & $+9.39^*$ \ci{+6.54}{+12.29} 0.0008 & $+9.51^*$ \ci{+7.11}{+11.99} 0.0008 & $+8.57^*$ \ci{+5.53}{+11.59} 0.0008 \\
$\Delta$ \ours{}+BM25+dense $-$ dense RAG + reranker & $+5.28$ \ci{+2.44}{+8.16} 0.0522 & $+7.58^*$ \ci{+4.70}{+10.53} 0.0010 & $+8.99^*$ \ci{+5.92}{+12.01} 0.0008 \\
$\Delta$ \ours{}+BM25+dense $-$ \ours{}+BM25 & $+1.85$ \ci{-0.31}{+3.95} 0.3086 & $+1.44$ \ci{-0.24}{+3.14} 0.2812 & $+1.26$ \ci{-0.57}{+3.01} 0.3920 \\
$\Delta$ \ours{}+BM25+dense $-$ BM25 & $+10.18^*$ \ci{+7.14}{+13.16} 0.0008 & $+9.04^*$ \ci{+6.92}{+11.21} 0.0008 & $+9.14^*$ \ci{+6.47}{+11.82} 0.0008 \\
$\Delta$ \ours{} $\cup$ dense RAG $-$ dense RAG + reranker & $+3.61$ \ci{+1.49}{+5.80} 0.0522 & $+6.20^*$ \ci{+4.42}{+8.06} 0.0008 & $+6.64^*$ \ci{+3.74}{+9.53} 0.0020 \\
$\Delta$ \ours{} $\cup$ dense RAG $-$ \ours{} & $+2.19$ \ci{-0.34}{+4.79} 0.3086 & $+1.31$ \ci{-0.82}{+3.50} 0.3264 & $+0.72$ \ci{-1.56}{+3.27} 0.6878 \\
\bottomrule
\end{tabular}}
\end{table}

\begin{table}[!htbp]
\centering
\caption{\textbf{\ours{} fused with dense retrieval} (Qwen2.5-7B, \bench{}, $B{=}2{,}048$, 400 questions per length). Notation as in Table~\ref{tab:stack-kv}.}
\label{tab:stack-dense-qwen}
\scriptsize
\setlength{\tabcolsep}{3.5pt}
\resizebox{\ifdim\width>\linewidth\linewidth\else\width\fi}{!}{%
\begin{tabular}{@{}lccc@{}}
\toprule
Method & 32k & 64k & 128k \\
\midrule
\ours{} & 66.6 & 65.5 & 60.9 \\
\ours{}+BM25 & 72.1 & 69.9 & 68.2 \\
\ours{}+dense & 73.0 & 70.8 & 73.6 \\
\ours{}+BM25+dense & 71.1 & 71.5 & 73.8 \\
\ours{} $\cup$ dense RAG & 69.9 & 69.8 & 70.7 \\
dense RAG + reranker & 67.0 & 66.7 & 61.8 \\
BM25 & 64.0 & 62.9 & 62.4 \\
\midrule
$\Delta$ \ours{}+dense $-$ dense RAG + reranker & $+5.96^*$ \ci{+2.75}{+9.18} 0.0132 & $+4.12$ \ci{+1.40}{+6.89} 0.1143 & $+11.80^*$ \ci{+7.75}{+15.81} 0.0008 \\
$\Delta$ \ours{}+dense $-$ \ours{} & $+6.38^*$ \ci{+3.09}{+9.54} 0.0077 & $+5.30^*$ \ci{+3.60}{+6.90} 0.0008 & $+12.65^*$ \ci{+9.80}{+15.62} 0.0008 \\
$\Delta$ \ours{}+dense $-$ BM25 & $+8.91^*$ \ci{+5.30}{+12.75} 0.0008 & $+7.91^*$ \ci{+5.16}{+10.64} 0.0018 & $+11.21^*$ \ci{+7.56}{+15.00} 0.0008 \\
$\Delta$ \ours{}+BM25+dense $-$ dense RAG + reranker & $+4.15$ \ci{+0.66}{+7.53} 0.1521 & $+4.81^*$ \ci{+2.27}{+7.22} 0.0472 & $+12.05^*$ \ci{+8.14}{+15.96} 0.0008 \\
$\Delta$ \ours{}+BM25+dense $-$ \ours{}+BM25 & $-1.00$ \ci{-3.31}{+1.29} 0.5246 & $+1.61$ \ci{-0.46}{+3.54} 0.2247 & $+5.59^*$ \ci{+3.27}{+7.85} 0.0008 \\
$\Delta$ \ours{}+BM25+dense $-$ BM25 & $+7.10^*$ \ci{+2.79}{+11.70} 0.0325 & $+8.60^*$ \ci{+6.11}{+11.19} 0.0008 & $+11.46^*$ \ci{+8.11}{+14.90} 0.0008 \\
$\Delta$ \ours{} $\cup$ dense RAG $-$ dense RAG + reranker & $+2.88$ \ci{+0.39}{+5.31} 0.1468 & $+3.05$ \ci{+0.68}{+5.42} 0.1143 & $+8.96^*$ \ci{+5.80}{+12.10} 0.0008 \\
$\Delta$ \ours{} $\cup$ dense RAG $-$ \ours{} & $+3.29$ \ci{+0.38}{+6.24} 0.1521 & $+4.22^*$ \ci{+1.66}{+6.81} 0.0460 & $+9.81^*$ \ci{+6.56}{+13.24} 0.0008 \\
\bottomrule
\end{tabular}}
\end{table}

\begin{table}[!htbp]
\centering
\caption{\textbf{\ours{} fused with dense retrieval on InfiniteBench} (Llama-3.1-8B, 351 En.QA and 229 En.MC questions, unit = question, Holm $p$ within each task and budget). Notation as in Table~\ref{tab:stack-kv}.}
\label{tab:stack-ib}
\scriptsize
\setlength{\tabcolsep}{3.5pt}
\resizebox{\ifdim\width>\linewidth\linewidth\else\width\fi}{!}{%
\begin{tabular}{@{}lcccc@{}}
\toprule
Method & En.QA 2{,}048 & En.QA 8{,}192 & En.MC 2{,}048 & En.MC 8{,}192 \\
\midrule
\ours{} & 13.9 & 20.4 & 62.9 & 67.7 \\
\ours{}+BM25 & 14.3 & 21.6 & 69.9 & 69.4 \\
\ours{}+dense & 15.5 & 22.0 & 70.7 & 73.8 \\
\ours{}+BM25+dense & 15.8 & 24.9 & 69.9 & 72.1 \\
\ours{} $\cup$ dense RAG & 18.1 & 25.5 & 68.6 & 76.4 \\
dense RAG + reranker & 17.1 & 25.1 & 73.8 & 76.4 \\
BM25 & 13.7 & 21.5 & 64.6 & 72.1 \\
\midrule
$\Delta$ \ours{}+dense $-$ dense RAG + reranker & $-1.64$ \ci{-3.54}{+0.21} 0.5387 & $-3.05^*$ \ci{-5.39}{-0.75} 0.0474 & $-3.06$ \ci{-9.17}{+3.06} 0.5966 & $-2.62$ \ci{-7.42}{+2.18} 1.0000 \\
$\Delta$ \ours{}+dense $-$ \ours{} & $+1.63$ \ci{-0.45}{+3.73} 0.5399 & $+1.63$ \ci{-0.49}{+3.79} 0.5415 & $+7.86$ \ci{+2.18}{+13.54} 0.1071 & $+6.11$ \ci{+0.87}{+11.79} 0.3264 \\
$\Delta$ \ours{}+dense $-$ BM25 & $+1.78$ \ci{-0.42}{+4.08} 0.5399 & $+0.51$ \ci{-2.35}{+3.36} 1.0000 & $+6.11$ \ci{+0.44}{+11.79} 0.3094 & $+1.75$ \ci{-3.49}{+6.99} 1.0000 \\
$\Delta$ \ours{}+BM25+dense $-$ dense RAG + reranker & $-1.39$ \ci{-3.61}{+0.75} 0.5399 & $-0.17$ \ci{-2.43}{+2.06} 1.0000 & $-3.93$ \ci{-10.04}{+2.18} 0.5966 & $-4.37$ \ci{-9.17}{+0.44} 0.7013 \\
$\Delta$ \ours{}+BM25+dense $-$ \ours{} & $+1.88$ \ci{-0.39}{+4.19} 0.5399 & $+4.50^*$ \ci{+1.93}{+7.14} 0.0048 & $+6.99$ \ci{+0.44}{+13.54} 0.3094 & $+4.37$ \ci{-0.87}{+9.61} 0.7619 \\
$\Delta$ \ours{}+BM25+dense $-$ BM25 & $+2.03$ \ci{+0.04}{+4.16} 0.3829 & $+3.39^*$ \ci{+0.93}{+5.93} 0.0474 & $+5.24$ \ci{+0.87}{+9.61} 0.2696 & $+0.00$ \ci{-4.37}{+4.80} 1.0000 \\
$\Delta$ \ours{} $\cup$ dense RAG $-$ dense RAG + reranker & $+0.91$ \ci{-0.95}{+2.86} 0.5399 & $+0.38$ \ci{-1.93}{+2.67} 1.0000 & $-5.24$ \ci{-10.04}{-0.87} 0.3094 & $+0.00$ \ci{-3.93}{+3.93} 1.0000 \\
$\Delta$ \ours{} $\cup$ dense RAG $-$ \ours{} & $+4.17^*$ \ci{+2.01}{+6.42} 0.0018 & $+5.05^*$ \ci{+2.50}{+7.66} 0.0009 & $+5.68$ \ci{-1.31}{+12.23} 0.5483 & $+8.73^*$ \ci{+3.49}{+13.97} 0.0153 \\
$\Delta$ \ours{} $\cup$ dense RAG $-$ BM25 & $+4.32^*$ \ci{+1.95}{+6.82} 0.0032 & $+3.94^*$ \ci{+1.13}{+6.82} 0.0350 & $+3.93$ \ci{-1.31}{+9.17} 0.5966 & $+4.37$ \ci{-0.44}{+9.17} 0.7013 \\
\bottomrule
\end{tabular}}
\end{table}

%% file: v3_sections/app_limits.tex
\section{Scope and limitations}
\label{app:limits}

This section collects the settings in which \ours{} does not lead, with the evidence behind each.

\paragraph{The full cache at short lengths.}
On Llama at 32k and 64k the full cache scores 7.75 and 7.56 points above \ours{}+BM25 (Table~\ref{tab:e2-deltas-pq-llama}).
It leads through variable tracking, whose chained assignments reward reading everything (96.0 and 87.8 against 38.5 and 36.5, Appendix~\ref{app:e2-full}).

\paragraph{Tuned text fusions.}
On Llama at 64k, \ours{}+BM25 scores above BM25 + dense, the strongest text retriever there, but the margin is not significant after Holm correction ($+3.36$, Table~\ref{tab:holm-main}).
At 128k the two are not significantly different ($+1.40$).

\paragraph{Qwen2.5 and the two-head variant.}
On Qwen2.5 the heads alone score above BM25 only at 64k ($+2.61$, CI excludes zero) and are not significantly different from it at 32k and 128k ($+2.54$, $-1.44$, Table~\ref{tab:e2-deltas-pq-qwen}), so the fusion is used on this model.
The fusion significantly beats BM25 + lexical density, the strongest text retriever there, at 32k ($+3.25$) and is not significantly different from it at 128k ($+0.38$).
At 128k \ours{} alone is significantly below BM25 + lexical density ($-6.94$) and below REFORM-style ($-3.65$, CI excludes zero).
\ourstwo{} relies on Llama's two strongest heads sharing one KV head of layer 2.
On Qwen2.5 two heads score 38.6, 36.7 and 29.6, or 63.2, 63.4 and 62.2 with BM25.
On Llama the two layer-2 heads of \ourstwo{} are not significantly different from all 16 heads at 32k and 128k ($+0.81$, $-1.70$) and fall 2.76 points below at 64k (Appendix~\ref{app:two-head}).

\paragraph{Natural text and chat histories.}
On the SQuAD and HotpotQA families, \ours{}+BM25 is not significantly different from reranked dense RAG on Llama ($-4.38$, $-2.50$, $-3.75$, Appendix~\ref{app:qa-subset}).
On NoLiMa, lexical scores add noise, and \ours{} alone beats its BM25 fusion by 4.61--12.11 points (Appendix~\ref{app:nolima}).
On LongMemEval-S, the answers of \ours{}+BM25 (43.2\%) are not significantly different from those of dense RAG (45.3\%, Appendix~\ref{app:lme}).

\paragraph{Documents and very long contexts.}
Table~\ref{tab:real} gives LongBench QA, and Appendix~\ref{app:beyond-full} gives the runs beyond the 128k window.
On LongBench, \ours{} is not significantly different from reranked dense RAG and the fusions on Llama and Qwen2, and on Qwen2 also from the truncated full cache.
Reranked dense RAG and the truncated full cache lead significantly on Mistral ($-3.83$, $-3.38$).
On Llama the truncated full cache scores 3.04 points above \ours{}, a significant gap.
On \bench{} at 128k with Mistral and Qwen2, the fusion is not significantly different from BM25.
On Llama at 512k, BM25 + dense, the strongest text retriever beyond the window, scores above \ours{}+BM25 ($-4.42$, CI excludes zero).
At 256k and 1M the two are not significantly different ($+0.25$, $-3.54$).
On whole books (InfiniteBench, Llama), reranked dense RAG scores significantly higher than \ours{} alone.
The union of the two is not significantly different from dense RAG in three of the four task and budget cells and scores below it on En.MC at $B{=}2{,}048$ ($-5.24$, CI excludes zero, Appendices~\ref{app:infbench} and~\ref{app:stack}).
\ours{}+BM25 is significantly below dense RAG in two of the four task and budget cells ($-3.45$, $-6.99$), and on En.QA the truncated full cache scores highest (F1 31.70).

\paragraph{Combination with KV compression.}
These runs use Llama with test episodes 0--5 of every family.
At 128k the combined prompts run up to 625 positions past the window.
Sharing the 2{,}048-token budget with a compressed cache is not significantly above giving all of it to \ours{} in any cell.
It is significantly below in four of the nine cells (Table~\ref{tab:stack-kv}).
On these episodes \ours{} alone scores 74.8, 72.1 and 72.4 with 2{,}048 tokens.
The combinations with StreamingLLM and SnapKV reach 66.4--71.6, and the KVzip combination falls to 23.5 at 128k.
The combination is therefore a way to add \ours{} to a pipeline that keeps a compressed cache, not a gain over \ours{} alone.

\paragraph{Privacy.}
\ours{} changes no model weights.
Its index keeps token ids and therefore carries the privacy obligations of the document.

%% file: v3_sections/app_scaling.tex
\section{Model size and input length}
\label{app:scaling}

\paragraph{Model size and head budget.}
We extend the frozen Qwen2.5-Instruct evaluation to 7B, 14B and 32B (YaRN $\times4$), keeping each model's head ranking, \pqc{}, 4,096-token index chunks, prompts and greedy decoding fixed.
Table~\ref{tab:scaling-mq} compares 16 heads with prespecified counts $K_p=16/39/52$, approximately 2\% of each model's query heads across all layers.
The dense and lexical-fusion baselines retain the paper's configurations, including the lexical-fusion weight of 0.8, without retuning.
The core set has 10 test episodes per family and length, each with eight questions.
On 14B at 128k, the 16-head \ours{}+BM25 gains $10.70$ points over dense RAG, CI $[6.90,14.40]$, Holm $p=0.0036$; the QA-only difference is $0.63$ with CI $[-6.25,7.50]$.
The separate development sweep over $K\in\{1,2,4,8,16,24,32,39,48,52,64\}$ uses two episodes per family and length and selects $16/32/39$ heads for \ours{} and $32/32/64$ for its BM25 fusion, averaging equally over the 15 family--length cells before testing.
For 32B, the selected 39 heads raise the held-out 128k score of \ours{} alone from 43.14 to 53.48 ($\Delta=10.34$, CI $[6.54,13.93]$, Holm $p=0.0108$).
The development answer-presence diagnostic already shows a localization gap with exact keys: 16, 32 and 64 heads give 48.75\%, 64.06\% and 58.75\% at 128k (16 questions per family).
Head allocation therefore matters when transferring the memory across models; parameter count alone does not determine retrieval quality.

\begin{table}[!htbp]
\centering
\caption{\textbf{Model size and a 1M-token stress test.} Each cell is five-family macro score / natural-text QA macro score ($\times100$); all methods read 2,048 tokens. The QA subset comprises SQuAD and HotpotQA. $K_p=16/39/52$ for 7B/14B/32B. Panel (a) uses the original core test episodes; panel (b) uses an independent corpus-expansion test, \texttt{mq\_unique\_docs\_v1}. Dense denotes dense RAG with reranking, and Lex denotes BM25 + lexical density. All values are held-out test scores.}
\label{tab:scaling-mq}
\small
\setlength{\tabcolsep}{4pt}
\renewcommand{\arraystretch}{1.12}
\resizebox{\ifdim\width>\linewidth\linewidth\else\width\fi}{!}{%
\begin{tabular}{@{}llccccc@{}}
\toprule
Model & Input & \shortstack{\ours{}\\16 heads} & \shortstack{\ours{}+BM25\\16 heads} & \shortstack{\ours{}+BM25\\$K_p$ heads} & Dense & Lex \\
\midrule
\multicolumn{7}{@{}l}{\textit{(a) Original core: 400 questions per length, including 160 QA questions}} \\
7B & 32k & 66.58 / 62.50 & 72.14 / 70.00 & 72.14 / 70.00 & 66.99 / 65.63 & 68.89 / 66.88 \\
 & 64k & 65.54 / 59.38 & 69.91 / 66.88 & 69.91 / 66.88 & 66.71 / 70.63 & 67.34 / 66.25 \\
 & 128k & 60.91 / 53.13 & 68.23 / 66.25 & 68.23 / 66.25 & 61.76 / 66.88 & 67.85 / 66.88 \\
14B & 32k & 65.79 / 65.00 & 71.16 / 69.38 & 71.14 / 71.25 & 69.04 / 72.50 & 70.88 / 73.75 \\
 & 64k & 65.63 / 55.63 & 71.25 / 64.38 & 70.33 / 65.63 & 65.34 / 68.75 & 67.19 / 65.00 \\
 & 128k & 67.11 / 60.63 & 71.80 / 70.00 & 73.04 / 73.75 & 61.10 / 69.38 & 69.61 / 73.75 \\
32B & 32k & 48.39 / 35.63 & 72.50 / 70.63 & 72.20 / 71.25 & 72.18 / 80.00 & 73.49 / 76.25 \\
 & 64k & 44.33 / 28.13 & 68.43 / 63.75 & 69.79 / 63.75 & 68.34 / 73.75 & 70.03 / 70.63 \\
 & 128k & 43.14 / 26.25 & 69.91 / 68.75 & 72.14 / 66.25 & 63.81 / 74.38 & 70.79 / 73.13 \\
\midrule
\multicolumn{7}{@{}l}{\textit{(b) Independent 1M stress test: 200 questions, including 80 QA questions}} \\
7B & 1M & 56.05 / 42.50 & 67.55 / 60.00 & 67.55 / 60.00 & 48.90 / 55.00 & 63.13 / 55.00 \\
14B & 1M & 66.75 / 55.00 & 68.63 / 61.25 & 67.33 / 65.00 & 49.38 / 58.75 & 66.50 / 67.50 \\
32B & 1M & 32.45 / 16.25 & 68.40 / 65.00 & 72.93 / 67.50 & 50.18 / 61.25 & 67.50 / 63.75 \\
\bottomrule
\end{tabular}}
\end{table}

\paragraph{Scaling the retrievable input.}
The independent five-family stress test expands inputs through 128k, 256k, 512k and 1M while preserving questions, gold documents and relative evidence positions.
It uses five test episodes per family and length; QA distractors are distinct paragraphs from the deduplicated SQuAD/HotpotQA pool, reaching 7,256--7,426 documents per 1M QA input.
At 1M, the 32B/52-head \ours{}+BM25 scores 72.93 versus 50.18 for dense RAG ($\Delta=22.75$, CI $[17.98,27.90]$, Holm $p=0.0032$) and 67.50 for BM25 + lexical density ($\Delta=5.43$, CI $[2.95,8.15]$, Holm $p=0.306$).
Its QA score is 67.50 versus 61.25 and 63.75, respectively, with neither QA difference significant after Holm correction.
These results establish useful answering over 1M indexed tokens with a fixed 2k reading budget and no additional neural retrieval model.

For natural-text transfer, Table~\ref{tab:scaling-books} separates the original InfiniteBench books (86 books, 351 En.QA and 229 En.MC questions) from a controlled distractor-book extension.
The latter keeps the same 27 complete target books and 191 questions (112 QA, 79 MC) at 128k/256k/512k/1M, adding distinct other books while keeping each target centered and each shorter input a contiguous part of the longer one.
This is an InfiniteBench-derived input expansion; the original books reach 745,586 Qwen tokens, and the derived 1M inputs contain 4--8 books.
At an 8k reading budget, 32B/52-head \ours{}+BM25 keeps MC accuracy at 77.22 across all four lengths, with QA F1 changing from 13.84 to 12.65; dense RAG reaches 82.28 and 14.13 at 1M.
The advantage over dense RAG does not grow significantly with input length in any of the twelve planned model--task--budget comparisons (all CIs include zero; all Holm $p=1$).
The extended input is an indexed retrieval pool, with the same frozen model answering from 2k or 8k selected tokens.

\begin{table}[!htbp]
\centering
\caption{\textbf{Whole books and controlled input expansion.} Each cell reports En.QA F1 / En.MC accuracy ($\times100$), with $K_p=16/39/52$ heads for the two \ours{} rows of 7B/14B/32B. Native uses all 580 original questions; the derived columns use the same 191 questions across lengths. $B$ is the number of context tokens read to answer. Native and derived columns contain different question sets and are not paired length comparisons. BM25 and dense RAG with reranking are evaluated on the identical questions and input for each column.}
\label{tab:scaling-books}
\small
\setlength{\tabcolsep}{4pt}
\renewcommand{\arraystretch}{1.12}
\resizebox{\ifdim\width>\linewidth\linewidth\else\width\fi}{!}{%
\begin{tabular}{@{}llcccc@{}}
\toprule
Model & Method & \shortstack{Native\\$B=8$k} & \shortstack{Derived 128k\\$B=8$k} & \shortstack{Derived 1M\\$B=8$k} & \shortstack{Derived 1M\\$B=2$k} \\
\midrule
7B & \ours{} & 14.37 / 63.32 & 12.64 / 63.29 & 8.84 / 46.84 & 7.50 / 41.77 \\
 & \ours{}+BM25 & 17.72 / 69.00 & 12.50 / 65.82 & 12.95 / 64.56 & 9.76 / 58.23 \\
 & BM25 & 19.85 / 69.43 & 14.43 / 64.56 & 9.02 / 63.29 & 8.34 / 55.70 \\
 & Dense RAG + reranker & 21.66 / 72.93 & 15.10 / 69.62 & 16.57 / 59.49 & 14.17 / 62.03 \\
\midrule
14B & \ours{} & 12.17 / 63.32 & 12.26 / 69.62 & 8.17 / 58.23 & 7.03 / 48.10 \\
 & \ours{}+BM25 & 14.83 / 73.36 & 12.95 / 68.35 & 10.49 / 67.09 & 9.62 / 60.76 \\
 & BM25 & 14.33 / 77.73 & 13.77 / 73.42 & 9.63 / 69.62 & 8.75 / 64.56 \\
 & Dense RAG + reranker & 15.63 / 80.79 & 14.98 / 79.75 & 13.08 / 75.95 & 12.58 / 77.22 \\
\midrule
32B & \ours{} & 12.94 / 67.25 & 12.46 / 72.15 & 5.99 / 40.51 & 4.19 / 39.24 \\
 & \ours{}+BM25 & 16.49 / 78.17 & 13.84 / 77.22 & 12.65 / 77.22 & 7.79 / 67.09 \\
 & BM25 & 16.52 / 79.48 & 14.33 / 78.48 & 11.95 / 81.01 & 9.00 / 72.15 \\
 & Dense RAG + reranker & 17.29 / 84.72 & 15.42 / 84.81 & 14.13 / 82.28 & 13.42 / 75.95 \\
\bottomrule
\end{tabular}}
\end{table}

For these new studies, paired CIs and sign-flip tests use 10,000 draws: episodes stratified by task for the multi-question tests, and whole books with question-weighted scores for InfiniteBench.
Core baseline comparisons apply Holm correction across both baselines, models, methods and lengths separately for each head-choice policy and for all-task versus QA scores (36 tests for fixed 16 heads); the corresponding 1M stress-test contrast belongs to the 32-test proportional-head family over all four lengths.
The selected-head versus 16-head core contrast uses its 18-test all-task family.
The book-length test corrects the twelve prespecified interactions $[\text{hybrid}-\text{dense}]_{1\mathrm{M}}-[\text{hybrid}-\text{dense}]_{128\mathrm{k}}$ across three models, two tasks and two budgets; the tabulated endpoint scores are descriptive.
\FloatBarrier